\documentclass[colophon,final,table,inline,utf8,russian,english]{phduio}
\usepackage{rotating}
\usepackage{phdstyle}   
\usepackage{tikz-dependency}
\usepackage{amsthm}
\usepackage{csquotes}
\usepackage{inputenc}
\DeclareUnicodeCharacter{0301}{\'{e}}

\newtheorem{question}{RQ}[chapter]
\newtheorem{Question}{RQ}

\counterwithin*{Question}{chapter}

\newcolumntype{P}[1]{>{\centering\arraybackslash}p{#1}}
\usepackage{booktabs}
\newcommand{\head}[1]{\textnormal{\textbf{#1}}}
\usepackage{enumerate}

\usepackage{etoolbox}
\AtBeginEnvironment{myexample}{\small}
\usepackage{relsize}
\usepackage{covington}
\usepackage[encapsulated]{CJK}

\usepackage[linesnumbered,ruled,vlined]{algorithm2e}
\usepackage{amsmath,amsfonts,amssymb}
\usepackage{algorithmic}
\usepackage{pifont}
\usepackage{stfloats}
\SetKwInput{KwInput}{Input} 
\SetKwInput{KwOutput}{Output}
\usepackage{rotating}
\usepackage{graphicx}
\usepackage{adjustbox}
\newcommand{\cev}[1]{\reflectbox{\ensuremath{\vec{\reflectbox{\ensuremath{#1}}}}}}
\usepackage{amsmath}
\DeclareMathOperator*{\argmax}{arg\,max}

\usepackage{nth}
\newcounter{eqnnosave}
{\end{enumerate}%
\setcounter{eqnnosave}{\arabic{enumi}}%
}
\usepackage{latexsym}
\usepackage{subfig}
\usepackage{url}
\usepackage{xcolor}
\usepackage{enumitem}
\usepackage{capt-of}
\usepackage{babel}

\newcommand{\egc}{e.g., }
\newcommand{\iec}{i.e., }

\newcommand{\Figref}[1]{Figure~\ref{#1}}  
\newcommand{\tabref}[1]{Table~\ref{#1}}

\newcommand{\secref}[1]{Section~\ref{#1}}
\newcommand{\equref}[1]{Eq.~(\ref{#1})}
\newcommand{\task}{\mathcal{T}}
\newcommand{\loss}{\mathcal{L}}

\newcommand{\learner}{\texttt{M}}
\newcommand{\lossi}{\loss_{\task_i}}

\newcolumntype{C}[1]{>{\arraybackslash}p{#1}}

\AtEveryBibitem{
    \clearname{editor}
    \clearfield{date}
    \clearfield{eprint}
    \clearfield{isbn}
    \clearfield{issn}
    \clearlist{location}
    \clearfield{month}
    \clearfield{series}
    \clearfield{url}%
    \clearfield{numpages}%
    \clearfield{language}%
    \clearfield{bibsource}%
    \clearfield{biburl}%
    \clearfield{note}%
    \clearfield{urldate}%
    \clearfield{review}%
    \clearfield{issn} 
    \clearfield{doi} 
    \clearfield{chapter}
    \ifentrytype{misc}{%
  }{%
    \clearfield{url}%
  }%
}

\author{Farhad Nooralahzadeh}
\title{Low-Resource Adaptation of Neural NLP Models}
\department{Department of Informatics}
\faculty{Faculty of Mathematics and Natural Sciences}
\ISSN{1501-7710}           
\dissertationseries{2299}  

\begin{document}

    \frontmatter        

    \uiotitle

    \thispagestyle{empty}
\vspace*{\stretch{1}}
\begin{flushright}
    \emph{In memory of my father,\\
    I miss you everyday! You always encouraged me.}
\end{flushright}
\vspace*{\stretch{3}}
    \chapter{Abstract}

Real-world applications of natural language processing (NLP) are challenging. NLP models rely heavily on supervised machine learning and require large amounts of annotated data.
These resources are often based on language data available in large quantities, such as English newswire. However, in real-world applications of NLP, the textual resources vary across several dimensions, such as language, dialect, topic, and genre. It is challenging to find annotated data of sufficient amount and quality.
The objective of this thesis is to investigate methods for dealing with such low-resource scenarios in information extraction and natural language understanding.
To this end, we study distant supervision and sequential transfer learning in various low-resource settings. We develop and adapt neural NLP models to explore a number of research questions concerning NLP tasks with minimal or no training data.
We first make use of sequential transfer learning in order to induce non-contextualized word embeddings to capture domain-specific semantics and benefit downstream tasks in NLP. We subsequently enhance these embeddings using a domain-specific knowledge resource and present a benchmark dataset for intrinsic and extrinsic evaluation of domain embeddings. In the information extraction field, we propose a hybrid model that combines a reinforcement learning algorithm with partial annotation learning to clean the noisy, distantly supervised data for low-resource named entity recognition in different domains and languages.
In the next step following entity detection in the information extraction pipeline, we design a neural architecture with syntactic input representation to alleviate domain impact in low-resource relation extraction. Finally, we introduce a cross-lingual meta-learning framework that provides further improvements in low-resource cross-lingual natural language understanding tasks in various settings and languages.

    \chapter{Acknowledgments}
This Ph.D. thesis and research project have been materialized with the support and sincere encouragement and guidance of several people who have inspired me to continue my path and settle for nothing less than the best. Thus, I would like to express my utmost gratitude to these people for their support.

First and foremost, I would like to express my sincere gratitude to my supervisors Lilja Øvrelid and Jan Tore Lønning, for the continuous support of my Ph.D. study and related research, for their patience, motivations, and immense knowledge. Their guidance helped me in all the time of research and writing of this thesis. I could not have imagined having better supervisors and mentors for my Ph.D. study. 

Besides my supervisors, I would like to thank the SIRIUS management team. I am especially grateful for the hard work and support of Arild Waaler, David Cameron, Ingrid Chieh Yu, and Lise Reang, as well as Tom Erling Henriksen for his advice during and after the mentoring program. 

I am thankful to Mara Abel, and her team at the Federal University of Rio Grande do Sul (UFRGS) for hosting me and many fruitful discussion on Ontology during my research visit in Brazil. Obrigado a todos.

I also would like to say a special thanks to Isabelle Augenstein, Johannes Bjerva, and CopeNLU research team who have hosted, supported, and encouraged me during my second research visit at the University of Copenhagen.
 
From the bottom of my heart, I would like to say big thank you for all the SIRIUS and Language Technology Group, past and present members, for the feedback and support on different occasions and for creating a friendly work and research environment. Tusen takk.

Last but not least, I would like to thank my family. Finishing this work would not have been possible without your love and support. I love you all.



\vskip\onelineskip
\begin{flushleft}
    \sffamily
    \uiocolon\textbf{\theauthor}
    \\
    Oslo,\MONTH\the\year
\end{flushleft}

    \cleartorecto
    \tableofcontents    
    \cleartorecto
    \listoffigures      
    \cleartorecto
    \listoftables       

    \mainmatter         

    \chapter{Introduction}
\label{sec:intro}
\section{Motivation}
There is a growing interest in real-world applications of natural language processing (NLP) for extracting, summarizing, and analyzing textual data. While NLP methods have led to many breakthroughs in practical applications, most notably perhaps in machine translation, question answering, and natural language inference, it is still challenging to use NLP in many real-world scenarios.
Since NLP relies heavily on supervised machine learning, the modeling of most NLP tasks requires large amounts of annotated data.
These resources are often based on language data available in large quantities, such as English newswire. However, in NLP's real-world applications, the textual resources may vary across several dimensions, such as language, dialect, topic, genre, etc. Considering the cross-product of these dimensions, it is difficult to find annotated data of sufficient amount and quality that spans all possible combinations and assists current advanced NLP techniques \parencite{DBLP:journals/corr/Plank16}.

In general, NLP application scenarios, can be classified into three categories according to their data resources \parencite{duong2017natural}:
\begin{enumerate*}[label=(\roman*)]
    \item \emph{High-} or \emph{Rich-resource} settings, where a large amount of annotated data is available;
    \item \emph{Low-resource} or \emph{Resource-poor} ones, where there is limited annotated data; and 
    \item \emph{Zero-resource} settings, where there is no annotated data available in the target context. 
\end{enumerate*}
Off-the-shelf resource-intensive NLP techniques tend to perform poorly where annotated data are not readily available (i.e., low-resource and zero-resource settings). An immediate solution is to create annotated data representative of new target scenarios. However, collecting and annotating corpora for each new variety requires experts and is usually expensive. Therefore, it is necessary to find techniques that can relieve the problem of creating training sets.
Our primary motivation in this thesis is based on the following argument in  \cite{DBLP:journals/corr/Plank16}:
\begin{displayquote}
"If we embrace the variety
of this heterogeneous data by combining it with
proper algorithms, in addition to including text covariates/latent factors, we will not only produce more robust models, but will also enable adaptive language technology capable of addressing natural language variation." 
\end{displayquote}
NLP for low-resource settings has recently received much attention, with dedicated workshops on the topic \parencite{ws-2018-deep, emnlp-2019-deep}. In general, most previous work associates the low resource property with the language dimension \parencite{king2015practical,tsvetkov2016linguistic,duong2017natural,Kann_2019}. In this work, we follow \cite{DBLP:journals/corr/Plank16} and consider the low-resource setting as fundamentally multi-dimensional, spanning over all kinds of variability within natural language, e.g., language, dialect, domain, genre. Therefore, the scope of this thesis is broader, and we explore how to adapt and improve the performance of NLP algorithms in a number of different low-resource settings, spanning across different domains, genres and languages and dealing with a number of central NLP tasks. 
We here make a distinction between domain, genre, and language. We call the variety aspect \emph{domain} when the source dataset defers in terms of topic (chapters \ref{sec:second}, \ref{sec:third}, and \ref{sec:fourth}). The term \emph{Topic} is the general subject of a document and ranging from very broad to more detailed such as oil and gas, biomedical, and e-commerce. Furthermore, we use the term \emph{genre}, where the source dataset covers non-topical text properties such as function, style, and text type in Chapter \ref{sec:fifth}. 
 
A number of approaches have been proposed to address the challenge of low-resource scenarios. They have significantly improved upon the state-of-the-art on a wide range of NLP tasks for various settings. In this thesis, we make use of adaptation techniques that fall into the following main paradigms: 
\begin{enumerate*}[label=(\roman*)]
    \item \emph{Distant Supervision:} A supervised learning
    paradigm where the training data is not manually annotated, but automatically generated using knowledge bases (KBs) and heuristics \parencite{Mintz:2009:DSR:1690219.1690287} 
        \item \emph{Transfer Learning:} Techniques for leveraging data from additional domains, tasks or languages to train a model with better generalization properties \parencite{ruder-etal-2019-transfer}.
\end{enumerate*}

Real-world applications of NLP typically incorporate a number of more specialized, task-specific systems, e.g., pre-processing,  various types of syntactic or semantic analysis, inference, etc. Here we focus mainly on NLP tasks from the areas of Information Extraction (specifically Named Entity Recognition and Relation Extraction) and Natural Language Understanding (more specifically Natural Language Inference and Question-Answering).

Before we delve into the theoretical and experimental study of our work, in the next few pages, we present our research questions and highlight some of our main contributions and, finally, provide a more detailed outline of the thesis.
\section{Research Questions}
At a high level of abstraction, we attempt to answer the following main research questions in this thesis:
\begin{Question}\label{RQ.1}
    What is the impact of different input representations in neural low-resource NLP?
\end{Question}
\noindent The vector representations of tokens instantiate the distributional hypothesis by learning representations of the meaning of words, called embeddings, directly from text corpora. These representations are crucial elements in the performance of downstream NLP systems and underlie the more powerful and more recent contextualized word representations.
We here study input representations trained on data from specific domains using sequential transfer learning of word embeddings. Concretely, we attempt to answer the following research questions:
\begin{enumerate}[label=(\roman*)]
    \item \emph{Can word embedding models capture domain-specific semantic relations even when trained with a considerably smaller corpus size?}
    \item \emph{Are domain-specific input representations beneficial in downstream NLP tasks?} 
\end{enumerate}
In order to address these questions, we study input representations trained on data from a low resource domain (Oil and Gas). We conduct intrinsic and extrinsic evaluations of both general and domain-specific embeddings. Further, we investigate the effect of domain-specific word embeddings in the input layer of a downstream sentence classification task in the same domain.
Domain-specific embeddings are further studied in the context of the relation extraction task on data from an unrelated genre and domain: scientific literature from the NLP domain.

In many NLP tasks, syntactic information is viewed as useful, and a variety of new approaches incorporate syntactic information in their underlying models. Within the context of this thesis, we hypothesize that syntax may provide a level of abstraction that can be beneficial when there is little available labeled data. We pursue this line of research particularly for low-resource relation extraction, and we look at the following question:
\begin{enumerate}[resume, label=(\roman*)]
    \item \emph{What is the impact of syntactic dependency representations in low-resource neural relation extraction?}
\end{enumerate}
We design a neural architecture over dependency paths combined with domain-specific word embeddings to extract and classify semantic relations in a low-resource domain. We explore the use of different syntactic dependency representations in a neural model and compare various dependency schemes. 
We further compare with a syntax-agnostic approach and perform an error analysis to gain a better understanding of the results.
\begin{Question}\label{RQ.2}
    How can we incorporate domain knowledge in low-resource NLP?
\end{Question} 
\noindent Technical domains often have knowledge resources that encode domain knowledge in a structured format. There is currently a line of research that tries to incorporate this knowledge encoded in domain resources in NLP systems. The domain knowledge can be leveraged either to provide weak supervision or to include additional information not available in text corpora to improve the model's performance. Here, we explore this line of research in low-resource scenarios by addressing the following questions: 
\begin{enumerate}[label=(\roman*)]
     \item \emph{How can we take advantage of existing domain-specific knowledge resources to enhance our models?}
 \end{enumerate}
We investigate the impact of domain knowledge resources in enhancing embedding models. We augment the domain-specific model by providing vector representations for infrequent and unseen technical terms using a domain knowledge resource and evaluate its impact by intrinsic and extrinsic evaluations.

Given the availability of domain-specific knowledge resources, distant supervision can be applied to generate automatically labeled training data in low-resource domains. In this thesis, we explore the use of distant supervision for low-resource Named Entity Recognition (NER) in various domains and languages. We here address the following question: 
\begin{enumerate}[resume, label=(\roman*)]
      \item \emph{How can we address the problem of low-resource NER using distantly supervised data?} 
\end{enumerate}
The outcome of distant supervision, however, is often noisy. To address this issue, we explore the following research question: 
\begin{enumerate}[resume, label=(\roman*)]
    \item \emph{How can we exploit a reinforcement learning approach to improve NER in low-resource scenarios?}
\end{enumerate}
We present a system which addresses the problem of noisy, distantly supervised data using reinforcement learning and partial annotation learning.
\begin{Question}\label{RQ.3}
    How can we address the challenges of low-resource scenarios using transfer learning techniques?\todo{sequential transfer learning: pre-trained Embeddings and meta-learning}
\end{Question}
\noindent Transfer learning has yielded significant improvements in various NLP tasks. The most dominant practice of transfer learning is to pre-train embedding representations on a large unlabeled text corpus and then to transfer these representations to a supervised target task using labeled data. We explore this idea, namely sequential transfer learning of word embeddings, in the first research question (i.e., RQ \ref{RQ.1} above).  

Further, we consider the transfer of models between two linguistic variants such as genre and language, when little (i.e., low-resource) or no data (i.e., zero-resource) is available for a target genre or language. 
We study this challenging setup in two natural language understanding tasks using meta-learning. Accordingly, we investigate the following research questions:
\begin{enumerate}[label=(\roman*)]
\item \textit{Can meta-learning assist us in coping with low-resource settings in natural language understanding (NLU) tasks?}
\item \textit{What is the impact of meta-learning on the performance of pre-trained language models in cross-lingual NLU tasks?}
\end{enumerate}
We here explore the use of meta-learning to perform the zero-shot and few-shot cross-lingual and cross-genre transfer in two different natural language understanding tasks: natural language inference and question answering.
  
\section{Structure of the Thesis}
This thesis is a collection of case studies with the unifying objective of addressing low-resource settings in NLP and is structured as follows:
\paragraph{Chapter \ref{sec:first}: Background}\hfill \break \hfill \break
This chapter contains the background that is necessary to understand the contributions of the thesis as a whole. It gives an overview of a particular family of machine learning models that will be employed in the thesis, Deep Neural Networks (DNNs). In this chapter, we describe two paradigms that have been proposed to address low-resource NLP: Distant supervision and Transfer learning. The general NLP areas of Information Extraction (IE) and Natural Language Understanding (NLU) are briefly described in this chapter, whereas details regarding specific NLP tasks are delegated to subsequent chapters.  
\paragraph{Chapter \ref{sec:second}: Evaluation of Domain-specific Word Embeddings}\hfill \break \hfill \break
In this chapter, we study input representations trained on data from a low resource domain (Oil and Gas) using sequential transfer learning of word embeddings. We conduct intrinsic and extrinsic evaluations of both general and domain-specific embeddings. We further adapt embedding enhancement methods to provide vector representations for infrequent and unseen terms.
\paragraph{Chapter \ref{sec:third}: Named Entity Recognition in Low-Resource Domains}\hfill \break \hfill \break
In this chapter, we explore the use of distant supervision for Named Entity Recognition (NER) in four low-resource scenarios. We apply distant supervision and present a system that addresses the problem of noisy, distantly supervised data in two ways. We study a \emph{reinforcement learning} strategy with a neural network policy to identify false positive instances at the sentence level. We further adopt a technique of incomplete annotation to address the false negative cases. Finally, we evaluate the proposed hybrid model on various benchmark datasets.

\paragraph{Chapter \ref{sec:fourth}: Low-Resource Relation Extraction}
\hfill \break \hfill \break
In this chapter, we focus on relation extraction in a low resource setting, namely scientific papers in NLP. We study the effect of varying input representations to a neural architecture, specifically Convolutional Neural Networks (CNN), to extract and classify semantic relations between entities in scientific papers. We investigate the effect of transfer learning using domain-specific word embeddings in the input layer and go on to provide an in-depth investigation of the influence of different syntactic dependency representations, which are used to produce dependency paths between the entities in the input to the system. We further compare our syntax-informed approach  with a syntax-agnostic approach. In order to gain a better understanding of the results, we perform manual error analysis.
\paragraph{Chapter \ref{sec:fifth}: Natural Language Understanding in Low-Resource Genres and Languages}
\hfill \break \hfill \break
In this chapter, we consider the transfer of models along two dimensions of variation, namely genre and language, when little or no data is available for a target genre or language, i.e. low-resource and zero-resource settings.
We explore \emph{meta-learning} to address this challenging setup, where, in addition to training a source model, another model learns to select which training instances are the most beneficial.
We experiment using standard supervised, zero-shot cross-lingual, as well as few-shot cross-genre and cross-lingual settings for different natural language understanding tasks (natural language inference, question answering).
We make use of an extensive experimental setup to investigate the effect of meta-learning in various low-resource scenarios.
We apply our proposed cross-lingual meta-learning framework on various pre-trained language models for zero-shot and few-shot natural language inference and question answering tasks.
We further conduct a comprehensive analysis to investigate the impact of typological sharing between languages in our framework.
\paragraph{Chapter \ref{sec:sixth}: Conclusion and Future work}
\hfill \break \hfill \break
In this chapter, we describe our proposed methods and findings. Our main contributions are summarized, and we provide an outlook into future directions in this chapter.
\section{Publications}
The part of the work presented in this thesis has been presented in the following scientific articles:
\begin{enumerate}
    \item Nooralahzadeh, Farhad; Øvrelid, Lilja and Lønning, Jan Tore (2018). "Evaluation of Domain-specific Word Embeddings using Knowledge Resources." In: \emph{Proceedings of the Eleventh International Conference on Language Resources and Evaluation,} European Language Resources Association (ELRA).
    \item Nooralahzadeh, Farhad; Øvrelid, Lilja and Lønning, Jan Tore (2018). "SIRIUS-LTG-UiO at SemEval-2018 Task 7: Convolutional Neural Networks with Shortest Dependency Paths for Semantic Relation Extraction and Classification in Scientific Papers." In: \emph{Proceedings of the 12th International Workshop on Semantic Evaluation. Association for Computational Linguistics.}
    
    \item Nooralahzadeh, Farhad and Øvrelid, Lilja (2018). "Syntactic Dependency Representations in Neural Relation Classification." In: \emph{Proceedings of the Workshop on the Relevance of Linguistic Structure in Neural Architectures for NLP.  Association for Computational Linguistics.}
    \item Nooralahzadeh, Farhad; Lønning, Jan Tore and Øvrelid, Lilja (2019). "Reinforcement-based denoising of distantly supervised NER with partial annotation." In:  \emph{Proceedings of the 2nd Workshop on Deep Learning Approaches for Low-Resource NLP (DeepLo 2019).  Association for Computational Linguistics.}
\end{enumerate}
The following preprint is also discussed:
\begin{enumerate}
\item Nooralahzadeh, Farhad; Bekoulis, Giannis; Bjerva, Johannes; and Augenstein, Isabelle (2020). "Zero-shot cross-lingual transfer with meta learning." \emph{In: CoRR} vol. abs/2003.02739.
\end{enumerate}
Finally, while not directly related, the following article has also been completed over
the course of the PhD:
\begin{enumerate}
    \item Nooralahzadeh, Farhad and Øvrelid, Lilja (2018). "SIRIUS-LTG: An Entity Linking Approach to Fact Extraction and Verification." In:\emph{Proceedings of the First Workshop on Fact Extraction and VERification (FEVER).  Association for Computational Linguistics.}
\end{enumerate}
    \chapter{Background}
\label{sec:first}
This chapter contains the background that is necessary to understand the contributions of the thesis as a whole.
We start by briefly discussing variation in textual data which is central to the thesis (Section \ref{ch1:domain}). After that, we give an overview of a particular family of machine learning models that will be employed in the thesis, Deep Neural Networks (Section \ref{sec:dnns}).
We subsequently describe two machine learning paradigms that have been proposed to address low-resource NLP: Distant supervision in Section \ref{subsec:ds} and Transfer learning in Section \ref{subsec:transfer}.
The general NLP areas of Information Extraction (IE) and Natural Language Understanding (NLU) are briefly described in Section \ref{sec:ie} and \ref{sec:NLU}, respectively. Whereas details regarding specific NLP tasks are delegated to subsequent chapters. 

\section{Dimensions of textual variation}\label{ch1:domain}
The notion of \emph{domain} is frequently used in low-resource NLP, although there is little common ground in what constitutes a domain \parencite{bplank2011phd}. The term is usually used to refer to textual data of the same topic, genre, or source under the assumption that this common denominator will have some systematic impact on the vocabulary or linguistic aspects of the text. Various definitions of the term \emph{domain} have been presented in previous research such as \cite{Lee2001GENRESRT,Finkel:2009,bplank2011phd,van-der-wees-etal-2015-whats,DBLP:journals/corr/Plank16} and \cite{aharoni2020unsupervised}.
\cite{Lee2001GENRESRT} notes that the terms \emph{genre, register, text type, domain, sub-language}, and \emph{style}
are often used differently in various communities or even interchangeably.
\cite{Finkel:2009} further describe the meaning of domain as  "It may refer
to a topical domain or to distinctions that linguists might term mode (speech versus writing) or register (formal written prose versus SMS communications)". It is defined in \cite{bplank2011phd} as a collection of texts from a certain coherent sort of discourse, and \cite{aharoni2020unsupervised} define domains by implicit clusters of sentence representations provided by pre-trained language models (see Section \ref{sec:pre-trained-LM}). However, \cite{DBLP:journals/corr/Plank16} argues that there are numerous other factors that should be taken into consideration, e.g., demographic factors, communicational
purpose, sentence type, style, technology/medium, language, etc.
She proposes to see a domain as a variety in a large dimensional variety space. In this view, most textual datasets are sub-spaces of this variety space. The dimensions in this space are fuzzy aspects such as language, dialect, topic, genre, social factors (age, gender, personality, etc.), including yet unknown aspects. "A domain forms a region in this space, with some members more prototypical than others" \parencite{DBLP:journals/corr/Plank16}.

In this thesis and as noted already in Chapter \ref{sec:intro}, we focus in particular on the textual varieties of  domain, genre, and language. We call the variety aspect \emph{domain} when datasets differ in terms of topic. Furthermore, we use the term \emph{genre} where dataset differences are characterized by non-topical text properties such as function, style, and text type.
\section{Deep Neural Networks in NLP} \label{sec:dnns}
Natural Language Processing (NLP) involves the engineering of computational models and processes to solve practical problems in understanding human languages. Processing natural language text encompasses a number of syntactic, semantic, and discourse-level tasks (e.g., word segmentation, part-of-speech tagging, phrase chunking, parsing, word sense disambiguation, named entity recognition, semantic role labeling, semantic parsing, anaphora resolution). For a long time, NLP systems were based on traditional machine learning approaches, centered around algorithms such as Perceptrons, linear Support Vector Machines (SVM), and Logistic Regression trained on sparse hand-crafted features \parencite{Yoav_book}. These methods are known to have some challenges. 
Recently, the re-emergence of artificial neural networks (ANNs), also known as deep neural networks (DNNs), provides a way to develop highly automatic features and representations to handle complex interpretation tasks. These approaches, with the pioneering work of \cite{DBLP:journals/corr/abs-1103-0398}, have yielded impressive results for many different NLP tasks. In general, a neural network with many hidden layers is often referred to as a deep learning model. In the following sections, we will briefly introduce several DNN models that have been widely employed in NLP and that are central also in this thesis.

\subsection{Deep Feed-Forward Networks}
\begin{figure}[t]
  \centering
   \includegraphics[width=.8\textwidth]{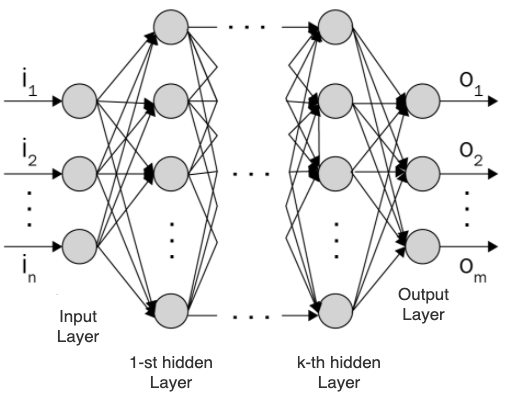}
   \caption[Structure of a feed-forward neural networks model \parencite{Goodfellow-et-al-2016}]{Structure of a feed-forward neural networks model \parencite{Goodfellow-et-al-2016}}
  \label{fig:mlps-background} 
 \end{figure}

Deep Feed-Forward Networks,  also known as Feed-Forward Neural  Networks (FFNNs) or Multi-layer Perceptrons (MLPs), are the simplified version of DNNs \parencite{Goodfellow-et-al-2016}. They are the foundation of most deep learning models and consist of many layers, with the first layer taking the input and last layer providing outputs.
The layers in the middle are known as hidden layers, and capture relations between the input and output.  Figure \ref{fig:mlps-background} shows a typical structure of a feed-forward neural networks model.

The hidden layer is used to transform the input layer values into values in a higher-dimensional space so that we can learn more features automatically from the input. The transformation is done by a collection of perceptron nodes in the hidden layer using a non-linear function, known as an \emph{activation function}. In such a model, information constantly flows from one layer to the next (i.e., input layer $\longrightarrow$ hidden layers$\longrightarrow$ output payer).
Training of the FFNNs model in supervised learning is done by two steps: (i) Forward propagation of information from the input to the output layer through hidden layers and compute a loss function (i.e., training errors), and (ii) Backward propagation of loss from the output to the input layer. The aim of backward propagation is to minimize the training error by measuring the margin of error of the output and then adjust the network parameters accordingly. We repeat both forward- and back-propagation to predict an output until the parameters of the model are calibrated.
\subsection{Convolutional Neural Networks} \label{ch01:cnns}
Convolutional Neural Networks (CNNs) \parencite{10.5555/303568.303704, NIPS2012_4824} are a specialized kind of deep neural networks where convolution operations, derived from mathematics and signal processing, are applied to capture indicative local patterns from the data. The resulting local aspects are most informative for the prediction task at hand.
In the field of NLP, The CNN's feature functions (i.e., the convolution filters) are applied to extract high-level features from adjacent words or n-grams regardless of their position, while taking local ordering patterns into account \parencite{Yoav_book}.
\begin{figure}[t]
  \centering
   \includegraphics[width=1\textwidth]{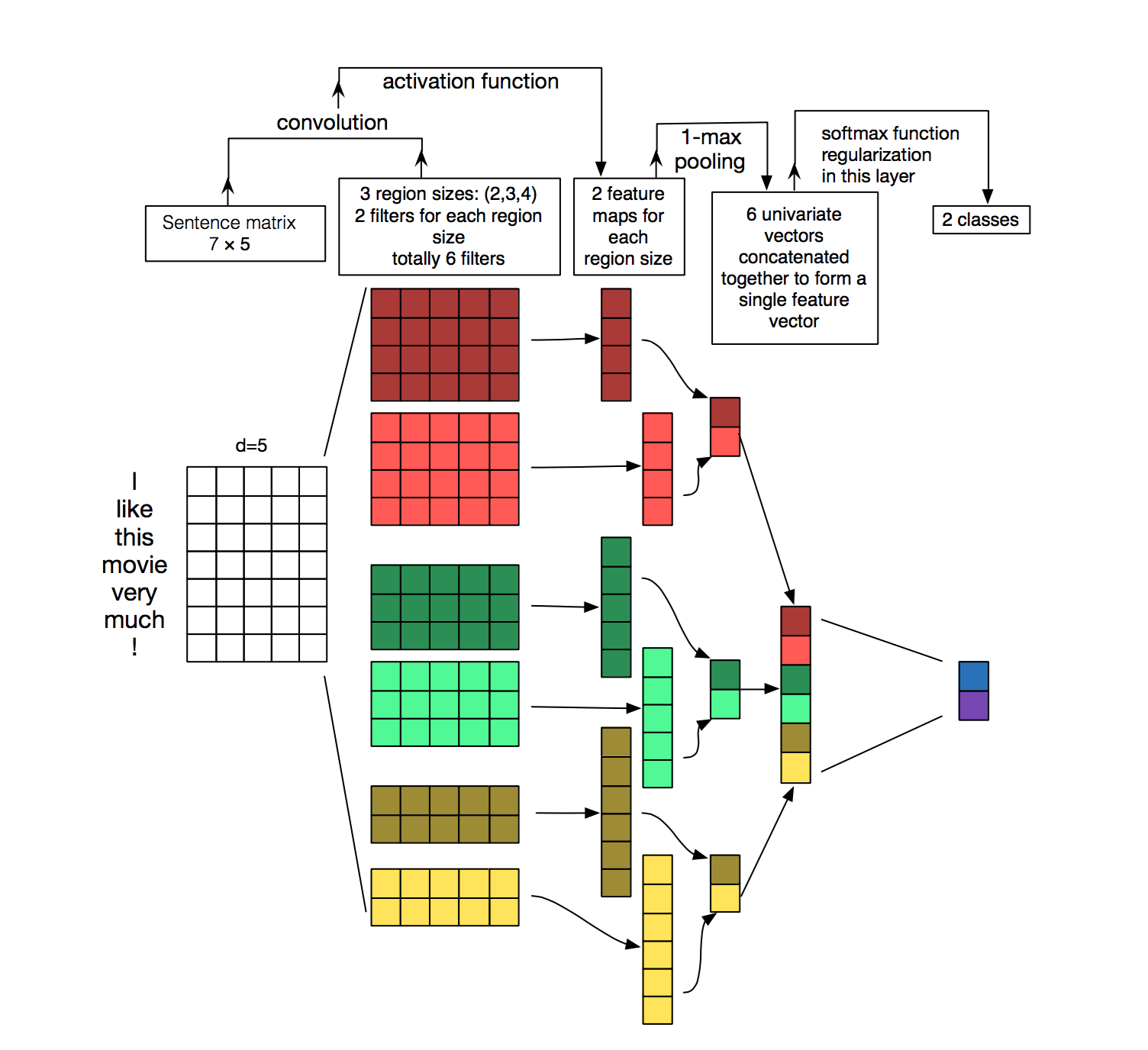}
   \caption[CNN architecture for a sentence classification task]{CNN architecture for a sentence classification task
   \parencite{DBLP:journals/corr/ZhangW15b}}
  \label{fig:CNN-background} 
 \end{figure}
The CNN architecture consists of multiple \emph{convolutions} and \emph{pooling} layers. The convolution layers aim to extract useful local features from the input, which results in multiple feature maps. Then, a pooling layer is applied to one or multiple convolution layers to reduce the spatial size of feature maps. In the end, usually, a fully connected layer outputs the probability distribution over each target class.

Figure \ref{fig:CNN-background} presents the CNN architecture applied to a sentence classification task and proposed by \cite{DBLP:journals/corr/Kim14f}. For each sentence, words are represented as a vector in the input layer. Word vectors can be initialized randomly or fetched from pre-trained embeddings (see Section \ref{sec:emb}).
The filter layer in Figure \ref{fig:CNN-background} includes three filter region sizes: 2, 3, and 4, each of which has two filters. This layer performs convolutions on the sentence matrix and generates (variable-length) feature maps. Subsequently, the 1-max pooling function performs pooling over each map (i.e., the largest number from each feature map is extracted). Thus a uni-variate feature vector is generated from all six maps, and a feature vector is formed by connecting these six features. Finally, the fully connected softmax layer receives this feature vector as input and uses it to classify the sentence as a binary classification task to output two possible output states \parencite{DBLP:journals/corr/ZhangW15b}.
We could provide different characteristics or views of a sentence in the input layer, referred to as a \emph{channel}. For instance, in the sentence classification task, one channel will be the sequence of words, while another channel is the sequence of corresponding POS tags.
It is common to apply a different set of filters to each channel, and then combine the multiple representations of input into a single vector.

The explained architecture in Figure \ref{fig:CNN-background} is just an example of CNNs (albeit a widely used architecture), and there are various designs where a different set of layers (i.e., filter, pooling, and fully connected layer) along with various hyper-parameters such as filter size, number of feature maps, activation function, pooling strategy, are employed.

CNN models have been effectively applied to position-invariant contextual features in various NLP tasks such as sentence and document classification. However, they have some challenges in maintaining sequential order and modeling long-distance dependencies, which is essential for many NLP tasks \parencite{DBLP:journals/corr/abs-1708-02709}. Recurrent Neural Networks and its variants are introduced as a suitable solution for such types of tasks, and we will now turn to these models in the next section.

\subsection{Recurrent Neural Networks} \label{sub:RNN}
\begin{figure}[t]
  \centering
   \includegraphics[width=1\textwidth]{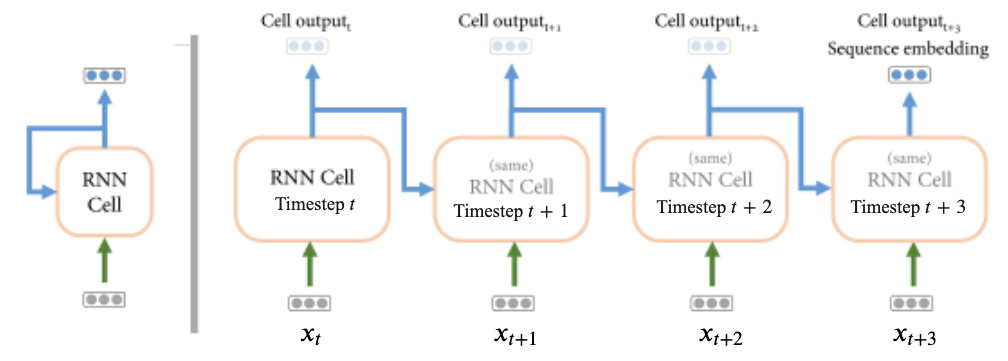}
   \caption[Recurrent Neural Networks (RNNs) architecture]{(left) Folded RNN with an input sequence and feedback loop. (right) Unfolded version of RNN through the time steps. The same RNN cell is applied in different time steps to the words in the sequence example. \parencite{EMB-NLP01}}
  \label{fig:simple-rnn} 
 \end{figure}

Recurrent Neural Networks (RNNs) \parencite{reason:RumHinWil86a,Elman90findingstructure} are designed to process sequential information. The model is recurrent since it performs the same task for each input sequence element, such that the current step's output is conditioned on the previous step. As shown in Figure \ref{fig:simple-rnn}, the input sequence is typically represented by a fixed-size vector of tokens and is passed sequentially (one by one) to a recurrent unit.
The main power of the RNN is the ability to memorize the outputs of previous computation steps and utilize them in the current computation. This capability made the RNNs a preferred neural architecture in solving sequential NLP tasks such as language modeling, machine translation, named entity recognition, textual similarity, and text generation.

\begin{figure}[t]
  \centering
   \includegraphics[width=1\textwidth]{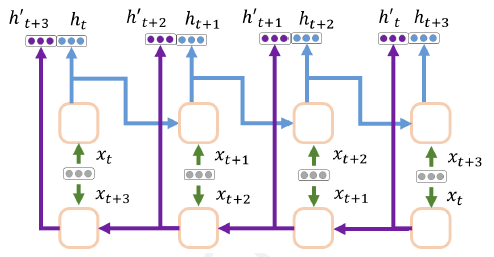}
   \caption[Bidirectional RNNs Model]{Bidirectional RNNs Model \parencite{EMB-NLP01}}
  \label{fig:bi-rnn} 
 \end{figure}

Some extensions of RNNs are introduced, such as Bidirectional RNN (Bi-RNNs) \parencite{schuster1997bidirectional}, which can be seen as stacking two RNNs on top of each other, one going forward, the other one going backward over the sequence input (Figure \ref{fig:bi-rnn}). 
The Bi-RNNs are based on the idea that the output at a specific time step depends on the previous elements in the sequence, as well as future elements.

For a long sequence length, however, the vanilla RNN models suffer from the vanishing gradient problem \parencite{Hochreiter:1997:LSM:1246443.1246450}. This means that the gradient shrinks as it back propagates through time. If a gradient value becomes extremely small, it does not contribute to the learning process. The vanishing gradient causes the model to ignore long-term dependencies and, hence, hardly learn the dependencies between temporally distant sequences. In other words, the RNNs tend to focus on short term dependencies, which are often not desired. This limitation is mitigated by alternative network architectures like long short-term memory and gated recurrent unit networks, which are the most widely used RNN variants in NLP applications.

\begin{figure}[t]
\centering
\includegraphics[width=.8\textwidth]{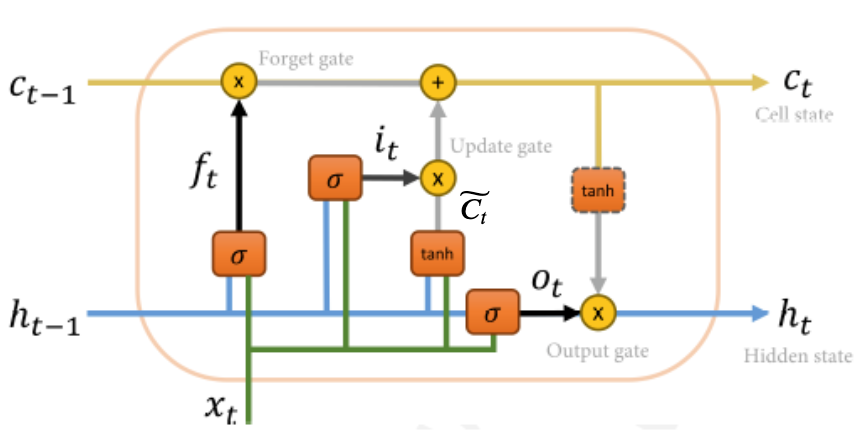}
\caption[LSTM module structure at time step $t$]{LSTM module structure at time step $t$ \parencite{EMB-NLP01}}\label{fig:rnn-lstm}
\end{figure}

\paragraph{Long Short-Term Memory Networks (LSTMs)} 
LSTMs \parencite{Hochreiter:1997:LSM:1246443.1246450,10.1162/089976600300015015} are explicitly designed to cope with the vanishing gradients problem using a gating mechanism. In the vanilla RNNs, the repeating modules (i.e., the rectangle boxes in Figures \ref{fig:simple-rnn} and \ref{fig:bi-rnn}) have a straightforward design, such as a single non-linearity.
The structure of LSTMs is not fundamentally different from the structure of RNNs, but the repeating module has a different setting. As shown in Figure \ref{fig:rnn-lstm}, the repeating cell consists of four neural layers: the input gate, forget gate, cell state, and the output gate.

These layers are calculated according to the following formula \parencite{Yoav_book}:
\begin{equation}
\begin{split}
i_t=\sigma (x_t W^{xi} + h_{t-1} W^{hi})\\
f_t=\sigma (x_t W^{xf} + h_{t-1} W^{hf}) \\
o_t=\sigma (x_t W^{xo}+ h_{t-1} W^{ho}) \\
\widetilde{C_t} =\tanh (x_t W^{xC} + h_{t-1} W^{hC}) \\
C_t=f_t \otimes  
C_{t-1} + i_t \otimes \widetilde{C_t} \\
h_t= o_t \otimes \tanh(C_t) \\
\end{split}
\end{equation}
\noindent Where $\otimes$ is element-wise  multiplication, $h_t$ is the hidden state in time-step $t$ and $i, f, o$ are the input, forget and output gates, respectively.
In the first step of the LSTM block, we decide what information should be retained or thrown away from the cell state (i.e., forget gate ). For example, in the language model, the cell state might decide to remember the singular or plural information of the present subject in order to predict the correct verb tense in the next related states. While, if it sees a new subject, it will forget the information about the old subject.
$\widetilde{C_t}$ is called the candidate cell state and is computed based on the current input and the previous cell state.
$C_t$ is the internal memory of the unit. It is a combination of the previous memory $C_{t-1}$ multiplied by the forget gate, and the newly computed cell state $\widetilde{C_t}$, multiplied by the input gate.
Given the memory $C_t$, the output hidden state $h_t$ is computed by multiplying the memory with the output gate (i.e., a filtered version of the cell state). Intuitively, in the LSTM cell, we compute the $C$ and $h$ at time $t$ and output them to the next cell.
\begin{figure}[t]
\centering
\includegraphics[width=.8\textwidth]{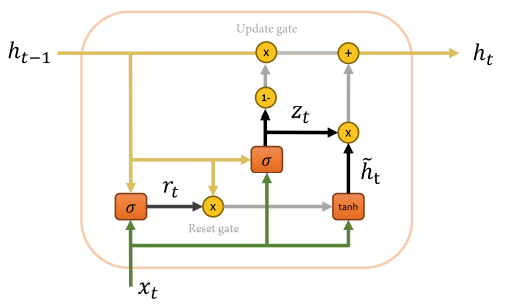}
\caption[GRU module structure at time step $t$]{GRU module structure at time step $t$ \parencite{EMB-NLP01}}\label{fig:gru}
\end{figure}
\paragraph{Gated Recurrent Unit Networks (GRUs)} The GRUs, introduced by
\cite{DBLP:journals/corr/ChoMBB14}, similar to LSTMs, follow the design of RNNs; however, each repeating block has a slightly simpler variant of the LSTM.
GRU combines the forget gate and input gate into a single update gate. It incorporates two gates, the reset gate and the update gate, and manages the flow of information similar to LSTM without a memory unit (Figure \ref{fig:gru}). The formulation of the GRU module is as follows \parencite{Yoav_book}:
\begin{equation}
\begin{split}
z_t=\sigma (x_t W^{xz}+ h_{t-1} W^{hz})\\
r_t=\sigma (x_t W^{xr} + h_{t-1} W^{hr}]) \\
\widetilde{h_t} =\tanh (x_t W^{x\widetilde{h}}+ (r_{t}\otimes h_{t-1})W^{h\widetilde{h}}]\\
h_t =(1-z_t) \otimes h_{t-1} +, z_t \otimes \widetilde{h_t}\\
\end{split}
\end{equation}
\noindent Because of this update, the final model is more straightforward than the standard LSTM and is also widely used in NLP.
\subsection{Attention Mechanisms and Transformer}
\begin{figure}[t]
\centering
\includegraphics[width=.7\textwidth]{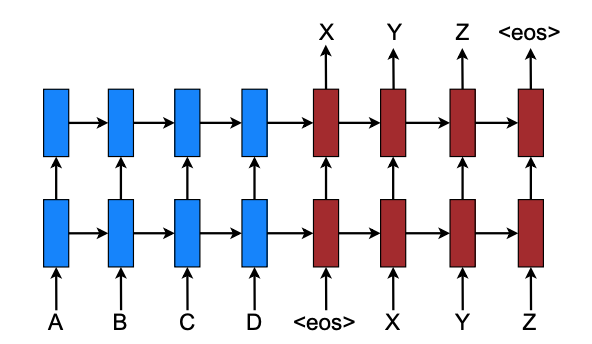}
\caption[The representation of the sequence to sequence (seq2seq) model]{The representation of the sequence to sequence (seq2seq) model - translating an input
sequence A B C D into a target sequence X Y Z. Here, <eos> indicates the end of a sequence. The blue boxes at the left show the encoder, and the red boxes construct the decoder \parencite{DBLP:journals/corr/LuongPM15}}\label{fig:seq2seq}
\end{figure}

One of the common applications of RNN architectures is in \emph{sequence-to-sequence} (seq2seq) models. Seq2seq models are DNNs that have been successfully employed in NLP tasks like machine translation between multiple languages, text summarization, and language generation. The seq2seq model tries to transfer an input sequence to a new output sequence where the length of input and output may vary.  
The seq2seq model (Figure \ref{fig:seq2seq}) normally has an encoder-decoder architecture, composed of:
\begin{itemize}
    \item {An encoder:} It compiles the incoming sequence and captures the information into a context vector (i.e., sentence embedding vector) of a fixed length. This representation is assumed to be a good summary of the meaning of the whole source sequence.
    \item {A decoder:} It is initialized with the context vector to produce the output sequence. The early implementation of the seq2seq model used the last state of the encoder network as the initial decoder state.
\end{itemize}
\begin{figure}[t]
\centering
\includegraphics[width=.4\textwidth]{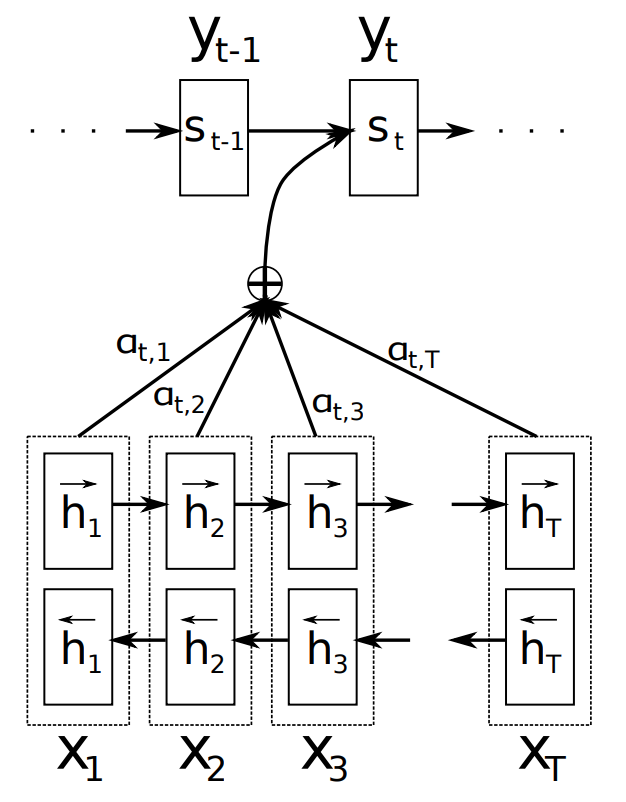}
\caption[The attention model]{The attention model proposed by \cite{DBLP:journals/corr/BahdanauCB14}. 
The decoder trying to generate the target word $y_t$ given a result of attention and encoder over the source sequence $X_1 X_2 \dots X_T $}\label{fig:attention}
\end{figure}
Both the encoder and decoder are RNNs with LSTM or GRU units. The naive seq2seq model works fine for short sequences. However, in a long sequence, it becomes problematic when the encoder compresses the entire input into a fixed-sized context vector and transmits it into the decoder as the contextual information of the input. This problem is addressed by attention mechanisms proposed by \cite{DBLP:journals/corr/BahdanauCB14} and \cite{DBLP:journals/corr/LuongPM15}.

The \emph{attention mechanism} (Figure \ref{fig:attention}) allows the decoder to refer back to the input sequence. Specifically, during decoding, it gives importance to specific parts of the input sequence instead of the entire sequence.
Therefore, in the attention, all the intermediate outputs from the encoder state are considered, and we utilize them to generate the context vector from all states. It allows the model to focus on essential elements by giving weight to each element in the sequence.

Different ways of constructing attention mechanisms have been introduced, including \textit{global} and \textit{local attention} \parencite{DBLP:journals/corr/LuongPM15} and \textit{self-attention} \parencite{DBLP:journals/corr/VaswaniSPUJGKP17}. Self-attention suggests implementing attention to words in the same sequence. For instance, while encoding a word in an input sentence, self-attention enables the encoder to look at other words in the input for clues that can further lead to a better encoding for the word.  During decoding to produce a resulting sentence, it makes sense to provide appropriate attention to words that have already been produced. This type of attention mechanism has become widely used in a state-of-the-art encoder-decoder model called \textit{transformer}.

\begin{figure}[t]
\centering
\includegraphics[width=.5\textwidth]{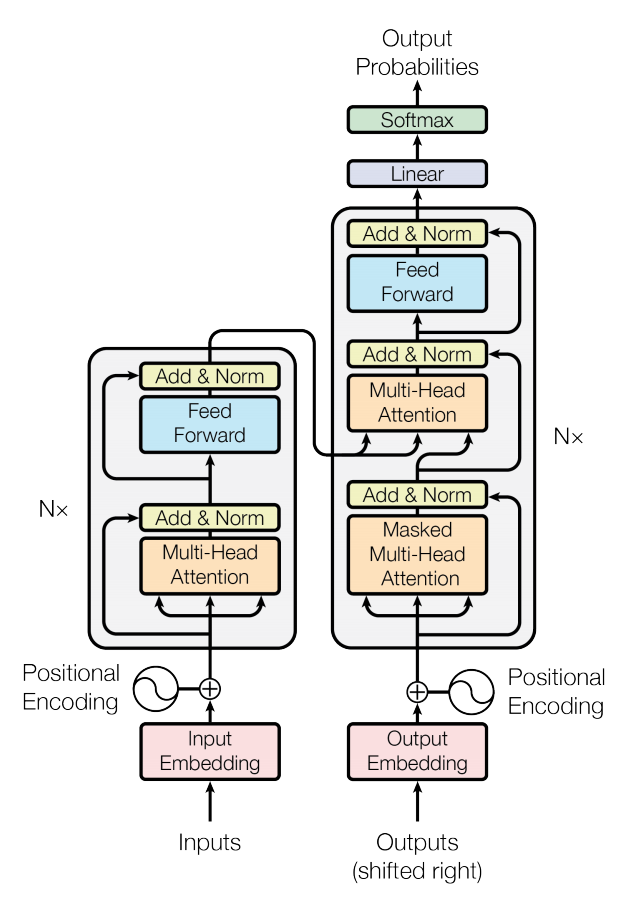}
\caption[The encoder-decoder model with Transformer architecture]{The encoder-decoder model with Transformer architecture  \parencite{DBLP:journals/corr/VaswaniSPUJGKP17}.}
\label{fig:transformer}
\end{figure}
The transformer model, shown in Figure \ref{fig:transformer}, has many stacked layers in both encoder and decoder components. It considers self-attention in the encoder and decoder modules, as well as
cross-attention between them. The proposed model is based entirely on an attention mechanism to capture the global relations between input and output, without including RNNs and CNNs.
It incorporates other techniques in its encoder and decoder components such as residual connections \parencite{DBLP:journals/corr/HeZRS15}, layer normalization \parencite{DBLP:journals/corr/BaKH16,}, dropouts and positional encodings.
The transformer becomes an essential component of pre-trained language models such as BERT and GPT (see Section \ref{sec:pre-trained-LM}).

\section{Distant Supervision} \label{subsec:ds}
Distant supervision has been proposed to deal with the lack of sufficient labeled data for training supervised machine learning methods by exploiting existing knowledge resources. \cite{CravenKumlien:99} initiated this idea as a \textit{weak supervision} method to populate a knowledge base in the biomedical domain. Subsequently, \cite{Mintz:2009:DSR:1690219.1690287} generalized the initial idea for the relation extraction task and formulated the distant supervision assumption as follow:
\begin{displayquote}
    "If two entities participate in a relation, any sentence that contains those two entities might express that relation."(Mintz et al., 2009, p. 1006) 
\end{displayquote}
The term \textit{distant} is used by assuming that no explicit labeled data is provided, however knowledge resources (e.g., Wikipedia, Freebase) are available for automated labeling of training instances in text corpora. Distant supervision can be formally defined as follows \parencite{10.1145/3241741}:
\begin{displayquote}
   "Given a text corpus $\mathcal{C}$ and a knowledge base $\mathcal{K}$, distant supervision assigns relations from $\mathcal{K}$ to sentences from $\mathcal{C}$. More specifically, the idea is to first collect those sentences from the corpus $\mathcal{C}$ that contain entity pair $(e_1,e_2)$ where both $e_1$ and $e_2$ exist in the knowledge base $\mathcal{K}$. If there exists one triple $(e_1,r, e_2)$ in the knowledge base, then the distant supervision set a label to the sentence as an instance of relation $r$." (Smirnova and Cudré-Mauroux, 2018, p. 106:4)
\end{displayquote}
For example, the following sentence contains the entity pair (\textit{Steven Spielberg, Saving Private Ryan}) 
\begin{covexample}
\textit{[Steven Spielberg]’s film [Saving Private
Ryan] is loosely based on the brothers’ story.}
\end{covexample}
Assuming the triple (\textit{Steven Spielberg, is director of, Saving Private Ryan}) exists in the knowledge base, the textual sentence is labeled with the \textit {is director of} relation label. It can be used as training data for subsequent relation extraction.

Distant supervision has been successfully applied to tasks like relation extraction 
\parencite{Riedel:2010:MRM:1889788.1889799, DBLP:conf/ekaw/AugensteinMC14} and entity recognition \parencite{DBLP:journals/corr/Fries0RR17,shang2018learning,yang-etal-2018-distantly}. We will return to this topic in Chapter \ref{sec:third} where we explore the use of distant supervision for named entity recognition in low-resource scenarios. 

\section{Transfer Learning}\label{subsec:transfer}
Humankind can learn new tasks faster and more efficiently if he/she has prior experience with similar tasks. For example, people who know how to ride a bike will likely manage to ride a motorcycle with little or no training. In short, we learn how to learn across tasks.
This statement brings the following question: Is it possible to design a machine learning model with similar properties, learning new tasks by leveraging prior knowledge gained in other learning processes? Recently, this question is answered by transfer learning \parencite{PanY09TKDE}.
Transfer learning makes use of knowledge acquired while solving one problem or more than one problem and applies it to a different but related problem/s. It refers to a set of
methods that extend the learning mechanism by leveraging data from additional languages, domains, or tasks to train a model with better generalization properties. Transfer learning has yielded to a significant improvement in various NLP tasks \parencite{ DBLP:journals/corr/abs-1902-05309,DBLP:journals/corr/abs-1810-04805,howard-ruder-2018-universal, DBLP:journals/corr/abs-1802-05365} and this is due to the fact that NLP tasks share common knowledge about language, such as linguistic representation and structural similarity \parencite{ruder-etal-2019-transfer}. Moreover, languages have common typological features such as phonological, grammatical, and lexical properties \parencite{wals}. 

Based on different scenarios that are  mostly encountered in NLP systems, \cite{PanY09TKDE} and \cite{ruder-etal-2019-transfer} proposed a taxonomy for transfer learning for the NLP field (Figure \ref{fig:TF-taxonomy}).
Following the notation of \cite{PanY09TKDE} and \cite{ruder-etal-2019-transfer}, transfer learning is defined as follows:
\begin{displayquote}
"Given a settings $\mathbb{S}=\{\mathcal{D},\mathcal{T}\}$ where, $\mathcal{D}$ is a dataset that contains a feature space $\mathcal{X}=\{x_1, \dots, x_n\}$ with a marginal probability distribution $\mathrm{P}(\mathcal{X})$  over the feature space. On the other hand, a task $\mathcal{T}=\{\mathcal{Y}, \mathrm{P}(\mathcal{Y}),\mathrm{P}(\mathcal{Y}|\mathcal{X}\})$ consists of a label space $\mathcal{Y}$, a prior distribution $\mathrm{P}(\mathcal{Y})$, and a conditional probability distribution $\mathrm{P}(\mathcal{Y}|\mathcal{X})$ which is usually learned using  the training data consisting pairs of $x_i \in \mathcal{X}$ and  $y_i \in \mathcal{Y}$.
Having a source setting $\mathbb{S}_s$ including $\mathcal{D}_s$ and a corresponding source task $\mathcal{T}_s$, as well as a target setting $\mathbb{S}_t$ with $\mathcal{D}_t $ and target task  $\mathcal{T}_t $, the aim of transfer learning is to perform the target task in order to learn the target $\mathrm{P}_t(\mathcal{Y}_t|\mathcal{X}_t)$ in $\mathcal{D}_t$ using the information provided by the elements in the source setting where $\mathcal{D}_s \neq \mathcal{D}_t$ or $\mathcal{T}_s \neq \mathcal{T}_t$. Typically, it is assumed that the target setting is either in low-resource or zero-resource mode."
\end{displayquote}
According to the proposed taxonomy and notation, transfer learning involves the following scenarios:
\begin{figure}[t]
  \centering
   \includegraphics[width=1\textwidth]{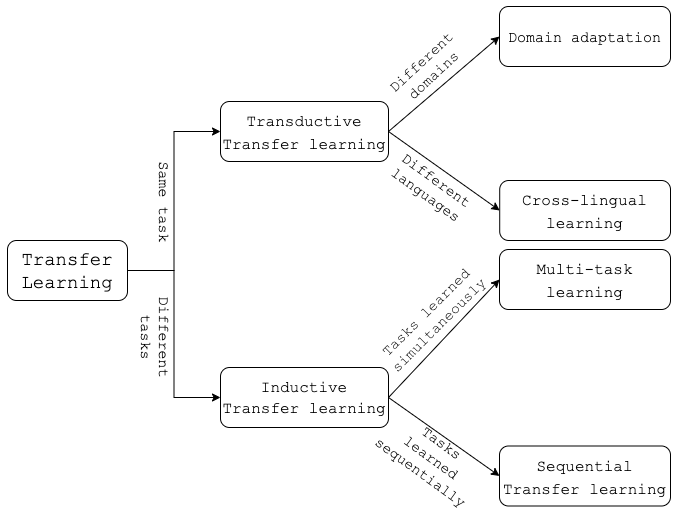}
   \caption[A taxonomy for transfer learning in NLP]{A taxonomy for transfer learning in NLP \parencite{ruder-etal-2019-transfer}}
  \label{fig:TF-taxonomy} 
 \end{figure}
\paragraph{Transductive transfer learning:} \label{domain-adaptation} The source and target tasks are the same, whereas the source and target dataset vary. If the marginal probability distribution of source and target dataset are different $\mathrm{P}_s(\mathcal{X}_s)\neq \mathrm{P}_t(\mathcal{X}_t)$, when the datasets come from different domains or genres, the scenario is known as {domain adaptation}. If there is a discrepancy in the feature spaces of the source and target datasets $\mathcal{X}_s \neq \mathcal{X}_t$, for example, when the datasets are in two various languages, we refer to {cross-lingual learning} scenario in NLP.  Cross-lingual learning can be viewed as an extreme case of adaptation \parencite{DBLP:journals/corr/Plank16}.

\paragraph{Inductive transfer learning:} The source and target tasks are different $\mathcal{T}_s \neq \mathcal{T}_t$, regardless of whether the source and target datasets are the same or not. In this case, if the tasks are learned simultaneously, the scenario is known as {multi-task learning}, while {sequential transfer learning} will be used if the learning process is performed sequentially. We want to stress that if in this category the variation on the source and target dataset is considered, both scenarios (i.e., multi-task learning and sequential transfer learning) can be applied to domain adaptation (i.e., $\mathrm{P}_s(\mathcal{X}_s)\neq \mathrm{P}_t(\mathcal{X}_t)$)  and cross-lingual learning (i.e., $\mathcal{X}_s \neq \mathcal{X}_t$).

In the context of this thesis, we focus on  sequential transfer learning.
\section{Sequential Transfer Learning}\label{sec:seqtr}
Sequential transfer learning is defined as a setting where a learning process is carried out in sequence \parencite{ruder-etal-2019-transfer}. It can be useful when (i) the target task is in a low- or zero-resource setting, (ii) the source task is in a high-resource setting, and (iii) the objective is the adaptation of many target tasks.  
This learning approach consists of the two following steps \parencite{ruder-etal-2019-transfer}:

\begin{enumerate}
    \item \textbf{Pre-training:} The general representation of the model is learnt on the source language, task, or domain. 
    \item \textbf{Adaptation:} The learned knowledge is transferred and adjusted to target languages, tasks, or domains. In other words, it involves copying the weights from a pre-trained network and tuning them on the targets.
\end{enumerate}
The most dominant practice of sequential transfer learning is to pre-train embedding representations on a large unlabeled text corpus and then to transfer these representations to a supervised target task using labeled data.
In the following section, we will give an overview of the pre-trained representations that are employed in this thesis.
\subsection{Embedding Representations} \label{sec:emb}
The distributed vector representations of tokens, called \emph{word embeddings}, are an essential component of neural methods for downstream NLP tasks.
Word embeddings are vectors based on the distributional hypothesis meaning that words appearing in a similar context have a similar meaning. In other words, they have learned representations of text where words with the same meaning have a similar representation. Learning useful word representations in a supervised setting with limited data is often difficult. Therefore, many unsupervised learning approaches have been proposed to take advantage of large amounts of unlabeled data that are readily available. It results in  more useful word embeddings \parencite{pennington2014glove,Mikolov2013a}. However, the differences in the meaning of a word in varying contexts are lost when it is associated with a single representation. Static pre-trained embeddings are limited in two respects \parencite{EMB-NLP01}: \begin{enumerate*}[label=(\roman*)]
\item the role of context is ignored, and
\item by providing the individual word vector representation, it is problematic to capture higher-order semantic phenomena, such as compositionality and long-term dependencies.
\end{enumerate*}
To alleviate the limitations of static word embeddings and to deal with varying word context, pre-trained language models \parencite{DBLP:journals/corr/abs-1802-05365, OpenAI-GP2,DBLP:journals/corr/abs-1810-04805,DBLP:journals/corr/abs-1906-08237,conneau2019cross} are proposed and create context-sensitive word representations.  The success of these approaches suggests that these representations capture highly transferable and task-agnostic properties of languages.  
\subsubsection{Static Word Embeddings} \label{w2v}
Word2vec, proposed by \cite{Mikolov2013a}, is one of the most popular approaches in learning
word representations from text inputs. It can effectively capture the semantics of words and is straightforwardly transferred into other downstream tasks. The proposed method consists of a single layer architecture based on the inner product between word vectors based on two different learning approaches as follows:
\paragraph{Continuous Bag-Of-Words (CBOW):} Learns the embeddings by estimating the conditional probability of a particular word based on its context (i.e., surrounding words within a specified window size). Specifically, given a sequence of words, the model receives as input a window of $\mathcal{C}$ context words and predicts the target word $w_i$ by minimizing the following objective \parencite{DBLP:journals/corr/Rong14}:
\begin{equation}
    E= - \frac{1}{|\mathcal{C}|} \sum_{t=1}^{|\mathcal{C}|} \log P(w_t| w_{t-\mathcal{C}},\dots,w_{t-1},w_{t+1},\dots,w_{t+\mathcal{C}})
\end{equation}
and 
\begin{equation}
P(w_t| w_{t-\mathcal{C}},\dots,w_{t-1},w_{t+1},\dots,w_{t+\mathcal{C}})=\frac{\exp(u_t^{\intercal} v_c)}{\sum_{i}^{|V|}\exp(u_i ^{\intercal} v_c)}
\end{equation}
\noindent where $V$ is the vocabulary size, $v_c$ is the sum of the embeddings vector of the context words $w_{t-\mathcal{C}},\dots,w_{t-1},w_{t+1},\dots,w_{t+\mathcal{C}}$, and $u$ is the embeddings vector of the target word.
   
\paragraph{Skip-gram:} Learns by predicting the surrounding words (context) given a current word. In other words, it minimizes the following objective \parencite{DBLP:journals/corr/Rong14}:
\begin{equation}
E= - \frac{1}{|\mathcal{C}|} \sum_{t=1}^{|\mathcal{C}|}  \sum_{-\mathcal{C}\leq j \leq \mathcal{C};j\neq 0} \log P(w_{t+j}| w_t)
\end{equation}
and 
\begin{equation}
P(w_{t+j}| w_t)= \frac{\exp(v_{t+j}^{\intercal} u_t)}{\sum_{i}^{|V|}\exp(v_i ^{\intercal} u_t)}
\end{equation}
\noindent where $u$ and $v$ are the current and context word embeddings, respectively. 

\begin{figure}[t]
  \centering
   \includegraphics[width=1\textwidth]{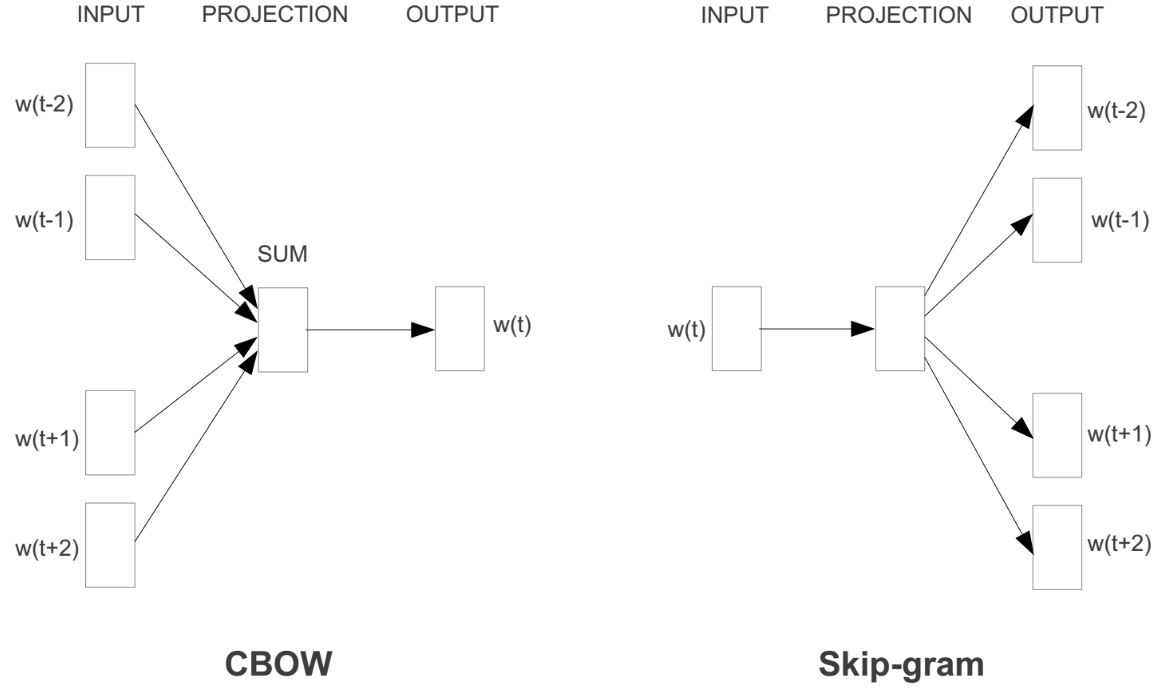}
   \caption[Word2vec Architectures]{Word2vec Architectures \parencite{Mikolov2013a}}
  \label{fig:w2v} 
 \end{figure}

Figure \ref{fig:w2v} depicts these two approaches
of the word2vec model where the window size $\mathcal{C}=2$. In order to train the word2vec model, we provide many word-context pairs where the window size parameter characterizes the context, and the weights learned by the models make up the actual word vector representations.
While the objective of both word2vec models is computationally expensive, the negative-sampling approach is presented as a more efficient way of deriving word embeddings. It means that for each
positive pair (i.e., output word and context word), it samples k words $w_j$ from the vocabulary and add it as a negative example $(w_o, w_j)$ to the $\mathcal{W}_{neg}$. The number of negative
samples k is a parameter of the algorithm. To this effect, the following simplified training objective is capable of producing high-quality word embeddings: 
\begin{equation}
    E= \log \sigma(u_{w_o}^{\intercal} v) - \sum_{w_j \in \mathcal{W}_{neg}} \log \sigma(u_{w_j}^{\intercal} v)
\end{equation}
\noindent where $u_{w_o}$ is the embeddings of output word (i.e., the positive sample), $\sigma$ is the sigmoid function, $\mathcal{W}_{neg}$ is the set of negative samples and $v$ in CBOW it is the mean of the embeddings vector of the context words $w_{t-\mathcal{C}},\dots, w_{t-1},w_{t+1},\dots,w_{t+\mathcal{C}}$, whereas, in the skip-gram model it is the input word embeddings.

\subsubsection{Contextualized Word Representations} \label{sec:pre-trained-LM}
\begin{figure}[t]
  \centering
   \includegraphics[width=1\textwidth]{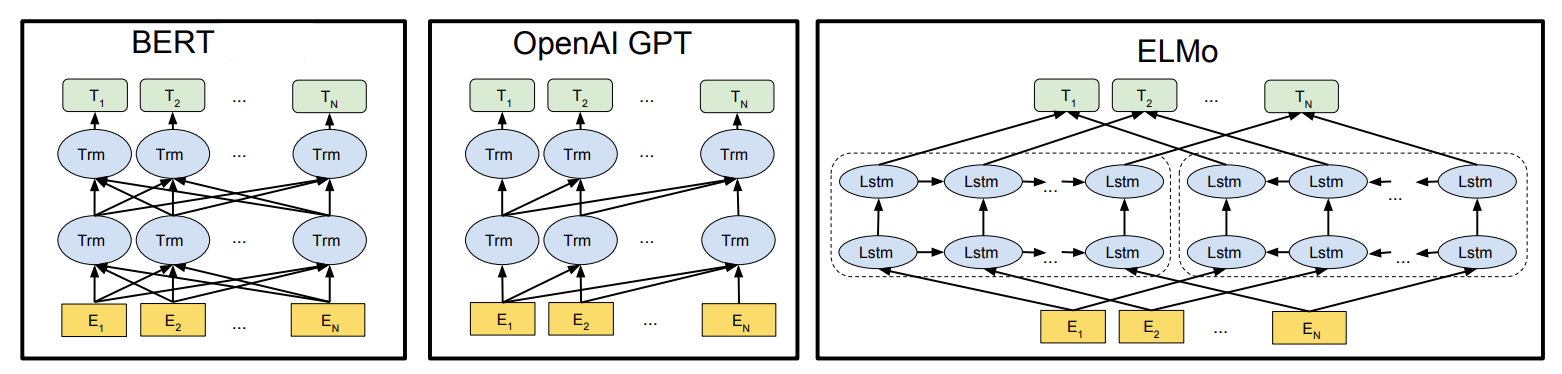}
   \caption[Comparison of the model architectures in BERT, GPT and ELMo]{Comparison of the model architectures in BERT, GPT and ELMo \parencite{DBLP:journals/corr/abs-1810-04805}. It can be seen that BERT is deeply bidirectional, GPT is unidirectional and ELMo is shallow bidirectional.}
  \label{fig:bert-gpt-elmo} 
 \end{figure}
Before pre-trained language models, context-independent or static vectors as described in the previous section, were generally used for transfer learning in NLP tasks. A common practice was employing them as a look-up table to structure the input layer of deep neural models where it results in high training efforts to learn a target task.
Considering the limitations of static word embeddings, recent work presents context-sensitive word representations using neural language models with two different transfer strategies \parencite{DBLP:journals/corr/abs-1810-04805}. The \textit{feature-based} approach, such as ELMo (Embeddings from Language Models proposed by \cite{DBLP:journals/corr/abs-1802-05365}), uses task-specific architectures that include the pre-trained representations as additional features. In contrast, in the \textit{fine-tuning} strategy such as GPT \parencite{OpenAI-GP2} and BERT \parencite{DBLP:journals/corr/abs-1810-04805}, the underlying network structure can be leveraged in the learning of a target task with simply fine-tuning all pre-trained parameters. These pre-trained language models show that despite being trained with only a language modeling task, they provide highly transferable and task-agnostic features of the language \parencite{liu-etal-2019-linguistic}. 
ELMo creates contextualized representations derived from a 2-layer bidirectional LSTM. It is trained with a coupled language model (LM) objective on a large text corpus \parencite{DBLP:journals/corr/abs-1810-04805}. In contrast, BERT (Bidirectional Encoder Representations from
Transformers) and GPT (Generative Pre-trained Transformer) are bi-directional and uni-directional language models, respectively, based on the transformer architecture \parencite{DBLP:journals/corr/VaswaniSPUJGKP17}. They create a contextualized representation of each token by attending to different parts of the input sentence. Unlike GPT, BERT integrates the concept of \textit{masked language model} in the pre-training phase, where the goal is to predict randomly masked tokens given their captured context from both directions. It is also trained on a \textit{next sentence prediction} task that further boosts the model's performance. Figure \ref{fig:bert-gpt-elmo} shows the differences in pre-training model architectures. It can be seen that BERT uses a bidirectional Transformer, while GPT
employs a left-to-right Transformer. ELMo uses the concatenation of independently trained left-to-right, and right-to-left LSTMs to generate features for downstream tasks. Among them, only BERT representations are jointly
conditioned on both left and right context in all layers and it is deeply bidirectional \parencite{DBLP:journals/corr/abs-1810-04805}. 

In the following section, we will discuss how GPT and BERT models are employed in the second step of sequential transfer learning, namely adaptation.
\begin{figure}[t]
  \centering
   \includegraphics[width=1\textwidth]{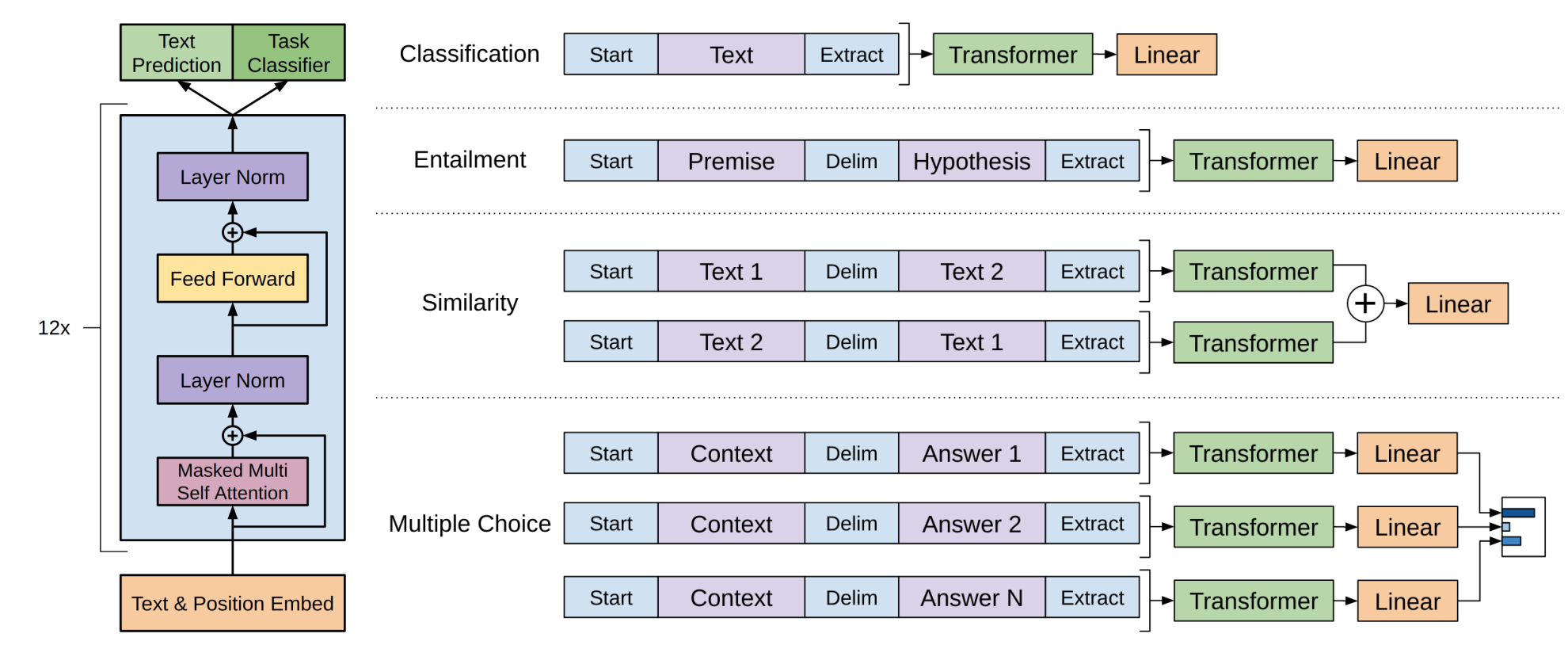}
   \caption[Adaptation stage using GPT]{Adaptation stage using GPT. (left) Transformer architecture. (right) Fine-tuning on different tasks. \parencite{OpenAI-GP2}}
  \label{fig:STL-GPT} 
 \end{figure}
 \begin{figure}[t]
  \centering
   \includegraphics[width=1\textwidth]{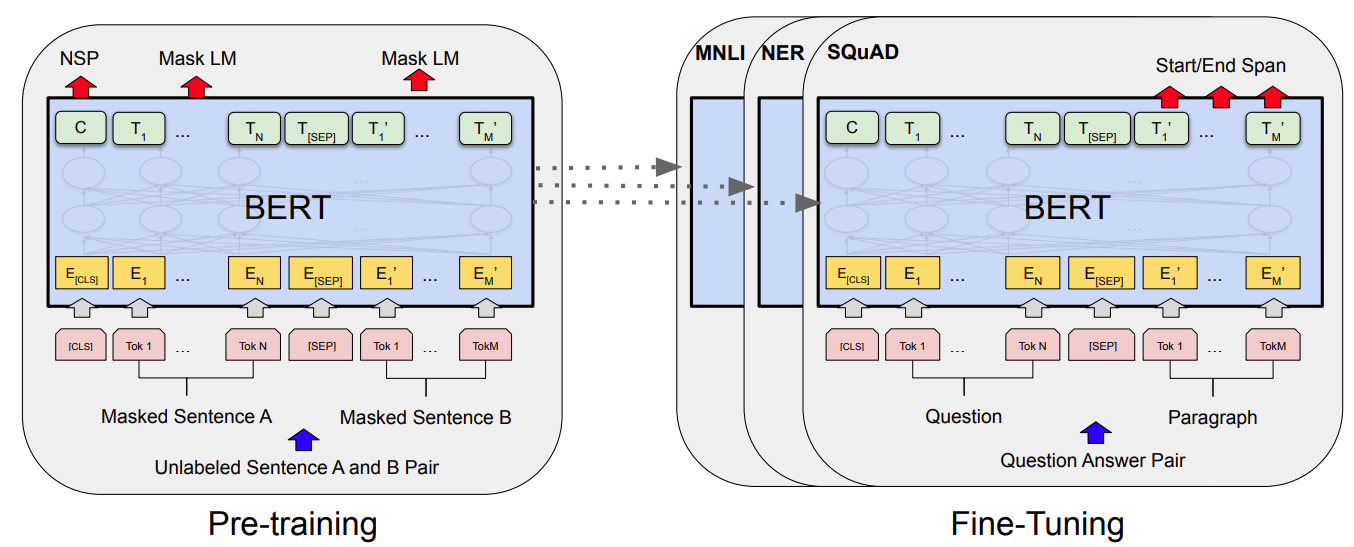}
   \caption[Sequential transfer learning using BERT]{Sequential transfer learning using BERT. (left) Pre-training stage using transformer, (right) Fine-tuning stage where the same pre-trained model parameters are used to initialize the BERT model for various NLP tasks. During fine-tuning, all parameters are updated. By excluding the output layers, the same architectures are used in both stages of sequential transfer leaning through BERT. \parencite{DBLP:journals/corr/abs-1810-04805}.}
  \label{fig:STL-BERT} 
 \end{figure}
\begin{figure}[t]
  \centering
   \includegraphics[width=1\textwidth]{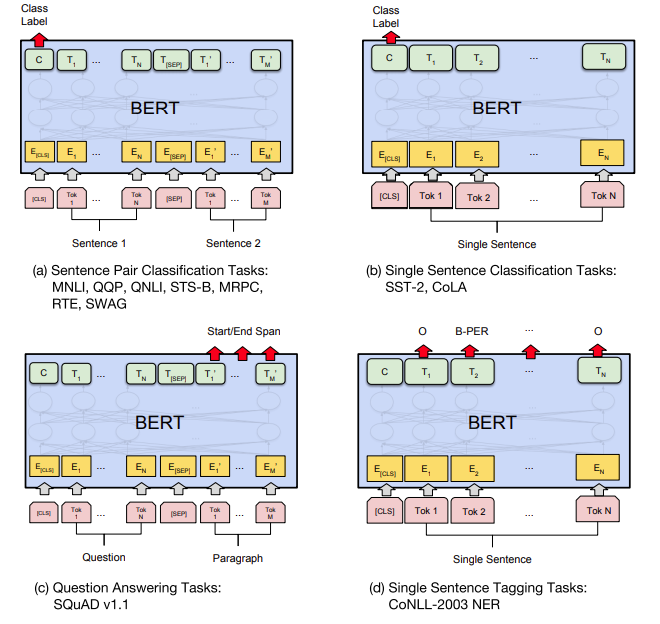}
   \caption[Fine-tuning in BERT for various NLP tasks]{Fine-tuning in BERT for various NLP tasks \parencite{DBLP:journals/corr/abs-1810-04805}.}
  \label{fig:BERT-Fine} 
 \end{figure}
\subsection{Adaptation} \label{ch1:adaptation}
Adaptation is the second stage of sequential transfer learning, in which the representation is transferred to a new task. In prior works, only input word embeddings are transferred to the down-stream task, however with GPT and BERT, all parameters are transferred to initialize end-task model parameters. It involves updating the pre-trained representations (i.e., Fine-tuning).

Figure \ref{fig:STL-GPT} depicts the adaptation stage using GPT and several input transformations to handle the inputs for different types of tasks. It can be seen that each task has a specific input transformation, and the discriminative fine-tuning step passes inputs through the GPT pre-trained model (i.e., transformer) to obtain the intermediate output. Then the transformer output is fed into an added softmax based linear output layer. 
BERT's fine-tuning is similar to GPT, and for each task, it processes the task-specific inputs and outputs with the pre-trained BERT and performs fine-tuning of all the parameters end-to-end (Figure \ref{fig:STL-BERT}). It uses a sentence separator ([SEP]) and
classifier token ([CLS]), in which their embeddings are learned during pre-training. For GPT, in contrast, these tokens are only introduced at fine-tuning time \parencite{DBLP:journals/corr/abs-1810-04805}. At the output level, as shown in Figure \ref{fig:BERT-Fine}, for token level tasks such as sequence labeling and question answering, the token representations are fed into the final layer. For classification tasks (e.g., entailment and sentiment analysis), on the other hand, the representation of the classifier token ([CLS]) is fed into the output layer.

\section{Information Extraction} \label{sec:ie}
{Information extraction (IE)} is the process of extracting desired knowledge in terms of names, entities, events, properties, and relations from a semi-structured or unstructured text by transforming them into a structured format. The structure is usually represented in the form of \emph {$<$subject, predicate, object$>$} or \emph {$<entity_1$, relation, $entity_2>$} triplets, known as facts. In this process, first, we have to define what constitutes a subject and object, then which type of relations should be considered. There are two paradigms of information extraction that have emerged recently \parencite{Nakashole2012}: 
\begin{itemize}
    \item \textbf{Schema-based IE}: In this approach, the process of extracting information from various information sources is guided by an ontology. The ontology typically consists of two kinds of information views; those that make up the assertion-view and those that make up the taxonomy-view. The taxonomy-view represents a topology or taxonomy of the domain at hand and includes the definition of the concepts, attributes, and their inter-relationships. The assertion-view describes the attributes of instances (or individuals), the roles between instances, and other assertions about instances regarding their concept membership within the taxonomy-view. The concepts and relations in the ontology are generally hand-specified by either developers of the ontology or by domain experts. Therefore the major flaw in schema-based IE is the limited number of classes and relations that can be populated from sentences.
    \item \textbf{Schema-free IE}: This approach to IE aims to extract assertions from a large volume of textual data, avoiding the restriction to a pre-specified vocabulary. It extracts all relations by learning a set of lexico-syntactic patterns in a supervised or unsupervised manner. While the schema-free IE answers the recall issue, it is highly susceptible to noise, due to the lack of tightly enforced semantics on relations and entities \parencite{Nakashole2012}.
\end{itemize}
Traditionally, the process of information extraction can be divided into a series of tasks. It typically begins with lexical analysis like assigning part-of-speech and features to words and phrases through morphological analysis and dictionary. Then it continues by entity recognition to identify names and instances of particular concepts of interest. The syntactic analysis comes along in most solutions to identify the noun, verb phrases, and dependency structure. Finally, relation extraction and classification tasks are applied to construct the facts of interests.  We can consider discourse analysis as a complementary step in the flow which resolves relations of co-reference and draws inferences from the document's explicitly stated facts.
Traditional IE systems are often based on a pipeline architecture where the tasks have been done sequentially using different task-specific patterns. The patterns are obtained using various statistical analysis and pattern recognition methods. 
Currently, most IE systems employ end-to-end neural networks by learning deep neural networks that map directly from the input to the output data naturally consumed and produced in IE tasks (e.g., \cite{DBLP:journals/corr/XuFHZ15}, \cite{gupta-etal-2016-table}, \cite{zheng-etal-2017-joint}, \cite{zeng-etal-2018-extracting} and \cite{cui-etal-2018-neural}). Even though most IE neural systems work end-to-end, there has been interest in incorporating various linguistic categories (PoS-tags, dependencies) into them to improve the performance, such as \cite{chiu-nichols-2016-named}, \cite{DBLP:journals/corr/XuMLCPJ15} and \cite{Noo:Ovr:Lon:18}.

Information Extraction has been explored under different areas such as {Fact \& Relation extraction}, {Knowledge base population} and {Unseen entity extraction}. 
\begin{itemize}
    \item \textbf{Fact \& Relation extraction.} 
This process generates structured data in the form of entity-relation triplets from natural language text documents. The conventional approach to fact extraction is to use pattern-based extraction and employ consistency constraint reasoning to provide proper facts.
In schema-based IE approaches such as BOA \parencite{Gerber2011}, PORSPERA \parencite{Nakashole:2011} , DARE \parencite{Xu:2010}, the patterns emerge by starting with a few seed facts from a knowledge base to bootstrap the extraction process. For example, BOA relies on distant supervision and existing facts from a knowledge base, in particular DBpedia. It applies a recursive procedure, starting with extracting triples from linked data, then extracting natural language patterns from sentences and constructing the facts in triplet format (i.e., RDF triples).
In schema-free IE approaches (e.g., TextRunner \parencite{banko2007open}, ReVerb \parencite{Fader:2011, 2015angeli-openie}) these patterns are constructed by leveraging linguistic structure, through syntactic and lexical rules. For instance, ReVerb extracts relations based on simple linguistic patterns (i.e., in terms of PoS-tags and noun phrase chunks). It extracts the facts by assuming that every relational phrase must be either a verb and a verb followed immediately by a preposition (e.g., located in), or a verb followed by nouns, adjectives, or adverbs ending in a preposition (e.g., has an atomic weight of) \parencite{Fader:2011}. ReVerb first looks for a matching relational phrase  and then finds the arguments (i.e., $entity_1$ and $entity_2$) of the relationship.  
Although these systems have been widely used in a variety of fact and relation extraction approaches, most of them were built on hand-crafted patterns from syntactic parsing, which causes errors in propagation and compounding at each stage. To alleviate extraction errors, various deep neural-based approaches have recently been proposed \parencite{stanovsky-etal-2018-supervised,cui-etal-2018-neural,jiang-etal-2019-improving,zhang-etal-2017-mt}. For instance, \cite{cui-etal-2018-neural} applied a seq2seq framework to provide a schema-free IE system. 
\item \textbf{Knowledge Base Population (KBP).}
Knowledge Base Population (KBP)  is the task of taking an incomplete knowledge base, and a large corpus of text, and completing the incomplete elements of the knowledge base. That is, the model has to interpret the text and get the desired information out of it.
Therefore in this task, we assume that we have prior but incomplete knowledge about the subject, and our aim is discovering its properties. Recent works like \cite{Lin15learningentity} and \cite{NIPS2013_5028}, exploit knowledge embeddings to infer new relational facts.
\item \textbf{Unseen entity extraction.} 
The previous areas rely on a common assumption in schema-based IE, namely the existence of a knowledge base that contains all entities and their types. However, during the extraction of facts from highly dynamic sources such as news, social media, and technical documents, new entities emerge that are not in the reference knowledge base. This area covers the problem of out-of knowledge base entities. Conventional named entity recognition tools have coarse-grained types and only deal with a limited set of entities such as a person, organization, and company. However, fine-grained  methods \parencite{Yukunma-etal-2016-label,mai-etal-2018-empirical, DBLP:journals/corr/abs-1904-10503} consider up to 200 types. In contrast, the proposed tools in this area, like PEARL \parencite{NakasholeTW13} and FINET \parencite{CorroAGW15}, deal with thousands of types. They are semi-supervised systems that leverage a repository of many relational patterns. Subjects and objects of each pattern carry the type information. They categorize entity mentions by the most likely type according to the pattern repository. The type system is based on a partial or the entire WordNet \parencite{Miller:1995:WLD:219717.219748} hierarchy.
\end{itemize}
In this thesis, we study two essential tasks in the area of Information Extraction, namely Named Entity Recognition (Chapter \ref{sec:third}) and Relation Extraction (Chapter \ref{sec:fourth}). 

\section{Natural Language Understanding} \label{sec:NLU}
Understanding of natural language is an essential and general goal of NLP.
Natural Language Understanding (NLU) comprises a wide range of diverse tasks, including, but not limited to, natural language inference, question answering, sentiment analysis, semantic similarity assessment, and document classification. In this thesis, we explore two central NLU tasks, including
natural language inference and question answering (Chapter \ref{sec:fifth}). We provide a brief description of these tasks in the following sections but go into details on related work in Chapter \ref{sec:fifth}.

\subsection{Natural Language Inference (NLI)}
NLI is the task of predicting whether a \textit{hypothesis} sentence is true (entailment), false (contradiction), or undetermined (neutral) given a \textit{premise} sentence.
NLI systems need some semantic understanding and models trained on entailment data can be applied to many other NLP tasks such as text summarization, paraphrase detection, and machine translation.  The task of NLI, also known as textual entailment, is well-positioned to serve as a benchmark task for research on NLU \parencite{williams-etal-2018-broad}.

\subsection{Question Answering (QA)}
The task of QA is often designed in the context of a reading comprehension task. This machine reading problem is formulated as extractive question answering, in which the answer is drawn from the original text \parencite{eisenstein2019introduction}. In this context, given a \emph{context} and a \emph{question}, the QA task aims to identify the span answering the question in the context.

    \chapter{Evaluation of Domain-specific Word Embeddings}
\label{sec:second}
In this chapter, we study input representations trained on data from a low resource domain (Oil and Gas) using sequential transfer learning of word embeddings (See Section \ref{sec:seqtr} in Chapter \ref{sec:first}). We conduct intrinsic and extrinsic evaluations of both general and domain-specific embeddings. We observe that constructing domain-specific word embeddings is worthwhile even with a considerably smaller corpus size. Although the intrinsic evaluation shows low performance in synonymy detection, an in-depth error analysis reveals the ability of these models to discover additional semantic relations such as hyponymy, co-hyponymy, and relatedness in the target domain. Extrinsic evaluation of the embedding models is provided by a domain-specific sentence classification task, which we solve using a convolutional neural network. We further adapt embedding enhancement methods to provide vector representations for infrequent and unseen terms. Experiments show that the adapted technique can provide improvements both in intrinsic and extrinsic evaluation.

\section{Introduction}
Domain-specific, technical vocabulary presents a challenge to NLP applications. Recently, word embedding models (See Section \ref{w2v-chapter2} in Chapter \ref{sec:first}) have been shown to capture a range of semantic relations relevant to the interpretation of lexical items \parencite{MikolovYZ13} and furthermore provide useful input representations and transferable knowledge for a range of downstream tasks \parencite{DBLP:journals/corr/abs-1103-0398}. The majority of work dealing with intrinsic evaluation of word embeddings has focused on general domain embeddings and semantic relations between common and generic terms. However, it has been shown that embeddings differ from one domain to another due to lexical and semantic variation \parencite{HamiltonCLJ16,BollegalaMK15a}. Domain-specific terms are challenging for general domain embeddings since there are few statistical clues in the underlying corpora for these items \parencite{BollegalaMK15a,pilehvar-collier:2016:BioNLP16}. On the other hand, domain knowledge resources, where the meanings of words are represented by defining the various relationships among those words, provide valuable prior knowledge for many NLP tools. Many works show that the encoded knowledge available in lexical resources can be exploited to improve the semantic coherence or coverage of existing word vector representations \parencite{faruqui:2014:NIPS-DLRLW,E17-2062}.

The following research questions related to the domain-specific data and model are investigated in this chapter:
\begin{question}\label{rq.2.1}
Can word embedding models capture domain-specific semantic relations even when trained with a considerably smaller corpus size?
\end{question}
\begin{question}\label{rq.2.2}
How can we take advantage of existing domain-specific knowledge resources to enhance the resulting models?
\end{question}

To answer these research questions, we train domain-specific embeddings and conduct a comprehensive study including a wide range of evaluation criteria against terminological resources, contrasting several general and domain-specific embedding models. We augment the domain-specific embeddings using a domain knowledge resource. We further adapt embedding enhancement methods to provide vector representations for infrequent and unseen terms by investigating the works of \cite{E17-2062} and \cite{faruqui:2014:NIPS-DLRLW}. We then go on to examine the contribution of these models in the performance of a downstream classification task. 
\section{Related Work}
Despite the pervasive use of word embeddings in language technology, there is no agreement in the community on the best ways to evaluate these semantic representations of language\footnote{RepEval @ACL 2016, 2017, and 2019: Workshop on Evaluating Vector Space Representations for NLP}. 
There exist a variety of benchmarks that are widely employed to assess the quality of word representations and to compare different distributional semantic models. Existing evaluation methods can largely be separated into two categories: \emph{intrinsic evaluation} and \emph{extrinsic evaluation}.
\subsection{Intrinsic Evaluation}
Intrinsic evaluation methods attempt to directly quantify how well various kinds of linguistic regularities can be detected
with a model-independent of its downstream applications \parencite{baroni-dinu2014,schnabel2015eval}. 
Existing schemes in intrinsic evaluation fall into two major scenarios \parencite{schnabel2015eval}: \emph{Absolute intrinsic evaluation} and \emph{Comparative intrinsic evaluation}, which will be described in the following.
\paragraph{Absolute intrinsic evaluation:}
\label{absIntrinsic}
This type of evaluation directly tests for syntactic or semantic relationships between words \parencite{schnabel2015eval} and analyzes the generic quality of embeddings \parencite{YaghoobzadehS16a}. Since it is computationally inexpensive and leads to fast prototyping and development of vector models, it has been the topic of many evaluation challenges. Several datasets have been developed to this end. Table \ref{tbl:dataset1} shows a compilation of datasets employed for intrinsic evaluation of word embeddings, organized by the semantic relation. It involves tasks such as the following \parencite{baroni-dinu2014,schnabel2015eval}: 
\begin{sidewaystable}
\scalebox{.80}{
\begin{tabular}{l|p{3.5cm}lr}
\midrule
{\bf Task} & {\bf Dataset name} & {\bf Dataset Inf.} & {\bf Reference} \\
\midrule
 \multirow{12}{*}{{Semantic Relatedness}}& RG & 65 word pairs & \cite{Rubenstein:1965}  \\
 & MC-30 & 30 word pairs  &\cite{miller1991contextual} \\
 & WordSim-353 & 353 word pairs  & \cite{Finkelstein:2001}   \\
& YP-130 &130 word pairs & \cite{Yang06verbsimilarity} \\
& WS-Rel & 252 word pairs & \cite{Agirre:2009}   \\
& WS-Sim & 203 word pairs & \cite{Agirre:2009}   \\
& MTruk-287 & 287 word pairs & \cite{Radinsky:2011}\\
& MTruk-771 & 771 word pairs & \cite{Halawi2012}  \\
& MEN & 300 word pairs  & \cite{Bruni:2012}  \\
& Rare Word & 2034 word pairs & \cite{Luong-etal2013}  \\
& Verb & 144 word pairs & \cite{BakerRK14}  \\
& SimLex-999 & 999 word pairs& \cite{HillRK14}\\
\midrule
Synonym  Detection
& TOFEL & 80 multi-choice questions (4 words)& \cite{Landauer97asolution} \\
\midrule
\multirow{6}{*}{{Categorization}}
&   &   &   \\
&  AP & 402 concepts, 21 categories & \cite{Almuhareb06}\\
& ESSLLI & 44 concepts, 6 categories & \cite{Baroni2008} \\
& BATTING & 83 concepts, 10 categories & \cite{Baroni:2010}\\
&   &   &   \\
\midrule
\multirow{4}{*}{{Selectional Preference}}
&   &   &   \\
&  UP & 211 noun-verb pairs & \cite{Pado07}\\
& MCRAE & 100 noun-verb pairs & \cite{McRae1998283} \\
&   &   &   \\
\midrule
\multirow{4}{*}{{Analogy}}
& AN & 19.5 K  analogy questions& \cite{Mikolov2013a}\\
& ANSYN & 10.5 K analogy questions & \cite{Mikolov2013a} \\
& ANSEM & 9 K analogy questions& \cite{Mikolov2013a}\\
& BATS  & 99.2 analogy questions  K  & \cite{Gladkova2016NAACL}  \\
\midrule
\multirow{4}{*}{{Coherence or Outlier detection}}
&   &   &  \\
& Intrusion & 100 of 3$+$1 words  & \cite{schnabel2015eval} \\
& 8-8-8  & 64 of 8$+$1 words & \cite{camacho2016find}\\
&    &    &   \\
\midrule

\multirow{5}{*}{QVEC-CCA}
&   &   &  \\
& SEM-Matrix & 4.2 K words with 41 features  & \cite{qvec:enmlp:15} \\
& SYN-Matrix  & 10.8 K  words with 45 features & \cite{TsvetkovFD16}\\
 &   &   &  \\
\midrule
\end{tabular}
}
\caption{Absolute intrinsic evaluation datasets.}
\label{tbl:dataset1}
\end{sidewaystable}

\begin{itemize}
\item {\bf Semantic Relatedness}: Given a ground truth of human assigned proximity scores to word pairs such as \emph{money:dollar $\approx$ 8.42, tiger:mammal $\approx$ 6.85}, the evaluation task aims to find the degree of correlation between the scores provided by the model and the human rating as a performance of the model. 
The cosine similarity of the corresponding vectors for word pairs provided by the model should be highly correlated with the gold standard (measured by Spearman or Pearson correlation).
 
\item {\bf Categorization}: 
 Given a set of words, the system needs to group them into different semantic categories (e.g., \emph{helicopters} and \emph{motorcycles} should go to the \emph{vehicle} class, \emph{dogs} and \emph{elephants} into the
\emph{mammal} class). By applying a clustering method to the corresponding vectors of all words in a dataset, the model's performance is calculated concerning the purity of the outcome clusters concerning the human-judgment labels. 
\item {\bf Synonym  Detection}: The ability of the embedding model to find the correct synonym for a word is assessed. For example, for the target word \emph{levied} the list of options to choose are \emph{imposed} (correct one), \emph{believed}, \emph{requested} and \emph{correlated}. For each target word, its cosine similarities with synonym candidates are calculated, and the one with the highest score is selected. The model performance is the accuracy of  the model prediction. 
\item {\bf Selectional Preference}:
The goal is to label a noun as a subject or object for a specific verb (e.g., people received a high average score as a subject of to eat, and a low score as an object of the same verb). \cite{baroni-dinu2014} and \cite{schnabel2015eval} followed the procedure of \cite{Baroni:2010} to perform this task. First, for each verb in the dataset, the 20 nouns which are most strongly associated as subject or object are selected, then a \emph{prototype} vector is calculated as the average of these nouns is calculated. Therefore we will have a subject and object type prototype vector for each verb. The performance of the model will be the correlation degree (\emph{spearman}) of the averaged human ratings for each type and the cosine scores between the target nouns vectors and the relevant prototype vectors of the verbs.

\item {\bf Analogy}: The analogy task asks the model to detect whether two pairs of words stand in the same relation. These relations fall into different types of linguistic relations, such as morphological and semantic relations.
Having two word pairs \emph{A:B}::\emph{C:D} where the \emph{D} is missing,  the goal is to find the missing word in the relation: \emph{A} is to \emph{B} as \emph{C} is to \emph{D}, in which \emph{C}, \emph{D} are related by the same relation as \emph{A}, \emph{B} . For example, \emph{France : Paris :: Germany : Berlin}.  Following the procedure proposed by \cite{Mikolov2013a}, the first term vector in the first pair is subtracted from the second term vector, then the test term is added to the result (\emph{B-A+C}). Afterward, the nearest neighbor to the final vector is requested from the model. The performance of the model is measured as the proportion of the questions where the nearest neighbor suggested by the model is the correct answer (accuracy).   

\item {\bf Coherence / Outlier Detection}: The goal is to assess whether the neighbor words in the embedding semantic space are mutually related. Therefore a good model should provide coherent neighborhoods for a target word. To tackle this task, groups of coherent words and intruder words are introduced, and the model should be able to spot the word that is an outlier and does not belong to the group of neighbor words. For example, among the following words: (a) \emph{finally} (b) \emph{eventually} (c) \emph{immediately} (d) \emph{put}, the query word is option (a), intruder is (d). \cite{schnabel2015eval} presented this intrinsic evaluation as an intrusion task and evaluated the performance of the models by the precision metric. On the other hand, in \cite{camacho2016find}, the task was introduced as outlier detection and solved as a clustering problem, in which each group of coherent words are clustered based on a compactness score and the intruder words are ranked by their positions (8 outlier positions: the 1st position has the lowest dissimilarity to the cluster and the 8th position has the highest dissimilarity). The model quality is measured by outlier position percentage and accuracy of the outlier detection.

\item {\bf QVEC and QVEC-CCA} \label{qvec-cca}:  The basic hypothesis of QVEC is that dimensions in distributional vectors encode the linguistic features of words. It measures the quality of a model by how well the embedding correlates with a matrix of features from manually crafted lexical resources. For example, the target word \emph{fish}	is  assigned to the following senses along with scores: \emph{animal: 0.684, food: 0.157, competition: 0.0526, contact: 0.105}. \cite{qvec:enmlp:15} introduced QVEC as a measure to quantify the linguistic regularities of an embedding model. For target words that are in the embedding model, it obtains an alignment between the word vector dimensions and the linguistic dimension in which it maximizes the correlation (Pearson correlation) between the aligned dimensions of the two matrices. The higher the correlation, the more salient the linguistic feature of the dimension. The QVEC-CCA \parencite{TsvetkovFD16} followed the same idea as  QVEC. However, to measure the correlation between the embedding matrix and the linguistic matrix, it employs canonical correlation analysis (CCA \parencite{Hardoon:2004}). CCA generates two basic vectors for the embedding and feature metrics such that the projections of these two metrics onto their basic vectors have a maximum correlation.
\end{itemize}
\paragraph{Comparative intrinsic evaluation:} This type of evaluation is based on direct feedback from the user on the model outcome using a crowd-sourcing environment (e.g., Amazon Mechanical Turk). For each target word, each embedding model is questioned to provide the nearest neighbors at ranks $k \in \{1,5,50\}$. Then the human annotators select the most similar answer, and the model that has the majority votes is considered to be the winner. The dataset is called \emph{Query Words} and includes 100 queries \parencite{schnabel2015eval}.  

\subsection{Extrinsic Evaluation} \label{r:extrinsic}
Given the widespread use of word embeddings as input representations in neural NLP systems, the quality of a word vector may also be assessed by performance in downstream tasks. This is done by measuring changes in performance metrics specific to the tasks by extrinsic evaluation. The downstream language technology tasks on which the quality of a word embedding has been examined, fall into syntactic (e.g., POS tagging, Chunking) and semantic (e.g., Named Entity Recognition, Sentiment Analysis) categories. However, by the definition of extrinsic evaluation, any downstream task could be considered as an evaluation method. Various downstream tasks and the related resources that are commonly used in the extrinsic evaluations of word embeddings are as follows (Table \ref{tbl:extrinsic}):

\begin{table}[t]
\scalebox{.66}{
\begin{tabular}{p{3cm}lll}

{\bf Task}& {\bf Data Set}&{\bf Dataset info.} & {\bf Evaluation }  \\
& &{\bf \footnotesize(Train/Dev/Test) }& {\bf Ref. }  \\
\midrule
\midrule
 POS Tagging & Penn Treebank   & 958K, 34K, 58K &\cite{GHANNAY16.392} \\
&  \parencite{Marcus:1993} &  &    \cite{ChiuBillyRepEval2016} \\
&   &   &    \cite{NayakRepEval2016}\\
\midrule
Chunking & CoNLL 2000   &191K, 21K, 47K  &\cite{schnabel2015eval}  \\
& \parencite{TjongKimSang:2000}  &    & \cite{ChiuBillyRepEval2016} \\
&   &  & \cite{GHANNAY16.392}\\
&      &    & \cite{NayakRepEval2016}\\
\midrule
 Named & CoNLL2003  & 205K, 52K, 47K &\cite{GHANNAY16.392} \\
 Entity Recognition &  \parencite{TjongKimSang:2003}    &  &\cite{ChiuBillyRepEval2016} \\
&   &    &\cite{NayakRepEval2016}\\
\midrule
 Sentiment Analysis & Stanford Sentiment Treebank  \parencite{socher-EtAl:2013}  & 8.5K, 1.1K, 2.2K &\cite{NayakRepEval2016} \\
  & Movie Reviews (aclimdb) \parencite{maas2011}& 25K, -, 25K &\cite{schnabel2015eval} \\
\midrule
 Question Classification &  TREC   \parencite{Li:2002}  & 15.5k, -, 500 &\cite{NayakRepEval2016} \\
\midrule
 Natural Language Inference & PPDB:Eng  \parencite{ganitkevitch2013ppdb}  & 221.4M &\cite{NayakRepEval2016} \\
\midrule
\end{tabular}
}
\caption{Extrinsic evaluation tasks and datasets.}
\label{tbl:extrinsic}
\end{table}
\paragraph{Part-Of-Speech (POS) Tagging} To identify the morpho-syntactic label of each word in the sentences. The evaluations are performed on the standard Penn treebank dataset \parencite{Marcus:1993}, using the neural method suggested by \cite{DBLP:journals/corr/abs-1103-0398} . 
\paragraph{Chunking} The chunking is a syntactic sequence labeling task where the goal is to locate phrases in the text. The pre-trained embedding models are used as an input for a noun phrase chunking task similar to those employed by \cite{Turian:2010:WRS:1858681.1858721} and  \cite{DBLP:journals/corr/abs-1103-0398} using the dataset of CONLL-2000 shared task \parencite{TjongKimSang:2000}. 
\paragraph{Named Entity Recognition} NER systems perform the task of detecting named entities in text (e.g., persons, locations and organization) as a sequence prediction task. The system proposed by \cite{DBLP:journals/corr/abs-1103-0398} is evaluated on the CoNLL 2003 benchmark \parencite{TjongKimSang:2003} using different embedding models.
\paragraph{Sentiment Analysis} The task of sentiment analysis aims to classify the polarity of a given text as positive, negative, or neutral. Several datasets like user reviews \parencite{maas2011} and Stanford Sentiment Treebank \parencite{socher-EtAl:2013} are used for downstream evaluation of this task.
\paragraph{Question Classification} Question classification refers to the task of mapping a given question to one of $k$ classes such as Person, Quantity, Duration, Location, and so on. For example, given the question “Who was the first woman killed in the Vietnam War?”, we would like to know that the target of this question is a person. A hierarchical classifier is used to classify questions into fine-grained classes described in \cite{Li:2002}. 
\paragraph{Natural Language Inference (NLI)} The NLI task is used to examine the ability of embedding models in propagating lexical relations \parencite{NayakRepEval2016}. NLI aims to infer the logical relationship, typically entailment or contradiction, between the given \textit{hypothesis} and  \textit{premise} sentences. The task is performed using the dataset presented in \cite{ganitkevitch2013ppdb}.
  
In the following sections, we will describe how we evaluate domain-specific word embedding models using both intrinsic and extrinsic evaluation schemes. 
\section{Intrinsic Evaluation Setup}
Intrinsic evaluation of word embeddings has two main requirements. First, we require a query inventory as a gold standard, and second, a word embedding model that has been trained on a specific corpus. In this section, we describe how we build a domain-specific query inventory for the Oil and Gas domain by exploiting a domain-specific knowledge resource. Then, the domain-specific corpus and the training of the embedding models will be described. We then go on to clarify the evaluation methodology. 

\subsection{Domain-specific query inventory} \label{slb}
As an intrinsic evaluation, we would like to assess the quality of representations independently of a specific NLP task. Currently, this type of evaluation is mostly done by testing the overall distance/similarity of words in the embedding space, i.e., it is based on viewing word representations as points and then computing full-space similarity. The assumption is that the high dimensional space is smooth and similar words are close to each other \parencite{YaghoobzadehS16a}. 
Computational models that could capture similarity as distinct from the association have many applications in language technology such as ontology and dictionary creation, language correction tools, and machine translation.
As described in detail above, for  the general domain, there exists a wide range of gold standard resources for evaluating distributional semantic models in their ability to capture semantic relations of different types, for instance, \emph{Simlex-999} \parencite{HillRK14}. WordSim-353 \parencite{Agirre:2009} and MEN \parencite{Bruni:2012} (See Table \ref{tbl:dataset1}).

\begin{figure}
\centering
\includegraphics[width=\textwidth]{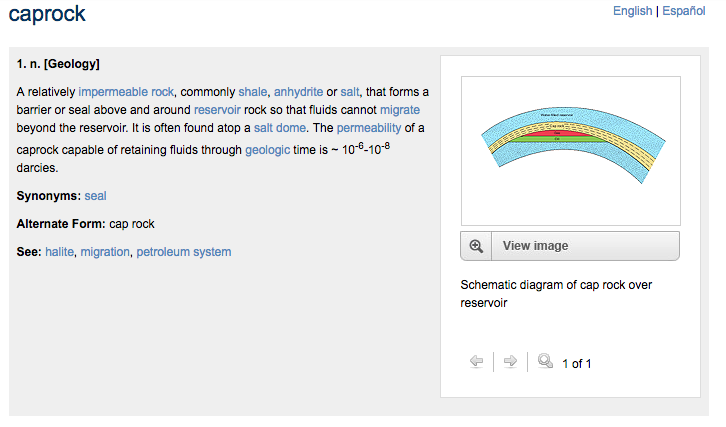}
\caption[Term structure in the slb glossary.]{Term structure in the slb glossary.}
\label{fig:slb:term}
\end{figure}
However, evaluating domain-specific embeddings by applying these gold standards will not provide an adequate picture of their quality since they do not share a common vocabulary and word meanings. The domain of oil and gas has received little previous work in NLP and there are no readily available resources. For this reason, we create a domain-specific gold standard using the Schlumberger oilfield glossary (\emph{slb}).\footnote{\url{http://www.glossary.oilfield.slb.com/}} The \emph{slb} is a reference that defines major oilfield activities and has been created by technical experts.

Figure \ref{fig:slb:term} shows the structure of the term \emph{caprock} in this glossary. 
Terms are described by their part of speech, their discipline (e.g., \emph{Geology}, \emph{Well Completions}), as well as a textual definition. Terms are linked to other terms in the glossary using semantic and lexical relations such as \emph{Synonyms}, \emph{Antonyms}, and \emph{Alternate forms}. It provides a network of related terms that can be navigated through the glossary. In our example entry for instance, we find that the \emph{synonyms} and \emph{alternate forms} of \emph{caprock} are \emph{seel} and \emph{cap rock}, respectively.
(see Figure \ref{fig:slb:term}). 
Finally, if the term has an image that clarifies its definition, it will appear in the image section next to the definition part. All these elements in the term's structure are located in the following tables inside a relational database:
\begin{description}
    \item [definitions:] All the main terms of the glossary are located in this table. They are defined by \emph{id}, \emph{name}, \emph{definition}, \emph{term\_type} (verb, noun, adjective, adverb, preposition and transitive verb) as part of speech tags, \emph{postdate}, and \emph{lastupdate}. It is possible that one term has more than one definition if it has different part of speech tags, or if it is assigned to different disciplines or both.
    For example, the term \emph{dry gas} has been assigned to the \emph{Geology} and \emph{Well Completions} disciplines and has different definitions in each discipline.
    \item [disciplines:] There are 20 main categories that describe disciplines in the glossary (e.g., \emph{Drilling, Geology, Geophysics} and \emph{Well Completion}). Each term in the definitions table is assigned to one or more disciplines and its definition varies based on the assigned discipline.
   \item [images:] To illustrate and clarify many definitions, high-quality and full-color photographs are assigned to the definition by \emph{image\_name}, \emph{image\_caption} and \emph{image\_url}.
    \item [links:] The inter-glossary relations among the terms in the glossary are specified in this table. These relations are characterized with type taken from the following set: \emph{Synonym}, \emph{Antonym}, \emph{Alternative form} and \emph{See}. These relations form the basis of our intrinsic evaluation dataset.
\end{description}
We construct a domain query inventory by extracting all terms and their inter-glossary relations from the relational database. The terms are converted to lowercase and assigned n-gram type (i.e., a contiguous sequence of a word, unigram where n=1, bigram where n=2, trigram where n=3 and $>$3 where n$>$3). 
The glossary consists of \emph{4,886} terms.
Table ~\ref{terms_type} shows the distributions of terms in the glossary concerning their n-gram type and part of speech tags (one term may be assigned to more than one tag).
\begin{table}[t]
    \centering
    \begin{tabular}{lcccccc} 
     \head {n-gram} & \head {\#} & \head{Noun} & \head{Verb} & \head{Adj.} & \head {Adv.} & \head{Pre-position} \\
     \midrule
       unigram & 1499 & 1261 & 98 & 189 & 1& 2 \\
   bigram & 2569 & 2505 &35 &36 &1 &1   \\
        trigram & 660 & 644& 13& 4& 0& 0 \\
        $>$3 & 158 & 149 & 3& 6 & 0& 0 \\
         \midrule
          All &  4886 &	4559 &	138&	246 & 2& 3  \\
    \end{tabular}
    \caption{N-grams \& Part of speech tags in the slb glossary.}
   \label{terms_type}
\end{table}

Following the symmetric nature of the \emph{Synonym}, \emph{Antonym}, and \emph{Alternative form} relations, we infer a relationship if it is missing between terms. The final query inventory contains \emph{878} synonym pairs, \emph{284} antonym pairs and \emph{934} alternative form pairs. We observe that the majority of terms in the query inventory are multi-word units (70\%) and nouns (72\%). This indicates that a large portion of the domain-specific vocabulary that we want to capture in our model consists of multi-word entities. Thus we need to take this into account during the training of embeddings.

\subsection{Training of Word Embeddings} \label{w2v-chapter2}

In order to train domain-specific embeddings, we need a domain-specific corpus. Therefore, we compile a corpus consisting of technical reports and scientific articles in the Oil and Gas domain \footnote{SIRIUS partners provided the sources for the corpus.}. Table \ref{tbl:oilgas_dataset} shows detailed information about these sources. As we can see, the corpus covers several different genres with a majority taken from the genre of scientific articles.  The corpus contains $47,423$ documents and $8,280,935$ sentences.

\begin{figure}[t]
\centering
\includegraphics[width=.5\textwidth]{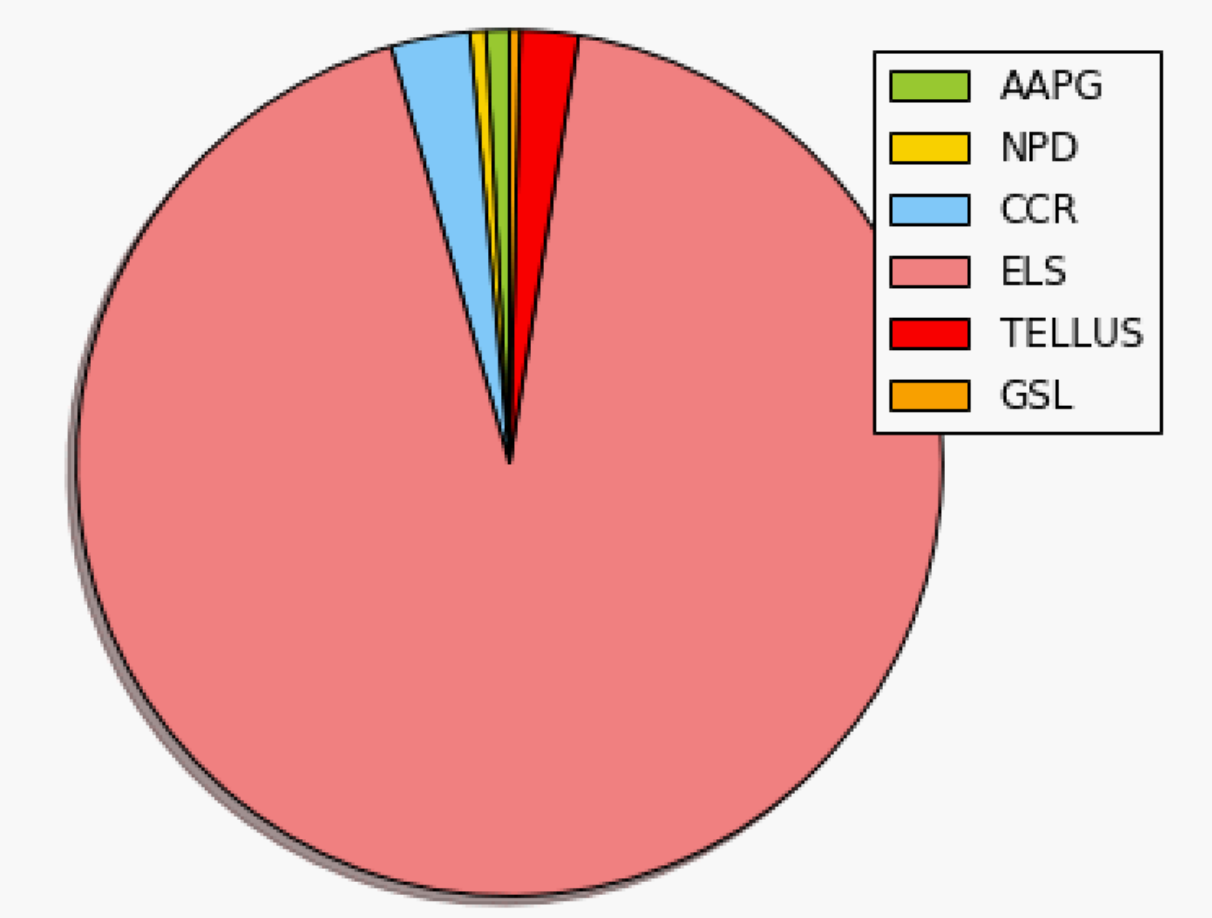}
\caption{Distribution of the sources in the domain corpora.}
\label{fig:source-01}
\end{figure}

\begin{table}[t]
    \centering
\scalebox{.7}{
   \begin{tabular}{@{}lllrr@{}}
   \head{Source} & \head{Abbr.} & \head{Description} & \head{Docs} & \head{Sentences}\\
       \midrule
        American Association of Petroleum Geologist & AAPG & Scientific articles & 3,382 & 72,243 \\
        
      C\&C Reservoirs-Digital Analogs &  CCR & Field evaluation reports &1,140 & 244,017 \\   
Elsevier & ELS & Scientific articles, magazines& 40,757 & 7,703,447\\
 Geological Society, London Memoirs & GSL & Scientific articles & 152 & 32,352
 \\

Norwegian Petroleum Directory& NPD& Norwegian Field info & 514 & 49,426
 \\
Tellus &TELLUS & Basin info& 1,478 & 179,450\\
\bottomrule
\multicolumn{3}{@{}l@{}}{Total}& 47,423 & 8,280,935
    \end{tabular}
   }
    \caption{Sources of the Oil and Gas corpus.}
    \label{tbl:oilgas_dataset}
\end{table}
As can be seen from Table \ref{tbl:oilgas_dataset}, the corpus contains different types of documents from various sources. Figure \ref{fig:source-01} depicts the distribution of sources in the domain corpora. We observe that most of the documents belong to \emph{Elsevier}, and the other sources cover a small proportion of the documents.
The domain-specific dataset is preprocessed using the following steps:
\begin{enumerate}
    \item Tokenization and lemmatization using Stanford-CoreNLP \parencite{manning-EtAl:2014:P14-5}. English stop words and sentences with less than three words are removed from the corpus.
    \item Shuffling: we randomly shuffle the text in the dataset. During the training of embedding models, the learning rate is linearly dropped as training progresses, text appearing early has a more significant effect on the model. Shuffling makes the effect of all text almost equivalent \parencite{ChiuACLBio}.
\end{enumerate}
For training of the word embeddings, we exploit the available word2vec (see Section \ref{w2v-chapter2} in Chapter \ref{sec:first}) implementation \emph{gensim} \parencite{gensim}. The elements that impact  the performance of the model are the input corpus, model architecture, and the hyper-parameters. In many articles \parencite{Velldal,DBLP:journals/corr/Camacho-Collados17aa} lemmatizing, case-folding, and shuffling input during training the word2vec are recommended; we carried out our experiments with these settings as detailed above.

We employ the phrase model of \emph{gensim}, which automatically detects common phrases (multi-word expressions). The phrases are collocations (frequently co-occurring tokens), and we consider bigrams and trigrams in this extraction process.
The phrase model has two main parameters: 

\begin{enumerate}
    \item \emph{min.count:} During training, all words and phrases with total count lower than this number are ignored.
    \item \emph{threshold:} Represents a threshold for forming the phrases (higher means fewer phrases).
\end{enumerate}
A phrase of words $a$ and $b$ is extracted if :
\begin{equation}
    \frac{\big(count(a, b) - min.count \big) * N} {count(a) * count(b)} > threshold
\end{equation}
\noindent Here N is the total vocabulary size and $count(a,b)$ is the total number of times where word $a$ and $b$ co-occur as a multi-word expression in the corpus.
We set the \emph{min.count} equal to 5 and the \emph{threshold} equal to 200 and 100 for bigrams and trigrams, respectively (these values were determined empirically). 
We further proceed with the domain-specific model generation by creating two sets of embeddings, employing both the \emph{CBOW} and the \emph{Skip-gram} architectures (see Section \ref{w2v-chapter2} in Chapter \ref{sec:first}) with default settings. In the initial evaluation step, we compare the outcomes of these two models to determine the better architecture. 

Embedding models consist of several parameters that can be tuned. We now go on to compare different settings for the hyperparameters, while keeping all other settings constant. 
It has been claimed in previous works that optimizations of hyper parameters and certain system choices constitute the leading causes of differences in performance rather than the algorithms themselves \parencite{LevyGD15}. 
Here we investigate the impact of various system design choices in the evaluation of domain-specific embeddings across the following parameters \footnote{Default values are in bold.}: 
\begin{itemize} 
\setlength{\itemsep}{2pt}
\item Vector size ($dim$): The dimensionality of the learned dense vector is determined by the vector size parameter. ($dim \in {50,{\bf 100},200,300,400,500,600}$) 
\item Context window size ($win$):
   The range of words included in the context of a target word is determined by the window size parameter. For instance, a size of 3, takes three words before and after a target word and injects into the training model as context words. ($win \in {2,3, {\bf5},10,15, 20}$).
\item  Negative sampling size ($neg$):  The idea of word2vec is to maximize the similarity between the word vectors, which appear close together (in the context of each other), and minimize the similarity of words that do not. However, this process includes an expensive computation to calculate the similarity between the target word and all other context words in the corpus. Negative sampling is one of the ways of addressing this problem, by simply selecting select a couple of contexts at random and calculating the similarity of target word to randomly chosen negative words. ($neg \in {3, {\bf 5},10,15}$) (See section \ref{w2v-chapter2} in Chapter \ref{sec:first}).
\item Frequency cut off ($min.count$): Words with a total frequency lower than the $min.count$ will be ignored from the corpus. This results in fewer words in the vocabulary of the model. ($min.count \in {2,3,{\bf 5},10}$).
\item $n$-most-similar: The parameter $n$ for top $n$-most-similar as output is fixed at the value $5$ (the maximum number of terms that are involved in each relation set in the query inventory).
\end{itemize}
We evaluate these different system design settings based on our intrinsic benchmark. We train different embedding models by varying values of one hyper-parameter and keeping others as default. After that, we perform evaluations over the domain-specific query inventory.

\subsection{Evaluation measures}  
For evaluation, we assume that for each term in the inventory, an embedding model should be able to propose similar words which are related semantically as either \emph{synonym}, \emph{alternative form} or \emph{antonym}. We will measure this by looking at a target word and its relation set in the query inventory, for instance, its synonyms and top $n$-most similar words predicted by the embeddings model. If the synonym of the target word is in the $n$-most-similar words, we will count it as a true positive. Otherwise, the target word and its synonym will be considered as a false negative, and the target word and the predicted word by the model will be counted as a false positive.
Since these relations are symmetric, the pairs $(t_i,t_j)$ and $(t_j,t_i)$ are considered equivalent in the evaluation. We calculate the \emph{accuracy} (A) as the number of target words for which the model provides at least one correct prediction, the \emph{recall} (R) as the number of correctly predicted word pairs over all word pairs (i.e., the sum of true positives and false negatives) and \emph{precision} (P) as the number of correctly predicted word pairs over all predicted word pairs
(i.e., the sum of true positives and false positives) for each relation category.

\section{Intrinsic Evaluation Experiments} \label{exp1}
In the following, we present experiments that evaluate the domain-specific word embedding models intrinsically. We first present tuning experiments and then present an experimental comparison between domain-specific and general domain embedding models.
\subsection{Model architecture: Skip-gram vs. CBOW}
\begin{table}[t]
    \begin{tabular}{p{2cm}|P{.6cm} P{.6cm} P{.6cm}|P{.6cm} P{.6cm} P{.6cm}|P{.6cm} P{.6cm} P{.6cm}}
      Model&\multicolumn{3}{P{1cm}|}{Synonymy}& \multicolumn{3}{P{1cm}}{Antonymy }   &\multicolumn{3}{|P{1.5cm}}{Alt. form  }\\
      \cmidrule{2-10}
       & $A$ &$R$ &$P$ & $A$ &$R$ & $P$ & $A$ &$R$ & $P$ \\
       \midrule
        Skip-gram & 9.8 & 8.0 & 2.2 & 46.4 &41.3 & 9.3 &12.1 &10.4 & 2.4  \\
        CBOW &{\bf 12.7} &{\bf 10.2} &  {\bf 2.7} & {\bf 55.3} &{\bf 49.2} & {\bf 11.1} & {\bf 12.8} &{\bf 11.0}&  {\bf 2.6}  \\
    \end{tabular}
    \caption{Evaluation results for different architectures.}
    \label{tbl:cbow,skip}
\end{table}

First, we compare the models obtained using the different word2vec architectures (CBOW and Skip-gram) with default values for hyper-parameters i.e. $dim=100 $, $win=5$, $min.count=5$ and $neg=5$. Table \ref{tbl:cbow,skip} presents the results for the two architectures broken down by semantic relation from the query inventory. The results show that the embedding models have higher scores for \emph{antonymy} prediction than \emph{synonymy}, see Table  \ref{tbl:cbow,skip}. This result is consistent with previous studies such as \cite{vanderPlas:2006} and \cite{Leeuwenberg2016} in which they reported that using distributional similarity some word categories like antonyms, (co)hyponyms or hypernyms show up more often than synonyms. In general, we find that the CBOW based model shows better results than the Skip-gram in all semantic relation tasks.

\subsection{Hyper-parameter tuning}
We go on to explore the impact of each hyper-parameter on the detection of semantic relations. We observe that the performance of the embedding models can be notably improved over the default hyper-parameters, but like the findings in other studies \parencite{Gladkova2016NAACL,ChiuACLBio}, the effects of different configurations are diverse and sometimes they are contradictory.
For example, different relation categories benefit from different context window sizes, and we find that the model with larger context windows tends to capture the \emph{antonymy} relation while a model with smaller windows, better captures \emph{synonymy} relation of the words. We also observe that negative sampling and frequency cut-off parameters have different impacts on the three relation categories.
\paragraph{Vector size (\emph{dim})}
The effect of vector size on the trained models is quite similar in all tasks (Table \ref{tbl:dim}). We observe a large improvement in all evaluations when the dimensionality is increased. However, the improvement peaks at $400$ for the \emph{synonymy} and \emph{antonymy} predictions and $500$ for \emph{alternative form}.

\begin{table}[t]
    \centering
    \scalebox{1}{
     \begin{tabular}{p{1cm}|P{.6cm} P{.6cm} P{.6cm}|P{.6cm} P{.6cm} P{.6cm}|P{.6cm} P{.6cm} P{.6cm}}
      $dim$&\multicolumn{3}{P{1cm}|}{Synonymy }& \multicolumn{3}{P{1cm}}{Antonymy }   &\multicolumn{3}{|P{1.5cm}}{Alt. form }\\
      \cmidrule{2-10}
       &$A$ & $R$ & $P$ & $A$ &$R$ & $P$ & $A$ &$R$ & $P$ \\
       \midrule
        50 & 12.7 &10.2 & 2.7  &48.2& 42.9 & 9.6 & 11.4 &9.8 & 2.3 \\
        100 &12.7 &10.2 & 2.7 &55.4 & 49.2 & 11.1 & 12.9 &11.0 & 2.6 \\
        200 &  14.7&12.4 & 3.3 &55.4& 49.2 & 11.1 &14.3 &12.3 & 2.9  \\
        300 & {\bf15.7}&{\bf 13.1}  & {\bf 3.5} & 55.4 &  49.2 & 11.1 & 13.6 & 11.7 & 2.7  \\
        400 &{\bf15.7}& {\bf 13.1} & {\bf 3.5} &{\bf 57.1} & {\bf 50.8 } & {\bf 11.4} & 13.6 & 11.7 & 2.7 \\
        500 &14.7 & 12.4 & 3.3 & 53.6 & 47.6 & 10.7 &  {\bf 15.0}&{\bf 12.9 } & {\bf 3.0 } \\
        600 &14.7 & 12.4 & 3.3 & 51.8 & 46.0 & 10.4 &12.9 & 11.0 & 2.6  \\
        700 &14.7 & 12.4 & 3.3 & 53.6 & 47.6 & 10.7 & 13.6 &11.7 & 2.7  \\
    \end{tabular}
    }
    \caption{Evaluation results for different vector size (default=100).}
    \label{tbl:dim}
\end{table} 
\paragraph{Context window size (\emph{win})}
Table \ref{tbl:win} depicts the impact of window size per evaluation task. We find that the embedding model can benefit from low window size ($w$=3) for the \emph{synonymy} task while in \emph{antonymy} and \emph{alternative} form tasks the model performance fluctuates between lower and higher window sizes.
\begin{table}[t]
    \centering
    \scalebox{1}{
          \begin{tabular}{p{1cm}|P{.6cm} P{.6cm} P{.6cm}|P{.6cm} P{.6cm} P{.6cm}|P{.6cm} P{.6cm} P{.6cm}}
      $win$&\multicolumn{3}{P{1cm}|}{Synonymy }& \multicolumn{3}{P{1cm}}{Antonymy }   &\multicolumn{3}{|P{1.5cm}}{Alt. form }\\
      \cmidrule{2-10}
       &$A$ & $R$ & $P$ & $A$ &$R$ & $P$ & $A$ &$R$ & $P$ \\
       \midrule
        2 &12.7 &10.2 &  2.7 & 55.4& 49.2  & 11.1& {\bf13.6} & {\bf 12.3 } & {\bf 2.9}  \\
        3 & {\bf13.7} & {\bf 12.4} & {\bf 3.3} & 48.2& 42.9 & 9.6  &11.4 &9.8 & 2.3   \\
        5 &12.7 &10.2  & 2.7& 55.4&  49.2 & 11.1 &12.9 &11.0 & 2.6  \\
        10 & 13.1& 10.9 & 2.9& 53.6& 47.6  & 10.7 &{\bf13.6}& {\bf 12.3}  & {\bf 2.9} \\
        15 & 12.7&10.2 & 2.7  & {\bf 67.1}&{\bf 50.8}  & {\bf 11.4} & 12.9 &11.0  & 2.6  \\
        20 &12.7& 10.2 & 2.7& 53.6& 47.6  & 10.7 &12.1& 10.4  & 2.4 \\
    \end{tabular}
    }
     \caption{Evaluation results for different context window size (default=5).}
    \label{tbl:win}
\end{table}

\paragraph{Negative sampling (\emph{neg})}
Unlike the practical recommendation in \cite{LevyGD15} who state that the skip-gram model prefers many negative samples, the CBOW model shows the contradictory result with respect to this parameter in our evaluation benchmarks. Table \ref{tbl:neg} presents that the results remain constant regardless of the negative sampling number in the synonym prediction task. On the other hand, we find that the performance correlates with an increase of this parameter in alternative form detection. For the antonym task, performance reaches a peak on $neg$ equal to 5 and 10 before dropping.  
\begin{table}[t]
    \centering
    \scalebox{1}{
\begin{tabular}{p{1cm}|P{.6cm} P{.6cm} P{.6cm}|P{.6cm} P{.6cm} P{.6cm}|P{.6cm} P{.6cm} P{.6cm} }
$neg$&\multicolumn{3}{P{1cm}|}{Synonymy }& \multicolumn{3}{P{1cm}}{Antonymy }   &\multicolumn{3}{|P{1.5cm}}{Alt. form }\\
      \cmidrule{2-10}
       &$A$ & $R$ & $P$ & $A$ &$R$ & $P$ & $A$ &$R$ & $P$ \\
       \midrule
        3 & 12.7 &10.2  & 2.7  &53.6& 47.6  & 10.7  &12.1 &10.4  & 2.4   \\
        5 &12.7& 10.2  & 2.7  &{\bf55.4}& {\bf 49.2} & {\bf 11.1}& 12.9 &  11.0  &  2.6 \\
        10 & 12.7&10.2  & 2.7 & {\bf55.4}&{\bf 49.2}  & {\bf 11.1} & 13.0 &11.7  &  2.7   \\
        15 &12.7& 10.2  & 2.7 &51.8& 46.0  & 10.4 &{\bf 13.6}& {\bf 12.3}  & {\bf 2.9}  \\
    \end{tabular}}
        \caption{Evaluation results for different number of negative samples  (default=5).}
    \label{tbl:neg}
\end{table} 

\paragraph{Frequency cut off (\emph{min.count})}
The impact of excluding words that are less frequent in response to variation of the $min.count$ parameter is summarized in Table \ref{tbl:min}. This parameter shows a different impact compared to the other parameters. While, ignoring more words has a positive effect in synonymy detection, improvement halts at $min.count= 3$ for  \emph{antonymy} and  \emph{alternative form} relations.

\begin{table}[t]
    \centering
    \scalebox{1}{
     \begin{tabular}{p{2cm}|P{.6cm} P{.6cm} P{.6cm}|P{.6cm} P{.6cm} P{.6cm}|P{.6cm} P{.6cm} P{.6cm} }
      $min.count$&\multicolumn{3}{P{1cm}|}{Synonymy }& \multicolumn{3}{P{1cm}}{Antonymy }   &\multicolumn{3}{|P{1.5cm}}{Alt. form }\\
      \cmidrule{2-10}
       &$A$ & $R$ & $P$ & $A$ &$R$ & $P$ & $A$ &$R$ & $P$ \\
       \midrule
        2 & 12.4&9.9 & 2.7  &54.4 &48.4  & 10.9  & 13.0 & 11.8  &  2.7   \\
        3 &  12.6&10.1  & 2.7  & {\bf 56.1}& {\bf 50.0}  & {\bf 11.2} & {\bf13.2}&{\bf 12.0} & {\bf 2.8} \\
        5 & 12.7&10.2  & 2.7 &55.4& 49.2 &  11.1 & 12.9&11.0    & 2.6    \\
        10 &{\bf 13.1}& {\bf 10.4 } & {\bf 2.8} &54.7& 48.3   & 10.9 & 13.0  &11.8 &   2.7 \\
    \end{tabular}}
     \caption{Evaluation results for different value for frequency cut off  (default=5).}
    \label{tbl:min}
\end{table}

Since the context window size (\emph{win}), negative sampling (\emph{neg}) and frequency cut off  (\emph{min.count}) parameters showed inconsistent results among the relations, we selected the CBOW model with vector size (\emph{dim}) equal to 400 and we fixed the other parameters to their defaults i.e. $win=5$, $min.count=5$ and $neg=5$. This configuration, hereinafter referred to as \emph{\textsc{OilGas}.d400}, showed the overall best performance during the tuning experiments.

\subsection{Comparative evaluation}
\label{comparative}
In order to compare the domain-specific embeddings with general domain embeddings,
we select two widely used pre-trained embedding models: \emph{Wiki+Giga} \footnote{\scriptsize\url{https://nlp.stanford.edu/projects/glove/}} and \emph{GoogleNews} \footnote{\scriptsize \url{https://code.google.com/archive/p/word2vec/}} to see how they perform in our evaluation benchmark. The pre-trained models were chosen to have similar settings to our models. The input data in the \emph{Wiki+Giga} has been tokenized and lowercased with the Stanford tokenizer, whereas the \emph{GoogleNews} model is trained on a part of the Google News dataset and it contains both words and phrases. The phrases are obtained using the same approach, as described in Section \ref{w2v-chapter2}. The words are not lemmatized in both models, and GoogleNews also contains capitalized words. 

The results of the comparative evaluation of the domain-specific and pre-trained models are summarized in Table \ref{tbl:compar1-2}. Since the words in the vocabularies of both pre-trained models are not in lemma form, we consider the surface form of terms for the evaluation.
We also report the proportion of query terms that are covered by the vocabulary of each model as coverage. We find that despite the large input and vocabulary size in both \emph{GoogleNews} and \emph{Wiki+Giga} models, they have less coverage than the domain-specific model. We further observe that (see Tables \ref{tbl:compar1-1}and  \ref{tbl:compar1-2}) despite the considerably smaller training data set, the \textsc{OilGas}.d400 performs better across all the tasks.

\begin{table}
\centering
\begin{tabular}{p{2.5cm}|p{2.8cm}|P{1cm}}
     \head{Model} &\head{Coverage} &\head{dim} \\
       \midrule
        Google News & 26\%  (100B, 3M) & 300\\
        Wiki+Giga & 23\%  (6B, 400K) & 300  \\
      \textsc{OilGas}.d400 &31\% (108M, 330K) &400  \\
        \midrule
        \midrule
        enwiki & 29\%  (1.8B, 2M)&400 \\
       enwiki+\textsc{OilGas}& 31\%  (1.9B, 2.3M)&400  \\
    \end{tabular}
    \caption{General domain and domain-specific embedding models.}
    \label{tbl:compar1-1}
\end{table}

\begin{table}
\centering
 
\begin{tabular}{p{2.5cm}|P{.6cm} P{.5cm} P{.6cm}|P{.6cm} P{.6cm} P{.6cm}|P{.6cm} P{.6cm} P{0.6cm}}
     \head{Model}  &\multicolumn{3}{P{.8cm}|}{Synonymy}& \multicolumn{3}{P{.9cm}}{Antonymy}   &\multicolumn{3}{|P{1.8cm}}{Alt. form}\\
      \cmidrule{2-10}
       &  $A$ & $R$ & $P$ & $A$ & $R$ & $P$ & $A$ & $R$ & $P$ \\
       \midrule
Google News  & 9.0 & 7.0 & 1.8 & 51.2 & 37.0 & 8.1 &4.1& 1.6 & 0.4\\

Wiki+Giga &  4.0 & 3.2 & 0.8 & 40.4 & 43.8 & 10.2  & 1.8 & 3.7 & 0.8  \\
        
\textsc{OilGas}.d400 & \head{15.7} & \head{13.1} & \head{3.5} &\head{57.1} &\head{50.8} & \head{11.4} &\head{13.6}&\head{11.7} & \head{2.7}  \\
\midrule
\midrule
enwiki & 8.2& 6.7  &  1.8 & 39.1 &33.3 & 7.5  & 8.3 & 8.1  & 1.9 \\

enwiki+\textsc{OilGas}& 11.1 & 7.8 &  2.1 &55.3& 47.7  & 10.7   & 8.6 & 8.9  & 2.0  \\
    \end{tabular}
    \caption{Results from the intrinsic comparative evaluation of general domain and domain-specific embedding models.}
    \label{tbl:compar1-2}
\end{table}
It is clear that, this comparison is somewhat unfair due to differences in pre-processing and hyperparameter tuning. Therefore, to investigate the impact of these differences, we apply the same pre-processing steps and hyperparameters to train the CBOW model over the English Wikipedia dump (20 September 2016), here dubbed \emph{enwiki}. Furthermore, we conduct a similar experiment with a data set consisting of both the general and domain-specific corpora (\emph{enwiki+\textsc{OilGas}}). However, these approaches do not show further improvements in our evaluation benchmark, as reported in Table \ref{tbl:compar1-2}. Surprisingly, the mixing of Wikipedia and \textsc{OilGas} does not increase the coverage rate. It can be attributed to the fact that the phrase extraction method (Section \ref{w2v-chapter2}) is not able to capture the missing multi-word expressions. In many cases in the mixed corpus (enwiki+\textsc{OilGas}) the relative increase in the frequency of tokens individually is higher than the relative increase of co-occurring tokens (e.g., the relative increase of the word "source" and the word "rock" in the \emph{enwiki+\textsc{OilGas}} is larger than the relative increase of the word "source rock" compared to the \emph{\textsc{OilGas}} corpus).

\section{Manual Analysis} \label{Serror}
The results in Section \ref{comparative} show that the domain-specific model provides better results than general domain models for a domain-specific benchmark. However, we also observe that performance is low for all three tasks, in particular for the synonymy detection task.
In this section, we explore the reasons behind these low scores and gain insight into the domain-specific model predictions, in particular the synonymy detection, through an in-depth error analysis.

As noted above, the primary cause of low performance is due to out of vocabulary (OOV) terms in the query inventory. As shown in Table \ref{tbl:compar1-1} the model vocabulary contains only 31\% of the evaluation dataset.
We find that the majority of terms that participate in synonymy relations are not included in the word embeddings model, this is in particular the case for multiword items.
The majority of these terms either do not occur or have a frequency lower than the cut off threshold in the domain dataset.
Excluding the OOV terms from the evaluation tasks has some impact on the model performance for synonymy detection, recall (R) is 29\%, and precision (P) is 6.5\%. Still, these scores are low, we therefore examine the model predictions closer.

We randomly choose 100 terms from the reference inventory, which are also in the model vocabulary, and we manually categorize their 10-most-similar words provided in the word embeddings. The manual analysis was performed by two domain experts as well as the author.

In this section, we are inspired by the work of \cite{Leeuwenberg2016}, where the authors categorized the result of embeddings for a synonym extraction task in the following categories (The categories with ${*}$ are added by us).
\begin{itemize}
\setlength{\itemsep}{2pt}
 \item {\bf Spelling Variant}: The prediction is an abbreviation or there are differences 
between prediction and target word because of hyphenation.
\item {\bf  Alternative or derived form}: The
prediction is an alternative or derived form of the target word.
\item {\bf Reference-Synonyms}:  The prediction is a synonym of the target word in the oilfield glossary.
\item {\bf Human-judged Synonyms}: The prediction is judged as true by the expert (but is not present in the glossary). 
\item {\bf Antonyms$*$}: The prediction is an antonym of a target term. 
\item {\bf Hypernyms}: The prediction is a more general category of the target term.
\item {\bf Hyponyms}: The prediction is a more specific type of the target term.  
\item {\bf Co-Hyponyms}: The prediction and target term share a common hypernym. 
\item {\bf Holonyms$*$} The prediction denotes a whole whose part is denoted by the target term. 
\item {\bf Meronyms$*$}: The prediction is a part of the target term. 
\item {\bf Related}: The prediction is semantically related to target.
\item {\bf Unrelated/Unknown}: The prediction and target terms are semantically unrelated.
\end{itemize}

\begin{figure}
\centering
\begin{minipage}[t]{1\textwidth}
    \includegraphics[width=1\textwidth]{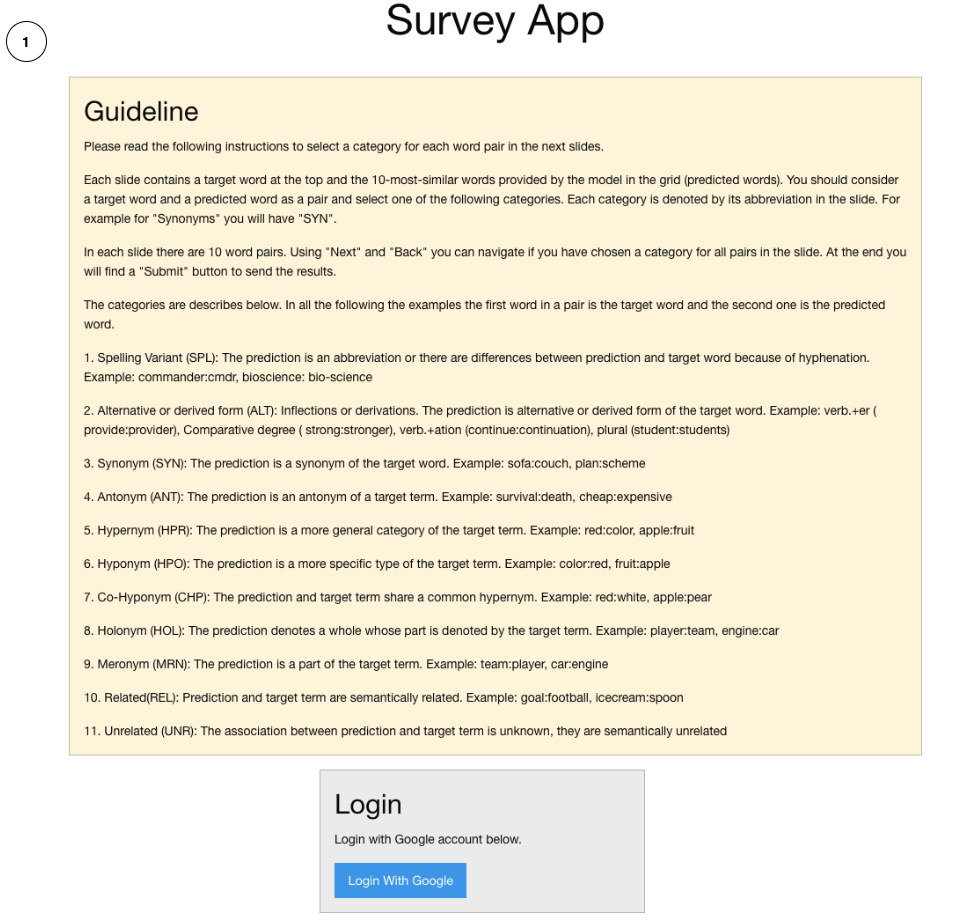}
\caption{Annotation interface (1): login and annotation guideline.}
\label{fig:annot1}
\end{minipage}
 \begin{minipage}[t]{.6\textwidth}
    \includegraphics[width=1\textwidth]{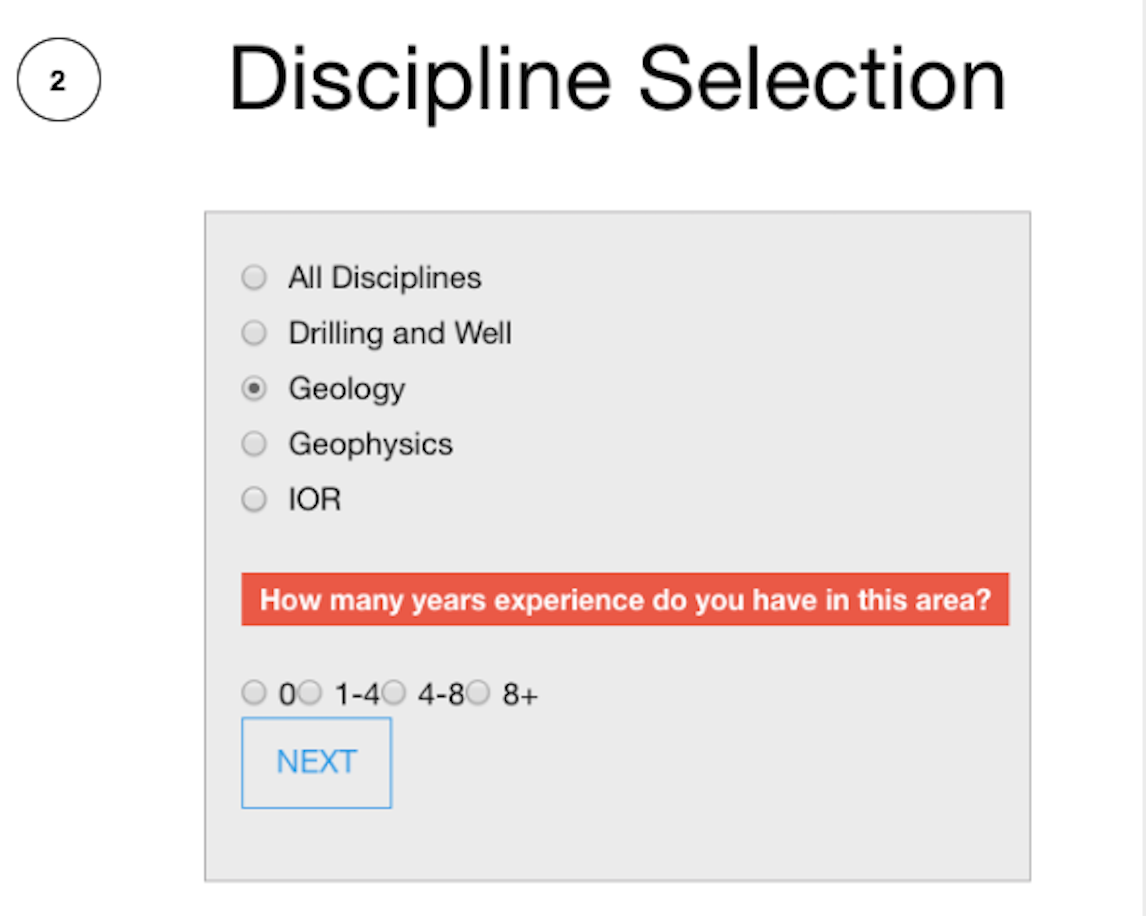}
    \caption{Annotation interface (2): disciplines  and years of experience.}
    \label{fig:annot2}
\end{minipage}
\end{figure}
\begin{sidewaysfigure}
\centering
\includegraphics[width=1\textwidth]{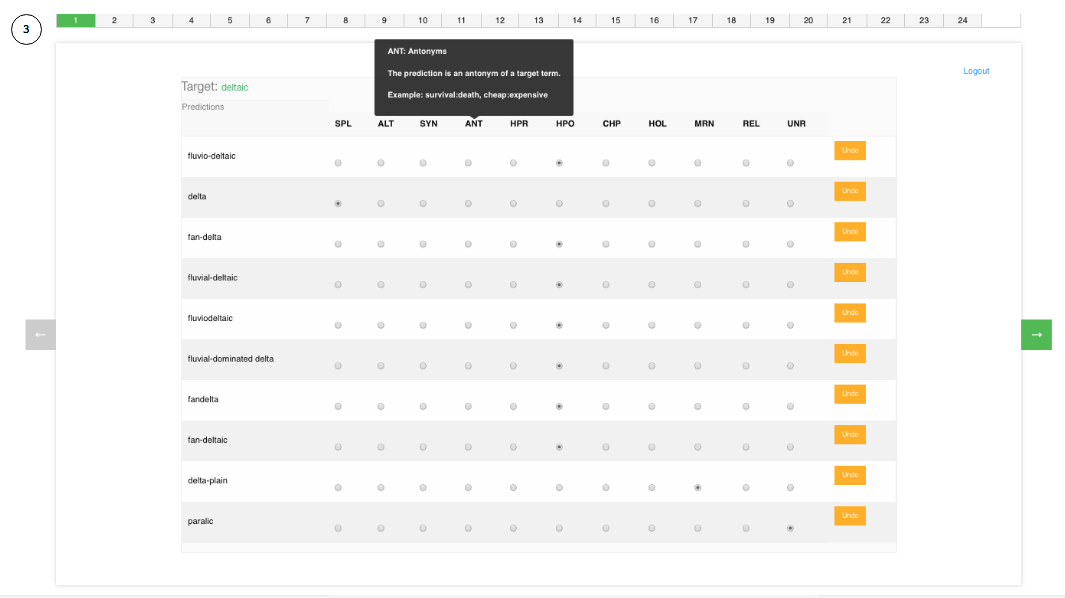}
    \caption{Annotation interface (3): target and prediction word pairs and relations.}
    \label{fig:annot3}
\end{sidewaysfigure}

\subsection{Annotation tools}

For manual analysis, the domain experts are asked to categorize the randomly selected terms (Section \ref{Serror}) and their 10-most-similar words provided by the domain-specific embedding model.
We implement a live web application \footnote{\url{http://obscure-tor-63439.herokuapp.com/}} using \emph{React.js} and the \emph{Firebase} platform.
Figures \ref{fig:annot1},  \ref{fig:annot2} and \ref{fig:annot3} depict screenshots of the annotation environment. The annotator can log in to the annotation environment using the first screen (Figure \ref{fig:annot1}). Then, the annotator selects the discipline that he/she wants to annotate (Figure \ref{fig:annot2}).  In the annotation interface (Figure \ref{fig:annot3}), we provide the target words according to their discipline in the Schlumberger oilfield glossary.
Each annotation page contains a target word at the top and the 10-most-similar words provided by the model in the grid (predicted words). The annotators should consider a target word and a predicted word as a pair and select one of the target categories. Each category is denoted by its abbreviation in the page (e.g., "Synonyms" as "SYN"). In each annotation page, there are 10-word pairs. Using "Next" and "Back", the annotators can navigate through the pages. In the end, they will find a "Submit" button to send the results.
Using this tool, the annotators assign each pair, i.e., target and predicted word to one of the categories. We also allow annotators to leave empty the assignment if they do not have sufficient knowledge about the relation between the terms. In order to encourage inter-annotator consistency, we provide a general annotation guideline (Figure \ref{fig:annot1}).
\subsection{Results and discussion}
\begin{figure}[t]
\centering
\includegraphics[width=1.1\textwidth]{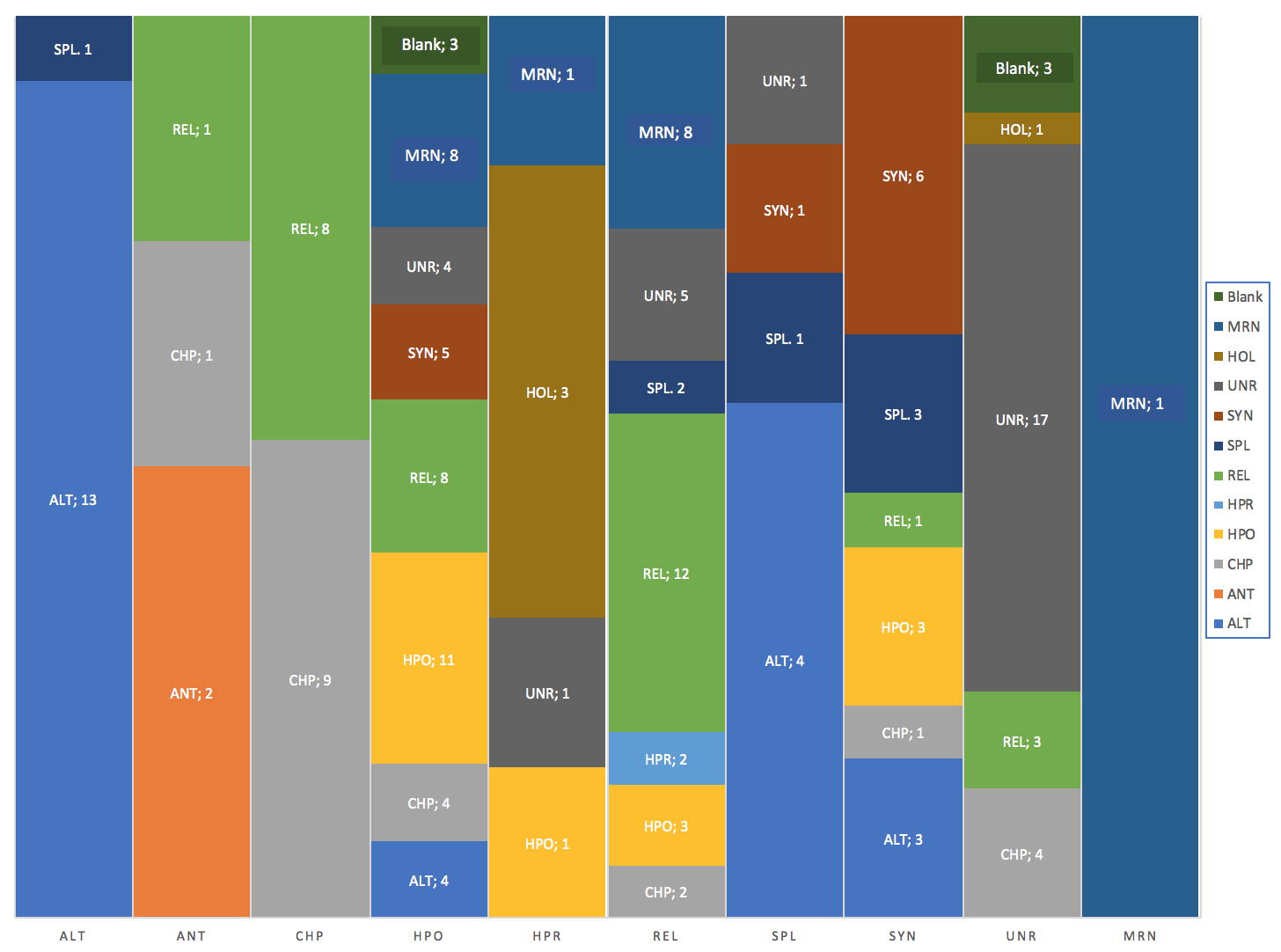}
\caption[Agreements/Disagreements categories among annotators.]{Agreements/Disagreements categories among annotators. Each column shows the number of pairs in which there is an agreement among the annotators. It also shows the category that has been assigned by one of the annotators in case of disagreement. (e.g., in alternative or derived form (ALT), all annotators agree on $13$ instances , while there is a disagreement in one instance and one of the annotators selects the spelling variant (SPL).}
\label{fig:confusion}
\end{figure}
We receive annotation results for 240-word pairs (24 target words) from the domain experts in the geology discipline. We report the inter-annotator agreement as a proportion of agreement. For this task, we do not use the Kappa statistic as the chance of agreement for the task is very low due to implicit similarity among the categories. We observe that in 175 cases, we have a majority agreement among three annotators. Moreover, all the annotators agree in 72 cases. By this observation, the inter-annotator agreement for the task is $\frac{175}{240}=0.729$.

To explore the disagreements of annotators in more detail, we extract the instances for which the two annotators determine the same category, whereas the third annotator selects another category, and we report it as a confusion matrix.

Figure \ref{fig:confusion} shows the resulting confusion matrix. In general, we find that the annotators disagree in terms of similar categories. For example the \emph{hyponyms} are annotated as \emph{meronyms}, \emph{related}, \emph{synonyms} and \emph{co-hyponyms}. The categories with the highest number of disagreements are \emph{hyponyms}, \emph{hypernyms}, \emph{related} and \emph{synonyms}.
The annotators agree mostly on \emph{alternative or derived form} and \emph{unrelated} categories. For further analysis, we consider the majority vote among the annotators as a final category for the word pairs.

Table \ref{tbl:analysis} shows the result of this analysis. We count the number of each category that has been considered as an error in the synonym extraction task. It means that a word is predicted as 10-most-similar words (i.e., $1^{st}$:$10^{th}$) by the model and is considered as a false positive. However, the prediction and target word are assigned to a specific category during the manual annotation process. 

In general, the result of this analysis shows that the model predictions are semantically meaningful in a majority of cases, and all categories except the \emph{Unrelated/Unknown} represent some type of morphosyntactic or semantic relation between terms. Less than 20\% of the errors are assigned to the \emph{Unrelated/Unknown} category. This reveals that if we consider the count of \emph{human-judged synonyms} as true positives, the actual scores for precision and recall will be considerably higher than those reported in the evaluation section. Moreover, the embeddings model proposes more synonyms that are not in the reference, even though the reference is provided by the manual procedure. The most frequent error type falls in the \emph{related} category.

The \emph{hyponym} and \emph{co-hyponym} relations are another frequent error type that were also reported in previous studies \parencite{vanderPlas:2006,Leeuwenberg2016}. 
The morphosyntactic type of relations such as \emph{Alternative forms}, \emph{spelling variant} cover another type of errors. The error analysis further reveals several meaningful relation types such as \emph{Hypernyms}, \emph{Meronyms}, and \emph{Holonyms} that are useful in many downstream  applications.
\begin{table}[t]
\centering
\scalebox{.85}{
\begin{tabular}{@{}llr@{}}
Category  & Example [target$\rightarrow$ prediction]& $1^{st}$:$10^{th}$(\%) \\ \midrule
1. Spelling Variant   &  borehole $\rightarrow$ bore-hole  & 2.4\\
2. Alternative or derived form   & acidizing$\rightarrow$ acidization  & 3.2 \\
3. Reference-Synonyms & filter cake $\rightarrow$ mud cake  & 2.8 \\
4. Human-judged Synonyms  & seismometer $\rightarrow$ seismograph  & 8.4\\
5. Antonyms  &  transgressive $\rightarrow$ regressive 	& 0.9\\
6. Hypernyms   &  acidizing $\rightarrow$ stimulation & 1.3 \\
7. Hyponyms  & EOR $\rightarrow$ In-situ combustion & 9.3  \\
8. Co-Hyponyms  & EOR $\rightarrow$ MEOR  & 13.1\\
9. Holonyms  & shoe$\rightarrow$ wellbore  	& 1.1  \\
10. Meronyms  & rig $\rightarrow$ wellhead   & 2.8 \\
11. Related & Kirchhoff migration $\rightarrow$ NMO correlation  & 35.2  \\
12. Unrelated/Unknown  & backflow $\rightarrow$ sediment-laden  & 19.5 
\end{tabular}
}
\caption{Manual analysis results for the 10-most-similar words.}
\label{tbl:analysis}
\end{table}

\section{Embedding Enrichment Using a Knowledge Resource} \label{retrofit}
Even though the word embeddings capture important semantic relations in the domain, the first experiment shows that the domain-specific technical vocabulary has many elements that are generally disregarded by the distributional representation techniques. In other words,
they still fall short of providing domain-specific representations for many terms. These approaches rely on the statistics derived from textual input; therefore, they are incapable of providing representations for words that are not seen frequently in the training process. Furthermore, they do not include the valuable information that is accommodated in domain knowledge resources such as semantic lexicons and glossaries. In this section, we address these issues by applying a set of techniques that exploit prior domain knowledge in enhancing the embedding models and induce representations for OOV terms. We then go on to evaluate the impact of the refinement methods over an unseen terminological resource.

\subsection{Related Work}
Although word embedding techniques have drawn significant interest in the field, they are not well equipped to deal with unseen and infrequent words, nor do they consider word relations found in knowledge resources. Improving the quality of embedding models has been an active area of research for the past few years. Based on the way they view the problem, these techniques can be classified into two main branches: (1) Improving word vectors using lexical resources, (2) Learning representations for rare words.

In the first line of techniques, researchers work on leveraging semantic knowledge resources such as WordNet \parencite{Miller:1995:WLD:219717.219748}, PPPD \parencite{ganitkevitch2013paraphrase}, and FrameNet \parencite{Baker:1998:BFP:980451.980860} as a relational semantics to improve the outcome of distributional semantics. Such semantic knowledge can provide the sense of the word, its relation with other words, such as synonyms, antonyms, hypernymy, and meronymy. The question is how to design new distributional semantic algorithms to leverage relational semantics and generate high-quality word embeddings given the availability of morphological, syntactic, and semantic knowledge. In \cite{yu2014improving}, and \cite{DBLP:journals/corr/FriedD14}, it was shown that the quality of word vectors could be improved by using semantic knowledge from lexicons.  In both works, they propose a new training objective that incorporates the word2vec language model objective and prior knowledge from semantic resources. These models use constraints among words as a regularization term on the learning objective during training. However, the proposed models are built on a specific distributional semantic technique i.e.,  word2vec.  Similarly, \cite{faruqui:2014:NIPS-DLRLW} and \cite{DBLP:journals/corr/PilehvarC16} proposed a refinement method as a post-processing step which exploits knowledge from the semantic network to apply to existing pre-trained embeddings. These approaches are general in that they can be readily applied to any set of word representations and any semantic network and they are not limited to particular methods for constructing vectors.

The Zipfian distribution \parencite{zipf1972human} is a characteristic of words in natural language, where some words are frequent, but most are rare. Learning representations for words in the "long tail" of this distribution is challenging for word embedding techniques since their learning methods require many occurrences of each word to generalize well. The second branch of related work includes methods that produce representations for words that were not encountered frequently during the training of the embeddings models. \cite{Botha:2014:CMW:3044805.3045104} and \cite{DBLP:conf/naacl/SoricutO15} have proposed methods that incorporate morphological information into word representations. They focus on morphologically complex rare words and tried to derive representations of words from morphemes using different composition functions. In another framework for the training of word embeddings, known as \emph{fasttext}, \cite{DBLP:journals/tacl/BojanowskiGJM17} improved the representation of words by taking into account subword information. The proposed approach incorporates character n-grams into the skip-gram model. It can construct a vector for a word from its character n-grams, even if a word does not appear in the training corpus. However, these techniques are incapable of deriving representations for words for which no sub-word information
might be available in the training corpus, such as infrequent domain-specific terms. To expand the vocabulary of the embedding model not only for morphological variation but also for unseen and infrequent domain-specific words, the approach proposed by \cite{E17-2062} exploits the knowledge encoded in lexical resources and induce vector representation for terms which either have low frequencies or are non-existent in domain corpora.
Here, we employ the techniques of \cite{E17-2062} and \cite{faruqui:2014:NIPS-DLRLW}, since they can be applied to vectors obtained
from any word vector training method as a post-processing step. In the following sections, these approaches are described further, and their impacts in our study are subsequently evaluated. 
\subsection{Embeddings for infrequent terms} \label{emb.rare}
To induce embeddings for rare and unseen words, \cite{E17-2062} recommend a technique that expands the vocabulary of pre-trained embedding models. The technique leverages knowledge encoded in external lexical resources that provides better coverage of rare words or comprises domain-specific terminologies. It assumes that there is a pre-trained word embedding model $W$ and a lexical resource $S=(V, E)$ in a graph structure, where $V$ is a set of nodes that correspond to words and $E$ is a set of semantic relations among the words. To produce an embedding for an infrequent word $w_r$ that does not exist in the vocabulary of $W$, but is covered by $S$, the following phases are implemented:
\begin{figure}
\centering
\includegraphics[width=1\textwidth]{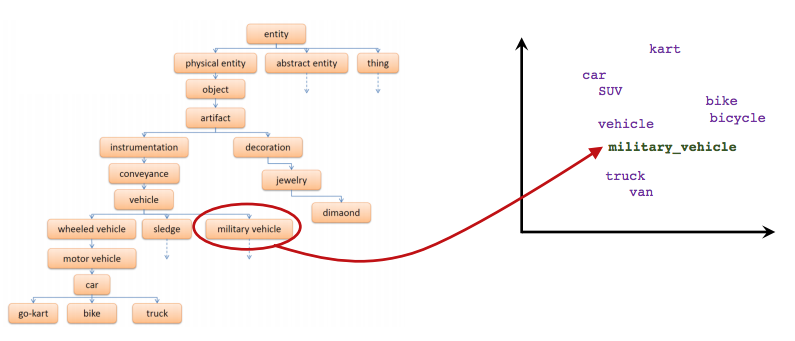}
\caption{Leveraging structure in external lexical resources to produce semantic landmark \parencite{E17-2062}.}
\label{fig:induce-01}
\end{figure} 
\paragraph{I. Word Semantic Landmarks Extraction}
A set of landmarks for word $w_r$ are the most semantically similar words, which can be extracted from $S$. As shown in Figure \ref{fig:induce-01}, it is achieved by viewing $S$ as a semantic network and analyzing its structure. 
The semantic landmarks for word $w_r$ are extracted by using the Personalized PageRank \parencite{Haveliwala:2003:TPC:1435677.858989} algorithm, here dubbed PPR. The PPR provides promising results in many NLP tasks such as Word Sense Disambiguation \parencite{Agirre:2009:PPW:1609067.1609070} , Named entity linking \parencite{10.1007/978-3-319-41754-7_7} and Word sense similarity \parencite{Pilehvar:2015:ST:2827893.2828161}. The PPR is a modified version of PageRank \parencite{Brin:1998:ALH:297810.297827} algorithm. PageRank is a method for ranking the nodes in a graph according to their relative structural importance. The main idea of PageRank is that whenever a link from node $i$ to node $j$ exists in a graph, a vote from node $i$ to node $j$ is produced, and hence the rank of node $j$ increases. Besides, the strength of the vote from $i$ to $j$ also depends on the rank of node $i$: the more important node $i$ is, the more strength its votes will have. Moreover, PageRank can also be viewed as the result of a random walk process, where the final rank of node $i$ represents the probability of a random walk over the graph ending in node $i$, for sufficiently many steps.

Let $S$ be a graph derived from external lexical resources, with  $n$  nodes  $\{w_1,\dots,w_n\}$ and $d_i$ be the outdegree of node $i$; let $P_{n \times n}$ be a transition probability matrix, where $P_{i,j}$ denotes the probability of shifting from node $i$ to node $j$. The $P_{i,j}$ is equal to the inverse of $d_i$ if there is a semantic link from $i$ to $j$ and zero otherwise. Therefore, to calculate PageRank vector $x^T$ over $S$ for each node $i$ , the power method can be used as follows: 
\begin{equation}
\label{emb-eq:induce-01}
x^{(t)T}=\alpha x^{(t-1)T}P+(1-\alpha)v
\end{equation}
\noindent In equation \ref{emb-eq:induce-01}, $v$ is the $n \times 1$ column vector in which the prior importance of each node is assigned to the cell corresponding to each node and $\alpha$ is the damping factor, a scalar value between 0 and 1. In the traditional PageRank, the vector $v$ is a stochastic normalized vector whose element values are all $\frac{1}{n}$ , thus assigning equal probabilities to all nodes in the graph in case of random jumps. However, in PPR, a modified version of $v$ is used. PageRank is calculated by applying an iterative algorithm that computes the equation successively until convergence below a given threshold is achieved, or, more typically, until a fixed number of iterations are executed. Once $x^{T}$ is calculated for a word $i$ in the semantic lexicon, we can obtain a list of most similar words to the $w_i$, i.e., semantic landmarks for word $i$, by sorting the $x^{T}$ according to their probabilities. 
\paragraph{II. Embedding Induction:}
Let $L_{i}$ be a sorted list of semantic landmarks for word $i$ and $q(x)$ be an embedding of word $x$ in the space of $W$. The induced embedding for $w_i$ in the same semantic space can be provided using the following equation:
\begin{equation}
\label{emb-eq:induce-02}
    \hat{q}(w_{i})=\theta q(w_i)+ \frac{1}{|L_i|} \mathlarger{\mathlarger{\sum}}_{j=1}^{|L_i|} e^{-i}q(l_{j,i})
\end{equation}
\noindent where $l_{j,i}$ is the $j^{th}$ word in $L_i$ and $q(w_i)$ is the observed embedding of word $w_i$ in $W$ . Here the intuition is to calculate a new embedding for $w_i$ by the weighted average of its semantic landmarks. The exponential weighting provides more importance to the top words in the semantic landmarks since these are more representative for word $i$. The parameter $\theta$ adjusts the contribution of the initial embeddings $q(w_i)$ in the final embeddings. To induce embeddings for unseen words, $\theta$ is set to zero.

In another work, \cite{faruqui:2014:NIPS-DLRLW} proposed the retrofitting method as a post-processing step to apply to existing pre-trained embeddings. The goal is to refine word vector representations to capture relatedness suggested by semantic lexicons while preserving their similarity to the corresponding embeddings. The objective of the retrofitting method is to minimize the following:
\begin{equation}
    \Psi(Q)=\mathlarger{\mathlarger{\sum}}_{i=1}^{n}\Big[\alpha_{i}\| q_{i}-\hat{q}_i\|^{2} + \sum_{(i,j)\in E}\beta_{ij}\|q_i-q_j\|^2\Big]
\end{equation}
\noindent where $\hat{q} \in \hat{Q}$ is the observed vector representation for each term in the semantic lexicon and $q \in Q$ is the corresponding retrofitted vector. 
\begin{figure}[t]
\centering
\includegraphics[width=.7\textwidth]{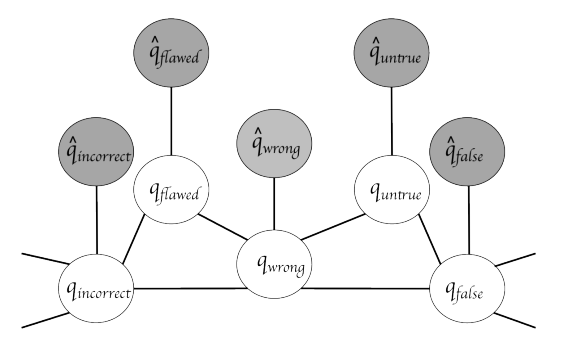}
\caption[Graph with links between related words
showing the observed (grey) and the inferred (white)
word vectors.]{Graph with links between related words
showing the observed (grey) and the inferred (white)
word vectors. \parencite{faruqui:2014:NIPS-DLRLW}.}
\label{fig:retro-01}
\end{figure} 

The method aims to learn ${Q}$ such that the $q$ is close to its counterparts in $\hat{Q}$ (i.e., $\hat{q}$) and to adjacent nodes in the semantic lexicon under a distance metric. Figure \ref{fig:retro-01} depicts an example graph with such connections; white nodes are labeled with the retrofitted vectors $q$, shaded
nodes are labeled with the observed ones $\hat{q}$.
$E$ is the set of relations among the terms in the semantic lexicon. $\alpha$ and $\beta$ correspond to the relative weights of the relation type. Since $\Psi$ is convex in $Q$, an efficient iterative updating method is used to solve the objective function. Retrofitted embeddings $Q$ are initialized to be equal to the observed ones $\hat{Q}$. Then by taking the first derivative of $\Psi$ with respect to $q_{i}$ the following online update is used for ten iterations to reach convergence:
\begin{equation} \label{emb-eq:modified}
q_i=\frac{\sum_{j:(i,j)\in E} \beta_{ij}q_j+\alpha_i\hat{q}_i}{\sum_{j:(i,j)\in E} \beta_{ij}+\alpha_i}
\end{equation}
\noindent The formula in the Eq. \ref{emb-eq:modified} computes a new embedding for a term $i$, which is in the pre-trained model and has relations of interest in the semantic lexicon, whereas its neighbors should be part of the pre-trained model. To provide an embedding for OOV words, we extend $\hat{Q}$ in each iteration by adding the terms that are in the semantic lexicon and connect to the terms that are in $\hat{Q}$ via relations of interest. Since there is no initial vector for these type of words in the observed model, $\alpha$ is set to zero, and the online update formula for the OOV terms will be as follows:
\begin{equation} \label{emb-eq:2}
q_{i}=\frac{\sum_{j:(i,j)\in E} \beta_{ij}q_j} {\sum_{j:(i,j)\in E}\beta_{ij}}
\end{equation}

\subsection{Additional domain-specific query inventory}
\begin{figure}[t]
\centering
\includegraphics[width=.9\textwidth]{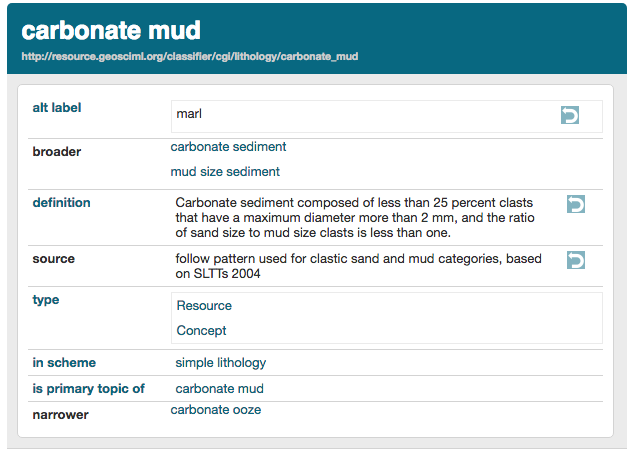}
\caption[Term structure in the Geoscience Vocabularies (GeoSci) data set.]{Term structure in the Geoscience Vocabularies (GeoSci) data set.}
\label{fig:geosci}
\end{figure}

We use another domain-related glossary to perform a quantitative comparison of domain-specific word embeddings before and after the inducing and retrofitting process. We create a test query inventory using the same approach as explained in Section \ref{slb} over the Geoscience Vocabularies data set \footnote{\url{http://resource.geosciml.org/}}, here dubbed \emph{GeoSci}. GeoSci was developed by the IUGS (CGI Commission for the Management and Application of Geoscience Information). GeoSci covers the domain of geology and describes geological features, geological time, mineral occurrences, and mining-related features. 
Figure \ref{fig:geosci} shows the structure of the term \emph{carbonate mud} in this domain-specific resource.
Each term is defined as a resource under \emph{geosciml} name-space with a specific \emph{URI\footnote{Uniform Resource Identifier}}. 
Resources are linked to other resources with syntactically and semantically aligned relations such as \emph{Abbreviated label}, \emph{Synonym} \footnote{GeoSic vocabulary specifies this relation as \emph{Alternative label}.}, \emph{Broader} and \emph{Narrower}.

We construct a test query inventory by extracting all terms and their inter-glossary relations from the RDF files. Table  \ref{tbl:geosci} shows the distributions of terms in the glossary with respect to their n-gram and participated relations. The test set consists of \emph{1,753} terms. It contains \emph{196} synonym pairs, \emph{1,639} broader pairs, \emph{1,584} narrower pairs and \emph{986} abbreviated label pairs. Like the slb glossary, the majority of terms are multi-word units (63\%).

\begin{table}[t]
    \centering
    \scalebox{.9}{
    \begin{tabular}{lccccc} 
     \head {n-grams} & \head {\#} & \head{Synonymy} & \head{Broader} & \head{Narrower} & \head {Abbr. label}  \\
     \midrule
       unigram & 651 & 110 & 588 & 387 & 730  \\
   bigram & 598 & 45 & 550 &737 &201   \\
        trigram & 328 & 28 & 326 &  305 &  47\\
        $>$3 & 176 & 13 & 175 & 155 & 8  \\
         \midrule
          All &  1753 &	196 & 1639 &  1584 &  986	  \\
    \end{tabular}}
    \caption[N-grams and Relations in the Geoscience Vocabularies (GeoSci) data set.]{N-grams and Relations in the Geoscience Vocabularies (GeoSci) data set.}
   \label{tbl:geosci}
\end{table}
\subsection{Experiments and Evaluation} \label{seml}
We use the structure of the \emph{slb} glossary as prior domain knowledge to enrich the \textsc{OilGas}.d400 embeddings model that was selected as the best embedding model following our experiments in Section \ref{exp1}.
To enrich the pre-trained embeddings, we employ the two techniques that are described in Section \ref{emb.rare} as follows:

\paragraph{Inducing embeddings for unseen words:} We create the semantic graph $G=(V,E)$ using the $slb$ glossary, where $V$ is the set of $slb$ terms and $E$ is the set of edges that denote semantic relationships i.e, \emph{synonymy}, \emph{antonymy} and \emph{alternative form} among terms in $V$. For each term in $V$, we extract semantic landmarks using the PPR algorithm by using Eq. \ref{emb-eq:induce-01}. We set the damping factor $\alpha$ to its default value ($0.85$), and the personalize vector $v$ is a one-hot initialization vector in which $1.0$ is assigned to the cell corresponding to the term. We choose top $n=10$ most similar words according to their probability in $x^{(t)}$ as a semantic landmark for each word. We induce new embeddings for observed terms in the \textsc{OilGas}.d400 embedding model by exploiting the Eq. \ref{emb-eq:induce-02} and setting $\theta$ to $1.0$ and for OOV to zero (".induced\#10").
\paragraph{Retrofitting Word Vectors to Semantic Lexicons} Experiments in \cite{faruqui:2014:NIPS-DLRLW} showed that including all semantic relations in the retrofitting process has a better impact than having only one of them. We, therefore, consider connections of a word to its synonyms, alternative forms, and antonyms. Moreover, similar to the origin, all $\alpha_i$ are set to 1 and $\beta_{ij}$ to be $degree(i)^{-1}$. The Eq. \ref{emb-eq:2} is used to retrofit the \textsc{OilGas}.d400 model by employing the structure of the semantic lexicon (".retrofitted"). To induce word vectors for OOV terms, we carry out the retrofitting process with Eq. \ref{emb-eq:2} (".retrofitted+OOV").

Table \ref{tbl:test.rtft} shows the performance of the model in the test dataset as well as the induced and retrofitted models with two different configurations. We observe that the inducing algorithm has negative impacts on the \emph{synonymy} and \emph{abbreviated label} relation.
It seems that with a small semantic lexicon, the inducing algorithm provides noisy semantic landmarks for each term, which leads the induced embeddings vector to be close to non-related terms. However,
the retrofitting process provides an improvement in the \emph{synonymy} relation. The improvement is highest when we consider the adapted version (retrofitted+OOV). Interestingly the inducing and retrofitted models have no impact in the \emph{narrower} and \emph{broader} relationships. This can be attributed to the fact that the employed semantic lexicon does not include these kinds of associations to lead the retrofitting process.
In the \emph{abbreviated label} relation, there is a slightly negative effect when we apply the original retrofitting process.
Expectedly, the applied retrofitting process encourages the terms in semantic lexicons to have similar vector representations with respect to their semantic and morphosyntactic relations (i.e., \emph{synonymy},  \emph{antonymy} and \emph{Alternative forms}). The improvement is biggest for retrofitted models when they are assessed in the query inventory generated from the input semantic lexicons. Covering OOV terms by using Eq.\ref{emb-eq:2},
empowers the model to provide high performance in each relation benchmark (Table \ref{tbl:train.rtft}).
\begin{table}[t]
    \centering
    \scalebox{.8}{
    \begin{tabular}{c|P{.5cm} P{.5cm} P{.5cm}|P{.5cm} P{.5cm} P{.5cm}|P{.5cm} P{.5cm} P{0.5cm}|P{.5cm} P{.5cm} P{0.5cm}}
      \head{Model}&\multicolumn{3}{P{1cm}|}{Synonymy}& \multicolumn{3}{P{.8cm}}{Narrower}   &\multicolumn{3}{|P{1cm}}{Broader} &\multicolumn{3}{|P{1.6cm}}{Abbr. label}\\
      \cmidrule{2-13}
        & $A$ & $R$ & $P$ & $A$ & $R$ & $P$ & $A$ & $R$ & $P$ & $A$ & $R$ & $P$ \\
       \midrule
         \textsc{OilGas} &25.5 & 15.5 & 5.1 & 12.5 & 5.0 & 2.7 &  4.7 & 4.4 & 0.9 & 2.4 &  2.3 & 0.5   \\
        
        \textsc{OilGas}.induced\#10 & 11.0 & 6.7 & 2.2 &  12.5 & 5.0 & 2.7 &   4.7 & 4.4 &  0.9 & 2.0 &  2.0 & 0.4 \\
        
        \textsc{OilGas}.retrofitted & 27.4 & 16.7 & 5.5 &  12.5 & 5.0 & 2.7 &  4.7 & 4.4 &  0.9 & 2.2 &  2.2 & 0.4 \\
         \textsc{OilGas}.retrofitted+OOV &\head{30.2} & \head{18.4} & \head{6.1}   & 12.5 & 5.0 & 2.7 & 4.7 & 4.4  &0.9 & 2.4 &2.3 & 0.5 \\
    \end{tabular}
    }
    \caption{Evaluation over the GeoSci knowledge resource.}
    \label{tbl:test.rtft}
\end{table}
\begin{table}[t]
    \centering
    \scalebox{.9}{
    \begin{tabular}{c|P{.6cm} P{.6cm} P{.6cm}|P{.6cm} P{.6cm} P{.6cm}|P{.6cm} P{.6cm} P{0.6cm}}
      \head{Model}&\multicolumn{3}{P{1cm}|}{Synonymy}& \multicolumn{3}{P{1cm}}{Antonymy} &\multicolumn{3}{|P{1.5cm}}{Alt. form} \\
      \cmidrule{2-10}
        & $A$ & $R$ & $P$ & $A$ & $R$ & $P$ & $A$ & $R$ & $P$  \\
       \midrule
        \textsc{OilGas}.retrofitted & 55.0 &  47.3 & 12.4  & 87.3 &  82.2 & 18.5 & 32.6 & 29.1& 6.8   \\
         \textsc{OilGas}.retrofitted+OOV  & 94.6 &  90.7 & 25.3 & 98.5 & 97.6 & 24.1 & 96.2 & 94.7 & 21.7 \\
    \end{tabular}}
    \caption{Evaluation in the slb data set (learning data set).}
    \label{tbl:train.rtft}
\end{table}

\section{Extrinsic Evaluation} \label{ch2:extrinsic}
While the intrinsic evaluation attempts to interpret the encoding content of an embedding model in terms of lexical-semantic relations, \emph{extrinsic} evaluation investigates the contribution of an embedding model to the performance of a specific downstream task (Section \ref{r:extrinsic}). In this section, we investigate the influence of our domain-specific model in a domain related classification task.

\subsection{Classification Data Set}
The task of the exploration department in the Oil and Gas industry is to find exploitable deposits of hydrocarbons (oil or gas). Geo-scientists in the exploration department model the subsurface geography by classifying rock layers according to multiple stratigraphic hierarchies using information from a wide range of different sources. The quality of the analysis depends on the availability and the ease of access to the relevant data. Previous technical studies, reports, and surveys are crucial resources in this process.

We were granted access to a dataset of $1,348$ sentences from exploration textual documents, which are then manually labeled with various geological type properties by domain experts. Example \ref{example} shows a sentence from the data set along with its assigned set of properties.

\begin{covexample} \label{example}
\emph{Submarine fans and deltaic/estuarine facies of the San Juan Formation were deposited during the Maastrichtian regression, which gave way during the Paleocene-Eocene to black marine shales and carbonates of the Vidoño Formation and the shelfal and pro-delta shales of the Caratas Formation.}\\
\vspace{10pt}
{\small \underline{Properties:} Lithology\_RockType (L$_R$), Lithology\_Main (L$_M$), DepEnv\_Sub (D$_S$), DepEnv\_General (D$_G$)}
\end{covexample}

The resulting data set contains 1,348 sentences in which experts assigned each sentence to 7 different properties. The sentences are pre-processed using the same approach, as described in Section \ref{w2v-chapter2}. Table \ref{tbl:markups} depicts the properties and number of sentences for each. 
\begin{table}[t]
\centering
    \begin{tabular}{llr}
     \head{Property} & \head{Label}& \head{\# Sentences}\\
       \midrule
        Lithology\_Main & L$_M$& 1,193\\
        DepEnv\_Main & D$_M$& 483 \\
        Facies & F& 387 \\
         DepEnv\_Sub & D$_S$& 298\\
        Lithology\_RockType & L$_R$ & 191\\
        BasinType & B& 49\\
        DepEnv\_General& D$_G$ & 38 \\
\bottomrule
    \end{tabular}
    \caption{Classification data set.}
    \label{tbl:markups}
\end{table}
It is clear that the data set is unbalanced regarding the properties, and the downstream task is a multi-label classification task.

\subsection{Multi-label Classification Model} We use a slight variant of the \emph{Convolutional Neural Networks} (CNNs) architecture (see Section \ref{ch01:cnns} in Chapter \ref{sec:first}) that is proposed by \cite{DBLP:journals/corr/Kim14f} for sentence classification tasks. 
We keep the value of hyperparameters equal to the ones that are reported in the original work, however we update the dimension of the embedding layer according to the dimension of the domain-specific embedding model. Furthermore, since the architecture aims to assign a single label to each sentence, we update the activation function to \emph{sigmoid} at the output layer. The \emph{sigmoid} is a non-linear function, is defined as $\sigma(x)=\frac{1}{1+e^{-x}}$ and produces a probability for each of the potential properties. 
During training, these probabilities are used to compute the error, while during testing, we round each of the probabilities  to 0 or 1 depending upon a set threshold (0.5).
\begin{table}[t]
   \scalebox{.93}{
    \begin{tabular}{l|p{.6cm}|p{.6cm}|p{.6cm}|p{.6cm}|p{.6cm}|p{.6cm}|p{.5cm}}
      \head{Model}
      &{D$_S$}
      &{L$_M$} 
      &{B} 
      &{D$_G$}
      &{F}
      &{D$_M$}
      &{L$_R$}\\
      \cmidrule{2-8}
        & $F1$ &  $F1$ & $F1$ &  $F1$&  $F1$&  $F1$& $F1$ \\
       \midrule
        CNN.rand &	28.9 &		90.6 &		0.0 &	0.0 &	63.0 &	57.1 &68.2   \\
        CNN.domain  & 	51.1 &		91.4 &		23.9 &	11.3 &	71.1 &	66.3 &	65.8   \\
        CNN.multi.rand & 38.0 &	91.4 &7.3 &	5.0 &	63.9 &	58.6 &	69.9   \\
        CNN.multi.enwiki & 43.9 & 90.5 &11.3 &	0.0 & 61.1 &	61.4 &	57.8  \\
        CNN.multi.domain & 56.2 &	92.2 &		\head{33.8} &		\head{15.0} &		71.7 &	69.4 &	\head{72.5}   \\
        \midrule
        \midrule
         CNN.multi.retrofitted+OOV & 64.0 &		91.3 & 11.3 &	0.0 & 67.6 &	68.8 &	72.2  \\
         CNN.multi.domain \&retrofitted+OOV & 	\head{68.2} &		\head{92.8} & 32.0 &	9.4 &	\head{73.4} &	\head{73.5} &	71.1   \\
         CNN.multi.retrofitted+OOV\&domain & 	53.4  &	92.6 &		20.9 &		10.0 &		71.8 &	67.0 &	70.7  \\
    \end{tabular}
    }
    \caption{Results of the classification task with various configurations.}
   \label{tbl:cnn}
\end{table}

\subsection{Extrinsic evaluation experiments} \label{ch2:extrinsic-exps}
Like \cite{DBLP:journals/corr/Kim14f}, we run experiments with several variants of the model to investigate the importance of domain-specific input as follows: 
\begin{itemize}
 \item {\bf CNN.rand}: As a baseline model, all words in the embedding layer are randomly initialized and updated in the training process.
 \item {\bf CNN.domain}: the embedding layer is initialized with a domain-specific model and fine-tuned for the target task.
 \item {\bf CNN.multi.rand}: There are two embedding layers as a 'channel' in the CNN architecture. Both channels are initialized randomly, and only one of them is updated during training while the other remains static. 
 \item {\bf CNN.multi.domain}: Same as before, but the channels are initialized with domain-specific vectors.
 \item {\bf CNN.multi.enwiki:} The channels consider the general domain word vectors from section \ref{comparative} using the English Wikipedia data.
 \end{itemize}
To deal with the effects of an unbalanced dataset and guarantee that each fold in 5-fold cross-validation will have the proportion of the same classes during training and testing, we apply the stratification of multi-label data proposed by \cite{Sechidis:2011:SMD:2034161.2034172}. 
The stratification is a sampling method that takes into account the existence of disjoint labels within a dataset and provides samples where the proportion of these labels is maintained. \cite{Sechidis:2011:SMD:2034161.2034172} introduce an iterative stratification
method that distributes samples based on how desirable a given label is in each fold, tackling the problem of lack of rare label evidence in folds.

Results of the classification task with various CNN configurations are presented in the first section of Table \ref{tbl:cnn}. In general, the multi-channel model performs better than the single-channel setting. The results suggest that having a significant amount of sentences per property helps the CNN model to classify better. The baseline model does not perform well on its own. The use of the pre-trained embeddings model helps the model in property assignment. Particularly, domain-specific embeddings provide higher performance gain in the task-at-hand when it is used in both channels. We further investigate the influence of the refined word embedding models in our classification task.
\begin{itemize}
  \item {\bf CNN.multi.retrofitted+OOV:} We used the retrofitted domain embeddings including the OOV vectors for two channels. One channel is static and the other is non-static.  \item  {\bf CNN.multi.domain\&retrofitted+OOV:} First channel is initialized with original domain-specific embeddings with static mode and the second makes use of the retrofitted embeddings with a non-static mode.
 \item  {\bf CNN.multi.retrofitted+OOV\&domain:} Same as previous setting, but the channels swap their input. 
\end{itemize}

In these experiments, because of having many multi-words as OOV terms in the model, we replaced tokens in the sentences with their bigram and trigram forms if their combination occurs in the model vocabulary (e.g., \emph{fracture porosity} is replaced by \emph{fracture\_porosity} as an input unit). The experiment (second section of Table \ref{tbl:cnn}) shows that the enhanced embedding models provide better input representations for classes with a sufficient number of instances.

\section{Summary} 
This chapter contains research and experiments on the evaluation of word embedding models trained on a low-resource domain, namely the Oil and Gas domain. The first research question is whether constructing domain-specific word embeddings is beneficial even with limited input data. This question is answered by conducting intrinsic and extrinsic evaluations of both general and domain-specific embeddings. The empirical evaluation shows that even though the distributional models have low performance in domain-specific synonymy detection, an in-depth manual error analysis reveals the striking ability of the embedding models to discover other semantic relations such as (co)hyponymy, hypernymy, and relatedness. Furthermore, we observe that domain-specific trained embeddings perform better compare to the general domain embeddings trained on much larger input data.

The second research question investigates the impact of existing domain knowledge resource on enhancing the embedding models. We augment the domain-specific model by providing vector representations for infrequent and unseen technical terms using a domain knowledge resource. Experiments show the importance of dealing with rare words in an embedding model in both intrinsic and extrinsic evaluation.

To summarize, we make the following contributions in this chapter: 
\begin{enumerate}[label=(\roman*)]
\item We create a domain-specific evaluation dataset: a corpus and a query inventory for the oil and gas domain,
\item  We train and release the domain-specific embeddings for the oil and gas domain \footnote{Link to the domain-specific model: \scriptsize{\url{http://vectors.nlpl.eu/repository/11/75.zip}}.}, 
\item  We conduct intrinsic evaluation including a manual analysis by domain experts,
\item We inject domain knowledge into domain embeddings and show that it produces advancements in the intrinsic and extrinsic evaluation.
\end{enumerate}

    \chapter{Named Entity Recognition in Low-Resource Domains}
\label{sec:third}
Named Entity Recognition (NER) is an important task in the information extraction pipeline as stated in Section \ref{sec:ie} of Chapter \ref{sec:first}.
Existing NER systems rely on large amounts of human-labeled data for supervision. However, obtaining large-scale annotated data in low-resource scenarios is challenging, particularly in specific domains like health-care, e-commerce, and so on. Given the  availability of domain specific knowledge resources (e.g., ontologies, dictionaries), distant supervision has become a solution to generate automatically labeled training data to reduce human effort, as explained in  Section \ref{subsec:ds} of Chapter \ref{sec:first}. 
The outcome of distant supervision for NER is often noisy however. False positive and false negative instances are the main issues that reduce performance on this kind of auto-generated data. 

In this chapter, we explore the use of distant supervision for NER in four low-resource scenarios. We present a system which addresses the problem of noisy data in two ways. We study a reinforcement learning strategy with a neural network policy to identify false positive instances at the sentence level. We further adopt a technique of incomplete annotation to address the false negative cases. The proposed hybrid model achieves competitive performance on benchmark datasets.

\section{Introduction}
Named Entity Recognition (NER) is one of the primary tasks in information extraction pipelines.
Traditional studies apply statistical techniques such as Hidden Markov Models (HMM) and Conditional Random Fields (CRF) using large amounts of features and extra resources \parencite{Ratinov:2009:DCM:1596374.1596399, DBLP:journals/corr/PassosKM14}.
In recent years, deep learning approaches achieve state-of-the-art results in the task without any feature engineering \parencite{ma-hovy-2016-end, DBLP:journals/corr/LampleBSKD16}.
Most of these works assume that there is a certain amount of annotated sentences in the training phase. 
However, the availability of large amounts of labeled data is problematic, particularly in specific domains.
In the low-resource setting, where the amount of data and the knowledge of the domain are insufficient for traditional approaches, distant supervision (see Section \ref{subsec:ds} in Chapter \ref{sec:first}) is proposed by \cite{Mintz:2009:DSR:1690219.1690287} to address the challenge of obtaining training data for new domains using existing knowledge resources (dictionaries, ontologies).
It has previously been successfully applied to tasks like relation extraction 
\parencite{Riedel:2010:MRM:1889788.1889799, DBLP:conf/ekaw/AugensteinMC14} and entity recognition \parencite{DBLP:journals/corr/Fries0RR17,shang2018learning,yang-etal-2018-distantly}.  
To create training data in a NER task, it identifies entity mentions if they exist in the knowledge base (e.g., domain-specific dictionary, glossary, ontology) and assigns the corresponding type according to the knowledge base.

However, distant supervision approaches encounter two main limitations.
First, due to limited coverage of the knowledge resources, unmatched tokens result in False Negatives (FNs). Second, since simple string matching is employed to detect entity mentions, ambiguity in the knowledge resource may lead to False Positives (FPs).
For the FN problem, \cite{tsuboi-etal-2008-training} incorporate partial annotations into CRFs  and propose a parameter estimation method for CRFs using partially annotated corpora 
(here-in after referred to as Partial-CRF). 
In order to reduce the negative impact of FPs for relation extraction, \cite{DBLP:journals/corr/abs-1805-09927} propose a deep reinforcement learning (RL) agent where the agent's goal is to decide whether to remove or keep the distantly supervised instance.

In this chapter, we combine the Partial-CRF approach with the RL approach to clean the noisy, distantly supervised data for NER.
More specifically, we explore the following research questions: 
\begin{question}\label{rq.3.1}
    How can we address the problem of low-resource NER using distantly supervised data?
\end{question}
\begin{question}\label{rq.3.2}
    How can we exploit a reinforcement learning approach to improve NER in low-resource scenarios?
\end{question}
\begin{question}\label{rq.3.3}
    Is the proposed solution beneficial for different low-resource scenarios?
\end{question}

\section{Background} \label{NE:backgroun}
In the following sections, we will describe the background of NER and models that have been proposed for the NER task.
\subsection{Named Entity Recognition}
The term \emph{Named Entity} was introduced in 1996 at the 6th Message Understanding Conference (MUC), as \emph{unique identifiers} of entities (organizations, persons, locations), times (dates, times), and quantities (monetary values, percentages) \parencite{chinchor98}. Most of the annotation datasets for the NER task contain these types of named entities, though with important variations. Example \ref{ex:NE} shows an example of named entities in the widely used CoNLL shared task \parencite{TjongKimSang:2003} dataset. 
\begin{covexample} \label{ex:NE}
\small
\digloss {\underline{Adams} and \underline{Platt} are both injured and will miss \underline{England} 's $\dots$}
{PER {} PER {} {} {} {} {} {} LOC {} $\dots$}
{}
\end{covexample}
At first glance, it seems that only proper names (e.g., Adams, Platt,  England) can be considered as named entities. However, depending on the application, it can be useful to recognize some other linguistic categories as named entities such as pronouns (e.g., it, who, she, he) or nominal mentions (e.g., the girl, mother, the company). Furthermore, the definition can vary depending on the genre and domain (e.g., Health care, Technical review, e-commerce). As an example, \ref{ex:NE-Bio} depicts another type of named entities in the bio-medical domain from BioCreative V CDR task corpus \parencite{10.1093/database/baw068}. Here we see that domain specific tags, such as Chemical and Disease, are used to label named entity mentions.
\begin{covexample} \label{ex:NE-Bio}
\small
\digloss{\underline{Selegiline}-induced {\underline{postural hypotension}} in {\underline{Parkinson' s disease}}}
{Chemical Disease {} Disease}
{}
\end{covexample}

The NER task, or the more general entity mention detection task, is the first step and an essential component of the information extraction pipeline. It involves detecting the boundaries of the phrases that correspond to entities and determining their entity types. Intuitively, given a sentence of words $W:w_1 w_2 ... w_n$, NER assigns a sequence of tags $y:y_1 y_2 ... y_n$ from a predefined set of categories $y_i \in \Phi, |\Phi|=k $.

A single named entity can span several tokens within a sentence. Therefore, sentences are usually represented in specific sequence labeling schemes in NER datasets. The two most popular ones are the following schemes:
\begin{itemize}
    \item {\bf BIO}: It stands for {\bf B}eginning, {\bf I}nside and {\bf O}utside (of a text segment). If the token is the beginning of a named entity, it will be labeled as {B-<type of NE>}, if it is inside a named entity, but not the first token within the token, it will be labeled {I-<type of NE>}, and if it is not a part of a named entity, it will be labeled as {O}.
    \item {\bf BIOES}: Similar but more detailed than BIO, BIOES encodes the beginning, the inside, and last token of multi-token chunks while differentiating them from unit-length chunks.
    It encodes a singleton entity as {S-<type of NE>} and explicitly marks the end token of multi-word named entities as {E-<type of NE>}. 
\end{itemize}   
Table \ref{ex:IOB-ES} shows an example sentence that is annotated in both of the {BIO} and BIOES labeling schemes.
\begin{table}[t]
\centering
\resizebox{.7\linewidth}{!}{
\begin{tabular}{*{3}{p{3cm}P{2cm}P{2cm}}}
\toprule
{} & \multicolumn{2}{c}{Label}  \\
\cmidrule{2-3}
Token&  BIO &  BIOES\\
\midrule
Adams   	 &       B-PER    &   S-PER\\
and    	&	O      &         O\\
Platt  	&	B-PER     &  S-PER\\
are    	&	O      &         O\\
both   	&	O      &         O\\
injured	&	O       &        O\\
and     	  &      O      &         O\\
will		&        O      &         O\\
miss		    &    O       &        O\\
England	&	B-LOC    &   S-LOC\\
's		   &     O      &         O\\
opening	&	O      &         O\\
World	&	B-MISC  &   B-MISC\\
Cup		      &  I-MISC   &    E-MISC\\
qualifier	    &    O      &         O\\
against	&	O       &        O\\
Moldova	&	B-LOC     &  S-LOC\\
on		    &    O   &            O\\
Sunday	&	O      &         O\\
.		    &    O       &       O  \\
\bottomrule

\end{tabular}
}
\caption{Example sentence from CoNLL03 in BIO and BIOES annotation schemes.}
\label{ex:IOB-ES}
\end{table}

\subsection{Neural NER models}
Recently, deep neural models have been employed in the NER task and reached state-of-the-art results on many NER datasets. They benefit from continuous vector representations and semantic composition through nonlinear processing to discover useful representations and underlying factors from input data (see Section \ref{sec:dnns} in Chapter \ref{sec:first}).
Existing models are composed of multiple processing layers to learn representations of data with multiple levels of abstraction. Their architecture (Figure \ref{ex:deepNER}) usually consists of three main components \parencite{DBLP:journals/corr/abs-1812-09449} as follows:
\begin{figure}[t]
\centering
\includegraphics[width=.8\textwidth]{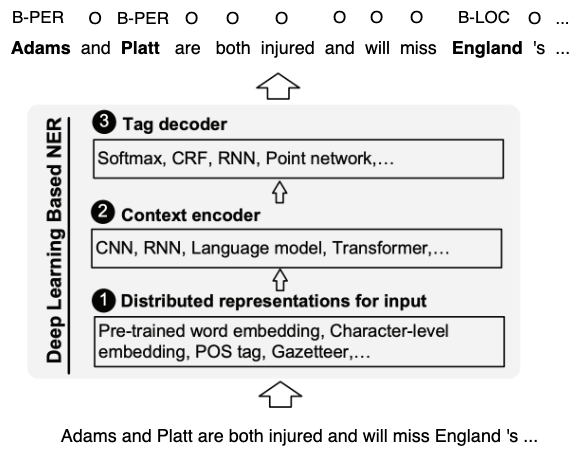}
\caption [General Architecture in the Deep Neural NER.]{General Architecture in the Deep Neural NER in the BIO labeling \parencite{DBLP:journals/corr/abs-1812-09449}.}
\label{ex:deepNER}
\end{figure}

\begin{itemize}
\item {\bf Distributed representations for input}: 
Distributed representations, as described in Section \ref{sec:emb} of Chapter \ref{sec:first},  present each word or character by a low dimensional dense vector where each dimension comprises a latent feature. 
Word-level distributed representations are considered inputs to the NER models. This representation is typically pre-trained over a large collection of text through unsupervised algorithms and captures the syntactic and semantic properties of its elements. The input layer can be either frozen or fine-tuned during the training phase. The pre-trained word embeddings that are used widely in English NER models are Google Word2vec \footnote{\url{https://code.google.com/archive/p/word2vec/}}, Stanford GloVe \footnote{\url{http://nlp.stanford.edu/projects/glove/}}, Facebook fastText \footnote{\url{https://fasttext.cc/docs/en/english-vectors.html}} and SENA \footnote{\url{https://ronan.collobert.com/senna/}}. Several NER models incorporate character-based word representations besides the word representations as an input layer. This representation is learned using an end-to-end neural model. It enables the NER model to learn the representations for unseen words and to share information of morpheme-level regularities. In addition to word- and character-level representations, some NER models include other syntactical, context, morphological, and lexical features into the input layer such as POS tags, word shape, dependency roles, word positions, and gazetteers. However, incorporating these types of features may affect the generality of the NER models. 
\item {\bf Context encoder}:
The second module of neural NER models is devoted to capturing the contextual dependency from the input representations. The widely-used contextual encoders are Convolutional Neural Networks (CNNs) and Recurrent Neural Networks (RNNs) (see Section \ref{sec:dnns} of Chapter \ref{sec:first} for more details). 
Lately, pre-trained language models, as explained in Section \ref{sec:pre-trained-LM} of Chapter \ref{sec:first}, such as ELMO \parencite{DBLP:journals/corr/abs-1802-05365} and BERT \parencite{DBLP:journals/corr/abs-1810-04805}, are used as context encoders and provide a pre-trained deep representation model from unlabeled text. It is empirically verified that the pre-trained model can be fine-tuned with one additional layer for various downstream tasks, including NER, and enhance their performance.
\begin{figure}[t]
\centering
\includegraphics[width=.9\textwidth]{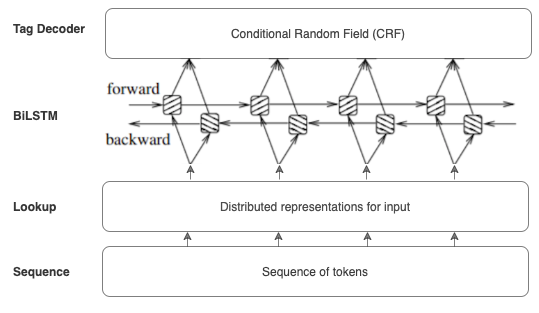}
\caption {Architecture of BiLSTM-CRF framework.}
\label{ex:bilstm}
\end{figure}
\item {\bf Tag decoder}: 
As a last part of the NER model, it takes the output of the context encoder and predicts a sequence of labels corresponding to the input sequence. A Conditional Random Fields (CRFs) framework \parencite{Lafferty:2001:CRF:645530.655813} is the most common choice for the tag decoder step. Most of the state-of-the-art NER models employ CRFs to capture inter-dependency among the labels and show that CRFs can provide higher tagging accuracy in general.
A Multi-Layer Perceptron and Softmax based decoder is another type of design choice for the tag decoder step in some NER models. It casts the sequence labeling task as a multi-class classification problem. In this arrangement, the label for each token is independently predicted without taking into account the adjacent label.
\end{itemize}

\subsection{BiLSTM-CRF framework} \label{bilstm-crf}
BiLSTM-CRF (Figure \ref{ex:bilstm}) is a commonly used neural framework that is exploited for NER. In the following, we introduce the components of the BiLSTM-CRF architecture employed in our work.
\paragraph{BiLSTMs Encoder.} Bidirectional Recurrent Neural Networks (Bi-RNNs) \parencite{schuster1997bidirectional} combines an RNN network (see Section \ref{sub:RNN} in Chapter \ref{sec:first}) which moves forward through time, beginning from the start of the sequence representation, along with another RNN that moves backward, starting from the end of the sequence and is trained using all available input information in the past and future of a specific time frame.

The BiLSTM context encoder employs a Long Short Term Memory (LSTM) \parencite{Hochreiter:1997:LSM:1246443.1246450} instead of RNN. LSTM, as described in Section \ref{sub:RNN} in Chapter \ref{sec:first}, is a special kind of RNN, capable of learning long-term dependencies. This network architecture can efficiently solve the long-term dependencies problem by introducing a gating mechanism and a memory cell. 

\begin{figure}[t]
\centering
\includegraphics[width=.9\textwidth]{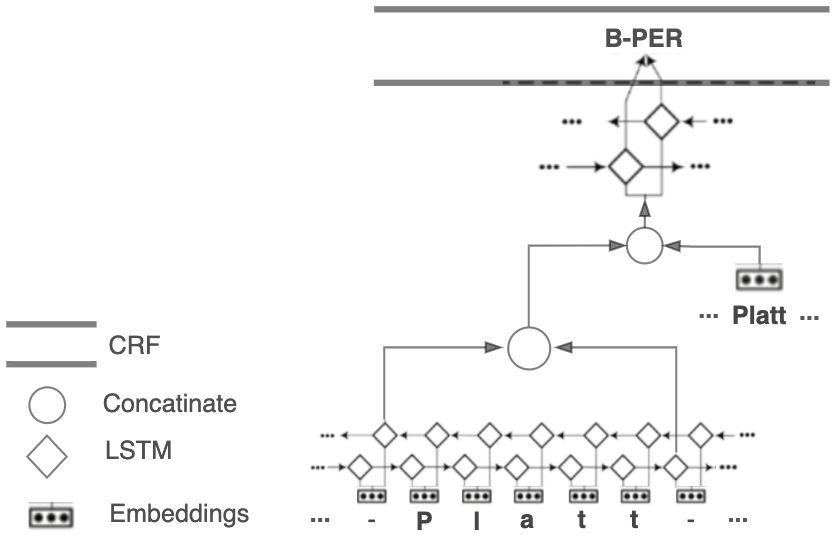}
\caption[Character- and word-level BiLSTM encoding of NER model.]{Character- and word-level BiLSTM encoding of NER model in the BIO labeling \parencite{liu-etal-2018-efficient-contextualized}.}
\label{fig:charWordbilstm}
\end{figure}

\paragraph{Character- and World level encoding.} The character-level BiLSTM networks process characters of word input and learn character-level features while training. Learning character-level embeddings has the advantage of learning representations specific to the task and domain at hand. It has been shown that this type of representation is useful for morphologically rich languages, and handles the out-of-vocabulary problem for some downstream tasks such as part-of-speech tagging and language modeling \parencite{ling-etal-2015-finding}. 
The randomly initialized embedding vector corresponding to each character in the input word is passed through the BiLSTM network in a forward and backward fashion. The forward and backward outputs from the BiLSTM are concatenated to form a character-level encoding for each word. As shown in Figure \ref{fig:charWordbilstm}, this character-level encoding is then concatenated with word embeddings from a word embeddings lookup-table and given to another BiLSTM network as a final context encoder layer. 
Let $X:x_1 x_2 \dots x_n $ be a word-level input representation , where $x_i$ is the embedding vector for the $i_{th}$ word and its character-level input is $C:c_{0,-} c_{1,1} c_{1,2} c_{1,3} c_{1,-} \dots c_{i,j} \dots c_{n,-}$, where $c_{i,j}$ is the dense vector of the $j_{th}$ character in word $w_i$ and $c_{i,-}$ is the representation for a space character after $w_i$.
The first BiLSTM encoder layer will receive a dense vector corresponding to each character. Formally, a LSTM cell will compute the current hidden state $h_t$ based on the current vector $c_{t}$, the previous hidden state $h_{t-1}$ and the previous cell state $s_{t-1}$, however we will only consider $h_t$ at word boundaries, namely space characters or $c_{i,-}$ (see the implementation of LSTM in Section \ref{sub:RNN} of Chapter \ref{sec:first}).
If $\vec{{}h}_{c_{i,-}}$ is the output of the forward character-level LSTM at $c_{i,-}$ and , $\cev{{}h}_{c_{i,-}}$ is the output of the backward character-level LSTM at $c_{i,-}$ , the character-level encoding result for the $i_{th}$ word is: 
\begin{equation}
    h^{c}_i= \vec{{}h}_{c_{i,-}} \oplus \cev{{}h}_{c_{i-1,-}}
\end{equation}
\noindent Subsequently, the character-level encoding output for $i_{th}$ word is concatenated to its word embedding vector and is fed into the second BiLSTMs network. $\vec{{}h_i}$ as the output of the forward word-level LSTM at the $i_{th}$ word and the output of backward word-level LSTM $\cev{{}h_i}$ are concatenated and provide the word-level encoding representation $h_i$ for a word $i$:
\begin{equation}
    h_i= \vec{h_i} \oplus \cev{h_i}
\end{equation}

\paragraph{CRF} In sequence labeling tasks, the dependencies between adjacent labels should be taken into account. Particularly in NER, the characteristics of tagging schemes (e.g. {BIO}, {BIOES}) impose various hard constraints such as \parencite{DBLP:journals/corr/LampleBSKD16}:
\begin{itemize}
    \setlength{\itemsep}{.2pt}
    \item The first word in a sentence should be annotated with a label that begins with {B-} or {O}, not {I-} in the {BIO} scheme.
    \item The label {I-ORG} cannot come after {B-LOC} or any other tag that is not {LOC}
    \item The possible tag that can take place after {B-ORG} is either {I-ORG} or {O} in the {BIO} scheme. 
\end{itemize}
In order to guarantee that the output label sequence is valid, we jointly decode a chain of labels. CRF, as a discriminative type of sequence-based model, considers the dependencies between labels in neighborhoods and defines a conditional probability distribution over a label sequence. It learns the dependency among labels automatically based on the annotated samples during the training process.
The CRF layer comes on top of the last layer (i.e, word-level BiLSTM) to model the dependencies across output tags and locate the best tag sequence by maximizing the log-probability in the following equation:
\begin{equation}\label{eq1-crf}
\log(p(y|W)) =\log \frac{e^{s(W,y)}}{\sum_{y'\in Y} e^{s(W,y')}}
\end{equation}
where 
\begin{equation}\label{eq1-1}
s(W,y)=\sum_{i=1}^{n} \mathbf{P}_{i,y_i} + \sum_{i=1}^{n} \mathbf{T}_{y_i,y_{i+1}}
\end{equation}
Where $W:w_1 w_2 \dots w_n$ is an input sequence, $y:y_1 y_2 \dots y_n$ is an annotated tag sequence and $Y$ is all possible tag sequences of the sentence. CRF takes $\mathbf{P}$ as an emission score matrix which is a ${k\times n}$ output tensor of a linear encoder applied to the last BiLSTM layer where  $P_{i,j}$ corresponds to the score of the $j^{th}$ tag of the $i^{th}$ word in a sentence. 
$\mathbf{T}$ is a ${(k+2)\times(k+2)}$ transition tensor 
which represents the transition probability from the $i^{th}$ tag to the $j^{th}$ tag. Two additional tags $<$BOS$>$ and $<$EOS$>$ are added at the start and end of a sequence, respectively. The transition matrix is randomly initialized and is learned by the CRF during the training phase. 
For training, we encourage the model to produce a valid sequence of output labels by defining the loss function as the negative log likelihood of the current sequence tag $y$:
\begin{equation} \label{eq:loss}
\mathcal{L}= -\log(p(y|W))= log(\sum_{y'\in Y} e^{s(W,y')})-s(W,y)
\end{equation}
While testing or decoding, the goal is to determine the best label sequence $y*$ that maximizes the likelihood conditioned on the input sentence $W$ and learnt model parameters $\Theta$ (e.g,  $\mathbf{T}$ and  $\mathbf{P}$) :
\begin{equation}\label{eq:argmax}
y*=\argmax_{y\in Y} p(y|W;\Theta)
\end{equation}
Since we model only bi-gram interactions (i.e, two adjacent labels) in the CRF model, both Eq. \ref{eq:loss} and Eq. \ref{eq:argmax} can be computed by adopting the Viterbi algorithm. 
 
\subsection{Reinforcement Learning} \label{RL}
Reinforcement learning (RL) differs from supervised and unsupervised learning in that the goal is to learn a set of actions without relying on a labeled training dataset to maximize a predefined reward function.
The learning process is based on the finite Markov Decision Process (MDP) framework where the RL model consists of various key elements: In a given \emph{state} ($s_t$) of a stochastic environment, an \emph{agent} as a learner and decision maker tries to find an optimal \emph{action} ($a_t$) in order to maximize the expected rewards ($r_t$), by following a policy ($\pi$). More specifically, at each time step $t=\{0,1,2,\dots, T\}$, the agent receives a state $s_t\in S$ as representation of the environment, and following the policy, the agent performs an action ($a_t \in A$). The RL model aims to maximize the following objective function \parencite{sutton1998reinforcement}:
\begin{equation} \label{eq:rl}
    \max_{\theta} E_{\tau \sim \pi_{\theta}(\tau)}[\sum_{t}^{T} r(s_t,a_t)]
\end{equation}

Where $r(s_t,a_t)$ is a reward in time step $t$, $\tau$ is a sequence of states, actions, and rewards known as a trajectory, and $\pi_{\theta}(\tau)$ is the joint probability of a sequence of actions that can be formulated in MDP as:
\begin{equation}\label{eq:r2}
    \pi_{\theta}(\tau)= \pi_{\theta}(s_1, a_1, \dots, s_T,a_T)= p(s_1)\prod_{t=1}^{T} \pi_{\theta}(a_t|s_t)p(s_{t+1}|s_t,a_t)
\end{equation}
$\pi_{\theta}(a_t|s_t)$ is a policy that tells the agent how to act from a particular state, and $p(s_{t+1}|s_t,a_t)$ known as the model in RL, is a transition function that predicts the next state after taking action.

The most successful RL techniques employ a neural network in conjunction with RL. Neural models enable the RL model to deal with unstructured environments, learn complex functions, solve complicated problems in an end-to-end fashion, or predict actions in unseen states. 
Policy Gradient introduced by \cite{Sutton:1999:PGM:3009657.3009806} is one of the RL algorithms that focuses on the policy. The policy is learned by directly differentiating the objective function in Eq. \ref{eq:rl} as follows:

\[ J(\theta)=E_{\tau \sim \pi_{\theta}(\tau)}\left[\sum_{t}^{T} r(s_t,a_t)\right]=E_{\tau \sim \pi_{\theta}(\tau)}[r(\tau)]=\int \pi_{\theta}(\tau) r(\tau) d\tau \]
\noindent Since:
\[ \nabla_{\theta}f(x)=f(x)\frac{\nabla_{\theta}f(x)}{f(x)}=f(x)\nabla_{\theta}\log f(x)
\]
\noindent Then:
\[\nabla_{\theta} J(\theta) =\int \nabla_{\theta} \pi_{\theta}(\tau)r(\tau) d\tau=\int \pi_{\theta}(\tau) \nabla_{\theta} \log \pi_{\theta}(\tau)r(\tau) d\tau=\]
\[E_{\tau \sim \pi_{\theta}(\tau)}[\nabla_{\theta}\log \pi_{\theta}(\tau)r(\tau)
\]

\noindent Considering Eq. \ref{eq:r2}:
\[\log\pi_{\theta}(\tau)=\log p(s_1)+ \sum_{t=1}^{T} \log \pi_{\theta}(a_t|s_t)+\log p(s_{t+1}|s_t,a_t)\] 

\[\nabla_{\theta} J(\theta)=E_{\tau \sim \pi_{\theta}(\tau)}\left[\left(\sum_{t=1}^{T}\nabla_{\theta} \log \pi_{\theta}(a_t|s_t)\right)\left(\sum_{t=1}^{T}r(s_t,a_t)\right)\right]
\]
\noindent Since: 
\[
J(\theta)=E_{\tau \sim \pi_{\theta}(\tau)}\left[\sum_{t}^{T} r(s_t,a_t)\right] \approx \frac{1}{N} \sum_i \sum_t r(s_{i,t},a_{i,t})
\]
\noindent Then:
\begin{equation} \label{eq:r3}
    \nabla_{\theta} J(\theta)\approx \frac{1}{N} \sum_{i=1}^{N} \left(\sum_{t=1}^{T}\nabla_{\theta} \log \pi_{\theta}(a_{i,t}|s_{i,t})\right) \left(\sum_{t=1}^{T}r(s_{i,t},a_{i,t})\right)
\end{equation} 

\begin{algorithm}[t]
\DontPrintSemicolon
  Initialize $\theta$ at random\\
 \For { Generate $\{\tau^i\}$, following $\pi_\theta$}
 {
 \For {$t=1$ to $T-1$}
 { $\nabla_{\theta} J(\theta)\approx  \sum_{i} \left(\sum_{t}\nabla_{\theta} \log \pi_{\theta}(a_{i,t}|s_{i,t})\right) \left(\sum_{t}r(s_{i,t},a_{i,t})\right)$\\
 $\theta \leftarrow \theta + \alpha \nabla_{\theta}J(\theta)$
 }
 }
 Return $\theta$
\caption{REINFORCE \parencite{Williams1992}.}
\label{alg.reinforce}
\end{algorithm}
\noindent Eq. \ref{eq:r3} computes how likely the trajectory is under the current policy. If the results of the trajectory lead to a high positive reward, it will increase the likelihood. On the other hand, it will decrease the likelihood of a policy if it outputs a high negative reward. In short, keep what has positive effects and throw out what does not. REINFORCE \parencite{Williams1992} is known as the Monte-Carlo policy gradient, which uses Monte Carlo rollout to compute the rewards (see Algorithm \ref{alg.reinforce}).  The agent collects a trajectory $\tau$ of one episode using its current policy and updates the policy parameter using the $\tau$. Since one full trajectory must be completed to construct a sample space, REINFORCE updates the policy network parameters (weights) in the direction of the gradient (see line 5 of Algorithm \ref{alg.reinforce}).
\section{Low-Resource NER}
The task of NER has been widely studied in the last decade and is usually formulated as a sequence labeling problem. Using neural techniques, many studies report state-of-the-art results on this task \parencite{DBLP:journals/corr/LampleBSKD16, ma-hovy-2016-end}. These studies utilize character and/or word embeddings to encode sentence-level features automatically. Recently, the use of contextualized word representation \parencite{DBLP:journals/corr/abs-1802-05365,akbik-etal-2018-contextual} significantly improves the state-of-the-art results in many sequence labeling tasks and specifically also in the NER benchmark.

In the supervised paradigm, NER suffers from a lack of large-scale labeled training data when moving to a new domain or new language. To alleviate the reliance on human annotated data, distant supervision is proposed by \cite{Mintz:2009:DSR:1690219.1690287}, to generate annotated data by heuristically aligning text to an existing domain-specific knowledge resource. It is widely used for relation extraction \parencite{Mintz:2009:DSR:1690219.1690287, Riedel:2010:MRM:1889788.1889799, DBLP:conf/ekaw/AugensteinMC14} and lately it has attracted attention also for NER \parencite{conf/kdd/RenEWTVH15, DBLP:journals/corr/Fries0RR17,shang2018learning,yang-etal-2018-distantly}. 
In this section, we look more closely at previous works that utilize the data generated by distant supervision in relation extraction and NER tasks and address the challenges of noisy generated data.

\cite{DBLP:journals/corr/abs-1808-08013} propose a model for sentence level relation classification in noisy sentences that are collected via distant supervision.
The model contains a relation classifier and an instance selector. The instance selector filters out the low-quality sentences by using reinforcement learning and provides the selected sentences for the relation classifier. They adopt REINFORCE in the instance filtering step. The relation classifier predicts the relation at the sentence level and produces rewards as a weak supervision signal to the filtering module. The two modules are trained jointly to optimize their objective functions.
In the relation classification module, a convolutional architecture determines the relation class for entity pairs in a given sentence. The instance selector is the agent that follows a feed-forward neural policy network to distill the training data for the relation classifier. At the same time, it refine its policy function using the feedback from the relation classifier. The reward is calculated based on the prediction probabilities in the CNN model when the selection of all training sentences is finished.

\cite{DBLP:journals/corr/abs-1805-09927} also explore deep reinforcement learning as a false positive removal tool for distantly supervised relation extraction. 
The policy-based agents are learned for each relation type, and they aim to determine and remove the false positive cases from auto-generated labeled data. 
In contrast to \cite{DBLP:journals/corr/abs-1808-08013}, the reward is intuitively reflected by the performance change of the relation classifier. They design the policy agent in a supervised fashion and use a pre-trained policy network in the RL module.
Here, we adapt their approach to the NER task. 
Unlike \cite{DBLP:journals/corr/abs-1805-09927}, we learn the policy agent in an unsupervised manner, where the parameters are learned by interaction with the environment. 

\cite{yang-etal-2018-distantly} make use of reinforcement learning to tackle false positives in distantly supervised NER. Similar to our work, \cite{yang-etal-2018-distantly} address the noisy automatic annotation in NER, by using partial annotation learning and reinforcement learning. However, unlike our approach, they train the NER model and reinforcement learning model jointly, calculating the reward based on the loss of the NER model. In contrast, we employ the RL module as a pre-processing/filtering step, incorporating the previous state to satisfy a Markov decision process (MDP). \cite{yang-etal-2018-distantly} evaluate only on a Chinese dataset, whereas we apply our model also to English datasets. 
Furthermore, after running their code \footnote{\url{https://github.com/rainarch/DSNER}}, we observe that to reach the reported results in their paper on the e-commerce dataset, the model needs more than 500 epochs and the reinforcement learning component removes all the distantly annotated sentences after some epochs. This means that after some epochs, the code in reality only applies the baseline NER model on the annotation dataset and ignores the RL module since there are no distantly annotated sentences. 
Their two datasets are included in our experiment in order to compare to their results.

\cite{shang2018learning} present the AutoNER model, which employs a new type of tagging scheme (dubbed Tie or Break) rather than the common ones (i.e., BIO, BIOES). The model does not have a final CRF layer but still achieves state-of-the-art unsupervised $F1$ scores on several benchmark datasets.
Instead of predicting the label of each token, they propose predicting whether two adjacent tokens are tied (i.e., Tie) in the same entity mention or not (i.e., Break). They find that even when the boundaries of an entity mention are mismatched by distant supervision, most of its inner ties are not affected, and thus more robust to noise. Accordingly, they design a neural architecture (AutoNER), that identifies all possible entity spans by detecting such ties and then predicts the entity type for each span. Crucially, they employ a set of high-quality phrases in distant supervision, using a phrase mining technique, AutoPhrase \parencite{shang2018automated}, to reduce the false-negative labels.  
The AutoPhrase framework leverages available high-quality phrases in general knowledge bases such as Wikipedia and Freebase for distant supervision to avoid additional manual labeling effort. Therefore, it independently creates samples of positive labels from general knowledge and negative labels from the given domain corpora and trains several classifiers. Then, it aggregates the predictions of the classifiers to reduce the noise from negative labels. In the first phase, AutoPhrase establishes the set of phrase candidates that contains all n-grams considering a threshold based on the raw frequency of the n-grams. Given a phrase candidate, the quality of the phrase is estimated by some statistical features such as point-wise mutual information, point-wise KL divergence, and inverse document frequency. Finally, it finds a complete semantic unit in some given context by using part-of-speech-guided phrasal segmentation. AutoPhrase can support any languages as long as a general knowledge base of the language is available, while benefiting from, but not requiring a POS tagger \parencite{shang2018automated}.

\section{Model} \label{NER+PA+RL}
\begin{figure}[t]
\centering
\scalebox{1}{
\includegraphics[width=.8\textwidth]{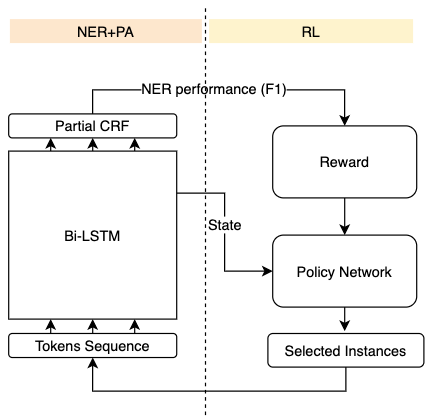}
}
\caption{NER+PA+RL architecture.}
\label{fig:NER+PA+RL}
\end{figure} 
In this section, we present the proposed model, which copes with the problems in distantly supervised NER. We implement Partial-CRF together with a performance-driven, policy-based reinforcement learning method to detect FNs and FPs in distantly supervised NER.
We here combine techniques that have been shown to be useful in previous work \parencite{yang-etal-2018-distantly}.
In our architecture, as shown in Figure \ref{fig:NER+PA+RL}, we first apply partial annotation learning (PA) using the annotation dataset ($A$) and distantly labeled data ($D$). Then, we apply reinforcement learning (RL) to clean FPs from the noisy dataset ($D$).
Our RL agent is rewarded based on the change in the NER's performance and is modeled as a Markov Decision Process (MDP).

Algorithm \ref{alg.1} describes the overall training procedure for our model, and in the following sections, we detail the various components of our model.
\begin{algorithm}[t]
\DontPrintSemicolon
  \KwInput{ Human Annotated ({A}) + Distantly Labeled Data ({D}) }
  Pre-train NER with Partial-CRF ({NER+PA}) on {A+D}\\
  Apply {RL} on {D}\\
  Train {NER+PA} using {A} + cleaned {D} \\
\caption{Overall Training Procedure {NER+PA+RL}.}
\label{alg.1}
\end{algorithm}
\subsection{Baseline NER model}
Our baseline model is a BiLSTM-CRF architecture \parencite{DBLP:journals/corr/LampleBSKD16,Habibi2017DeepLW}, which is described in Section \ref{bilstm-crf}. The first layer takes character embeddings for each word sequence and then merges the output vector with the word embedding vector to feed into a second BiLSTM layer. We modify the top element (CRF layer) of the baseline model as follows.
\begin{figure}[t]
\centering
\scalebox{1}{
\includegraphics[width=1\textwidth]{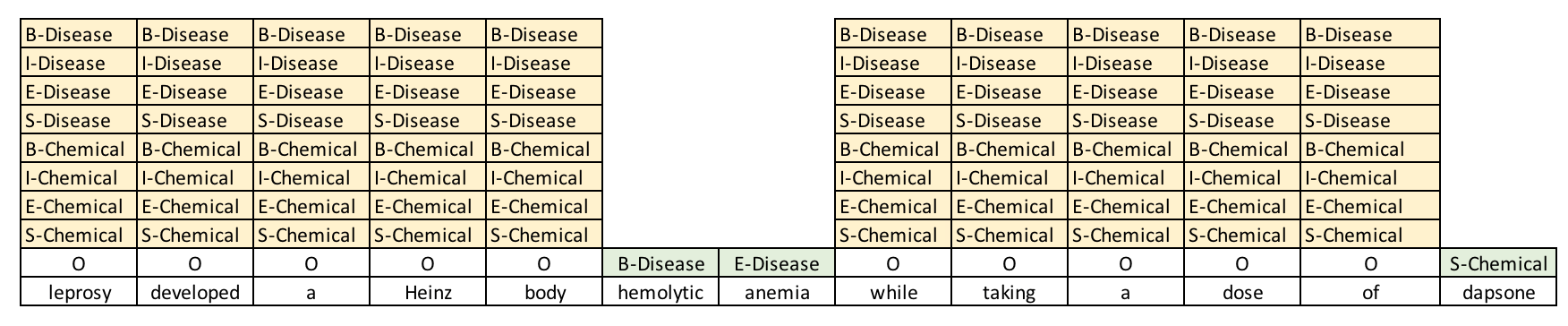}
}
\caption[Annotation of the distantly labeled example in Partial-CRF.]{Annotation of the distantly labeled example in Partial-CRF based on the {BIOES} labeling. The words with green tags are found in the dictionary and assigned to the corresponding entity types, and the ones that are not found in the dictionary are assigned to all possible tags (yellow).}
\label{fig:partial-crf}
\end{figure} 
\subsection{Partial-CRF layer (PA)} \label{NER+PA}
As mentioned above, FN instances constitute a common problem in distantly annotated datasets. It is caused by limited coverage of the knowledge base resource when some of the entity mentions are not found in the resource and followingly labeled as non-entities ('O'). 
We follow \cite{tsuboi-etal-2008-training} and treat the result of distant supervision as a partially annotated dataset where non-entity text spans are annotated as any possible tag.
Figure \ref{fig:partial-crf} illustrates the annotation of distantly supervised examples using the BIOES labeling scheme that we employ.

Let $Y_L$ denote all the possible tag sequences for a distantly supervised sentence $W$. Then, the conditional probability of the subset $Y_L$ given $W$ is:
\begin{equation}\label{eq-partial}
 p(Y_L|W)= \sum_{y \in Y_L}p(y|W)
\end{equation}
Extending the original equation of the CRF layer (Eq. \ref{eq1-crf})
provides the log-probability for the distantly supervised instance:
\begin{equation}\label{eq2}
\log(p(Y_L|W)) =\log \frac{\sum_{y'\in Y_L}e^{s(W,y')}}{\sum_{y'\in Y} e^{s(W,y')}}. 
\end{equation}
Using partial annotation learning, non-entity text spans are annotated as any possible tag. It gives a chance for non-entity text spans to be considered and scored properly in the updated version of CRF (Partial-CRF) and become a part of the most optimal tag sequence. 

\subsection{Reinforcement Learning (RL)} \label{NER+RL}
The RL agent is designed to determine whether the distantly supervised instance is a true positive or not. There are two main components in RL: 
\begin{enumerate*}[label=(\roman*)]
    \item the environment, and
    \item the policy-based agent.
\end{enumerate*}
Following \cite{DBLP:journals/corr/abs-1805-09927}, we model the environment as a Markov Decision Process (MDP), where we add information from the previous state to the current state. The policy-based agent is formulated based on the Policy Gradient Algorithm \parencite{Sutton:1999:PGM:3009657.3009806}, as explained in Section \ref{RL}, where we update the policy model by computing the reward after finishing the selection process for the whole training set. Algorithm 
\ref{alg.2} presents additional details of the RL strategy in our NER model. The following subsections describe the elements of the RL agent.

\begin{algorithm}[t]
\DontPrintSemicolon
 \KwInput{Training dataset ($A_{train}$) + Distantly Labeled Data ($D$) , Pre-train \emph{NER+PA} on  $A_{train}+D$, Validation dataset ($A_{val}$)}
 
 Initialize $\theta$ in policy network \\
 Initialize $s^{*}$ as all-zero vector with the same dimension of $s_j$ \\
 
  \For{epoch $i=0\rightarrow N$} 
  {
  \For {instance $d_j \in D$}
  {
  Provide $s_j$ using \emph{NER+PA} model
  $\tilde{s}_{j}=$\emph{concatenation}($s_j,s^{*}$)\\
  Randomly sample $a_j \sim \pi (a; \theta,\tilde{s}_{j})$; compute $p_j=\pi (a; \theta,\tilde{s}_{j})$, save $(a_j,p_j)$\\
  \If {$a_j==0$} {
    save $\tilde{s}_{j}$ into $\Psi_i$ 
  }
  }
  Recompute the $s^{*}$ as an average of $\forall \tilde{s}_{j} \in \Psi_i$ \\
  $D_i=D- (\forall{d_j}$; $j \in \Psi_i)$\\
  Train \emph{NER+PA} on $A_{train}+D_i$\\
  Calculate $F_{1}^{i}$ on $A_{val}$  and save $F_{1}^{i}$ and $\Psi_i$\\ 
  $r_i= F_{1}^{i}-F_{1}^{i-1}$\\
  Find $\Omega_{i}, \Omega_{i-1}$ (Eq. \ref{omega})\\
  Update Policy network (Eq. \ref{gradient})
  }
Update $D=D- (\forall{d_j}$; $j \in \Psi_N)$\\
Re-train NER+PA on $A+D$
\caption{Reinforcement learning Algorithm to clean FPs on $D$.}
\label{alg.2}
\end{algorithm}

\subsubsection{State} 
The RL agent interacts with the environment to decide about instances at the sentence level. A central component of the environment is the current and previous state in the selection process. The state $S_i$ in step $i$ represents the current instances as well as their label sequences. Following \cite{yang-etal-2018-distantly} the state vector $S_i$ includes: 
\begin{itemize}
    \item The vector representation of instances before the Partial-CRF layer, where we concatenate the outputs of the first and last nodes in the BiLSTM layer of the base NER model.
    \item The label sequence scores calculated by the linear encoder before the Partial-CRF model. (i.e, $P_{i,j}$ in Eq. \ref{eq1-1}).%
\end{itemize} 
If a word is annotated with a specific label, the score will be the corresponding value of the label. Otherwise, the score will be the mean of all possible word labels in the linear encoder. These two vectors are concatenated to represent the current state. To satisfy the MDP, the average vector of the removed instances in the earlier step $i-1$ is concatenated to the current state and represents the state for the RL agent.
\subsubsection{Reward} 
The NER model will achieve improved performance if the RL agent filters out the FP instances from the noisy dataset. Accordingly, the RL agent will receive a positive reward; otherwise, the agent will receive a negative reward. 
Following \cite{DBLP:journals/corr/abs-1805-09927}, we model the reward as a change of the NER performance; particularly, we adapt the $F1$ score to calculate the reward as the difference between $F1$ scores of the adjacent epochs (i.e., $r_i= F_{1}^{i}-F_{1}^{i-1}$). 
\subsubsection{Policy Network} 
The policy network $\pi (a_j;\theta_i,s_j)$ is a feed forward network with two fully-connected hidden layers. It receives the state vector for each distantly supervised instance and then determines whether the instance is a false positive or not. The $\pi$ as a classifier with parameter $\theta$ decides an action $a_j \in \{1,0\}$ for each $s_j \in S_j$.
The loss function for the policy network is formulated based on the policy gradient method and the REINFORCE algorithm (Section \ref{RL}). Since we calculate the reward as a difference between $F1$ scores in two contiguous epochs, the agent will be compensated for a set of actions that has a direct impact on the performance of the NER model in the current epoch. In other words, the different parts of the removed instances in each epoch are the reason for the change in $F1$ scores. 
Accordingly, the policy will update using the following gradient:
\begin{equation} \label{gradient}
\begin{split}
\theta =\theta + \alpha [  \bigtriangledown_{\theta}\sum _{a_j,s_j \in \Omega_{i}}\log \pi (a_j|S_j;\theta)r_{i}\\ + \bigtriangledown_{\theta}\sum _{a_j,s_j \in \Omega_{i-1}} \log \pi (a_j|S_j;\theta)(-r_{i})]
\end{split}
\end{equation}
According to \cite{DBLP:journals/corr/abs-1805-09927}, assuming $\Psi_{i} $ is removed in epoch $i$ : 
\begin{equation}\label{omega}
\begin{split}
\Omega_{i} =\Psi_{i}-(\Psi_{i}\cap \Psi_{i-1})\\
\Omega_{i-1} =\Psi_{i-1}-(\Psi_{i}\cap \Psi_{i-1})
\end{split}
\end{equation}
This means that if there is an increase in $F_1$ at the current epoch $i$, we will assign a positive reward to the instances that have been removed in epoch $i$ and not in epoch $i-1$ and negative reward to the instances that have been removed in epoch $i-1$ and not in the current epoch.  
\section{Experiments}
We perform experiments on four benchmark datasets to compare our method to similar techniques and investigate the impact of the number of available annotated sentences for our approach. In this section, we describe the experimental setup and various components of the model.
\subsection{Datasets} 
\begin{table}[t]
  \centering
\resizebox{1\linewidth}{!}{
  \begin{tabular}{*{6}{p{2cm}p{3cm}cccc}}
    \toprule
    \multirow{2}{*}{\head{Name}} & \multirow{2}{*}{\head{Domain} } &  \multirow{2}{*}{\head{Entity Types}}&\head{Size}& \head{Dictionary} &\multirow{2}{*}{\head{\# Raw Sent.}} \\
    &&&(Train/Dev./Test)& \head{Size}&\\
    \midrule
    \midrule
    {BC5CDR} & Bio-Medical	& Disease, Chemical 	& 4,560/4,581/4,797 &	322,882 & 20,217  \\
    \midrule
     {LaptopReview} & Technical Reviews & Aspect  terms	& 2,445/609/800 &13,457 & 15,000 \\ 
     \midrule
    \multirow{3}{4em}{EC}& \multirow{2}{10cm}{E-commerce} & Brand, Product &  & & \\
    &\multirow{2}{10cm}{(Chinese)}& Model, Material&1,200/400/800 & 927& 2,500\\
    &&Specification&&&\\
     \midrule
     {NEWS} & news (Chinese)&  Person& 3,000/3,328/3,186	& 71,664 & 3,722 \\ 
     \midrule
  \end{tabular}
  }
\caption{Overview of datasets in our experiments.}
\label{tbl:ner-dataset}
\end{table}
Our approach requires an annotated dataset, a knowledge resource , and a corpus of raw text. We rely on the resources used by \cite{shang2018learning} and \cite{yang-etal-2018-distantly} for English and Chinese, respectively, as well as their train-test splits. As is shown in Table \ref{tbl:ner-dataset}, these datasets are from several different domains (biomedical, e-commerce, technical reviews, and news) as well as two different languages. For all datasets, the distant supervision is performed on the raw data to create a distantly annotated dataset using the knowledge resource (i.e., dictionary). The annotation is based on the BIOES labeling scheme. Below we briefly describe the datasets.
\paragraph{BC5CDR.} This dataset is from BioCreative V Chemical Disease Relation task \parencite{10.1093/database/baw068} and contains 12,852  \emph{Disease} and 15,935 \emph{Chemical} entity mentions in 1,500 articles. Example \ref{fig:bc5cdr-example} shows an annotated sentence in this dataset with the BIOES tags.
The BC5CDR is already partitioned randomly into a training, a development and a test set (500 articles each). The related dictionary is constructed from the MeSH database\footnote{\url{https://www.nlm.nih.gov/mesh/download_mesh.html}} and the CTD chemical and Disease\footnote{\url{http://ctdbase.org/downloads/}}
vocabularies and contains 322,882 \emph{Disease} and \emph{Chemical} entities. As raw text, we use a corpus consisting of 20,217 sentences that is provided in \cite{shang2018learning} and extracted from PubMed papers.

\begin{covexample}\label{fig:bc5cdr-example}
\digloss[tlr=true,fsii={\footnotesize}]
{Selegiline - induced postural hypotension in Parkinson ' s disease :}
{S-Chemical  O O B-Disease E-Disease O B-Disease I-Disease I-Disease E-Disease O}
{} 
\end{covexample}

\paragraph{LaptopReview.} 
The LaptopReview dataset contains laptop aspect terms taken from the SemEval 2014 Challenge, Task 4 Subtask 1 \parencite{pontiki-etal-2014-semeval}. The 3,845 review sentences are annotated with 3,012 \emph{AspectTerm} mentions (e.g., disk drive). We extract 15,000 sentences from the Amazon laptop review dataset \footnote{\url{http://times.cs.uiuc.edu/\~wang296/Data/}} as raw text. \cite{Wang:2011:LAR:2020408.2020505}
designed this dataset for aspect-based sentiment analysis. Thanks to \cite{shang2018learning}, they provide
a dictionary of 13,457 computer terms crawled from a public website \footnote{\url{https://www.computerhope.com/jargon.htm}}. An example sentence from the training in the BIOES tags is shown in example \ref{fig:lptr-example}.

\begin{covexample}\label{fig:lptr-example}

\digloss[fsii={\footnotesize}]
{I love the operating system and the preloaded software .}
{O O O B-AspectTerm E-AspectTerm O O B-AspectTerm E-AspectTerm O}
{} 
\end{covexample}

\paragraph{EC.} The EC dataset is a Chinese dataset from the e-commerce domain. We choose this dataset in order to compare our results to the approach by \cite{yang-etal-2018-distantly}. There are 5 entity types: \emph{Brand}, \emph{Product}, \emph{Model}, \emph{Material} and \emph{Specification} on user queries. An example sentence of the EC dataset is represented in example \ref{fig:ec-example}. This corpus contains 1,200 training instances, 400 in development set, and 800 in the test set. \cite{yang-etal-2018-distantly} provide a small dictionary of 927 entries and 2,500 sentences as raw text.

\begin{covexample}\label{fig:ec-example}
\begin{CJK*}{UTF8}{gbsn}
\begin{tabular}{lllllllll}
我&	要&	买&一&台&游&戏&	本&。 \\
\footnotesize{O}& \footnotesize{O} &\footnotesize{O}&\footnotesize{O}&\footnotesize{O}&\footnotesize{B-Product}&\footnotesize{I-Product}&\footnotesize{E-Product}&\footnotesize{O}
\end{tabular}\\
\hspace{3mm} 'I want to buy a gaming computer.'
\end{CJK*}
\end{covexample}

\paragraph{NEWS.} The NEWS dataset is another Chinese dataset from the news domain and is annotated with \emph{Person} type (PER) and provided by \cite{yang-etal-2018-distantly}, as shown by the sentence taken from this dataset in example \ref{fig:news-example}. The NEWS dataset contains 3,000 sentences for training, 3,328 for development, and 3,186 for testing. \cite{yang-etal-2018-distantly} apply distant supervision to raw data, and obtain 3,722 annotated sentences. The dataset and raw text are taken from the MSRA corpus \parencite{levow-2006-third}.
\begin{covexample}\label{fig:news-example}
\begin{CJK*}{UTF8}{gbsn}
\begin{tabular}{llllllllll}
巫&昌&桢&、&罗&涵&先&委&员&$\dots$ \\
\footnotesize{B-PER}&\footnotesize{I-PER}&\footnotesize{E-PER}&\footnotesize{O}&\footnotesize{B-PER}&\footnotesize{I-PER}&	\footnotesize{E-PER}&\footnotesize{O}&\footnotesize{O}&$\dots$
\end{tabular}
\\
\hspace{3mm}'Committee members Wu Changzhen and Luo Hanxian $\dots$'
\end{CJK*}
\end{covexample}

\subsection{Pre-trained Embeddings}
The pre-trained embeddings have been used as initialization for the embedding layer of the LSTM layers of the BiLSTM model described in Section \ref{NER+PA+RL}.
Standard pre-trained GloVe 100-dimensional word vectors are employed for the {LaptopReview} dataset.
In our experiments on the {EC} dataset, we use the 100-dimensional Chinese character embeddings provided by \cite{yang-etal-2018-distantly}, which is trained on one million sentences of user-generated text. 
For the biomedical dataset, we use pre-trained 200-dimensional word vectors trained on PubMed abstracts, all PubMed Central (PMC) articles, and English Wikipedia \parencite{Pyysalo:2013b}. We here employ an embedding model that is domain-specific since we observe that this type of model provides an improvement in our previously studied domain-specific downstream task (see Chapter \ref{sec:second}). In addition, \cite{DBLP:journals/corr/abs-1801-09851} show that the domain-specific embeddings are beneficial for tagging performance on the BC5CDR dataset.

\subsection{Evaluation}
We report the performance of the model on the test set as the micro-averaged precision, recall, and $F1$ score. According to CoNLL-2003 \parencite{TjongKimSang:2003},
a predicted entity is counted as a true positive if both the entity boundary and entity type is the same as the ground-truth (i.e., exact match).
To alleviate the randomness of the scores, the mean of five different runs are reported. 
\subsection{Model Variants}
We use slightly different variants of our model for English and Chinese. For English we follow  \cite{DBLP:journals/corr/abs-1709-04109} in leveraging a language model to extract character-level knowledge. 
We keep the parameters in the model the same as in the original work. In order to compare to state-of-the-art models, we follow the same approach during training (i.e., by merging the training and development data as a training set in BC5CDR and randomly selecting $20$\% from the training set as the development set in LaptopReview). For the Chinese EC dataset, we only use character-based LSTM and CRF layers and discard the word-based LSTM and language model. For a fair comparison, the model parameters are set to be the same as in \cite{yang-etal-2018-distantly}, as well as the batch size, optimizer, and learning rate for RL module.
We use 100 epochs in RL and initialize the average vector of the removed sentences as an all-zero vector.
\subsection{High-Quality Phrases} Considering all non-entity spans (i.e., 'O' type) as a potential entity, provides noise in the Partial-CRF process. 
To address this issue, we use a set of quality multi-word and single-word phrases, provided by \cite{shang2018learning} and obtained using their AutoPhrase method \parencite{shang2018automated}.
Note that this resource is available only for the English datasets; therefore, it is not included in the experiments on the Chinese datasets. 
When using these phrases, we assign all possible tags only for the token spans that are matched with this extended list.
In our model, we treat high-quality phrases as potential entities, and we assign all possible entity types in the annotation of distantly supervised sentences. For example, in Figure \ref{fig:partial-crf}, we could only find the word 'leprosy' in this list, therefore, in annotation, we assign all possible tags to this token, and the other non-entity tokens remain as 'O'. 
\begin{table}
\centering
\begin{tabular}{*{4}{p{6cm}|c|P{1cm}P{1cm}P{1cm}}}
\toprule
\head{Model Variant} &\head{Data}& Pr.& Re. & F1  \\
\midrule
 {NER+PA}& \parbox[t]{2mm}{\multirow{4}{*}{\rotatebox[origin=c]{90}{{BC5CDR}}}}& 85.82 & 88.58& 87.18\\
{{NER+PA}\ding{74}} && 91.28 & 87.07 &	 89.13  \\
  {NER+PA+RL} && 87.00 &\textbf{89.04} &  88.01  \\
  {NER+PA+RL}\ding{74}&&  \textbf{92.05}  &  {87.91}  &  \textbf{89.93} \\\hline
{NER+PA}&\parbox[t]{4mm}{\multirow{4}{*}{\rotatebox[origin=RB]{90}{{Laptop}}\rotatebox[origin=RB]{90}{{Review}}}}&61.00	&70.80&	65.53 \\
 {NER+PA}\ding{74} && 66.36 &  66.06 & 66.21 \\
 {NER+PA+RL} &&80.47&73.70&76.94\\
 {NER+PA+RL}\ding{74} && \textbf{81.07}&	\textbf{74.01} &	\textbf{77.38}\\
\bottomrule
\end{tabular}
\caption[Result with different setting of the distantly supervised NER model.]{Result with different setting of the distantly supervised NER model. \ding{74} indicates that we use the list of high-quality phrases along with the dictionary to annotate raw text. The PA and RL denote the use of partial annotation learning and reinforcement-based components, respectively.}
\label{tbl.2}
\vspace{1cm}
\begin{tabular}{*{5}{p{4.3cm}|c|P{1.5cm}P{1.5cm}P{1.5cm}}}
\toprule
\head{Model} & \head{Data}& Pr.& Re. & F1 \\
\midrule

\cite{DBLP:journals/corr/abs-1709-04109} *& \parbox[t]{2mm}{\multirow{4}{*}{\rotatebox[origin=c]{90}{{BC5CDR}}}}& 88.84 & 85.16& 86.96 \\
\cite{DBLP:journals/corr/abs-1801-09851} & & 89.10 & \textbf{88.47}& 88.78 \\
\cite{DBLP:journals/corr/abs-1903-10676}** && - & - & 88.94\\
{NER+PA+RL} (This work) & & \textbf{92.05} & 87.91 & \textbf{89.93} \\
\midrule
\cmidrule(r){2-4}
Winner system in \cite{pontiki-etal-2014-semeval}&\parbox[t]{6mm}{\multirow{3}{*}{\rotatebox[origin=RB]{90}{{Laptop}}\rotatebox[origin=RB]{90}{{Review}}}}& \textbf{84.80}  & 66.51 & 74.55\\
{NER+PA+RL} (This work) && 81.07 & \textbf{74.01}& \textbf{77.38} \\
\midrule
\cite{yang-etal-2018-distantly}&\parbox[t]{2mm}{\multirow{2}{*}{\rotatebox[origin=c]{90}{{EC}}}}& {61.57} & 61.33& 61.45 \\
{NER+PA+RL} (This work) && \textbf{61.86} & \textbf{65.36}& \textbf{63.56} \\
\midrule

\cite{yang-etal-2018-distantly}&\parbox[t]{2mm}{\multirow{2}{*}{\rotatebox[origin=c]{90}{\footnotesize{NEWS}}}}& \textbf{81.63} & 76.95 & 79.22 \\
{NER+PA+RL} (This work) && 80.20 & \textbf{79.88}& \textbf{80.04} \\
\bottomrule
\end{tabular}
\caption[NER models comparison.]{NER models comparison. 
*: is the base NER model in our approach and results are reported by \cite{DBLP:journals/corr/abs-1801-09851}.  **: They use Pretrained Contextualized Embeddings for Scientific Text (SciBERT) with an in-domain vocabulary (\textsc{SciVocab}) in \cite{ma-hovy-2016-end} for NER.}
\label{tbl.1}
\end{table}

\section{Performance Comparison}\label{ch3:comparision}
We investigate the impact of the different components of the model (Table \ref{tbl.2}) in the two English datasets via ablation experiments, where we contrast the use of partial annotation learning ({PA}) (see Section \ref{NER+PA}) and the reinforcement-based component ({RL}) (see Section \ref{NER+RL}), with and without the high-quality phrases (the high-quality phrases (\ding{74}) are available only for the English datasets). 

The experiments confirm the efficiency of the {PA} and {RL} modules in resolving FN and FP issues in the distantly labeled datasets. We observe that compared with NER+PA+RL, NER+PA+RL\ding{74} obtains absolute improvements of +1.92 and +0.44 F1 points on the BC5CDR and LaptopReview datasets, respectively. 
Overall, our final system (NER+PA+RL\ding{74}) achieves an improvement of +2.75 and +11.85 F1 on the BC5CDR and LaptopReview respectively over the baseline system NER+PA. The results also corroborate \cite{shang2018learning}, showing that incorporating high-quality phrases always leads to a boost in precision and, subsequently, F1 scores. 

Table \ref{tbl.1} depicts the comparison of our model to the previous NER models. We observe that our final system, the {NER+PA+RL} model with high-quality phrases, achieves higher $F1$ scores on the different datasets compared to the other models. In order to compare to the RL based approach in \cite{yang-etal-2018-distantly}, we run the model without high-quality phrases on the Chinese {EC} and {NEWS} dataset. Our design provides higher F1 scores than \cite{yang-etal-2018-distantly}, where it boosts the reported $F1$ score with $+2.11$ and $+0.82$ points on the {EC} and {NEWS} datasets, respectively. These experiments show that the different design of the RL module leads to improved results.

Following this work, there are some new studies on the BC5CDR and LaptopReview datasets such as \cite{DBLP:journals/corr/abs-1903-10676} and \cite{liu2020hamner}. These approaches generally rely on the use of large, pre-trained language models. \cite{DBLP:journals/corr/abs-1903-10676} achieve $90.01\%$ F1 score in the BC5CDR by fine-tuning \textsc{SciBERT} and \cite{liu2020hamner} report $82.80\%$ F1 score in the LaptopReview using ELMO \parencite{DBLP:journals/corr/abs-1802-05365} trained on the corresponding dataset.

\section{Size of Gold Dataset} \label{ch3:size-data} 
\begin{figure*}
\resizebox{1\linewidth}{!}{
\begin{minipage}[c]{\textwidth}
\centering
 \includegraphics[width=.75\textwidth]{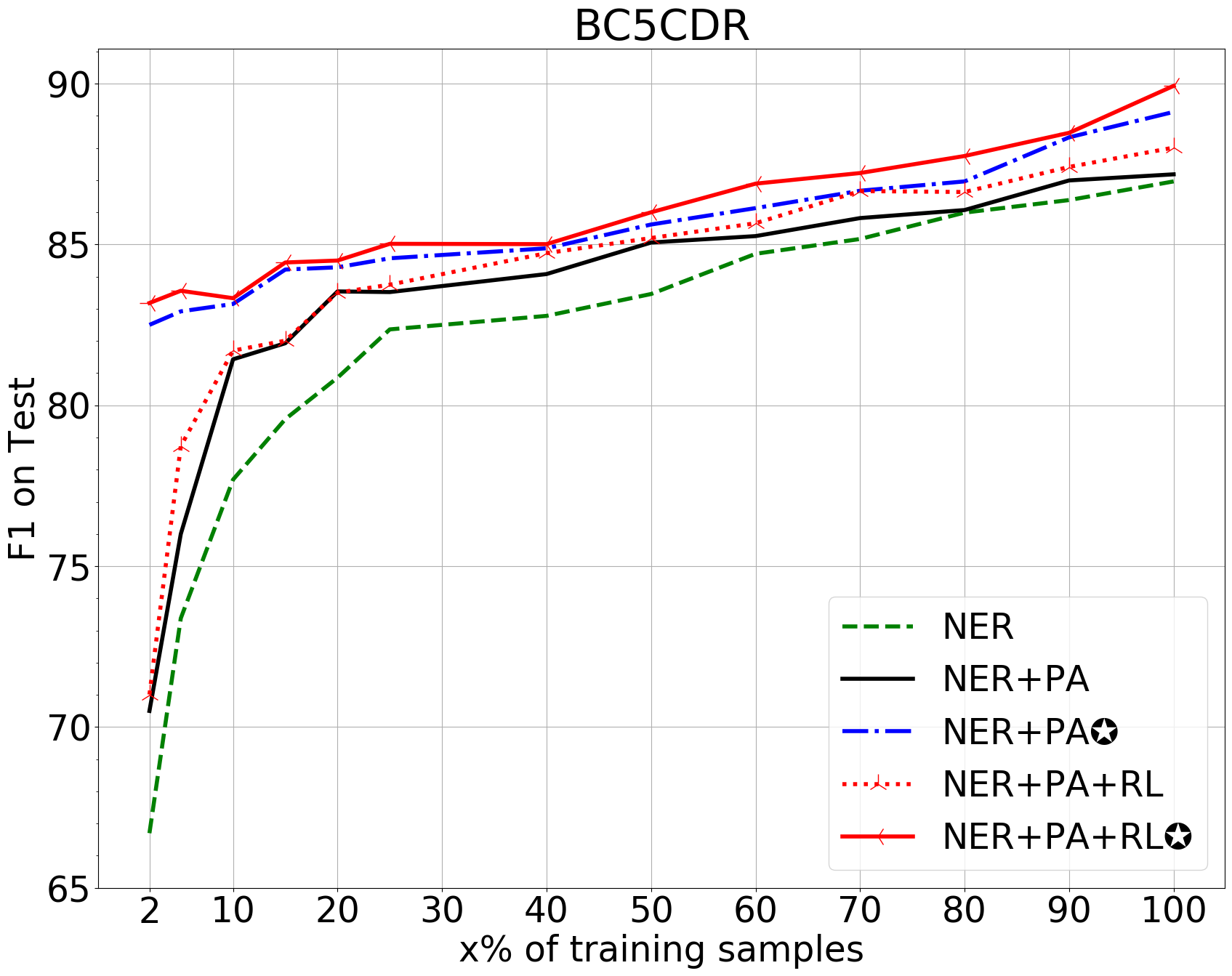}
    \caption[Performance of the different configuration of our model trained on various sizes of annotated dataset in the BC5CDR.]{Performance of the different configuration of our model trained on various sizes of annotated dataset in the BC5CDR. F1 Score on Test vs, the percentage of human annotated sentences.}
    \label{fig:size-1}
\vspace{2mm}
\begin{minipage}{\textwidth}
\centering
\medskip
\centering
  \centering
    \includegraphics[width=.8\textwidth]{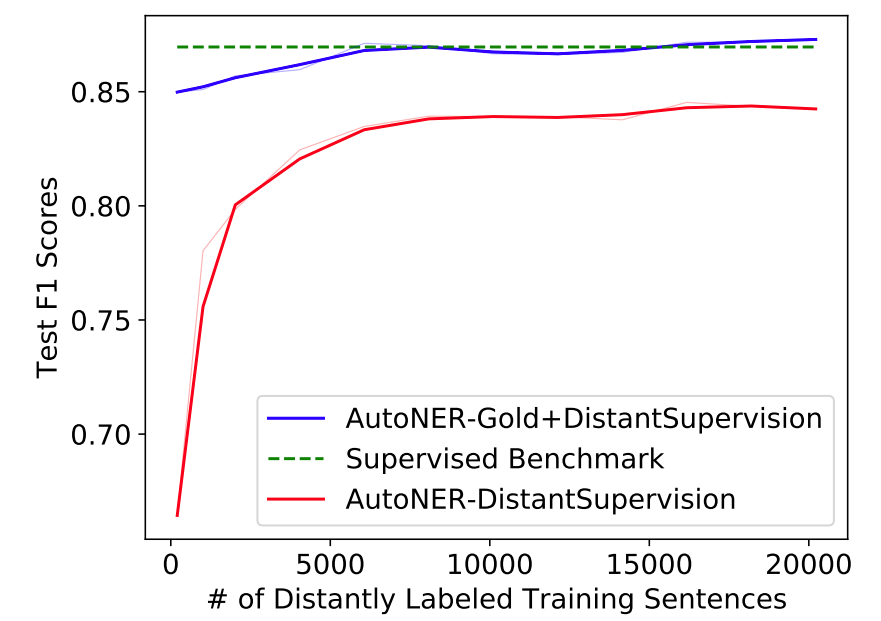}
 \caption[Test F1 score vs. the number of distantly supervised sentences in the BC5CDR dataset.]{Test F1 score vs. the number of distantly supervised sentences in the BC5CDR dataset. The supervised benchmarks with $86.96$ F1 score, is LM-LSTM-CRF ~\parencite{DBLP:journals/corr/abs-1709-04109}  trained with all human-annotate sentences (NER in our experiment). AutoNER-DistantSupervision is the AutoNER  model \parencite{shang2018learning} trained on the selected sentences from distantly labeled data. AutoNER-Gold+DistantSupervision is the AutoNER model trained on both human-annotated and selected distantly labeled sentences.}
 \label{fig:AutoNER-BC5CDR}
\end{minipage}
\end{minipage}
}
\end{figure*}

\begin{figure*}
\resizebox{1\linewidth}{!}{
\begin{minipage}[c]{\textwidth}
\centering
 \includegraphics[width=.75\textwidth]{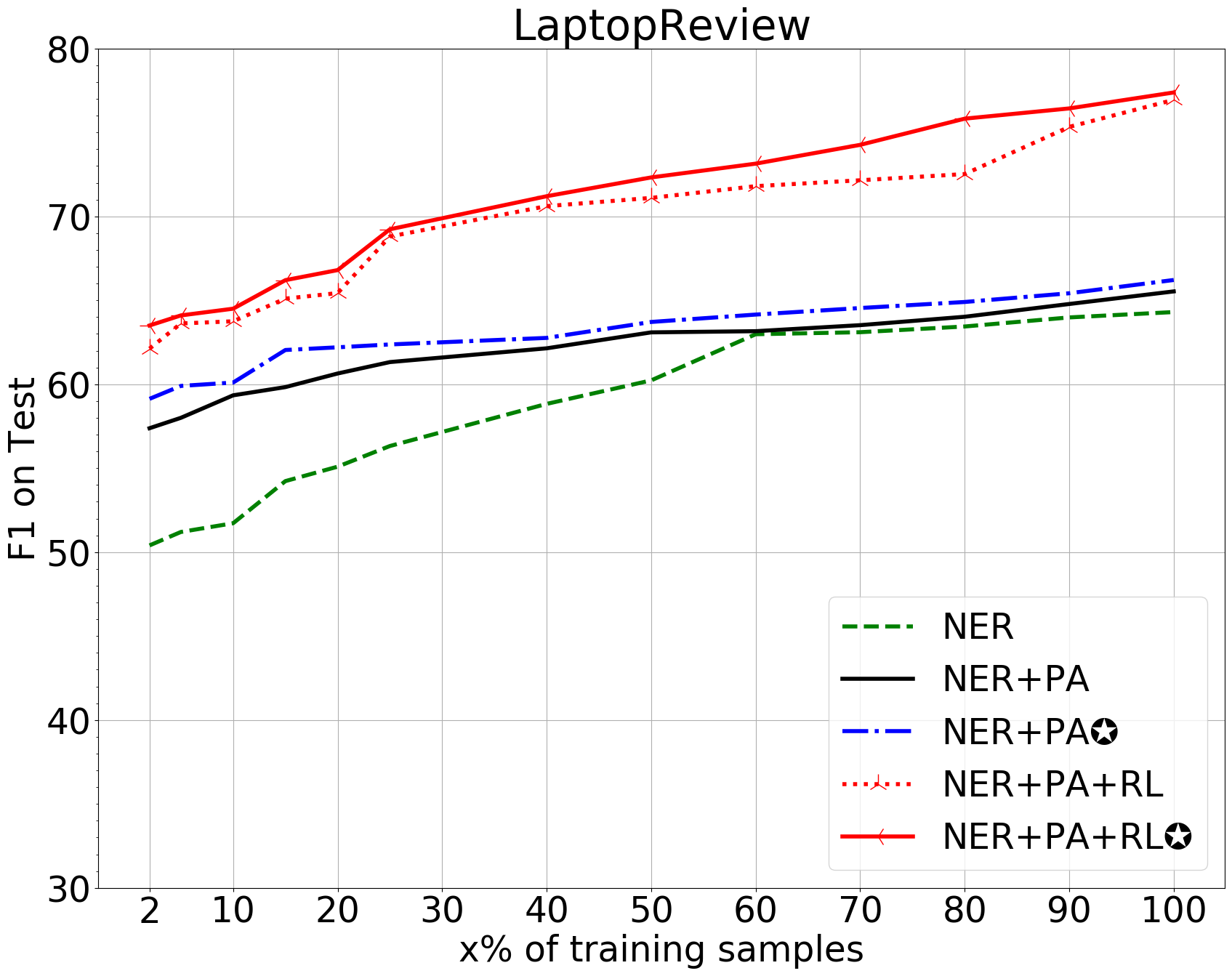}
    \caption[Performance of the different configuration of our model trained on various sizes of annotated dataset in the LaptopReview.]{Performance of the different configuration of our model trained on various sizes of annotated dataset in the LaptopReview. F1 Score on Test vs, the percentage of human annotated sentences.}
    \label{fig:size-2}
\vspace{2mm}
\begin{minipage}{\textwidth}
\centering
\medskip
\centering
  \centering
    \includegraphics[width=.8\textwidth]{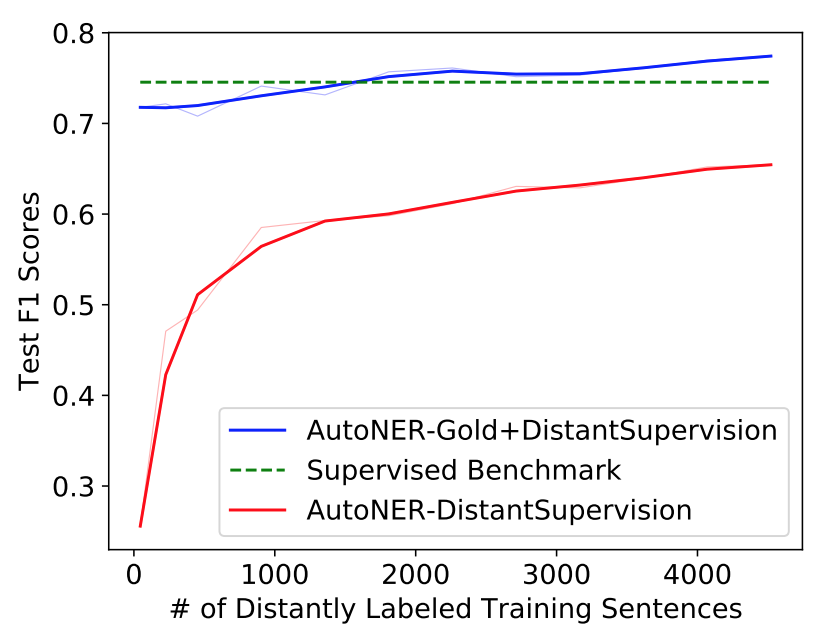}
 \caption[Test F1 score vs. the number of distantly supervised sentences in the LaptopReview dataset.]{Test F1 score vs. the number of distantly supervised sentences in the LaptopReview dataset. The supervised benchmarks with the scores of $74.55$ is the score of the winner system in the SemEval2014 Challenge Task 4 Subtask 1 \parencite{pontiki-etal-2014-semeval}. AutoNER-DistantSupervision is the AutoNER  model \parencite{shang2018learning} trained on the selected sentences from distantly labeled data. AutoNER-Gold+DistantSupervision is the AutoNER model trained on both human-annotated and selected distantly labeled sentences.}
 \label{fig:AutoNER-laptop}
\end{minipage}
\end{minipage}
}
\end{figure*}

In all the previous experiments, we take advantage of the availability of an annotated dataset. However, one of the challenges in domain-specific NER is the availability of gold supervision data. We here examine the performance of our proposed model under settings using different sizes of human-annotated data. In order to conduct this examination with our final method, we focus on the English datasets because of the availability of high-quality phrases.
We proportionally select sub-samples x\% $\in[2,10,20,30,40,50,60,70,80,100]$ from the training data of the BC5CDR and LaptopReview (with random sampling). Figures \ref{fig:size-1} and \ref{fig:size-2} show the performances of the models trained on the selected sentences. The X-axis is the corresponding proportions (x\%) of the human-annotated dataset, while the Y-axis is the F1 scores on the testing set. We observe that the performance of all models (including baselines) improves as more training instances become available. However, as shown in Figures \ref{fig:size-1} and \ref{fig:size-2}, the final method (NER+PA+RL\ding{74}) achieves a performance of $83.18$ and $ 63.50$ with only $2\%$ of the annotated dataset in the BC5CDR and LaptopReview, respectively. Whereas the base NER model requires almost $45\%$ of the ground truth sentences to reach the same performance. This indicates that with a small set of human annotated data, our model can deliver a relatively good performance.

\begin{table}[t]
\centering
\begin{tabular}{*{4}{p{6cm}|c|p{1cm}p{1cm}p{1cm}}}
\toprule
\head{Method} &\head{Data}& Pr.& Re. & F1  \\
\midrule
 {Dictionary Match}& \parbox[t]{2mm}{\multirow{4}{*}{\rotatebox[origin=c]{90}{{BC5CDR}}}}& 93.93 & 58.35 & 71.98 \\
 \cite{DBLP:journals/corr/Fries0RR17}&& 84.98 & 83.49 &	\textbf{84.23}  \\
  \cite{shang2018learning}& & 88.96 & 81.00& 84.80  \\
 {NER+PA+RL}\ding{74} && 88.73 & 77.51 & 82.74  \\ 
  \midrule
  {Dictionary Match}&\parbox[t]{4mm}{\multirow{4}{*}{\rotatebox[origin=RB]{90}{{Laptop}}\rotatebox[origin=RB]{90}{{Review}}}} & 90.68 & 44.65 & 59.84  \\
  \cite{giannakopoulos-etal-2017-unsupervised}&& 74.51 & 31.41 & 44.37 \\
  \cite{shang2018learning} &&  72.27 & 59.79& \textbf{65.44}  \\
  {NER+PA+RL}\ding{74}& & 68.63	&56.88&	62.21  \\
\bottomrule
\end{tabular}
\caption[Unsupervised NER Performance Comparison.]{Unsupervised NER Performance Comparison. The proposed method is trained only on distantly labeled data.}
\label{tbl.3}
\end{table}
We further carry out experiments on the {BC5CDR}  and {LaptopReview} test sets, where our model is trained exclusively on distantly annotated data. We report the outcome together with the scores of the other  state-of-the-art unsupervised methods in Table \ref{tbl.3}, where we also compare to simple dictionary matching. It is clear that the model of \cite{shang2018learning} (AutoNER) is still the best performing NER method on the {BC5CDR} and {LaptopReview} datasets in an unsupervised setup. However, if we compare the performance of our model (NER+PA+RL\ding{74} in Figure \ref{fig:size-1}) with AutoNER trained with both gold training and distantly labeled sentences in the BC5CDR dataset (i.e., AutoNER-GOLD+DistantSupervision in Figure \ref{fig:AutoNER-BC5CDR} taken from \cite{shang2018learning}), we observe that our method provides slightly higher performance ($F$1 score) compared to the AutoNER system \footnote{The absolute F1 score is not reported in the original work. Therefore, we compare our result with the corresponding F1 in Figure \ref{fig:AutoNER-BC5CDR}.} in a similar training scenario (i.e., training with both human annotated and distantly labeled sentences). Furthermore, comparing the performance of our model on the LaptopReview dataset (NER+PA+RL\ding{74} in Figure \ref{fig:size-2}) with AutoNER  (i.e., AutoNER-GOLD+DistantSupervision in Figure \ref{fig:AutoNER-laptop} taken from \cite{shang2018learning}) shows that both systems have quite similar results (i.e., F1 scores) on this dataset.  
It is also worth noting that the approach proposed by \cite{DBLP:journals/corr/Fries0RR17} utilizes extra human effort to design regular expressions and requires specialized hand-tuning.
\section{Summary}
This chapter presents an approach to alleviate the critical shortcoming of auto-generated data in low-resource NER. We propose a performance-driven, policy-based reinforcement learning module that removes the sentences with FPs, whereas the adapted Partial-CRF layer deals with FNs. We examine the impact of each component in ablation experiments. We also found that the proposed model and methodology lead to competitive results on four benchmark datasets from different domains and languages in a supervised setting.

To summarize, the contribution of our model is three-fold. Concretely, we:
\begin{enumerate}[label=(\roman*)]
\item combine the Partial-CRF approach with performance-driven, policy-based  reinforcement learning to clean the noisy, distantly supervised data for low-resource NER in a pre-processing step; 
\item formulate the reward function in RL based on the change in the performance of the NER module where the policy of RL is trained in an unsupervised manner by interaction with the environment; 
\item show that our approach can boost the neural NER system's performance on four datasets from different domains and for two different languages (English and Chinese).
\end{enumerate}

    \chapter{Low-Resource Relation Extraction}
\label{sec:fourth}
Relation extraction is the next step following entity detection in the information extraction pipeline, where semantic relationships are extracted from an input text. Extracted relationships usually occur between two or more named entities (see Section \ref{NE:backgroun} in Chapter \ref{sec:third}) and are classified based on a set of predefined semantic categories. Relation extraction allows us to acquire structured knowledge from unstructured text as explained in Section \ref{sec:ie} of Chapter \ref{sec:first}.
In this chapter, we focus on
relation extraction in a low resource setting, namely the genre and domain of scientific papers in NLP. We study the effect of varying input representations to a  neural architecture, specifically CNN (see Section \ref{ch01:cnns} of Chapter \ref{sec:first}), to extract and classify semantic relations between entities in scientific papers. We investigate the effect of transfer learning using domain-specific word embeddings in the input layer and go on to provide an in-depth investigation of the influence of different syntactic dependency representations which are used to produce dependency paths between the entities in the input to the system. We compare the widely used CoNLL, Stanford Basic, and Universal Dependencies schemes and further compare them with a syntax-agnostic approach. In order to gain a better understanding of the results, we perform manual error analysis.

\section{Introduction}
Over the past years, natural language technology has been used increasingly in computational research for humanities and sciences. It provides an intelligent way for search engines to access scientific literature, and it enables the search engines to respond to complex queries such as \emph{Which papers address a problem using a specific method}, or \emph{What materials or resources have been utilized for a specific problem in the articles}? One of the critical elements of this type of technology is relation extraction and classification.

The neural advances in the NLP field, as explained in Section \ref{sec:dnns} of Chapter \ref{sec:first}, challenge long-held assumptions regarding system architectures. The classical NLP systems, where components of increasing complexity are combined in a pipeline architecture, are being challenged by end-to-end architectures trained on distributed word representations to directly produce different types of analyses traditionally assigned to downstream tasks. Syntactic parsing has been viewed as a crucial component for many tasks aimed at extracting various aspects of meaning from text, but recent work challenges many of these assumptions. For the task of semantic role labeling, for instance, systems that make little or no use of syntactic information, have achieved state-of-the-art results \parencite{Mar:Fro:Tit:17}. For tasks where syntactic information is still viewed as useful, a variety of new methods for the incorporation of syntactic information have been employed, such as recursive models over parse trees \parencite{Ebrahimi2015ChainBR,Soc:Per:Wu:13,}, tree-structured attention mechanisms \parencite{Kok:Pot:17},  multi-task learning \parencite{Wu:Zha:Yan:17}, or the use of various types of syntactically aware input representations, such as embeddings  over syntactic dependency paths \parencite{DBLP:journals/corr/XuMLCPJ15}.

In this chapter, we continue this line of work and present a system based on a CNN architecture over the shortest dependency paths combined with domain-specific word embeddings to extract and classify semantic relations in scientific papers. We investigate the use of different syntactic dependency representations in a neural relation classification task and compare the widely used CoNLL, Stanford Basic, and Universal Dependencies schemes. 
We further compare with a syntax-agnostic approach and perform an error analysis to gain a better understanding of the results. Accordingly, the contributions of this chapter lie in investigating the following research questions:
\begin{question}\label{rq.4.1}
   Are domain-specific input representations beneficial for relation extraction task?
\end{question}
\begin{question}\label{rq.4.2}
    What is the impact of syntactic dependency representations in low-resource neural relation extraction?
\end{question} 
\begin{question}\label{rq.4.3}
 Which kind of syntactic dependency representation is most beneficial for neural relation extraction and classification?
\end{question}
\section{Previous Work}
Relation extraction and classification can be defined as follows: given a sentence where entities are manually annotated, we aim to identify the pairs of entities that are instances of the semantic relations of interest and classify them based on a pre-defined set of relation types.
Different approaches have been applied to solve the task of relation extraction and classification in previous work.
The traditional studies mainly focus on feature-based methods. Almost all systems submitted to SemEval 2010 task 8 \footnote{SemEval 2010 task 8: Multi-Way Classification of Semantic Relations Between Pairs of Nominals} \parencite{Hendrickx:2010:STM:1859664.1859670}, used either Maximum Entropy or Support Vector Machine classifiers. These systems made use of contextual, lexical, and syntactic features combined with richer linguistic and background knowledge such as WordNet and FrameNet \parencite{Hendrickx:2010:STM:1859664.1859670, Rink:2010:UCS:1859664.1859721}.

The re-emergence of neural networks provides a way to develop highly automatic features and representations to handle complex interpretation tasks. These approaches have yielded impressive results for many different NLP tasks.  
In the relation classification task, the use of deep neural networks has been investigated in several studies \parencite{Socher:2012:SCT:2390948.2391084,P16-1200,DBLP:conf/acl/ZhouSTQLHX16}.
There are three widely used deep neural networks (DNNs) architectures (see Section \ref{sec:dnns} of Chapter \ref{sec:first} for more details) used for relation extraction: Convolutional neural networks (CNNs), Recurrent neural networks (RNNs), and hybrid models which combine these two types of models.
Recently, pre-trained language model, as explained in Section \ref{sec:pre-trained-LM} of Chapter \ref{sec:first}, such as BERT \parencite{DBLP:journals/corr/abs-1810-04805} is used for several tasks in scientific text. For example \textsc{SciBERT} \parencite{DBLP:journals/corr/abs-1903-10676} leverages BERT on a large multi-domain corpus
of scientific publications to improve performance on downstream scientific NLP tasks.
Subsequently, \cite{wang-etal-2019-extracting} and \cite{jiang2020improving} employed BERT and \textsc{SciBERT} for the scientific relation classification task.

In previous work, CNNs have been effectively applied to extract lexical and sentence level features for relation classification \parencite{DBLP:journals/corr/ZhangW15a,DBLP:journals/corr/LeeDS17,W15-1506}.
Sentences or the context between two target entities are used as input for the CNNs. Such representations suffer from irrelevant sub-sequences or clauses when target entities occur far from each other, or there are other target entities in the same sentence. To avoid negative effects from irrelevant chunks or clauses and capture the relation between two entities, the researchers proposed methods that can embed syntactic tree features within a neural architecture. 
The \emph{shortest dependency path} (\emph{sdp}) between two entities is frequently used for relation classification. The \emph{sdp} between two entities in the dependency graph captures a condensed representation of the information required to assert a relationship between two entities \parencite{Bunescu:2005:SPD:1220575.1220666}. \cite{DBLP:journals/corr/XuFHZ15}, \cite{DBLP:journals/corr/LiuWLJZW15} and \cite{DBLP:journals/corr/XuMLCPJ15} employ DNNs such as CNNs and RNNs to learn more robust and effective relation representations from the \emph{sdp} between two entities. Their experiments on the SemEval 2010 relation classification data set show that \emph{sdp} can be a valuable resource for relation classification by covering highly relevant information of target relations.

Dependency representations have by now become widely used representations for syntactic analysis, often motivated by their usefulness in downstream applications. There is currently a wide range of different types of dependency representations in use, which vary mainly in terms of choices concerning syntactic head status.  Some previous studies have examined the effects of these choices in various downstream applications \parencite{Miy:Sae:Sag:08,Elm:Joh:Kle:13}. Most recently, two Shared Tasks on Extrinsic Parser Evaluation \parencite{Oep:Ovr:Bjo:17,Oep:Far:Ovr:18} were aimed at providing better estimates of the relative utility of different types of dependency representations and syntactic parsers for downstream applications. However, the downstream systems in this previous work have been limited to traditional (non-neural) systems, and there is still a need for a better understanding of the contribution of syntactic information in neural downstream systems. 
\section{SemEval 2018 Task 7}
In this chapter, we employ the data sets released for the SemEval 2018 task 7 \parencite{SemEval2018Task7}, which encode relation instances between scientific concepts. The relations belong to a set of semantic categories that are related to the science domain, and their instances are frequently used in abstracts and introductions of scientific articles. The shared task provides systematic evaluation steps that are essential for complete information extraction from scientific text. The concepts represent domain entities specific to the scientific discipline of Natural Language Processing (NLP). The task consists of three sub-tasks, where the first two sub-tasks are dedicated to the classification of relation instances, and the last one is devoted to the full task of extracting the relation instances, as well as classifying them. Our system participated in this task and ranked third (out of 28) participants in the overall evaluation.

The data that is provided in the task is extracted from the abstract section of scientific papers from the ACL Anthology corpus \parencite{GBOR16.870}. 
Each sub-task makes available 350 annotated abstracts and 150 abstracts as training and test data, respectively. The training and test data sets contain pre-annotated domain entities. Furthermore, the relation instances along with their directionality, are provided in both the training and the test data sets of the classification sub-tasks.
The test data provided for the extraction sub-task, on the other hand, does not contain the relation instances. Below, we will describe the sub-tasks in more detail.
\begin{sidewaystable}
  \centering
    \scalebox{1}{
  \begin{tabular}{*{3}{lll}}
    \toprule
    \head{Relation Type} & \head{Explanation} & \head{Example} \\
    \midrule
    USAGE & Methods, tasks, and data are linked by usage relations. &     \\
   {used by} &{\emph{ARG1}: method, system \emph{ARG2}: other method} &{approach - model}\\
    {used\_for\_task} & {\emph{ARG1}: method,system \emph{ARG2}: task} & { approach - parsing}\\
    {used\_on\_data} & {\emph{ARG1}: method applied to \emph{ARG2}: data} & {MT system - Japanese}\\
    task\_on\_data & {\emph{ARG1}: task performed on \emph{ARG2}: data} &{ parse - sentence}\\
    \midrule
    RESULT & An entity affects or yields a result.     \\
   {affects} &{\emph{ARG1}: specific property of data \emph{ARG2}: results} &{order - performance}\\
    {problem} & {\emph{ARG1}:  phenomenon is a problem in a \emph{ARG2}:  field/task} & { ambiguity - sentence}\\
    {yields} & {\emph{ARG1}: : experiment/method \emph{ARG2}: result} & { parser - performance}\\ 
     \midrule
    MODEL & An entity is a analytic characteristic or abstract model of another entity.    \\
   {char} &{\emph{ARG1}: observed characteristics of an observed \emph{ARG2}: entity} &{order - constituents}\\
    {model} & {\emph{ARG1}:  abstract representation of an  \emph{ARG2}:  observed entity} & {interpretation - utterance}\\
    {tag} & {\emph{ARG1}: tag/meta-information associated to an \emph{ARG2}: entity} & { categories - words}\\ 
     \midrule
    PART\_WHOLE & Entities are in a part-whole relationship.    \\
   {composed\_of} &{\emph{ARG1}: database/resource \emph{ARG2}: data} &{  ontology - concepts}\\
    {datasource} & {\emph{ARG1}: information extracted from  \emph{ARG2}:  kind of data} & {knowledge - domain}\\
    {phenomenon} & {\emph{ARG1}: : entity, a phenomenon found in \emph{ARG2}: context} & {expressions - text}\\ 
     \midrule
      TOPIC & This category relates a scientific work with its topic.   \\
   {propose} &{\emph{ARG1}: : paper/author presents \emph{ARG2}: an idea} &{paper - method}\\
    {study} & {\emph{ARG1}:  analysis of a  \emph{ARG2}:  phenomenon} & { research - speech}\\
     \midrule
     COMPARISON & An entity is compared to another entity.    \\
   {compare} &{\emph{ARG1}: result, experiment compared to \emph{ARG2}: result, experiment} &{result, standard}\\
  
    \bottomrule
  \end{tabular}}
   \caption[Semantic relation typology and the coarse relations in SemEval 2018 Task 7.]{Semantic relation typology and the coarse relations from a finer grained ones that are used in annotation process (Source: \cite{SemEval2018Task7}).}
   \label{tbl:relationtypology}
\end{sidewaystable}
\subsection{Relation classification scenario} \label{rel-classes}
The task of relation classification on this data set is to predict the semantic relation between a given pair of entities within the abstract of a scientific paper. The semantic relation set contains five asymmetric relations ({\small \emph{USAGE, RESULT, MODEL-FEATURE, PART\_WHOLE,
TOPIC}}) and one symmetric relation ({\small \emph{COMPARE}}). Each abstract in the training dataset contains pairs of entities that are assigned to one of these six relations.
Table \ref{tbl:relationtypology} shows the semantic relation typology of the six major relation types and their definitions, along with some example entity pairs. 

There are two sub-tasks in this relation classification scenario: classification on clean data and classification on noisy data.
\paragraph{Sub-task 1.1: Relation classification on clean data:} \label{sub-task1.1} The entities and corresponding relations have been manually annotated in the training and test dataset. The test dataset contains the unlabeled relation instances, and the task is to predict the label for each entity pair. For example in the text snippet in example \ref{ex:snippet}, the relation instance holds between two entity identifiers, i.e., \emph{(P05-1057.3, P05-1057.4)} and the relation label should be predicted as : \emph{{\small USAGE} (P05-1057.3, P05-1057.4)}.
\begin{covexample}\label{ex:snippet} {\small <entity id='P05-1057.3'> All knowledge sources </entity> \\are treated as <entity id='P05-1057.4'> feature functions </entity>}
\end{covexample}

\paragraph{Sub-task 1.2: Relation classification on noisy data:}
In sub-task 1.2 the entities have been automatically annotated based on a combination of terminology extraction (Saffron Knowledge Extraction Framework \parencite{bordea2013domain}) and available ontological resources such as WordNet \parencite{Miller:1995:WLD:219717.219748} and BabelNet \parencite{navigli2012babelnet}. Therefore, the dataset contains a fair amount of noise (verbs, irrelevant words). The terms include high-level terms (e.g. \emph{algorithm}, \emph{paper}, \emph{method}) and are not always full NPs.
 The example sentence in \ref{ex:snippet-noisy} shows an instance from the test dataset for this sub-task, where we observe the noisy entity assignments which incorrectly predicts an entity label for {\it challenging}.
\begin{covexample}\label{ex:snippet-noisy} {\small Morphological <entity id='N06-1042.14'> ambiguity </entity>\\ (e.g. lives = live+s or life+s) is a <entity id='N06-1042.15'> challenging </entity> <entity id='N06-1042.16'> problem </entity> for agglutinative <entity id='N06-1042.17'> languages </entity>}
\end{covexample}
The relation instance in example \ref{ex:snippet-noisy} is \emph{(N06-1042.14, N06-1042.17)} and the task is to predict the label as : \emph{ {\small MODEL-FEATURE} (N06-1042.14, N06-1042.17)}.

\subsection{Relation extraction and classification scenario}\label{sub-task2}
Given an abstract and pre-annotated entities, the goal of the extraction and classification task is: 1) To find entity pairs that are in a relation 2) To predict the relation label (as in the classification sub-tasks) and its direction. The training data contains manually annotated entities and labeled semantic relations that hold between these along with the directionality of the relation. The dataset is identical to the one provided for sub-task 1.1. In the test dataset, only abstracts and annotated entities are given, and participants are asked to predict the entity pairs, their relation types, and the direction of the relations. For instance, in the following sentence in example \ref{ex:snippet-task2}, the entity pairs \emph{(H01-1001.5, H01-1001.7, {\small REVERSE})} and \emph{(H01-1001.9, H01-1001.10)} should be identified and classified with the \emph{{\small USAGE}} label.
\begin{covexample}\label{ex:snippet-task2} 
{\small Traditional <entity id='H01-1001.5'> information retrieval </entity> use a <entity id='H01-1001.6'> histogram </entity> of <entity id='H01-1001.7'> keywords </entity> as the <entity id='H01-1001.8'> document representation </entity> but <entity id='H01-1001.9'>oral communication</entity> may offer additional <entity id='H01-1001.10'> indices </entity> such as the time and place of the rejoinder and the attendance.}
\end{covexample}
\section{Evaluation Metrics}
Following the SemEval 2018 task 7 \parencite{SemEval2018Task7}, each sub-task is evaluated differently.
For sub-tasks 1.1 and 1.2, which are classification tasks, the following evaluation metrics are used:
\begin{itemize}
\setlength{\itemsep}{4pt}
    \item {\bf Relation class-wise:} Precision, recall, and F1-measure ($\beta$=1) for each semantic relation label.
    \item {\bf Global:} Macro-average and Micro-average F1 scores evaluated for every distinct relation label.
\end{itemize}
For sub-task 2, evaluation is conducted for each step (i.e., Extraction and Classification).
The quality of the extraction step is evaluated based on the standard measures of Precision, Recall, and F1, where the label and directionality of the relations are ignored in the calculation.
In the classification step, the same evaluation metrics as sub-task 1.1 and 1.2 are employed. However, only correctly connected entities with correct directions (when relevant) and labels are considered as a correct instance. 
Here, we report the official scores of each experiment's task, i.e., for the classification tasks, we report Macro-average F1, and we report the F1 score for the extraction task.
\section{System Design}\label{ch4:model}
In this section, we describe the various components of our
system. We introduce the input data's specifics in terms of pre-processing, label encoding, and word embeddings and further introduce the architecture of our CNN system for relation extraction and classification.
\subsection{Dataset preparation}
For each relation instance, in the training data set, the sentence containing the participant entities is considered as a text representation of the relation instance.
Therefore, if two relations appear in one sentence, they will have the same text representation. 
Sentence and token boundaries are detected using the Stanford CoreNLP tool \parencite{manning-EtAl:2014:P14-5}. Since most of the entities are multi-word units, in order to obtain a precise dependency path between entities, we replace the participant entities in the relation instance with their codes prior to parsing. The example sentence in \ref{ex:sent} below is thus transformed to (\ref{ex:ent}).\footnote{Preliminary results showed that this replacement technique improved results for relation extraction classification.}
\begin{covexamples}
\setlength{\itemsep}{5pt}
\item\label{ex:sent} {\small All knowledge sources are treated as feature functions.}
\item\label{ex:ent} {\small All P05\_1057\_3 are treated as P05\_1057\_4.}
\end{covexamples}
\noindent Given an encoded sentence, we obtain the shortest dependency path connecting two target entities for each relation instance using a syntactic parser, see below.

For syntactic parsing, we employ the parser described in \cite{Boh:Niv:12}, a transition-based parser that performs joint PoS-tagging and parsing. We train the parser on the standard training sections 02-21 of the Wall Street Journal (WSJ) portion of the Penn Treebank \parencite{Marcus:1993}. The constituency-based treebank is converted to dependencies using two different conversion tools: (i) the pennconverter
software\footnote{\url{http://nlp.cs.lth.se/software/treebank-converter/}} \parencite{Joh:Nug:07}, which produces the CoNLL dependencies\footnote{The pennconverter tool is run using the \emph{rightBranching=false} flag.}, and (ii) the Stanford parser using either the option to produce basic dependencies \footnote{The Stanford parser is run using the \emph{-basic} flag to produce the basic version of Stanford dependencies.} or its default option which is Universal Dependencies v1.3\footnote{Note, however, that the Stanford converter does not produce UD PoS-tags, but outputs native PTB tags.}.
The parser achieves a labeled accuracy score of 91.23 when trained on the CoNLL08 representation, 91.31 for the Stanford basic model, and 90.81 for the UD representation, when evaluated against the standard evaluation set (section 23) of the WSJ. 
We acknowledge that these results are not strictly speaking state-of-the-art parse results for English. However, the parser is straightforward to use and re-train with the different dependency representations. We also compare to another widely used parser, namely the pre-trained parsing model for English included in the Stanford CoreNLP toolkit \parencite{manning-EtAl:2014:P14-5}, which outputs Universal Dependencies only. However, it was clearly outperformed by our version of the \cite{Boh:Niv:12} parser in the initial development experiments.

Based on the dependency graphs output by the parser, we extract the shortest dependency path connecting two entities. The path records the direction of arc traversal using left and right arrows (i.e., $\leftarrow$ and $\rightarrow$) as well as the dependency relation of the traversed arcs and the predicates involved, following \cite{DBLP:journals/corr/XuFHZ15}. The entity codes in the final path are replaced with the corresponding word tokens at the end of the pre-processing step. For the sentence in (\ref{ex:sent}) and the two entities \textit{knowledge sources} and \textit{feature functions} we thus extract the path in (\ref{ex:path}) below.

\begin{covexample}
\label{ex:path}
{\small knowledge sources $\leftarrow$ SBJ $\leftarrow$ are $\rightarrow$ VC $\rightarrow$ treated $\rightarrow$ ADV $\rightarrow$ as $\rightarrow$ PMOD $\rightarrow$ feature functions}
\end{covexample}
\noindent Since the related entity pairs and the relation types are provided for the full dataset, we extend the dataset for sub-task 1.1 and 2 by extracting the related entities and their corresponding \emph{sdp} from the sub-task 1.2 dataset. In order to train a model for sub-task 2, we also augment the dataset by extracting \emph{NONE} relation instances (see Section \ref{encoding}), from the corresponding dataset. Table \ref{dataset} shows the number of instances for each relation class. As we can see, the class distribution is clearly unbalanced.
\begin{table}[t]
  \centering
  \begin{tabular}{*{6}{lrrrrr}}
    \toprule
    &\multicolumn{2}{c}{\head{Sub-task}} & \multicolumn{2}{c}{\head{Reverse}} & \\
    \cmidrule(ll){2-3}
    \cmidrule(ll){4-5}
    \head{Relation} &1.1 \& 2 & 1.2& False & True & \head{Total} \\
    \midrule
    {USAGE} &483 & 464 &615 &332 &947    \\
        {MODEL-FEATURE} & 326& 172&346 & 152    & 498 \\  
        {RESULT}&72&121 & 135 & 58 & 193 \\
         {TOPIC} &18&240& 235 & 23 &  258\\
         {PART\_WHOLE}&233& 192& 273 & 152 & 425 \\
          {COMPARE} &95&41& 136 & - &136 \\
          {NONE} & 2315&-&2315& - & 2315 \\
    \bottomrule
  \end{tabular}
   \caption{Number of instances for each relation in the final dataset.}
   \label{dataset}
\end{table}
\subsection{Label encoding}\label{encoding} The classification sub-tasks contain five asymmetric and one symmetric classes (see Section \ref{rel-classes}). The relation instances, along with their directionality, are provided in both the training and the test data sets. For these sub-tasks, we therefore use the same labels in our system. 
For sub-task 2, which combines the extraction and classification tasks, however, we construct an extra set of relation types. First, we collect every pair of entities within a single sentence that are not included in the annotated relation set. To minimize the noise, we retain only the entity pairs which are not further away than $6$ tokens. From these entity pairs, we generate negative instances with the \emph{{\small NONE}} class and extract the corresponding \emph{sdp}. Second, to preserve the directionality in the asymmetric relations, we add the $\neg$ symbol to the instances with reverse directionality (e.g.,\emph{{\small USAGE}(e1,e2,{\small REVERSE})} becomes  \emph{$\neg${\small USAGE}(e1,e2)}). The final label set for sub-task 2 thus consists of 12 relations.  
\subsection{Word embeddings} 
In our system, following the sequential transfer learning of word embeddings (see Section \ref{sec:emb} in Chapter \ref{sec:first}), two different sets of pre-trained word embeddings are used for initialization. One is the 300-d pre-trained embeddings provided by the NLPL repository \footnote{\url{http://vectors.nlpl.eu/repository/}}\parencite{Velldal}, trained on English Wikipedia data with word2vec \parencite{Mikolov2013a}, here dubbed wiki-w2v. In Chapter \ref{sec:second}, we saw that domain-specific embeddings perform better compared to the general domain embeddings trained on much larger input data. Therefore, we train a second set of domain-specific embeddings on the ACL Anthology corpus. We obtain the XML versions of 22,878 articles from ACL Anthology \footnote{\url{https://acl-arc.comp.nus.edu.sg/}}. After extracting the raw texts, for training of the 300-d word embeddings (acl-w2v), we exploit the available word2vec \parencite{Mikolov2013a} implementation \emph{gensim} \parencite{gensim} for training.  
\subsection{Classification Model}

\begin{figure}[t]
\centering
\includegraphics[width=1\textwidth]{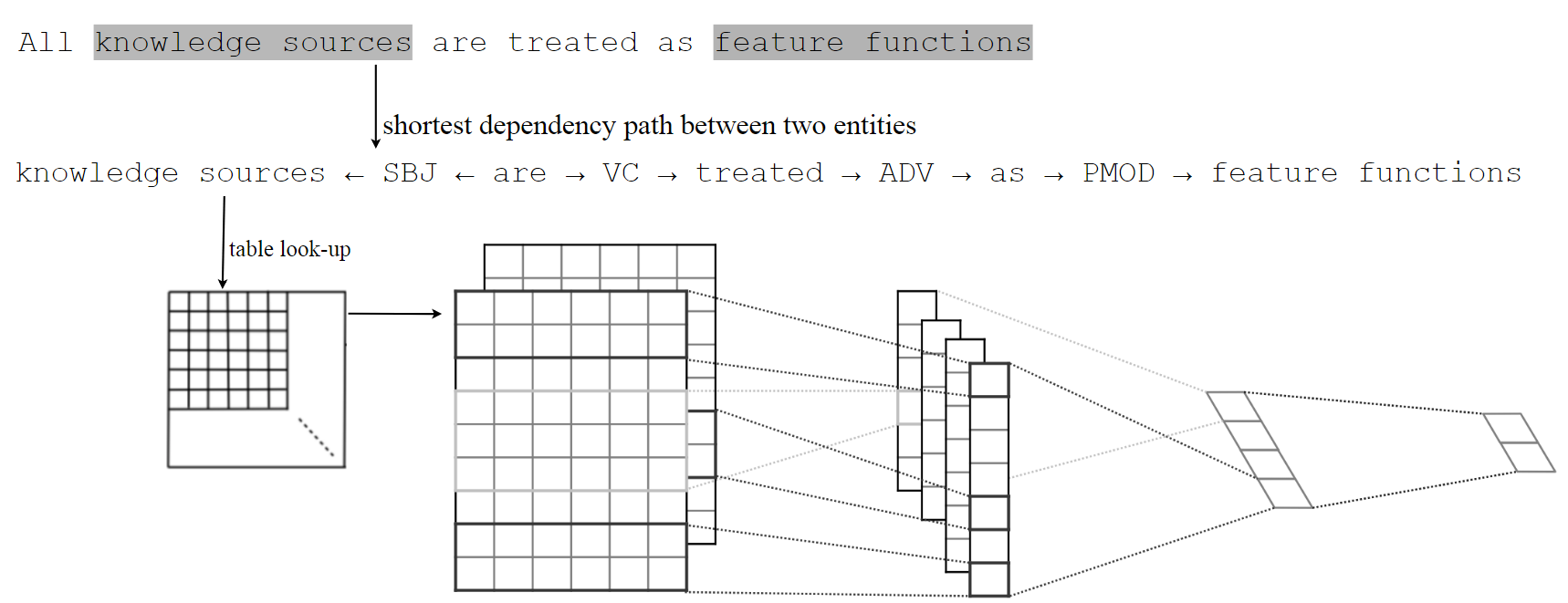}
\caption{Model architecture with two channels for an example shortest dependency path.}
\label{fig:cnn}
\end{figure} 
At the time of the SemEval task, the dominant approaches in relation extraction using the shortest dependency path as an input representation generally involve a CNN architecture. 
We design our system based on a CNN architecture similar to the one used for sentence classification by \cite{DBLP:journals/corr/Kim14f} (see Section \ref{ch01:cnns} in Chapter \ref{sec:first} and also employed for sentence classification in Chapter \ref{sec:second}). 
Figure ~\ref{fig:cnn} provides an overview of the proposed model. 
It consists of 4 main layers as follows:
\begin{enumerate}
\setlength{\itemsep}{4pt}
    \item {\bf Look-up Table and Embedding layer:} In the first step, the model takes the shortest dependency path (i.e., the words, dependency edge directions, and dependency labels) between entity pairs as input and maps it into a feature vector using a look-up table operation. Each element of the dependency path (i.e., word, dependency label, and arrow) is transformed into an embedding layer by looking up the embedding matrix $ M \in \mathcal{R}^{d\times V} $, where $d$ is the dimension of the CNN embedding layer, and $V$ is the size of the vocabulary. Each column in the embedding matrix can be initialized randomly or with pre-trained embeddings. The dependency labels and edge directions are always initialized randomly and fine-tuned during model training.
    \item  {\bf Convolutional Layer}: The next layer performs convolutions with ReLU activation over the embeddings using multiple filter sizes, and extracts feature maps.
    \item {\bf Max pooling Layer}: By applying the \emph{max} operator, the most effective local features are generated from each feature map. 
    \item {\bf Fully connected Layer}: Finally, the higher-level syntactic features are fed to a fully connected \emph{softmax} layer, which outputs the probability distribution over each relation.
\end{enumerate}
\section{Initial Experiments}
In an initial round of experimentation, we assess the influence of different word embedding models for our task. Specifically, we contrast the use of general domain embeddings with domain-specific word embeddings. We also assess the use of a two-channel architecture (see Section \ref{ch01:cnns} in Chapter \ref{sec:first}) for the incorporation of pre-trained word embeddings in our model.
\subsection{Model settings} \label{modelvariant}
In the initial experiments, we keep the value of the model hyper-parameters equal to the ones that are reported by \cite{DBLP:journals/corr/Kim14f}, i.e., $128$ filters for each window size, a dropout rate of $\rho=0.5$ and $l_2$ regularization of $3$.
To deal with the effects of class imbalance,  we weight the cost by the ratio of class instances. Thus each observation receives a weight, depending on the class it belongs to. The effect of the minority class observations is thereby increased simply by a higher weight of these instances and is decreased for majority class observations. Furthermore, to guarantee that each fold in $n$-fold cross-validation will have the same distribution of classes during training, development, and test, we apply the stratification technique proposed by \cite{Sechidis:2011:SMD:2034161.2034172}.
We use the development set to detect when overfitting starts during
the training of our model; using \emph{early stopping}, training is then stopped before convergence to avoid overfitting \parencite{Prechelt:1998:ES:645754.668392}. As described above, the official evaluation metric is the macro-averaged F1-score. Therefore we implement early-stopping with \emph{patience}= $20$ (i.e., the number of epochs to wait before early stop if no progress on the development set) based on the macro-F1 score in the development set.
\subsection{Model variants}
We run experiments with several variants of the model. In particular, we here contrast the use of pre-trained (general vs. domain-specific) and randomly initialized word embeddings in the input layer, and the use of one or two channels (see Section \ref{ch01:cnns} in Chapter \ref{sec:first}). Specifically, we compare the following model variants:
\begin{itemize}
\setlength{\itemsep}{4pt}
    \item \textbf{cnn.rand:} A baseline model, where all elements in the embedding layer are randomly initialized and updated in the training process.
    \item  \textbf{cnn.wiki-w2v:} The embedding layer is initialized with the pre-trained Wikipedia word embeddings and fine-tuned for the target task.
    \item \textbf{cnn.acl-w2v}: The embedding layer is initialized with the pre-trained ACL Anthology word embeddings and fine-tuned for the target task.
    \item \textbf{cnn.multi.rand:} There are two embedding layers as a 'channel' in the CNN architecture. Both channels are initialized randomly, and only one of them is updated during training while the other remains static.
    \item \textbf{cnn.multi.wiki-w2v:} Same as before, but the channels are initialized with Wikipedia embedding vectors.
    \item \textbf{cnn.multi.acl-w2v:} The two channels are initialized with ACL embedding vectors.
    \item \textbf{cnn.multi.wiki-w2v.rand:} First, the channel is initialized with Wikipedia embeddings in static mode and the second initialized randomly with a non-static mode.
    \item \textbf{cnn.multi.acl-w2v.rand:} Same  as  previous  setting, but  the  first channel makes use of ACL embeddings. 
\end{itemize}
\subsection{Results}
During development, we first investigated the performance of different model variants (see Section \ref{modelvariant}) using the Universal Dependency representation output by the Stanford CoreNLP toolkit; by running $5$-fold cross-validation. The data set is split into five folds. In the first iteration, the first fold is used to test the model, and the rest is used to train the model (i.e., three folds for training and one fold for development set to perform early stopping). In the second iteration, the second fold is used as the testing set, while the rest serve as the training set. This process is repeated until each fold of the five folds has been used as the testing set. The experiments  (Table \ref{result-model-vairant}) show that the multi-channel mode performs better only in the classification sub-tasks compared to the single-channel setting. 
The use of the pre-trained embeddings helps the model in class assignments. Notably, the domain-specific embeddings (i.e., acl-w2v) provide higher performance gains when used in the model.
\begin{table}[t]
  \centering
 \resizebox{1\linewidth}{!}{
  \begin{tabular}{*{3}{p{4cm}P{1.5cm}P{1.5cm}P{1.5cm} P{1.5cm}}}
    &\multicolumn{4}{c}{\head{Sub-task}}\\
    \cmidrule(lr){2-5}
    \head{Model}&\multirow{ 2}{*}{\head{1.1}}&\multirow{ 2}{*}{\head{1.2}}&\multicolumn{2}{c}{\head{2}}\\
    \cmidrule(lr){4-5}
    &&&Ext.&Class.\\
    \midrule
    cnn.rand& 68.86   & 73.47 &  72.33 &	54.62 \\
    cnn.wiki-w2v& 70.14 & 74.20& 72.50 &	54.20 \\
    cnn.acl-w2v& 72.74 &75.69 & {\bf 72.74}&	{\bf57.56} \\
    cnn.multi.rand& 68.30 &74.11&72.56 &	55.16 \\
    cnn.multi.wiki-w2v& 68.07 &75.01&72.59 &	55.30\\
    cnn.multi.acl-w2v&72.85&75.83&72.63 &	55.45\\
    cnn.multi.wiki-w2v.rand& 69.85& 75.58&72.70 &	56.69\\
    cnn.multi.acl-w2v.rand&{\bf73.06}&{\bf 76.36}&72.05 &	56.99 \\
     \bottomrule
  \end{tabular}
 }
   \caption{F1.(avg. in 5-fold) scores for different model setting during training.}
   \label{result-model-vairant}
\end{table}

Further, we experiment with the selected configuration for each task using different dependency representations to produce the shortest paths between entities.  Table \ref{result-different-dep} presents the F1-score of each dependency representation for each sub-task via $5$-fold cross-validation on the training data. In the evaluation period, we re-run $5$-fold cross-validation using the selected model for each sub-task. However, in this setting we use four folds as training and one fold as a development set, and we apply the output model to the evaluation dataset. The results indicate that the Stanford Basic scheme performs best in the classification subtask, whereas the CoNLL representation provides the highest result in the full extraction task.

The comparison of different syntactic representations is potentially problematic; however, given that the default hyper-parameters may favor one of the representations simply by chance. Ideally, the hyper-parameters should be tuned for each dependency representation in turn to enable a fair comparison. In the next sections, we apply Bayesian Optimization to tune our hyper-parameters and provide an analysis of the influence of syntax and various syntactic dependency representations in our system.

\begin{table}[t]
  \centering
  \begin{tabular}{*{3}{p{4cm}P{1cm}P{1cm}P{1cm} P{1cm}}}
  \toprule
   \multirow{4}{*}{\head{Representation}} &\multicolumn{4}{c}{\head{Model/Sub-task}}\\
    \cmidrule(lr){2-5}
    &\multicolumn{2}{c}{\head{cnn.multi.acl-w2v.rand}}&\multicolumn{2}{c}{\head{cnn.acl-w2v}}\\
    \cmidrule(lr){2-5}
     &\multirow{ 2}{*}{\head{1.1}}&\multirow{ 2}{*}{\head{1.2}}&\multicolumn{2}{c}{\head{2}}\\
     \cmidrule(lr){4-5}
    &&&Ext.&Class.\\
    \midrule
    Stanford Basic& {\bf 74.16}  &{\bf  77.70}  &  72.91 & 58.11	 \\
    CoNLL08 &  72.65  & 76.83 & {\bf 74.26} &	{\bf 60.31 }	 \\ 
    UD v1.3 &  69.55   & 76.60  & 71.09&  54.53	  \\
    UD (Stanford CoreNLP) & 73.06    & 76.36  & 72.74 &	57.56 	  \\
    
     \bottomrule
  \end{tabular}
   \caption{F1.(avg. in 5-fold) scores for different dependency representation during training.}
   \label{result-different-dep}
\end{table}
\subsection{Participating systems and results in SemEval 2018}
\begin{table}[t]
\centering
\begin{tabular}{*{7}{l}}
\toprule
   &  \multicolumn{2}{c}{\head{\nth{1}}} & \multicolumn{2}{c}{\head{\nth{2}}} 
      & \multicolumn{2}{c}{\head{\emph{mv}}}\\
\cmidrule(r){2-3}
\cmidrule(r){4-5}
\cmidrule(r){6-7}
\head{Sub-task}  & Ext.& Class. &Ext.& Class.&Ext.& Class. \\
\midrule
1.1  &- &72.1 &-& 74.7 &-&{\bf76.7} \\
1.2  &  - & {\bf 83.2} &-& 82.9 &-& 80.1  \\
2  & {\bf 37.4} & {\bf 33.6} & 36.5& 28.8 & 35.6& 28.3 \\
\bottomrule
\end{tabular}
\caption{Official evaluation results of the submitted runs on the test set.}
\label{semeval-1}
\end{table}

The SemEval 2018 task attracted 32 participants. The subtask 1.1, subtask 1.2, and subtask 2 received around 28, 19, and 11 participants, respectively.
The DNNs methods, including CNNs and LSTMs were widely used by the participating teams. Only five teams applied non-neural approaches such as Support Vector Machines (SVM) \parencite{SemEval2018Task7}. 

We select the first (\nth{1}) and second (\nth{2}) best performing models on the development datasets as well as the majority vote (mv) of 5 models for the final submission.
The overall results of our system, as evaluated on the SemEval 2018 shared task dataset are shown in Table \ref{semeval-1}. 

Our system ranks third in all three sub-tasks of the shared task \parencite{SemEval2018Task7}. We compare our system to the baseline and the winning systems in Table \ref{semeval-2}. We also report the most recent works on the SemEval 2018 dataset, \cite{wang-etal-2019-extracting} and \cite{jiang2020improving}, where the pre-trained transformers such as BERT \parencite{DBLP:journals/corr/abs-1810-04805} and \textsc{SciBert} \parencite{DBLP:journals/corr/abs-1903-10676} have been exploited in the relation classification task.

\begin{table}[t]
\centering
\resizebox{1\linewidth}{!}{
\begin{tabular}{*{5}{l}}
\toprule
   &  \multirow{2}{*}{\head{1.1}} & 
   \multirow{2}{*}{\head{1.2}}
      & \multicolumn{2}{c}{\head{2}}\\
\cmidrule(r){4-5}
 &  & & Ext.& Class. \\
\midrule
{Baseline} \parencite{SemEval2018Task7} & 34.4 & 53.5 & 26.8 & 12.6  \\
{ETH-DS3Lab} \parencite{rotsztejn-etal-2018-eth} & \head{81.7} & \head{90.4} & 48.8 & \head{49.3} \\
{UWNLP} \parencite{luan-etal-2018-uwnlp}& 78.9 & - &\head{50} & 39.1 \\
{Talla} \parencite{pratap-etal-2018-talla}& 74.2 &84.8 & - & - \\
\midrule
{SIRIUS-LTG-UiO} (Our system) &76.7  & 83.2 &37.4 & 33.6 \\
\midrule
Entity-Aware BERT$_{sp}$ \parencite{wang-etal-2019-extracting}
&81.4  & - &- & - \\
MRC-\textsc{SciBert} \parencite{jiang2020improving}
&80.5  & - &- & - \\

\bottomrule
\end{tabular}
}
\caption{Results on SemEval Task 2018 Task 7 \parencite{SemEval2018Task7}.}
\label{semeval-2}
\end{table}
\section{Syntactic Dependency Representations} 
In this section, we further examine the use of syntactic representations as input to our neural relation classification system.
We hypothesize that the shortest dependency path as a syntactic input representation provides an abstraction that is somehow domain independent. Therefore, the use of this type of structure may lessen domain effects in low-resource settings.
We quantify the influence of syntactic information by comparing to a syntax-agnostic approach and further compare different syntactic dependency representations that are used to generate embeddings over dependency paths.
\subsection{Dependency representations}
\begin{figure}[t]
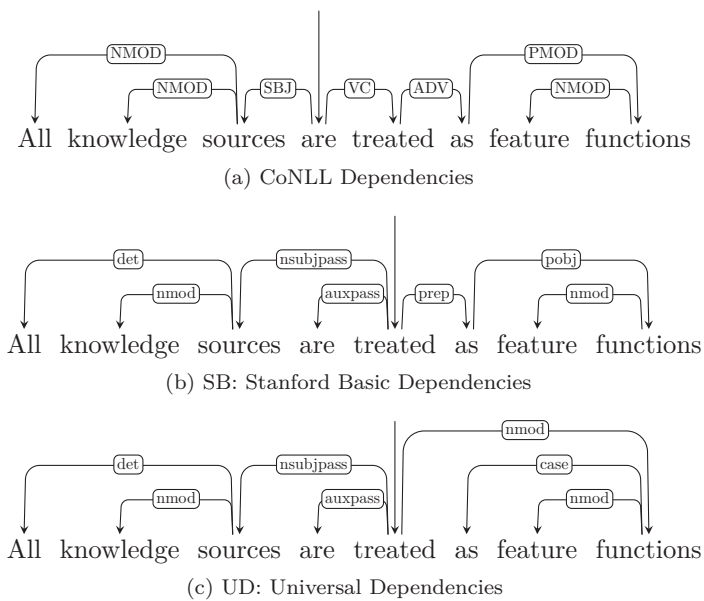

\centering
\captionsetup[subfloat]{farskip=1pt,captionskip=0.5pt}
\subfloat[CoNLL Dependencies\label{fg:conll08}]
{
\begin{dependency}[edge below, edge slant=0.15ex,edge horizontal padding=0.6ex]
\begin{deptext}[column sep=.05cm]
All \& knowledge \& sources \& are \& treated \& as \& feature \& functions\\
\end{deptext}
\depedge[edge above]{3}{1}{NMOD}
\depedge[edge above]{3}{2}{NMOD}
\depedge[edge above]{4}{3}{SBJ}
\depedge[edge above]{4}{5}{VC}
\depedge[edge above]{5}{6}{ADV}
\depedge[edge above]{6}{8}{PMOD}
\depedge[edge above]{8}{7}{NMOD}
\deproot[edge above,hide label]{4}{ROOT}
\end{dependency}}\\
\subfloat[SB:\ Stanford Basic Dependencies\label{fg:stanford}]{
  \begin{dependency}[edge below, edge slant=0.15ex,edge horizontal padding=0.6ex]
  \begin{deptext}[column sep=.6ex]
All \& knowledge \& sources \& are \& treated \& as \& feature \& functions\\ 
  \end{deptext}
\depedge[edge above]{3}{1}{det}
\depedge[edge above]{3}{2}{nmod}
\depedge[edge above]{5}{3}{nsubjpass}
\depedge[edge above]{5}{4}{auxpass}
\depedge[edge above]{5}{6}{prep}
\depedge[edge above]{6}{8}{pobj}
\depedge[edge above]{8}{7}{nmod}
\deproot[edge above,hide label]{5}{root}
 \end{dependency}}\\
\subfloat[UD:\ Universal Dependencies \label{fg:ud}]{
  \begin{dependency}[edge below, edge slant=0.15ex,edge horizontal padding=0.6ex]
  \begin{deptext}[column sep=.6ex]
All \& knowledge \& sources \& are \& treated \& as \& feature \& functions\\ 
  \end{deptext}
\depedge[edge above]{3}{1}{det}
\depedge[edge above]{3}{2}{nmod}
\depedge[edge above]{5}{3}{nsubjpass}
\depedge[edge above]{5}{4}{auxpass}
\depedge[edge above]{5}{8}{nmod}
\depedge[edge above]{8}{6}{case}
\depedge[edge above]{8}{7}{nmod}
\deproot[edge above,hide label]{5}{root}
  \end{dependency}}\\
\caption{Dependency representations for the example sentence.}
\label{fg:dep_graphs}
\end{figure}

Figure \ref{fg:dep_graphs} illustrates the three different dependency representations we compare: the so-called CoNLL-style dependencies \parencite{Joh:Nug:07} which were used for the 2007, 2008, and 2009 shared tasks of the
Conference on Natural Language Learning (CoNLL), the Stanford `basic' dependencies (SB) \parencite{de-marneffe-etal-2006-generating} and the Universal Dependencies (v1.3) (UD; \cite{McD:Niv:Qui:13,de-marneffe-etal-2014-universal,Niv:Mar:Gin:16}). We see that the analyses differ both in terms of their choices of heads vs. dependents and the inventory of dependency types.
Where CoNLL analyses tend to view functional words as heads (e.g., the auxiliary verb \textit{are}), the Stanford
scheme capitalizes more on content words as heads (e.g., the main verb \textit{treated}). UD takes the tendency to select contentful heads one step further, analyzing the prepositional complement \textit{functions} as a head, with
the preposition {\it as} itself as a dependent case marker. This is in contrast to the CoNLL and Stanford scheme, where the preposition is head.

\subsection{Experiments}
We run all the experiments with the multi-channel setting described above \footnote{As we recall, our initial rounds of experiments show that the multi-channel model works better than the single-channel model.} in which the first channel is initialized with the pre-trained ACL embeddings in static mode (i.e., it is not updated during training) and the second channel is initialized randomly and is fine-tuned during training (non-static mode). The macro F1-score is measured by 5-fold cross-validation and, once again, to deal with the effects of class imbalance, we weight the cost by the ratio of class instances; thus, each observation receives a weight, depending on the class it belongs to. 

\begin{table}[t]
  \centering
  \begin{tabular}{*{4}{lrrr}}
    \toprule
    & \multicolumn{2}{c}{\head{best F1} (in 5-fold)} & \\
    \cmidrule(ll){2-3}
    \head{Relation} & without \emph{sdp}  & with \emph{sdp}  & \head{Diff.} \\
    \midrule
    {USAGE} &60.34	&80.24	&+ 19.90  \\
     {MODEL-FEATURE} & 48.89 &	70.00	& + 21.11  \\ 
         {PART\_WHOLE}& 29.51&	70.27&	+40.76 \\
         {TOPIC} &45.80 &	91.26	& +45.46 \\
          {RESULT}& 54.35	& 81.58	& +27.23  \\
         {COMPARE} & 20.00  &61.82&	+ 41.82 \\
    \bottomrule
     macro-averaged & 50.10 &  76.10  & +26.00 \\
  \end{tabular}
   \caption[Effect of using the shortest dependency path on each relation type in sub-task 1.1.]{Effect of using the shortest dependency path on each relation type in sub-task 1.1 (see Section \ref{sub-task1.1}).}
   \label{tbl.sdp}
\end{table}

\subsection{Assessing the effect of syntactic information} \label{ch4:syntactic-effect}
To evaluate the effects of syntactic information in general for the relation classification task, we compare the model's performance with and without the dependency paths. In the syntax-agnostic setup, a sentence that contains the participant entities is used as input for the CNN. In addition to the word embeddings, to specify the position of each entity pair, we also use position embeddings for all words in the sentence. The position embeddings encode the relative distances of each word
to the entity mentions. 
We here keep the value of hyper-parameters equal to the ones used in the initial experiments. To provide the shortest dependency path (\emph{sdp}) for the syntax-aware version we compare to, we use our parser with Stanford dependencies, as described above.
Table \ref{tbl.sdp} shows the effect of using syntactic information through the shortest dependency path for each relation type.
We find that the effect of syntactic structure varies between the different relation types. However, the \emph{sdp} information has a clear positive impact on all the relation types, ranging from improvements of 20 to 45 percentage points depending on the specific relation. This can be attributed to the fact that the context-based representations suffer from irrelevant sub-sequences or clauses when target entities occur far from each other, or there are other target entities in the same sentence. The \emph{sdp} between two entities in the dependency graph captures a condensed representation of the information required to assert a relationship between two entities \parencite{Bunescu:2005:SPD:1220575.1220666}.
\begin{table}[t]
\resizebox{1\linewidth}{!}{
  \begin{tabular}{*{8}{llP{1cm}P{1.5cm}P{1.5cm}cP{1.5cm}P{1.5cm}}}
  \toprule
    &&\multicolumn{6}{c}{\head{Hyper parameters (optimal)}}\\
    \cmidrule(r){3-8}
    \head{Sub-task}&\head{Repr.} & Filter size & Feature maps& Activation func.& L2 Reg.& Learning rate& Dropout Prob.\\
    \midrule
    \multirow{3}{*}{1.1}& CoNLL  & 4-5 & 1000 & Softplus & 1.15e+01 & 1.13e-03 & 1  \\
    & SB  & 4-5 & 806 & Sigmoid & 8.13e-02 & 1.79e-03 & 0.87 \\
    &  UD v1.3  &5 & 716 & Softplus & 1.66e+00 &  9.63e-04 & 1  \\
    \midrule
    \multirow{ 3}{*}{2}& CoNLL  & 3-4-5 & 667 & ReLU & 4.96e+00 & 1.26e-03 & 0.88  \\
    
    & SB  & 6-7 & 339 & Sigmoid & 1.00e-04 & 6.96e-04 & 0.48 \\
    
    &  UD v1.3  &3-4-5 & 549 & Iden & 5.22e-01 &  5.09e-04 & 0.81   \\
    \midrule
    \midrule
    \multicolumn{2}{c}{Default values}  & 3-4-5 & 128 & ReLU & 3 &  1e-3 & 0.5   \\
    \bottomrule
  \end{tabular}
  }
   \caption[Hyper parameter optimization results for each model with different representation.]{Hyper parameter optimization results for each model with different representation. The $max$ pooling strategy consistently performs better in all model variations. We also report the default value for each hyper parameter.}
   \label{hyperparam-1}
\end{table}

\begin{table}[t]
\centering
  \begin{tabular}{*{4}{llcc}}
  \toprule
    &&\multicolumn{2}{c}{\head{F1}.(avg. in 5-fold)}\\
    \cmidrule(r){3-4}
    \head{Sub-task}&\head{Repr.} & Default& Optimized\\
    \midrule
    \multirow{ 3}{*}{1.1}& CoNLL  & 72.65  & 74.49\\
    & SB  & 74,16  & {\bf 75.05}\\
    &  UD v1.3  & 69.55 & 69.57 \\
    \midrule
    \multirow{ 3}{*}{2}& CoNLL  & 60.31   & 60,54 \\
    
    & SB  & 58.11   & {\bf 61.18} \\
    
    &  UD v1.3  & 54.53  & 56.80   \\
  
    \bottomrule
  \end{tabular}
   \caption[Performance of each model with optimized hyper parameters for different representation.]{Performance of each model with optimized hyper parameters for different representation. In optimized column we evaluate each model with the optimal value for each hyper parameter, given in Table \ref{hyperparam-1}. In default column, the default value for each hyper parameter is used.}
   \label{hyperparam-2}
\end{table} 

\subsection{Comparison of different dependency representations} \label{ch4:different-reps} 
To assess model performance with various syntactic dependency representations, we create a \emph{sdp} for each training example using the different parse models and exploit them as input to both the relation extraction and classification model. With the use of default parameters, there is a risk that these favor one of the representations simply by chance. In order to perform a fair comparison between the different dependency representations, we make use of Bayesian optimization \parencite{DBLP:journals/corr/abs-1012-2599} in order to locate optimal hyper-parameters for each of the dependency representations. We construct a Bayesian optimization procedure using a Gaussian process with 100 iterations and Expected Improvement (EI) for its acquisition functions. We set the objective function to maximize the macro F1 score over 5-fold cross-validation on the training set. Here we investigate the impact of various system design choices with the following parameters\footnote{Default values are \{3-4-5,  128,  ReLU, max,  3,  1e-3, 0.5\}}: 
\begin{itemize}
\setlength{\itemsep}{4pt}
    \item \emph{Filter region size: $\in$ \{3, 4, 5, 6, 7, 8, 9, 3-4, 4-5, 5-6, 6-7, 7-8, 8-9, 3-4-5, 4-5-6, 5-6-7, 6-7-8, 7-8-9\}}
  \item \emph{Number of feature maps for each filter region size: $\in \{10: 1000\}$ }
  \item \emph{Activation function: $\in \{Sigmoid, ReLU, Tanh, Softplus, Iden \}$.}
  \item \emph{Pooling strategy: $ \in \{max, avg\}$. }
  \item \emph{L2 regularization: $ \in \{1e-4: 1e+2\}$.}
  \item \emph{Learning rate: $\in \{1e-6: 1e-2\}$.}
  \item \emph{Dropout probability \footnote{The probability that each element is kept, in which $1$ implies that none of the nodes are dropped out}: $\in \{0.1: 1\}$. }
 \end{itemize}
Table \ref{hyperparam-1} presents the optimal values for each configuration using different dependency representations. We see that the optimized parameter settings vary for the different representations, showing the importance of tuning for these types of comparisons. The results (Table \ref{hyperparam-2}) furthermore show that the \emph{sdp}s based on the Stanford Basic (SB) representation provide the best performance for both subtasks following hyperparameter tuning, followed by the CoNLL08 representation. We also observe that for the extraction subtask (subtask 2), the best representation changes following tuning of the system.
It can be seen that the results for the UD representation are consistently quite a bit lower than the two others. This is perhaps somewhat surprising given the fact that downstream usefulness is one of the motivations behind this dependency framework.

\section{Error Analysis} \label{ch4:errors}
\begin{table}[t]
  \centering
  \begin{tabular}{*{5}{lcrrr}}
    \toprule
    & &\multicolumn{3}{c}{\head{best F1} (in 5-fold) } & \\
    \cmidrule(ll){3-5}
    \head{Relation} & \head{Freq.}& CoNLL  & SB  & UD \\
    \midrule
    {USAGE}&947 & 76.84 & 82.39  & 77.56    \\
   
    {MODEL-FEATURE}&498 &  68.27  &  68.54   & 66.36  \\  
    {PART\_WHOLE}&425&  75.32 & 71.28  &    67.11\\
     {TOPIC} &258& 89.32 & 90.57  &    87.62\\
     {RESULT}&193 &82.35 & 81.69 & 82.86   \\
     {COMPARE} & 136& 66.67  & 66.67 &   54.24\\

    \bottomrule
    macro-averaged && 76.94 & 77.57  & 72.83
  \end{tabular}
   \caption[Effect of using the different parser representation on each relation type in sub-task 1.1.]{Effect of using the different parser representation on each relation type in sub-task 1.1 (see Section \ref{sub-task1.1}).}
   \label{tbl.repsf1}
\end{table}
\begin{sidewaystable}[ph!]
  \centering
    \scalebox{.7}{
 \begin{tabular}{lcr}
    \toprule
    \head{Sentence 1} & \multicolumn{1}{l}{{\small 
    This indicates that there is no need to add {\color{white}\colorbox{gray}{punctuation}} in transcribing {\color{white}\colorbox{gray}{spoken corpora}} simply in order to help parsers.}} & class: {\small PART\_WHOLE} \\
    \midrule
      {\small CoNLL} & {\small punctuation $\leftarrow$ obj $\leftarrow$ add $\rightarrow$ adv $\rightarrow$ in $\rightarrow$ pmod $\rightarrow$ transcribing $\rightarrow$ obj $\rightarrow$ spoken corpora} &\\
        {\small SB} &{\small punctuation $\leftarrow$ dobj $\leftarrow$ add $\rightarrow$ prep $\rightarrow$ in $\rightarrow$ pcomp $\rightarrow$ transcribing $\rightarrow$ dobj $\rightarrow$ spoken corpora} & \\  
        {\small UD v1.3}& {\small punctuation  $\leftarrow$ dobj  $\leftarrow$ add $\rightarrow$ advcl $\rightarrow$ transcribing $\rightarrow$ dobj $\rightarrow$ spoken corpora} & \\
    \midrule
    \midrule
    {\head{Sentence 2}} & \multicolumn{1}{l}{{\small In the process we also provide a {\color{white}\colorbox{gray}{formal definition}} of {\color{white}\colorbox{gray}{parsing}} motivated by an informal notion due to Lang .}} & class: {\small MODEL-FEATURE}\\
    \midrule
    {\small CoNLL} & {\small formal definition $\rightarrow$ nmod $\rightarrow$ of $\rightarrow$ pmod $\rightarrow$ parsing} &\\
    {\small SB} &{\small formal definition $\rightarrow$ prep $\rightarrow$ of $\rightarrow$ pobj $\rightarrow$ parsing}&\\
    {\small UD v1.3}& {\small formal definition $\rightarrow$ nmod $\rightarrow$ parsing}&\\
    \midrule
    \midrule
    \head{Sentence 3} & \multicolumn{1}{l}{{\small This paper describes a practical {\color{white}\colorbox{gray}{"black-box" methodology}} for automatic evaluation of {\color{white}\colorbox{gray}{question-answering NL systems}} in spoken dialogue.}} & class: {\small USAGE}\\
    \midrule
    {\small CoNLL} & {\small " "black-box" methodology $\rightarrow$ nmod $\rightarrow$ for $\rightarrow$ pmod $\rightarrow$ evaluation $\rightarrow$ nmod $\rightarrow$ of $\rightarrow$ pmod $\rightarrow$ question-answering NL systems} &\\
   {\small SB} &{\small "black-box" methodology $\rightarrow$ prep $\rightarrow$ for $\rightarrow$ pobj $\rightarrow$ evaluation $\rightarrow$ prep $\rightarrow$ of $\rightarrow$ pobj $\rightarrow$ question-answering NL systems} & \\
    {\small UD v1.3}& {\small  "black-box" methodology $\rightarrow$ nmod $\rightarrow$ evaluation $\rightarrow$ nmod $\rightarrow$ question-answering NL systems} & \\
    \bottomrule
  \end{tabular}}
   \caption{The examples for which the CoNLL/SB-based models in the classification sub-task correctly predict the relation type in 5-fold trials, whereas the UD based model has an incorrect prediction.}
   \label{tbl.examples}
   \bigskip \bigskip
 \scalebox{.68}{
  \begin{tabular}{lcr}
    \toprule
    \head{Sentence 4} & \multicolumn{1}{l}{{\small However , for {\color{white}\colorbox{gray}{grammar formalisms}} which use more {\color{white}\colorbox{gray}{fine-grained grammatical categories}} , for example tag and ccg , tagging accuracy is much lower .}} & class: {\small USAGE}\\
 \midrule
      {\small CoNLL} & {\small fine-grained grammatical categories $\leftarrow$ prd $\leftarrow$ use $\leftarrow$ nmod $\leftarrow$ grammar formalisms} &\\
        {\small SB} &{\small fine-grained grammatical categories $\leftarrow$ dobj $\leftarrow$ use $\leftarrow$ rcmod $\leftarrow$ grammar formalisms} & \\  
         {\small UD v1.3}& {\small fine-grained grammatical categories $\leftarrow$ advmod $\leftarrow$ use $\rightarrow$ nmod $\rightarrow$ grammar formalisms} & \\
         \midrule
         \midrule
             \head{Sentence 5} & \multicolumn{1}{l}{{\small We consider the case of multi-document summarization , where the input {\color{white}\colorbox{gray}{documents}} are in {\color{white}\colorbox{gray}{Arabic}} , and the output summary is in English .}} & class: {\small  MODEL-FEATURE}\\
 \midrule
      {\small CoNLL} & {\small Arabic $\leftarrow$ pmod $\leftarrow$ in $\leftarrow$ prd $\leftarrow$ are $\rightarrow$ sbj $\rightarrow$ documents} &\\
       {\small SB} &{\small Arabic $\leftarrow$ pobj $\leftarrow$ in $\leftarrow$ prep $\leftarrow$ are $\rightarrow$ nsubj $\rightarrow$ documents} & \\  
         {\small UD v1.3}& {\small Arabic $\rightarrow$ nsubj $\rightarrow$ documents} & \\
\bottomrule
  \end{tabular}}
   \caption[The examples based on the CoNLL-, SB- and UD-based models in the extraction sub-task.]{The examples for which the CoNLL/SB-based models in the extraction sub-task correctly predict the relation type in 5-fold trials, whereas the UD based model has an incorrect prediction.}
   \label{tbl.examples2}
\end{sidewaystable}
Our results show that the best performing dependency framework in our system is the Stanford Basic scheme and furthermore that the widely used Universal Dependencies scheme consistently provides somewhat lower results in both relation classification and full extraction. To gain a better understanding of the reasons behind these differences in performance, we perform error analysis.

Table \ref{tbl.repsf1} presents the effect of each parser representation in the classification task, broken down by relation type. Firstly, we note that in general, the results differ between the different relation types, where the \emph{{\small TOPIC}} relation has the highest score (90.57 F1 with the SB representation). In contrast, the most infrequent \emph{{\small COMPARE}} relation has the lowest F-score (66.67 F1 with SB). 
We further observe that the UD-based model falls behind the others on most of the relation types (i.e., \emph{{\small COMPARE, MODEL-FEATURE, PART\_WHOLE, TOPICS}}).  
To explore these differences in more detail, we manually inspect the instances for which the CoNLL/SB-based models correctly predict the relation type in 5-fold trials, whereas the UD-based model has an incorrect prediction.

Table \ref{tbl.examples} shows some of these examples for the classification sub-task, marking the entities and the gold class of each instance and also showing the \emph{sdp} from each representation. We observe that the UD paths are generally shorter. A striking similarity between most of the instances is the fact that one of the entities resides within a prepositional phrase. Whereas the SB and CoNLL paths explicitly represent the preposition in the path, the UD representation does not. Clearly, the difference between, for instance, the \emph{{\small USAGE}} and \emph{{\small PART\_WHOLE}} relation may be indicated by the presence of a specific preposition ({\it X for Y} vs. {\it X of Y}). This is also interesting since this particular syntactic choice has been shown in previous work to have a negative effect on intrinsic parsing results for English \parencite{Sch:Abe:Rap:12}.

We go on to examine the errors in the full extraction sub-task (task 2, see Section \ref{sub-task2}) where the system trained using UD-based paths has an incorrect prediction, whereas the two other systems (CoNLL-based and SB-based) do not. We note that here as well the exclusion of the preposition from the path in the UD representation is problematic. A high proportion of the errors contain an entity that resides within a prepositional phrase, as exemplified by the sentences in Table \ref{tbl.examples}. We also observe some parse errors in the UD parse. Sentence 4 in Table \ref{tbl.examples2}, for instance, gives an example where the UD parser incorrectly assigns the embedded verb ({\it use}) of a relative clause status as a main verb with two modifier dependents, rather than recognizing that the relative clause ({\it which use more fine-grained grammatical categories}) depends on the entity {\it fine-grained grammatical categories}. We also find another difference between the representations which shows up in the errors, namely the combination of the aforementioned UD treatment of prepositions as dependent case markers and the copula construction, e.g., for Sentence 5 in Table \ref{tbl.examples2}, where the CoNLL and SB parsers assign head status to the copula verb {\it are} in combination with the PP complement {\it in Arabic}, whereas the UD parser assigns head status to {\it Arabic} of which the {\it documents} is a subject dependent.

\section{Summary}
This chapter presents a CNN model over the shortest dependency paths between entity pairs for relation extraction and classification in scientific text. We examine several variants of this architecture for the proposed model. The experiments demonstrate the effectiveness of domain-specific word embeddings for all sub-tasks as well as sensitivity to the specific dependency representation employed in the input layer. 
We compared three widely used dependency representations (CoNLL, Stanford Basic, and Universal Dependencies) and find that representation matters and that certain choices have clear consequences in downstream processing. Our experiments also underline the importance of performing hyperparameter tuning when comparing different input representations.

To summarize, the following contributions are made in this chapter: 
\begin{enumerate}[label=(\roman*)]
    \item We evaluate the effect of syntax in a neural relation extraction and classification system,
    \item We study the impact of domain-specific embeddings, 
    \item We assess the effect of varying the syntactic input representations, and
    \item We perform a manual error analysis that helps understand the most important aspects of syntactic representation for these tasks.
\end{enumerate}

    \chapter{Natural Language Understanding in Low-Resource Genres and Languages}
\label{sec:fifth}
Learning what to share between tasks (i.e., transfer learning as explained in \ref{subsec:ds} of Chapter \ref{sec:first}) has been a topic of great importance recently, as strategic sharing of knowledge has been shown to improve the performance of downstream tasks.
In multilingual applications, sharing knowledge between languages is important when considering that most languages in the world suffer from being under-resourced.
In this chapter, consider the transfer of models along two dimensions of variation (see Section \ref{domain-adaptation} in chapter \ref{sec:first}), namely genre and language, when little or no data is available for a target genre or language.
These scenarios are known as low-resource and zero-resource settings.
We show that this challenging setup can be approached using meta-learning, where, in addition to training a source model, another model learns to select which training instances are the most beneficial.
We experiment using standard supervised, zero-shot cross-lingual, as well as few-shot cross-genre and cross-lingual settings for different natural language understanding tasks (natural language inference, question answering).
Our extensive experimental setup demonstrates the consistent effectiveness of meta-learning in various low-resource scenarios.
We improve the performance of pre-trained language models for zero-shot and few-shot NLI and QA tasks on two NLI datasets (\iec MultiNLI and XNLI), and on the MLQA dataset.
We further conduct a comprehensive analysis, which indicates that the correlation of typological features between languages can further explain when parameters sharing learned via meta-learning is beneficial.

\section{Introduction}\label{xmaml:rqs}
There are more than 7000 languages spoken in the world, over 90 of which have more than 10 million native speakers each \parencite{ethnologue:19}.
Despite this, very few languages have proper linguistic resources when it comes to natural language understanding tasks.
Although there is a growing awareness in the field, as evidenced by the release of datasets such as XNLI \parencite{conneau2018xnli}, most NLP research still only considers English \parencite{bender2019benderrule}.
While one solution to this issue is to collect annotated data for all languages, this process is both too time consuming and expensive to be feasible.
Additionally, it is not trivial to train a model for a task in a particular language (\egc English) and apply it directly to another language where only limited training data is available (\iec low-resource languages).
Therefore, it is essential to investigate strategies that allow us to use a large amount of training data available for English for the benefit of other languages.

Meta-learning has recently been shown to be beneficial for several machine learning tasks   \parencite{koch:15,vinyals:16,santoro:16,finn:17,ravi:17,nichol:18}.
In the case of NLP, recent work has also shown the benefits of this sharing between tasks and domains \parencite{dou-etal-2019-investigating,gu-etal-2018-meta,qian-yu-2019-domain}.
Although meta-learning for cross-lingual transfer has been investigated in the context of machine translation \parencite{gu-etal-2018-meta}, in this chapter, we attempt to study the meta-learning effect for natural language understanding tasks.
We investigate cross-lingual meta-learning using two challenging evaluation setups, namely:
\begin{enumerate}[label=(\roman*)]
    \item \emph{Few-shot learning:} where only a limited amount of training data is available for the target domain or genre.
    \item \emph{Zero-shot learning:} where no training data is available for the target domain or genre.
\end{enumerate}
Specifically, in \secref{sec:experiments}, we evaluate the performance of our model on two natural language understanding tasks, as follows:

\begin{itemize}
\item  Natural Language Inference (NLI) by experimenting on the MultiNLI (cross-genre setup) and the XNLI (cross-lingual setup) datasets   \parencite{conneau2018xnli}.
\item  Question Answering (QA) on the MLQA as a multilingual question answering dataset  \parencite{lewis2019mlqa}.
\end{itemize}
Accordingly, we investigate the following research questions in this chapter:
\begin{question}\label{rq.5.1}
    Can meta-learning assist us in coping with low-resource settings in natural language understanding (NLU) tasks?
\end{question}
\begin{question} \label{rq.5.2}
    What is the impact of meta-learning on the performance of pre-trained language models such as BERT \parencite{DBLP:journals/corr/abs-1810-04805}, XLM \parencite{conneau2019cross} and  XLM-RoBERTa \parencite{conneau2019unsupervised} in cross-lingual NLU tasks \footnote{At the time of writing, these are the top performing models in cross-lingual NLU benchmarks.}?
\end{question}
\begin{question} \label{rq.5.3}
    Can meta-learning provide a model- and task-agnostic framework in low-resource NLU tasks?
\end{question}
\begin{question} \label{rq.5.4}
    Are typological commonalities among languages beneficial for the performance of cross-lingual meta-learning?
\end{question}

\section{Natural Language Understanding (NLU)}
Understanding of natural language is an essential and challenging goal of NLP.
Natural language understanding comprises a wide range of diverse tasks, including, but not limited to, natural language inference, question answering, sentiment analysis, semantic similarity assessment, and
document classification. In this thesis,
we explore the use of transfer learning by leveraging meta-learning to perform various NLU tasks, including
natural language inference and question answering. We provide a brief description of these tasks in the following sections.
\subsection{Natural Language Inference (NLI)}
\begin{table}
	\centering
	\begin{tabular}{lrrr}
		\toprule
		 & \multicolumn{3}{c}{\bf \#Examples}  \\
		\bf Genre & \bf Train& \bf Dev. & \bf Test  \\
		\midrule
		\sc Fiction & 77,348 & 2,000 & 2,000  \\
		\sc Government & 77,350 & 2,000 & 2,000 \\
		\sc Slate & 77,306 & 2,000 & 2,000 \\
		\sc Telephone & 83,348 & 2,000 & 2,000 \\
		\sc Travel & 77,350 & 2,000 & 2,000 \\
		\midrule
		\sc 9/11 & 0 & 2,000 & 2,000 \\
		\sc Face-to-face & 0 & 2,000 & 2,000 \\
		\sc Letters & 0 & 2,000 & 2,000 \\
		\sc OUP & 0 & 2,000 & 2,000 \\
		\sc Verbatim & 0 & 2,000 & 2,000 \\
		\midrule
		\bf MultiNLI Overall & \textbf{392,702} & \textbf{20,000} & \textbf{20,000} \\
		\bottomrule
	\end{tabular}
	\caption[Statistics for the MultiNLI corpus by genre.]{Statistics for the MultiNLI corpus by genre. The first five genres represent  in the training, development and test sets, and the remaining five represent in the development and test set \parencite{williams-etal-2018-broad}.}
	\label{tbl:mnli-stat}
\vspace{1cm}
\resizebox{1\linewidth}{!}{
 \begin{tabular}{p{7cm}p{2.1cm}p{5.35cm}}
  \toprule
\bf Premise & \bf Label & \bf Hypothesis\\
  &&\\
 \multicolumn{3}{l}{\textsc{\Large{Fiction}}}\\
 \texttt{The Old One always comforted Ca'daan, except today.}	& \emph{neutral} & \texttt{Ca'daan knew the Old One very well.}\\
  &&\\
 \multicolumn{3}{l}{\sc \Large Letters}\\
 \texttt{Your gift is appreciated by each and every student who will benefit from your generosity.} & \emph{neutral} & \texttt{Hundreds of students will benefit from your generosity.}\\
 &&\\
  \multicolumn{3}{l}{ \sc \Large Telephone}\\
  \texttt{yes now you know if if everybody like in August when everybody's on vacation or something we can dress a little more casual or} & \emph{contradiction} & \texttt{August is a black out month for vacations in the company.}\\
   &&\\
    \multicolumn{3}{l}{ \sc \Large 9/11 }\\
    \texttt{At the other end of Pennsylvania Avenue, people began to line up for a White House tour.}& \emph{entailment}& \texttt{People formed a line at the end of Pennsylvania Avenue.}\\
    \bottomrule
  \end{tabular}
  }
  \caption[Examples from the MultiNLI corpus.]{Examples from the MultiNLI corpus, shown with their genre and selected gold labels \parencite{williams-etal-2018-broad}.}
  \label{tbl:mnli-Ex}
\end{table}
NLI is the task of predicting whether a \textit{hypothesis} sentence is true (entailment), false (contradiction), or undetermined (neutral) given a \textit{premise} sentence.
NLI systems need some semantic understanding and models trained on entailment data can be applied to many other NLP tasks such as text summarization, paraphrase detection, and machine translation.  The task of NLI, also known as textual entailment, is well-positioned to serve as a benchmark task for research on NLU \parencite{williams-etal-2018-broad}.
Here, we present some of the datasets that have been provided for NLI tasks and are exploited in this chapter.

\begin{figure*}
\begin{minipage}{\textwidth}
\centering
\small
\resizebox{1\linewidth}{!}{
        \begin{tabular}{lp{11cm}cc}
        \toprule
        Language
        & Premise / Hypothesis
        & Genre
        & Label\\
        \midrule
        English (en)
            & \begin{tabular}[x]{@{}l@{}}
                You don't have to stay there.\\
                You can leave.
            \end{tabular}
            & Face-To-Face
            & Entailment\\
        \midrule
        French (fr)
            & \begin{tabular}[x]{@{}l@{}}
                La figure 4 montre la courbe d'offre des services de partage de travaux.\\
                Les services de partage de travaux ont une offre variable.
            \end{tabular}
            & Government
            & Entailment\\
        \midrule
        Spanish (es)
            & \begin{tabular}[x]{@{}l@{}}
                Y se estremeció con el recuerdo.\\
                El pensamiento sobre el acontecimiento hizo su estremecimiento.
            \end{tabular}
            & Fiction
            & Entailment\\
        \midrule
        German (de)
            & \begin{tabular}[x]{@{}l@{}}
                Während der Depression war es die ärmste Gegend, kurz vor dem Hungertod.\\
                Die Weltwirtschaftskrise dauerte mehr als zehn Jahre an.
            \end{tabular}
            & Travel
            & Neutral\\
        \midrule
        Swahili (sw)
            & \begin{tabular}[x]{@{}l@{}}
                Ni silaha ya plastiki ya moja kwa moja inayopiga risasi.\\
                Inadumu zaidi kuliko silaha ya chuma.
            \end{tabular}
            & Telephone
            & Neutral\\
        \midrule
        Russian (ru)
            & \begin{tabular}[x]{@{}l@{}}
           \begin{otherlanguage*}{russian} И мы занимаемся этим уже на протяжении 85 лет.\end{otherlanguage*}\\
           \begin{otherlanguage*}{russian}
           Мы только начали этим заниматься.
           \end{otherlanguage*}
            \end{tabular}
            & Letters
            & Contradiction\\
        \midrule
        Chinese (zh)
            & \begin{tabular}[x]{@{}l@{}}
         \begin{CJK}{UTF8}{gbsn}让我告诉你，美国人最终如何看待你作为独立顾问的表现。\end{CJK}\\
         \begin{CJK}{UTF8}{gbsn}美国人完全不知道您是独立律师。\end{CJK}
            \end{tabular}
            & Slate
            & Contradiction\\
        \midrule
        Arabic (ar)
            & \begin{tabular}[x]{@{}l@{}}
            \raisebox{-0.1\totalheight}{\includegraphics[scale=0.3]{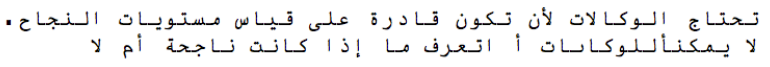}}\end{tabular}
            & Nine-Eleven
            & Contradiction\\
        \bottomrule
        \end{tabular}
        }
        \captionof{table}{Examples (premise and hypothesis) from various languages and genres from the XNLI corpus \parencite{conneau2018xnli}.}
        \label{tab:xnli}
\end{minipage}
\newline
\vspace*{1cm}
\newline
\begin{minipage}[c]{\textwidth}
\centering
 \includegraphics[width=1\textwidth]{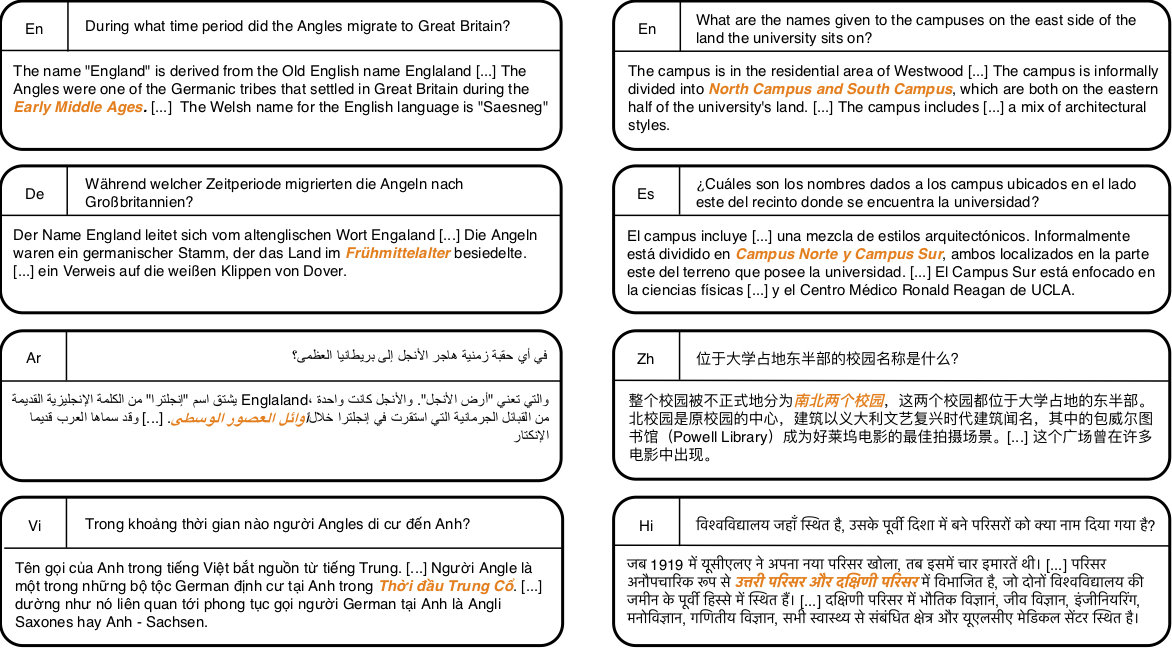}
   \caption[QA instances in the MLQA dataset.]{QA instances in the MLQA dataset. Answers shown as highlighted spans in contexts. Contexts shorten for clarity with  "[$\dots$]" \parencite{lewis2019mlqa}.}
  \label{fig:mlqa-examples}
\end{minipage}
\newline
\vspace*{1cm}
\newline
\begin{minipage}{\textwidth}
\centering
\small
\resizebox{1\linewidth}{!}{
\begin{tabular}{c|ccccccc}
\toprule
Dataset& English (en)  & Arabic (ar) & German (de) & Spanish (es) & Hindi (hi) & Vietnamese  (vi)& Chinese (zh)\\
\toprule
Dev& 1148  & 517 & 512 & 500& 507 & 511& 507\\
Test& 11590  & 5335 & 4517 & 5253& 4918 & 5495& 4918\\
\toprule
\end{tabular}
}
        \captionof{table}{Overview of the number of QA instances in the development and test portions of the MLQA dataset across the different languages.}
    \label{tbl:mlqa-dataset}
\end{minipage}
\end{figure*}

\paragraph{MultiNLI}
The Multi-Genre Natural Language Inference (MultiNLI) corpus has 433k sentence pairs annotated with textual entailment information \parencite{williams-etal-2018-broad}.
It covers a range of different genres of spoken and written text and offers an explicit setting for cross-genre evaluation.
The NLI premise sentences are derived from 10 different resources to cover a maximally broad range of genres of American English, such as:
\textsc{facetoface}, \textsc{telephone}, \textsc{verbatim}, \textsc{state}, \textsc{government}, \textsc{fiction}, \textsc{letters}, \textsc{9/11}, \textsc{travel} and \textsc{oup}.
All of the genres appear in the test and development sets, but only five are included in the training set (See Table \ref{tbl:mnli-stat}, which presents the statistics for the MultiNLI dataset by genre).
Table \ref{tbl:mnli-Ex} depicts randomly chosen examples from the MultiNLI dataset, shown with their genre labels, and both the selected gold labels.
\paragraph{XNLI}
The Cross-lingual Natural Language Inference (XNLI)  dataset \parencite{conneau2018xnli} consists of 5000 test data and 2500 development hypothesis-premise pairs with their textual entailment labels for English.
The pairs are annotated and translated, by employing professional translators, into 14 languages French (fr), Spanish (es), German (de), Greek (el), Bulgarian (bg), Russian (ru), Turkish (tr), Arabic (ar), Vietnamese (vi), Thai (th), Chinese (zh), Hindi (hi), Swahili (sw) and Urdu (ur). XNLI provides a multilingual benchmark to evaluate how to perform inference in low-resource languages such as Swahili or Urdu, in which only training data for the high-resource language English is available from MultiNLI. Some  examples from the XNLI corpus are shown in
Table \ref{tab:xnli}.

\subsection{Question Answering (QA)}  \label{xmaml-qa-data}
The task of QA is often designed in the context of a reading comprehension task. This machine reading problem is formulated as extractive question answering, in which the answer is drawn from the original text \parencite{eisenstein2019introduction}. In this context, given a \emph{context} and a \emph{question}, the QA task aims to identify the span answering the question in the context.
We study the QA task using the following two datasets:
\paragraph{SQuAD} Stanford Question Answering Dataset (SQuAD v1.1), provided by \cite{rajpurkar-etal-2016-squad,}, is a reading comprehension dataset and contains 107,785 question-answer pairs obtained from  536 English Wikipedia articles.

\paragraph{MLQA}
\cite{lewis2019mlqa} introduce a Multilingual Question Answering dataset (called MLQA) containing QA instances in 7 languages, namely English (en), Arabic (ar), German (de), Spanish (es), Hindi (hi), Vietnamese (vi) and Simplified Chinese (zh). Figure \ref{fig:mlqa-examples} shows some examples from the MLQA dataset. MLQA is split into development and test splits, with detailed numbers in Table \ref{tbl:mlqa-dataset}. Recently, it has been used in many benchmarks for the evaluation of cross-lingual transfer learning, e.g., \cite{hu2020xtreme} and \cite{Liang2020XGLUEAN}.

\section{NLU Models} \label{xmaml:nlu-models}
We perform experiments on a variety of models that have been proposed for NLU tasks, including Enhanced Sequential Inference Model (ESIM), Bidirectional Encoder Representations from Transformers (BERT), Cross-Lingual Language Model (XLM) and  XLM on RoBERTa (XLM-RoBERTa). These models have become competitive baselines on NLI and QA tasks. In the following sections, we will briefly describe the models.
 \begin{figure}[t]
    \centering
    \includegraphics[width=1\textwidth]{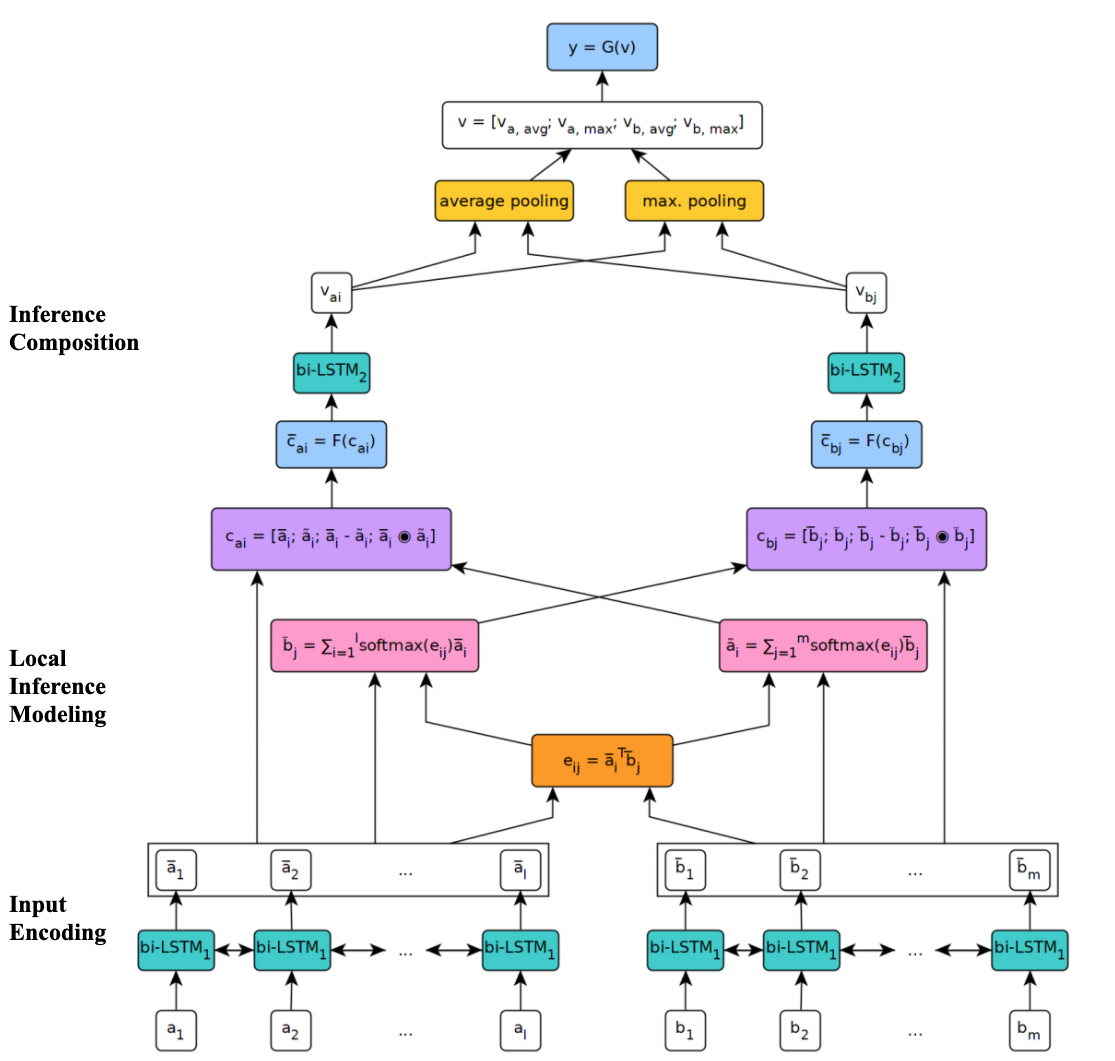}
    \caption[Architecture of the ESIM model.]{Architecture of the ESIM model \parencite{DBLP:journals/corr/ChenZLWJ16}.}
    \label{fig:esim-xmalm}
 \end{figure}
\paragraph{ESIM} \label{xmalm-esim}
Enhanced Sequential Inference Model (ESIM), proposed by \cite{DBLP:journals/corr/ChenZLWJ16}, is commonly used for textual entailment problems. ESIM employs LSTMs with attention to create a rich representation, capturing the relationship between premise and hypothesis sentences. It introduces local inference modeling, which models the inference relationship between premise and hypothesis after the two fragments have been aligned locally. Figure \ref{fig:esim-xmalm} shows the architecture of the ESIM model. It consists of three layers. The input encoding layer, which is the first layer, uses BiLSTM to provide a contextual representation of each word element in the input premise and hypothesis. Then, the local inference modeling collects information to perform local inference between words and phrases in the second layer. This layer computes a form of soft attention computed between the words in the two sentences to model their interactions.
The $softmax$ function is applied to transfer the attention weights computed between each word in the premise and the hypothesis to a probability distribution.
The inference between sentence pairs is modeled by concatenation of the encoded and conditioned representations of words, their difference, and component-wise product.
The last layer is devoted to inference composition to perform composition and aggregation over local inference output and to make the global judgment.
Since the previous layer introduces a lot of new dimensions, the outputs from inference modeling are first passed through a mapping function $F$ consisting of a simple feed-forward layer with $ReLU$ activation to control the model’s complexity. Then, the second BiLSTM layer provides two new vectors. To merge these two vectors, average and max pooling operations are applied, and the results are concatenated in a final representation to predict the probabilities of the classes associated to the input sentences. The prediction step contains a two-layer perceptron $G$ with $tanh$ and $softmax$ activation functions.
\paragraph{BERT} The Bidirectional Encoder Representations from Transformers (BERT) \parencite{DBLP:journals/corr/abs-1810-04805} is designed to pretrain deep bidirectional representations from
unlabeled text by jointly conditioning on both
left and right context in all layers (see Section \ref{sec:pre-trained-LM} in Chapter \ref{sec:first}).
In our study, we employ the original English BERT model (En-BERT) and  Multilingual BERT (Multi-BERT) models. Like the original English BERT model, Multi-BERT is a 12 layer transformer, but instead of being trained only on monolingual English data with
an English-derived vocabulary, it is trained on the Wikipedia pages of 104 languages with a shared word piece vocabulary.
\paragraph{XLM}
 \begin{figure}[t]
    \centering
    \includegraphics[width=1\textwidth]{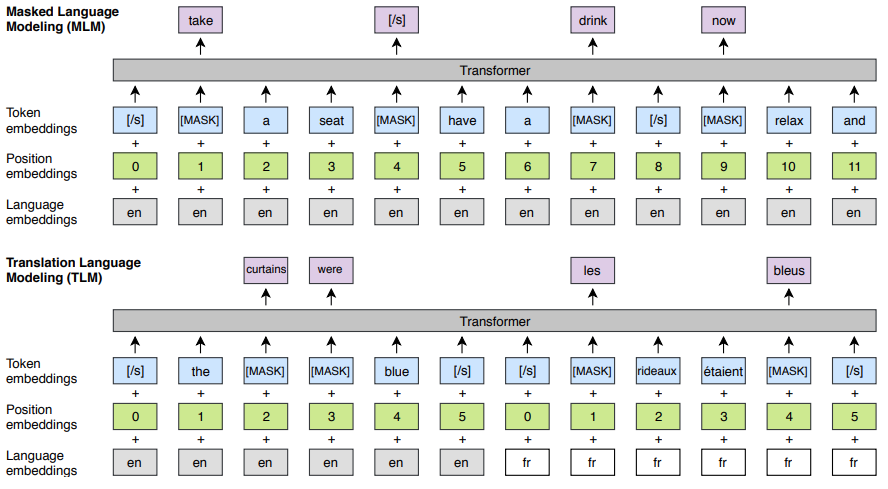}
    \caption[Modified Masked Language Model and Translation Language Model in XLM.]{Modified Masked Language Model and Translation Language Model in XLM  \parencite{conneau2019cross}.}
    \label{fig:xlm}
 \end{figure}
XLM, proposed by \cite{conneau2019cross}, uses a similar pre-training objective as Multi-BERT with a larger model, a  more extensive shared vocabulary, and leverages both monolingual and parallel data.
XLM modifies BERT in the following way:
First, instead of using word or characters as the input of the model, it uses Byte-Pair Encoding (BPE), introduced by \cite{sennrich-etal-2016-neural}, that splits the input into the most common sub-words across all languages, thereby increasing the shared vocabulary between languages. This setting is denoted as Masked Language Modeling (MLM). Second, the Translation Language Modeling (TLM) modifies the BERT architecture as follows:
\begin{enumerate*}[label=(\roman*)]
    \item It extends the masked language model to pairs of parallel sentences. Unlike BERT, each training sample consists of the same text in two languages. Therefore, the model can use the context from one language to predict tokens in the other, as different words are randomly masked words in each language, and
    \item The model is also informed about the language ID and the order of the tokens (i.e., the Positional Encoding) in each language as input metadata. It helps the model learn the relationship between related tokens in different languages.
\end{enumerate*}
The complete XLM model is trained by both MLM and TLM and alternating between them (Figure \ref{fig:xlm}). We make use of a variant of the XLM-15 that is trained with MLM + TLM on the 15 XNLI languages.

\paragraph{XLM-RoBERTa (XML-R)}
Robustly Optimized BERT Pre-training Approach (RoBERTa) \parencite{DBLP:journals/corr/abs-1907-11692} has the same architecture as BERT, but uses SentencePiece \parencite{kudo-richardson-2018-sentencepiece} as a tokenizer. It modifies key hyperparameters, removing the next-sentence pretraining objective and training with much larger mini-batches and learning rates.  XLM-RoBERTa (XML-R) is a RoBERTa-version of XLM trained based on a much larger
multilingual corpus (i.e., more than two terabytes of publicly available CommonCrawl data in 100 different languages) and has become the new state-of-the-art
on cross-lingual benchmarks \parencite{hu2020xtreme}. The biggest update that XLM-R offers over the original is a significantly increased amount of training data. XLM-R$_{base}$ with 125M parameters and XLM-R$_{large}$ with 355M parameters are the variations of XLM-R that are trained on 2.5 TB of CommonCrawl data in 100 languages and have been used in our work.

\section{Model-Agnostic Meta-Learning (MAML)} \label{maml-desc}
Meta-learning, or learning to learn, can be seen as an instance of sequential transfer learning (see Section \ref{sec:seqtr} of Chapter \ref{sec:first}).
It trains a high-level model sequentially based on the sub-models that are typically optimized  \parencite{ruder-etal-2019-transfer}.
Meta-learning tries to tackle the problem of fast adaptation to a handful of new training data instances. It discovers the structure among multiple tasks such that learning new tasks can be done quickly. In NLP, this has been done by repeatedly simulating the learning process on low-resource tasks using various high-resource tasks \parencite{gu-etal-2018-meta}.
There are several ways of performing meta-learning:
\begin{enumerate}[label=(\roman*)]
    \item Metric-based:
    It aims to learn similarities between feature representations of instances from different training sets given a similarity metric. The idea is to learn a metric space and then use it to compare low-resource testing to high-resource training samples. The representative works in this category include Siamese Network \parencite{koch:15}, Matching Network \parencite{vinyals:16}, and Relation Network \parencite{8578229}.
    \item Model-based: The idea is to use an additional meta-learner to learn and to update the original learner with a few training examples. The focus has been on adapting models that learn fast (\egc memory networks) for meta-learning \parencite{santoro:16}.
    \cite{ravi:17} introduce an LSTM-based meta-learner to learn the optimization algorithm used to train the original network.
    \item Optimization-based:  The optimization algorithm itself is
    designed in a way that favors fast adaption \parencite{finn:17,nichol:18}. The optimization-based methods introduce no additional architectures nor parameters. They can find good initialization parameters of the model using a small training set and adapt to new tasks quickly.
\end{enumerate}
In this chapter, we focus on optimization-based methods due to their superiority in several tasks (\egc computer vision  \parencite{finn:17}) over the above-mentioned meta-learning architectures. They achieved stat-of-the-art performance by directly optimizing the gradient towards a proper parameters initialization and fine-tuning on low-resource scenarios,
We investigate the idea of meta-learning for transferring knowledge in a cross-genre and cross-lingual setting for natural language understanding, particularly for NLI and QA tasks.
Specifically, we exploit the usage of Model Agnostic Meta-Learning (MAML), which uses gradient descent and achieves a good generalization for a variety of tasks \parencite{finn:17}. MAML can quickly adapt to new target tasks by using only a few instances at test time, assuming that these new target tasks are drawn from the same distribution.

\begin{figure}[t]
    \centering
    \includegraphics[width=1\textwidth]{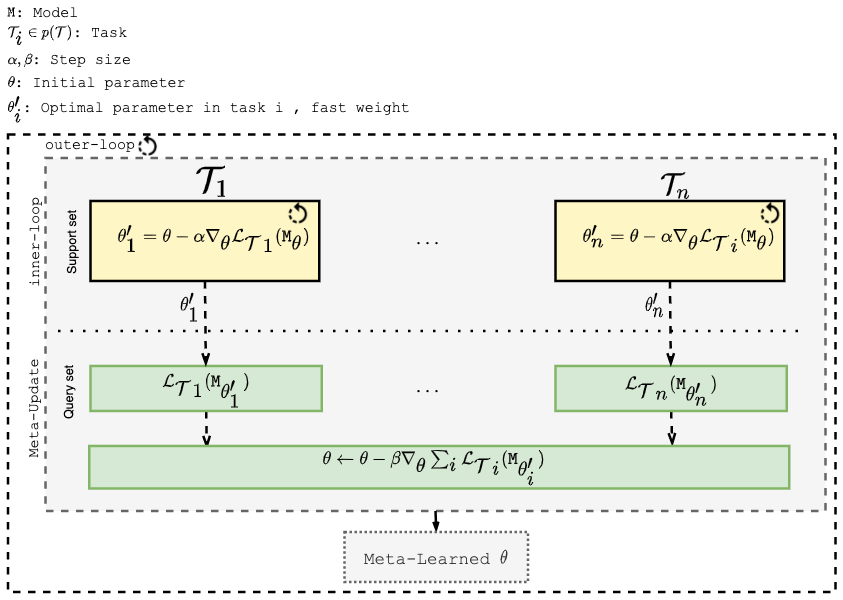}
    \caption{Model Agnostic Meta-Learning (MAML) in supervised learning.}
    \label{fig:MAML-SL}
 \end{figure}
 \begin{figure}[t]
    \centering
    \includegraphics[width=1.01\textwidth]{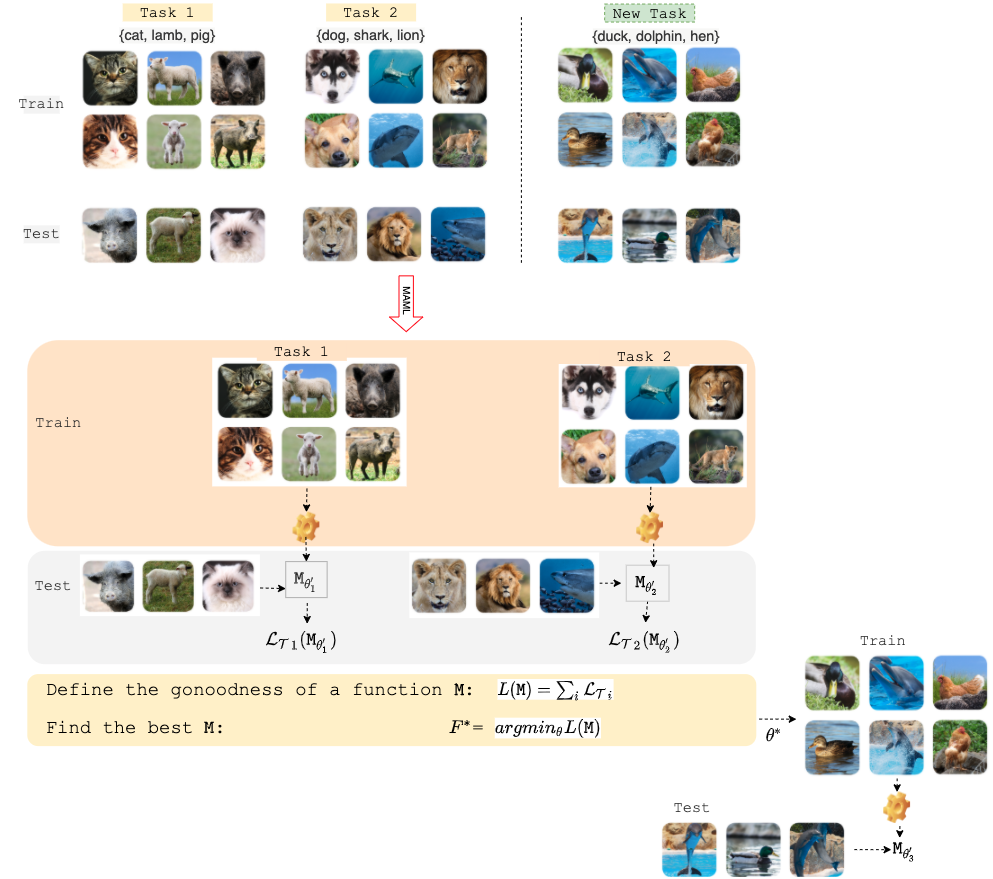}
    \caption{Example of applying Model Agnostic Meta-Learning (MAML) in supervised learning.}
    \label{fig:MAML-example}
 \end{figure}
Formally, MAML (see Figure \ref{fig:MAML-SL}) is applied in supervised learning step-by-step as follows \parencite{finn:17}:
\begin{enumerate}
    \item Let us assumes that there is a model $\learner$ with parameters $\theta$ and a distribution $p(\mathcal{T})$ over tasks.
    \item We sample a batch of tasks $\mathcal{T}_i$ from the distribution $p(\mathcal{T})$. Let us say we sample $n$ tasks as $\{\mathcal{T}_1, ..., \mathcal{T}_n\}$. \label{malm-step2}
    \item In the inner loop, for each task  $\mathcal{T}_i$ in tasks $\mathcal{T}$, we prepare the support set ($\iec D_{i}^{train}$) and query set (\iec $D_{i}^{test}$). The parameters $\theta$ is updated using one or a few iterations of gradient descent steps on the training examples in the support set (\iec $D_{i}^{train}$) of task $\mathcal{T}_i$. For example, for one gradient update,
    \begin{equation}\label{eq:fast-up}
    {\theta_i}^{'} = \theta - \alpha \nabla_\theta \mathcal{L}_{\mathcal{T}_i} (\learner_\theta)
    \end{equation}
where $\alpha$ is the step size, the $\learner_\theta$ is the learned model from the neural network and $\mathcal{L}_{\mathcal{T}_i}$ is the loss on the specific task $\mathcal{T}_i$. \label{malm-step1}
\item The parameters of the model $\theta$ are trained to optimize the performance of $\learner_{\theta_i'}$ on the unseen test examples (\iec $D_{i}^{test}$) across tasks $p(\mathcal{T})$. The meta-learning objective is:
\begin{equation}\label{maml-2}
\min_\theta \sum_{\task_i \sim p(\task)}  \lossi ( \learner_{\theta_i'})
= \min_\theta \sum_{\task_i \sim p(\task)}  \lossi ( \learner_{\theta - \alpha \nabla_\theta \lossi(\learner_\theta)})
\end{equation}
The MAML algorithm aims to optimize the model parameters via a small number of gradient steps on a new task, which we
refer to as the meta-update.
The meta-update across all involved tasks is performed for the $\theta$ parameters of the model using stochastic gradient descent (SGD) as:
\begin{equation}
\label{eq:metaupdate}
\theta \leftarrow \theta - \beta \nabla_\theta \sum_{\task_i \sim p(\task)}  \lossi ( \learner_{\theta_i'})
\end{equation}
where $\beta$ is the meta-update step size. \label{malm-step3}
\item We repeat  step 2 to 4  for $N$ number of iterations as outer-loop. \label{malm-step4}

\item The final parameters $\theta$ is the optimal parameters that can be used to initialize the model $\learner$ in a new task. \label{malm-step5}
\end{enumerate}
The  meta-update step in MAML (Eq. \ref{eq:metaupdate}) involves a gradient
through a gradient which can be both computationally and memory intensive. A modified version of MAML ignores the second derivative \parencite{finn:17}, resulting in a simplified and cheaper implementation, known as First-Order MAML (FOMAML):
\begin{equation}
\label{eq:metaupdate-fomaml}
\theta \leftarrow \theta - \beta  \sum_{\task_i \sim p(\task)} \nabla_{\theta_i'} \lossi ( \learner_{\theta_i'})
\end{equation}
\noindent We further illustrate MAML with a simple example, shown in Figure \ref{fig:MAML-example}. We here have two image classification tasks: One training task (Task 1), to label images as a cat, lamb, or pig, and another task, to label images as a dog, shark, or lion. The aim is to train a neural network model $\learner$ towards parameters that can adapt quickly and with few examples to a novel classification task (i.e., to label images as a duck, dolphin, or hen). First, we randomly initialize our model parameters $\theta$. We train our model on Task 1 using the train set and minimize the loss using gradient descent and find the optimal parameters ${\theta_1}^{'}$ (see Eq. \ref{eq:fast-up}). Similarly, for Task 2, we start with a randomly initialized model parameters $\theta$ and minimize the loss by finding the optimal parameters ${\theta_2}^{'}$ using gradient descent. In the next step, we perform meta-optimization in each task's test set by minimizing the loss in the test set. We calculate the losses (i.e., $\loss_{\task_{1}}(\learner_{{\theta_1}^{'}})$ and $\loss_{\task_{2}}(\learner_{{\theta_2}^{'}})$ ) by taking the gradient with respect to our optimal parameters calculated in the previous step ${\theta_1}^{'}$ and ${\theta_2}^{'}$. Then, we update the original parameters $\theta$ using the test sets of Tasks 1 and 2 (see Eq. \ref{eq:metaupdate}). During meta-training, the MAML learns the optimal initialization parameters that allow the model $\learner$ to adapt quickly and efficiently to a new few-shot task with new, unseen classes (i.e., Task 3).

\section{Related Work}
This chapter's primary motivation is the low availability of labeled training datasets for most of the different text genres and languages. To alleviate this issue, several methods, including so-called few-shot learning approaches, have been proposed.
Few-shot learning methods have initially been introduced within the area of image classification \parencite{vinyals:16,ravi:17,finn:17}, but have recently also been applied to NLP tasks such as relation extraction \parencite{han:18}, text classification \parencite{yu:18} and machine translation \parencite{gu-etal-2018-meta}. Specifically, in NLP, these few-shot learning approaches include either: (i) the transformation of the problem into a different task (\egc relation extraction is transformed to question answering  \parencite{abdou-etal-2019-x,levy-etal-2017-zero}); or (ii) meta-learning \parencite{andrychowicz:16,finn:17}.

\subsection{Meta-Learning}
Meta-learning has recently received much attention from the NLP community.
It has been applied to the task of machine translation \parencite{gu-etal-2018-meta}, where they propose to use meta-learning for improving the machine translation performance for low-resource languages by learning to adapt to target languages based on multilingual high-resource languages.
They show that the use of meta-learning significantly outperforms the multilingual, transfer learning-based approach proposed by \cite{zoph-etal-2016-transfer} and enables them to train a competitive neural machine translation system with only
a fraction of training examples.
However, in the proposed framework, they include 18 high-resource languages as auxiliary languages and five diverse low-resource languages as target languages. In this chapter, we assume access to only English as a high-resource language.

For the task of dialogue generation, \cite{qian-yu-2019-domain} address domain adaptation using meta-learning. They introduce
an end-to-end trainable dialog system
that learns from multiple resource-rich tasks and is adapted to new domains with minimal training samples using meta-learning. Model-agnostic meta-learning (MAML) \parencite{finn:17} is applied to the dialog domain and adapts a dialog system model using multiple resource-rich single domain dialog datasets. They show that the meta-learning enables the model to learn general features across multiple tasks and is capable of learning a competitive dialog system on a new domain with only a few training examples in an efficient manner.

\cite{dou-etal-2019-investigating} explore model-agnostic meta-learning (MAML) and variants thereof for low-resource NLU tasks in the GLUE dataset \parencite{DBLP:journals/corr/abs-1804-07461}. They consider different high-resource NLU tasks such as MultiNLI \parencite{williams-etal-2018-broad} and QNLI \parencite{rajpurkar-etal-2016-squad} as auxiliary tasks to learn meta-parameters using MAML. Then, they fine-tune the low-resource tasks using the adapted parameters from the meta-learning phase. They demonstrate the effectiveness of model-agnostic meta-learning in NLU tasks and show that the learned representations can be adapted to new tasks effectively.

\cite{obamuyide-vlachos-2019-model} show that framing relation classification as an instance of meta-learning improves the performance of supervised relation classification models, even
with limited supervision at training time. They apply model-agnostic meta-learning to explicitly learn model parameters initialization for enhanced predictive performance across all relations with limited supervision in relation classification.

Recently, model-agnostic meta-learning has been applied to the task of Natural Language Generation (NLG) \parencite{Mi2019MetaLearningFL}. They formulate the problem from a meta-learning perspective and propose a generalized optimization-based approach. They show that the meta-learning based approach significantly outperforms other training procedures since it adapts fast and well to new low-resource
settings.

All the works mentioned above on meta-learning in NLP assume that there are multiple high-resource tasks or languages, which are then adapted to new target tasks or languages with a handful of training samples.
However, in a cross-lingual NLI and QA setting, the available high-resource language is usually only English.
\subsection{Cross-Lingual NLU}
Cross-lingual learning has a fairly short history in NLP, and has mainly been restricted to traditional NLP tasks, such as PoS tagging and parsing. In contrast to these tasks, which have seen much cross-lingual attention \parencite{plank2016multilingual,bjerva:2017,delhoneux:2018}, there has been relatively little work on cross-lingual NLU, partly due to a lack of benchmark datasets. Existing work has mainly been focused on NLI \parencite{agic2017baselines,conneau2018xnli}, and to a lesser degree on RE \parencite{faruqui2015multilingual,verga2015multilingual} and QA \parencite{lewis2019mlqa, abdou-etal-2019-x}.
Previous research generally reports that cross-lingual learning is challenging and that it is hard to beat a machine translation baseline (\egc \cite{conneau2018xnli}). Such a baseline is (for instance) suggested by \cite{faruqui2015multilingual}, where the text in the target language is automatically translated to English.
For many language pairs, a machine translation model may be available, which can be used to obtain data in the target language. To evaluate the impact of using such data, in much of previous research work, the English training data is translated into the target language using a machine translation system. Then, the model  is fine-tuned on the translated data and evaluated on the test set of target languages and reported as a TRANSLATE-TRAIN baseline. Alternatively, after fine-tuning the model on the English training data, a TRANSLATE-TEST baseline is introduced by evaluating the model on the test data that is translated from the target language to English using the machine translation system.

In this chapter, we show that our meta-learning based framework can achieve competitive performance compared to a machine translation baseline (for XNLI), and propose a method that requires no training instances for the target task in the target language.

\section{Cross-Lingual Meta-Learning}
\label{sec:mode_xmaml}

The underlying idea of using MAML in NLP tasks \parencite{gu-etal-2018-meta,dou-etal-2019-investigating,qian-yu-2019-domain} is to employ a set of high-resource auxiliary tasks or languages to find an optimal initialization from which learning a target task or language can be done using only a small number of training instances. In a cross-lingual setting (\iec XNLI, MLQA), where only an English dataset is available as a high-resource language, and a small number of instances are available for other languages, the training procedure for MAML requires some non-trivial changes.
For this purpose, we introduce a cross-lingual meta-learning framework (X-MAML), which uses the following training steps (a more formal description of the proposed model X-MAML is given in Algorithm \ref{alg:X-MAML}):
\begin{enumerate}
\item Pre-training on the high-resource language \texttt{h} (\iec English): Given all the training samples in the high-resource language \texttt{h}, we first train the model \texttt{M} on \texttt{h} to initialize the model parameters $\theta$. \label{step1}
\item Meta-learning using low-resource languages \texttt{L}: This step consists of choosing one or more auxiliary languages \texttt{A} from the low-resource set \texttt{L}. Using the development set of each auxiliary language in \texttt{A}, we construct a randomly sampled batch of tasks $\task_i$. Then, we update the model parameters using $K$ data points of $\task_i$ ($D_{i}^{train}$) by one gradient descent step (see ~\equref{eq:fast-up}). After this step, we can calculate the loss value using $Q$ examples ($D_{i}^{test}$) in each task.  It should be noted that the $K$ data points used for training ($D_{i}^{train}$) are different from the $Q$ data points used for constructing $D_{i}^{test}$. We sum up the loss values from all tasks to minimize the meta-objective function and to perform a meta-update using \equref{eq:metaupdate}. This step is performed in multiple iterations. \label{step2}
\item Zero-shot or few-shot learning on the target languages $\{\texttt{L} \smallsetminus \texttt{A}\}$: In the last step of X-MAML, we first initialize the model parameters with those learned during
meta-learning. We then continue by evaluating the model on the test set of the target languages (\iec zero-shot learning) or fine-tuning the
model parameters with standard supervised learning using the development set of target languages and evaluate on the test set (\iec few-shot learning). \label{step3}
\end{enumerate}
\begin{algorithm}[t]
\DontPrintSemicolon
 \KwInput{high-resource language \texttt{h}, set of low-resource languages \texttt{L},\\ Model $
 \learner$, step size $\alpha$ and learning rate $\beta$}

 Pre-train $\learner$ on \texttt{h} and provide initial model parameters $\theta$\\
 Select one or more languages from \texttt{L} as a set of auxiliary languages (\texttt{A}) \\
  \While{not done}
    {
    \For{\texttt{l} $\in$ \texttt{A}}
    {
    Sample batches of tasks $\task_i$ using the development set of the auxiliary language $l$ \\
    \For {each $\task_i$}
    {
    Sample $K$ data-points to form $D_{i}^{train}=\{(X^{k},Y^{k})\}_{k=1}^{K} \in \task_i $\\
    Sample $Q$ data-points to form $D_{i}^{test}=\{(X^{q},Y^{q})\}_{q=1}^{Q} \in \task_i$~for meta-update\\
    Compute $\nabla_{\theta}\lossi(\learner_{\theta})$ on $D_{i}^{train}$ \\
    Compute adapted parameters with gradient descent:
    $\theta^{'}= \theta - \alpha\nabla_{\theta}\lossi(\learner_{\theta})$ \\
    Compute $\lossi(\learner_{\theta^{'}})$ using $D_{i}^{test}$  \\
    }
    }
    Update $ \theta \leftarrow \theta - \beta \nabla_{\theta} \sum_{i} \lossi(\learner_{\theta^{'}})$ \\
  }
Perform either (i) zero-shot or (ii) few-shot  learning on \{\texttt{L} $\smallsetminus$ \texttt{A}\} using meta-learned parameters $\theta$

\caption{X-MAML.}
\label{alg:X-MAML}
\end{algorithm}
\section{Experiments}
\label{sec:experiments}
In this section, we address our research questions (i.e., RQ \ref{rq.5.1} and RQ \ref{rq.5.2} are explored across Sections \ref{ch5:mnli}, \ref{ch5:xnli} and \ref{X-MAML:QA}, and RQ \ref{rq.5.3} is addressed in Section \ref{X-MAML:QA}) by conducting a set of experiments.
We perform experiments on the MultiNLI, XNLI, and MLQA datasets using different NLU models, as explained in Section \ref{xmaml:nlu-models}. We report results for {few-shot} as well as {zero-shot} cross-genre and cross-lingual learning. To examine the model- and task-agnostic features of X-MAML, we conduct experiments with various models for both tasks.

\subsection{Experimental Setup:}
We implement X-MAML using the \textit{higher} library.\footnote{\url{https://github.com/facebookresearch/higher}}
We use the Adam optimizer \parencite{Adam} with a batch size of 32 for both zero-shot and few-shot learning. We fix the step size $\alpha$ and learning rate $\beta$ to $1e-4$ and $1e-5$, respectively. We experimented using  $[10,20,30,50,100,200,300]$ meta-learning iterations in X-MAML. However, $100$ iterations led to the best results in our experiments.  We sample two sets of 16 data points from the batch to construct $D^{train}$ and $D^{test}$ (i.e.,  The sample sizes $K$ and $Q$ in X-MAML are equal to 16 for each dataset). We report results for each experiment by averaging the performance over ten different runs (i.e., various random seeds). An evaluation of NLI benchmarks is performed
reporting accuracy on the respective test sets. For the evaluation of the QA dataset, we use the F$_1$ score following the multilingual evaluation script available with the MLQA data \footnote{\url{https://github.com/facebookresearch/MLQA}}.
\paragraph{Baselines:}
In order to evaluate the impact of meta-learning on various scenarios, we create our baseline for each scenario.
We create:
\begin{enumerate*}[label=(\roman*)]
\item zero-shot baselines: directly evaluate the model on the test set of the target languages and genres (for each task), and
\item few-shot baselines: fine-tune the model on the development set and evaluate on the test set of the low-resource languages and genres.
\end{enumerate*}

\subsection{Few-Shot Cross-Genre NLI} \label{ch5:mnli}
To verify our learning routine more generally and to address the research question RQ \ref{rq.5.1}, we define $\task_i$ as an NLI task in each genre. We exploit MAML, in its original setting (see Section \ref{maml-desc}), to investigate whether meta-learning encourages the model to learn a good initialization for all target genres, which can then be fine-tuned with limited supervision for each genre's development instances (2000 examples) to achieve a good performance on its test set.
In MultiNLI, which is a cross-genre dataset, we employ the Enhanced Sequential Inference Model (ESIM), as explained in Section \ref{xmalm-esim}.
We train ESIM on the MultiNLI training set to provide initial model parameters $\theta$. We evaluate the pre-trained model on the English test set of XNLI (since the MultiNLI test set is not publicly available) to set the baseline for this scenario.
Since MultiNLI is already split into genres, we use each genre as a task within MAML.
We then include either the training set (5 genres) or the development set (10 genres) during meta-learning (similar to Step~\ref{step2} in  ~\secref{sec:mode_xmaml}).
\begin{table}[t]
 \centering
 	\resizebox{.5\linewidth}{!}{
\begin{tabular}{ccccc}
\toprule
& \textbf{Baseline} &\multicolumn{2}{c}{\textbf{MAML}}\\
\cmidrule(l){3-4}
\Large{x}\% & &  $\task_{Train}$ &  $\task_{Dev}$   \\
\midrule
\texttt{1}& 38.60 &	49.78 &	{\bf 50.92}  \\
\texttt{2} &37.80 &	48.58 &	{\bf 50.66}  \\
 \texttt{3} &47.09 &	51.40 &	{\bf 52.85}   \\
 \texttt{5} &49.88 &	{\bf 52.22} &	51.40   \\
 \texttt{10} &51.02 &	52.51 &	{\bf 53.95}  \\
 \texttt{20} &59.14&	{\bf 61.38} &	58.16  \\
 \texttt{50}&63.37&	{\bf 63.85} &
 61.74  \\
 \texttt{100}&64.35&{\bf 64.99} &64.61  \\
\bottomrule
\end{tabular}
}
        \caption[Test accuracies with different settings of MAML on MultiNLI.]{Test accuracies with different settings of MAML on MultiNLI. \texttt{x}\%: the percentage of training samples. {\bf Baseline:} The test accuracy of trained ESIM using \texttt{x}\% of training data. {\bf MAML:} The test accuracy of ESIM after meta-learning, where $\task_{Train}$: 5 tasks are defined in MAML using the training set, and $\task_{Dev}$: 10 tasks are included in MAML using the development set. Bold font indicates best results for the various proportions of the used training data.}
\label{tbl:mnli}
\end{table}
In the last phase (similar to Step~\ref{step3} in X-MAML), we first initialize the model parameters with those learned by MAML. We then continue to fine-tune the model using the development set of MultiNLI and report the accuracy on the English test set of XNLI.
We proportionally select sub-samples $x=[1\%,2\%,3\%,5\%,10\%,20\%,50\%,100\%]$ from the training data (with random sampling).
The results obtained by training on the corresponding proportions ($x\%$) of the MultiNLI dataset using ESIM (as the learner model \texttt{M}) are shown in \tabref{tbl:mnli}.

We observe that for both settings (\iec MAML on training (5 tasks) and on development (10 tasks)), the performances of all models (including baselines) improve as more training instances become available. However, as demonstrated by our experimental study, the effectiveness of MAML is larger when only limited training data is available (improving by $12\%$ in accuracy when $2\%$ of the data is available on the development set).

\subsection{Zero- and Few-Shot Cross-Lingual NLI}\label{ch5:xnli}
We now aim to answer the questions RQ \ref{rq.5.1} and RQ \ref{rq.5.2} and proceed to investigate zero- and few-shot X-MAML for the cross-lingual NLI task.
In XNLI, which is a cross-lingual dataset, we employ the PyTorch version of BERT \footnote{\url{https://github.com/huggingface/transformers}} \parencite{DBLP:journals/corr/abs-1810-04805} as the underlying model \texttt{M} (see Section \ref{xmaml:nlu-models}). However, since our proposed meta-learning method is model-agnostic, it can easily be extended to any other architecture.
Following our X-MAML framework, we first study the impact of meta-learning with one low-resource language to serve as an auxiliary language. We evaluate the performance of a cross-lingual NLI model on the set of languages provided in the XNLI dataset. In the following sections, we study the effect of X-MAML on the performance of the En-BERT and Multi-BERT models (see Section \ref{xmaml:nlu-models}) for the cross-lingual NLI task.
\subsubsection{Zero-Shot Learning}
In this set of experiments, we employ the proposed framework (\iec X-MAML) within a zero-shot setup, in which we do not fine-tune after the meta-learning step.

\begin{figure}[t]
\centering
\includegraphics[width=1\textwidth]{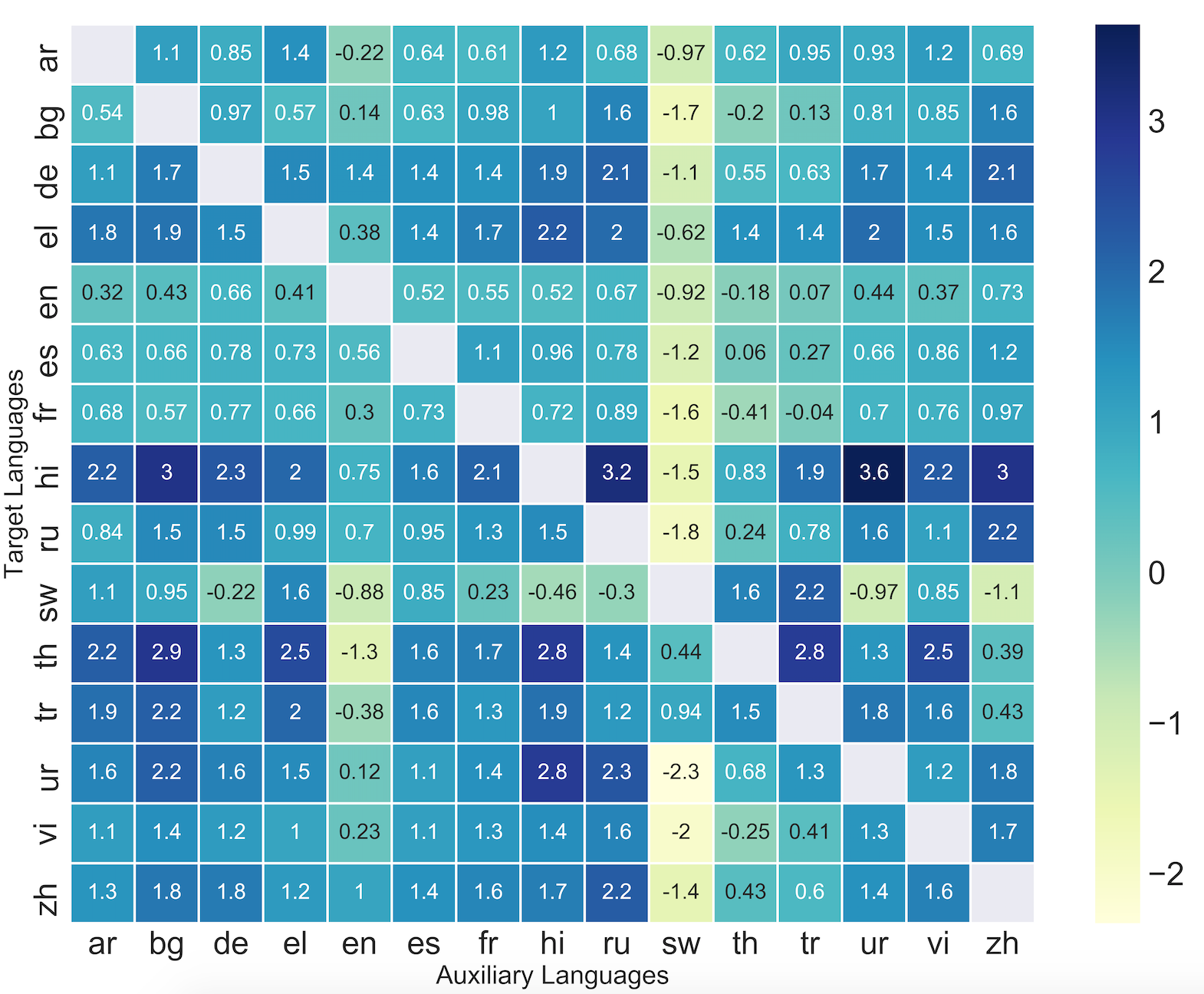}
\caption[Differences in performance in terms of accuracy scores on the test set for zero-shot X-MAML on XNLI using the Multi-BERT model.]{Differences in performance in terms of accuracy scores on the test set for zero-shot X-MAML on XNLI using the Multi-BERT model. Rows correspond to target and columns to auxiliary languages used in X-MAML. Numbers on the off-diagonal indicate performance differences  between X-MAML and the baseline model in the same row. The coloring scheme indicates the differences in performance (\egc blue for large improvement).}
\label{fig:heat-map-mBERT-zero}
\end{figure}
\paragraph{Zero-shot X-MAML with Multi-BERT}
\setlength{\tabcolsep}{4pt}
\begin{sidewaystable}
\begin{center}
\resizebox{1\linewidth}{!}{
\begin{tabular}[b]{p{4.5cm}|ccccccccccccccc|c}
\toprule
& \textbf{en} & \textbf{fr} & \textbf{es} & \textbf{de} & \textbf{el} & \textbf{bg} & \textbf{ru} & \textbf{tr} & \textbf{ar} & \textbf{vi} & \textbf{th} & \textbf{zh} & \textbf{hi} & \textbf{sw} & \textbf{ur} & \textbf{avg} \\
\midrule
\rowcolor{gray!20}\multicolumn{16}{l}
{\it  Zero-shot cross-lingual transfer} \\
\midrule
\cite{DBLP:journals/corr/abs-1810-04805} &81.4& - &74.3 &70.5 &- &-& -& - &62.1& -& -& 63.8 &- &-& 58.35 & - \\
\cite{DBLP:journals/corr/abs-1904-09077} &  82.1 & 73.8 & 74.3 & 71.1 & 66.4 & 68.9 & 69.0 & 61.6 & 64.9 & 69.5 &  55.8 & 69.3 & 60.0 & 50.4 & 58.0  & 66.3 \\
Multi-BERT (Our baseline) & 81.36 & 73.45&73.85& 69.74& 65.73 & 67.82& 67.94 & 59.04 & 64.63 & 70.12 & 52.46 & 68.90 & 58.56 & 47.58 & 58.70 & 65.33 \\
\midrule
\multicolumn{16}{l}{\it \textbf{X-MAML} (One aux. lang.)} \\
\midrule
AVG & 81.69	& 73.86 &	74.43	& 71.00 &	67.16 &	68.39 &	68.90 &	60.41 &	65.33 &	70.95 &	54.08 &	70.09	& 60.51&	47.97&	59.94 & -\\
MAX & 82.09 &	 74.42 &	 75.07 &	 71.83 &	67.95 &	 69.45 &	 70.19 &	61.20 &	 66.05&	{ 71.82}&	55.39&	{ 71.11}&	{ 62.20} &	49.76 &	{61.51}&{ 67.33} \\
$hi\rightarrow X$ & 81.88	& 74.17 &	74.81 &	71.59 &	67.95 &	68.86 & 69.44 &	60.93 &	65.86 &	71.57 &	55.26 &	70.59 &	- &	47.12 &	61.51 & - \\
\midrule
{\it \textbf{X-MAML} (Two aux. lang.)} & {\it (hi,de)} & {\it (hi,ar)} & {\it (fr,de)} & {\it (bg,zh)} & {\it (ur,ru)} & {\it (hi,ru)} & {\it (de,bg)} & {\it (ur,sw)} & {\it (el,tr)} & {\it (de,bg)} & {\it (bg,tr)} & {\it (ru,el)} & {\it (ur,ru)} & {\it (el,tr)} & {\it (hi,de)}\\
\midrule
$(l_1,l_2) \rightarrow X$& {\bf 82.59} & {\bf 75.69} & {\bf 75.97} & {\bf 73.45} & {\bf 69.16} & {\bf 71.42} & {\bf 71.44}& {\bf 62.57} & {\bf 67.19} & {\bf 72.63} & {\bf 62.57} &{\bf 73.13} & {\bf 63.53} & {\bf 50.42} & {\bf 62.93} & {\bf 68.98}
\\
\midrule
\midrule
\rowcolor{gray!20}\multicolumn{16}{l}{\it  Few-Shot learning} \\
\midrule
Multi-BERT (Our baseline) & 81.94 & {  75.39} & 75.79 & 73.25 & 69.54 &71.60 &70.84 & 64.85 &67.37 &73.23 & 61.18  & 73.93  &  64.37 & 57.82  &  63.71   &  69.65 \\
\midrule
\multicolumn{16}{l}{\it \textbf{X-MAML} (One aux. lang.)} \\
\midrule
AVG & 82.22&	75.24&	76.06	&73.34&	69.97&	71.80&	71.28&	64.76&	67.82&	73.41&	61.57&	74.02&	64.83&	58.02&	63.66& -\\
MAX& { 82.39}	&  75.32& 	{  76.18}	& {  73.46}& 	{  70.03}& 	{  71.94}& 	{  71.45}& 	{  64.92}& 	{  67.95}& 	{  73.52}& 	{  61.74}& 	{  74.21}& {  	64.97}& 	{  58.23}& 	{  63.81} & { 70.01}\\
$sw\rightarrow X$ & 82.24	&75.31&	75.94&	73.34&	69.98&	71.77&	71.31&	64.89&	67.87	&73.38&	61.5&	73.99&	64.94&	- &	63.63 & -\\ 
\midrule
{\it \textbf{X-MAML} (Two aux. lang.)} & \textit{ (ar,ru)}&\textit{ (ru,th)}&\textit{ (ru,th)}&\textit{ (el,hi)}&\textit{ (sw,vi)}&\textit{ (ar,zh)}&\textit{ (de,tr)}&\textit{ (es,sw)}&\textit{ (bg,hi)}&\textit{ (bg,ru)}&\textit{ (el,vi)}&\textit{ (ar,th)}&\textit{ (sw,vi)}&\textit{ (ar,tr)}&\textit{ (en,ru)}\\
\midrule
$(l_1,l_2) \rightarrow X$& {\bf 82.71} & 75.97 & 76.51 & 74.07 & 70.66 & 72.77 & 72.12 & 65.69 & 68.4 & {\bf 73.87} & 62.5 & 74.85 & {\bf 65.75} & 59.94 & {\bf 64.59} & 70.69
\\
\midrule
\rowcolor{gray!20}\multicolumn{16}{l}{\it Machine translate at test (TRANSLATE-TEST)} \\
\midrule
\cite{DBLP:journals/corr/abs-1810-04805} & 81.4& - &74.9 &74.4 &- &-& -& -& 70.4& -& -& 70.1& - &- &62.1 & -\\
\midrule
\rowcolor{gray!20}\multicolumn{16}{l}{\it Machine translate at training (TRANSLATE-TRAIN)} \\
\midrule
\cite{DBLP:journals/corr/abs-1904-09077} &  82.1&	{\bf 76.9}&	{\bf 78.5}&	{\bf 74.8}&	{\bf 72.1}&	{\bf 75.4}&	{\bf 74.3}&	{\bf 70.6}&	{\bf 70.8}&	67.8&	{\bf 63.2}&	{\bf 76.2}&	65.3&	65.3&	60.6 & {\bf 71.6} \\
\bottomrule
\end{tabular}
}
\caption[Accuracy results on the XNLI test set for zero- and few-shot X-MAML. Columns indicate the target languages.]{Accuracy results on the XNLI test set for zero- and few-shot X-MAML. Columns indicate the target languages.
The models of  \cite{DBLP:journals/corr/abs-1810-04805} and \cite{DBLP:journals/corr/abs-1904-09077} are also Multi-BERT models. For our Multi-BERT baseline model for (i) zero-shot learning, we evaluate the pre-trained model on the test set of the target language; and for (ii) few-shot learning, we fine-tune the model on the development set and evaluate on the test set of the target language. The avg column indicates row-wise average accuracy. We also report the average (AVG) and maximum (MAX) performance by using one auxiliary language for each target language.
$(l_1,l_2)$ are the most beneficial auxiliary languages for X-MAML in improving the test accuracy of each target language $X$. In TRANSLATE-TEST  \parencite{DBLP:journals/corr/abs-1810-04805}, the target language test data is translated to English and then the model is fine-tuned on English. In TRANSLATE-TRAIN  \parencite{DBLP:journals/corr/abs-1904-09077}, the English training data is translated to the target language and the model is fine-tuned using the translated data.}
\label{tab:originalBERT}
\end{center}
\end{sidewaystable}
As the first training step (\iec pre-training on a high-resource language, see Step~\ref{step1} in~\secref{sec:mode_xmaml} for more information) in X-MAML for XNLI, we fine-tune Multi-BERT on the MultiNLI dataset (English) to obtain the initial model parameters $\theta$ for each experiment.
We go on to apply the second and third steps of X-MAML in the zero-shot scenario. We report the impact of meta-learning for each target language as a difference in accuracy with and without meta-learning on top of the baseline model (Multi-BERT) on the test set (\Figref{fig:heat-map-mBERT-zero}). Each column corresponds to the performance of Multi-BERT after meta-learning with a single auxiliary language, and evaluation on the target language of the XNLI test set. In general, we observe that our zero-shot approach with X-MAML outperforms the baseline model without MAML and results reported by \cite{DBLP:journals/corr/abs-1810-04805}. This way, we improve the performance of Multi-BERT in zero-shot cross-lingual NLI. We observe the largest difference in performance when transferring from Urdu (ur) as an auxiliary language to Hindi (hi) as a target (\egc $+3.6\%$ in accuracy). We also detect strong gains when transferring from Urdu (ur),  Russian (ru), and Bulgarian (bg) as auxiliary languages in X-MAML.

Furthermore, Hindi (hi) is the most effective auxiliary language and provides the highest average accuracy in the zero-shot setting. \tabref{tab:xnli-mBERT-zeroshot} shows the average accuracy over ten runs of X-MAML on the XNLI dataset using Multi-BERT as the base model. Each column corresponds to the performance of the Multi-BERT system after meta-learning with a single auxiliary language, and evaluation on the target language of the XNLI test set. 

We hypothesize that the degree of typological commonalities among the languages affects (\iec positive or negative) on the performance of X-MAML and will return to this below. It can be observed that the proposed learning approach provides positive impacts across most of the target languages. However, including Swahili (sw) as an auxiliary language in X-MAML is not beneficial for the performance on the other target languages.

\begin{table}[t]
\resizebox{1\linewidth}{!}{
\begin{tabular}{lccccccccccccccc|r}
\toprule
 &\multicolumn{15}{c|}{\textbf{Auxiliary language}} & \textbf{baseline}\\
{} &     \textbf{ar} &     \textbf{bg} &     \textbf{de} &     \textbf{el} &     \textbf{en} &     \textbf{es} &     \textbf{fr} &     \textbf{hi} &     \textbf{ru} &     \textbf{sw} &     \textbf{th} &     \textbf{tr} &     \textbf{ur} &     \textbf{vi} &     \textbf{zh}  & \\
\midrule
\textbf{ar} &   - &  65.76 &  65.48 &  66.05 &  64.41 &  65.27 &  65.24 &  65.86 &  65.31 &  63.66 &  65.25 &  65.58 &  65.56 &  65.84 &  65.32 & 64.63\\
\textbf{bg} &  68.36 &   - &  68.79 &  68.39 &  67.95 &  68.45 &  68.80 &  68.86 &  69.41 &  66.10 &  67.62 &  67.95 &  68.63 &  68.67 &  69.45 & 67.82 \\
\textbf{de} &  70.88 &  71.46 &   - &  71.26 &  71.09 &  71.12 &  71.11 &  71.59 &  71.83 &  68.65 &  70.29 &  70.37 &  71.42 &  71.15 &  71.83 & 69.74 \\
\textbf{el} &  67.53 &  67.58 &  67.25 &   - &  66.11 &  67.13 &  67.39 &  67.95 &  67.71 &  65.11 &  67.12 &  67.15 &  67.69 &  67.19 &  67.34 & 65.73\\
\textbf{en} &  81.68 &  81.79 &  82.02 &  81.77 &   - &  81.88 &  81.91 &  81.88 &  82.03 &  80.44 &  81.18 &  81.43 &  81.80 &  81.73 &  82.09 & 81.36\\
\textbf{es} &  74.48 &  74.51 &  74.63 &  74.58 &  74.41 &   - &  74.95 &  74.81 &  74.63 &  72.66 &  73.91 &  74.12 &  74.51 &  74.71 &  75.07 & 73.85\\
\textbf{fr} &  74.13 &  74.02 &  74.22 &  74.11 &  73.75 &  74.18 &   - &  74.17 &  74.34 &  71.87 &  73.04 &  73.41 &  74.15 &  74.21 &  74.42 & 73.45\\
    \textbf{hi} &  60.75 &  61.59 &  60.84 &  60.61 &  59.31 &  60.18 &  60.66 &   - &  61.75 &  57.10 &  59.39 &  60.47 &  62.20 &  60.76 &  61.56 & 58.56\\
\textbf{ru} &  68.78 &  69.47 &  69.47 &  68.93 &  68.64 &  68.89 &  69.25 &  69.44 &   - &  66.11 &  68.18 &  68.72 &  69.52 &  69.02 &  70.19 & 67.94\\
\textbf{sw} &  48.71 &  48.53 &  47.36 &  49.13 &  46.70 &  48.43 &  47.81 &  47.11 &  47.28 &   - &  49.20 &  49.76 &  46.61 &  48.43 &  46.50 & 47.58\\
\textbf{th} &  54.65 &  55.39 &  53.80 &  54.98 &  51.14 &  54.09 &  54.15 &  55.26 &  53.82 &  52.90 &   - &  55.24 &  53.79 &  54.99 &  52.85 & 52.46 \\
\textbf{tr} &  60.94 &  61.20 &  60.22 &  61.09 &  58.66 &  60.60 &  60.32 &  60.93 &  60.29 &  59.98 &  60.53 &   - &  60.82 &  60.68 &  59.47 & 59.04\\
\textbf{ur} &  60.30 &  60.87 &  60.34 &  60.20 &  58.82 &  59.81 &  60.12 &  61.51 &  61.02 &  56.37 &  59.38 &  60.02 &   - &  59.87 &  60.46 & 58.70 \\
\textbf{vi} &  71.27 &  71.56 &  71.32 &  71.14 &  70.35 &  71.22 &  71.42 &  71.57 &  71.73 &  68.11 &  69.87 &  70.53 &  71.43 &   - &  71.82 & 70.12\\
\textbf{zh} &  70.24 &  70.68 &  70.65 &  70.12 &  69.91 &  70.29 &  70.47 &  70.59 &  71.11 &  67.47 &  69.33 &  69.50 &  70.29 &  70.54 &   -  & 68.90\\
\bottomrule
\end{tabular}
}
\captionof{table}[The performance in terms of average test accuracy for the zero-shot setting over 10 runs of X-MAML on the XNLI dataset using Multi-BERT.]{The performance in terms of average test accuracy for the zero-shot setting over 10 runs of X-MAML on the XNLI dataset using Multi-BERT (multilingual BERT), as base model. Each column corresponds to the performance of the Multi-BERT system after meta-learning with a single auxiliary language, and evaluation on the target language of the XNLI test set. The auxiliary language is not included during the evaluation phase. Results of the Multi-BERT model without X-MAML (baseline) are also reported.}
\label{tab:xnli-mBERT-zeroshot}
\end{table}

In Table \ref{tab:originalBERT}, we include the original baseline performances reported in \cite{DBLP:journals/corr/abs-1810-04805}\footnote{\url{https://github.com/google-research/bert/blob/master/multilingual.md}} and \cite{DBLP:journals/corr/abs-1904-09077}. We report the average and maximum performance by using one auxiliary language for each target language. We also report the performance of X-MAML by also using Hindi (which is the most effective auxiliary language for the zero-shot setting, as shown in ~\Figref{fig:heat-map-mBERT-zero}). We Once again, suspect that this may be because of the typological similarities between Hindi (hi) and other languages.

Now we conduct the zero-shot X-MAML using two auxiliary languages (see Step~\ref{step3} in~\secref{sec:mode_xmaml}). The results (Table \ref{tab:originalBERT}) show that X-MAML using two auxiliary languages obtains the largest benefit in the zero-shot experiments. It improves our internal Multi-BERT baseline by $+3.65\%$ points in terms of average accuracy \footnote{We consider only the best auxiliary languages for each target language, and then take the average.} on the zero-shot scenario. We report the most beneficial pair of auxiliary languages for the zero-shot X-MAML in improving the test accuracy of each target language in Table \ref{tab:originalBERT}.

\begin{figure}[t]
\includegraphics[width=1\textwidth]{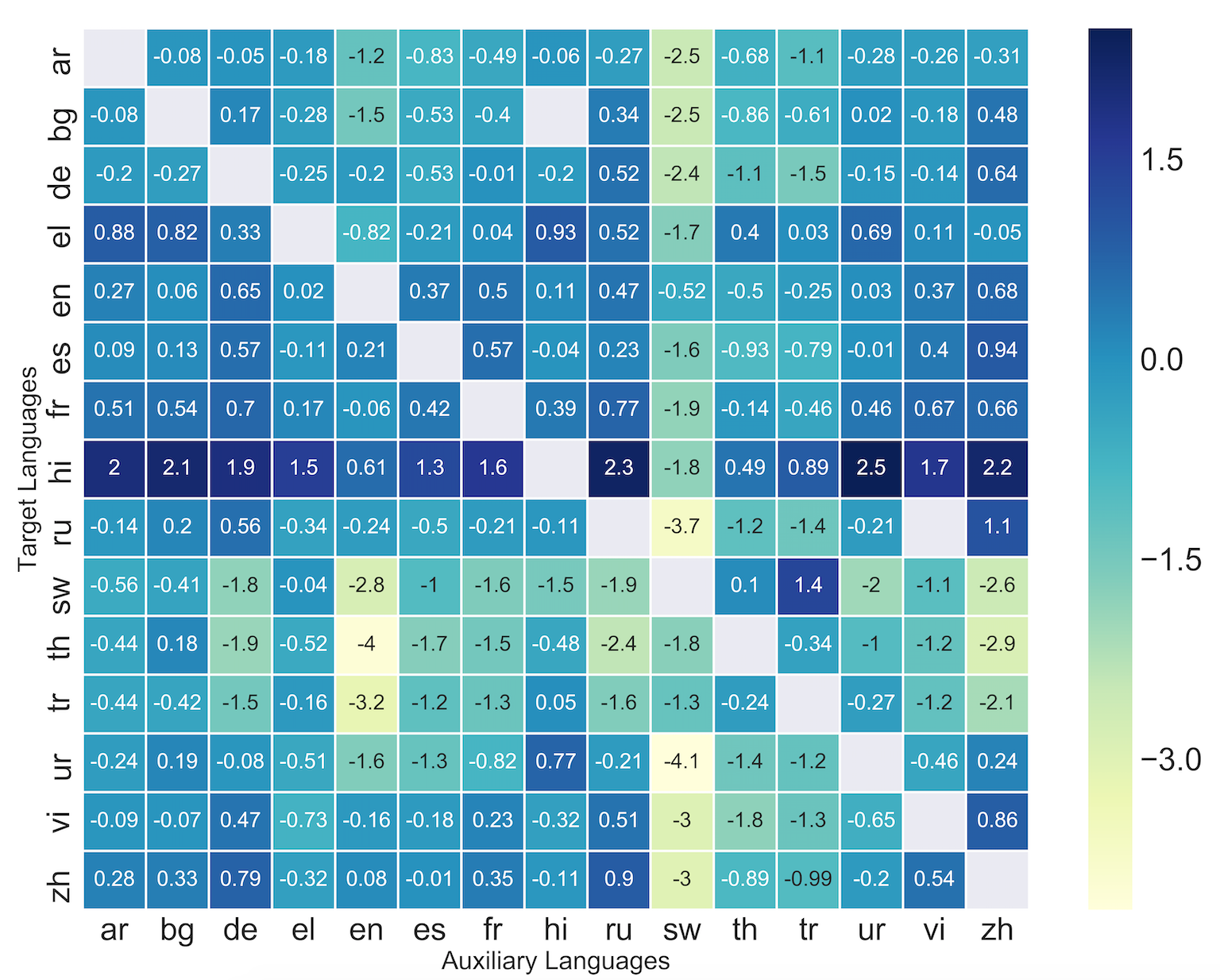}
    \caption[Differences in performance in terms of accuracy scores on the test set for the zero-shot case using training (without meta-learning) on XNLI with the Multi-BERT model.]{Differences in performance in terms of accuracy scores on the test set for the zero-shot case using training (without meta-learning) on XNLI with the Multi-BERT model. Rows correspond to target and columns to auxiliary languages used in X-MAML. Numbers on the off-diagonal indicate performance differences between training on the auxiliary languages (without meta-learning) and the baseline model in the same row. The coloring scheme indicates the differences in performance (\egc blue for large improvement).}
    \label{fig:heatmap-NORMAL-MBERT}
\end{figure}

We further experiment with regular training of the model using an auxiliary language, instead of performing meta-learning (step~\ref{step2} in~\secref{sec:mode_xmaml}), followed by zero-shot learning on the target languages.
In other words, we apply all steps of X-MAML as explained in~\secref{sec:mode_xmaml}, however instead of step~\ref{step2}, we perform regular supervised learning using the development set of the auxiliary language. We evaluate the final model on the test sets of the target languages. From this experiment, we observe that meta-learning has a strongly positive effect on predictive performance (see ~\Figref{fig:heatmap-NORMAL-MBERT}). Comparing the results in~\Figref{fig:heat-map-mBERT-zero} and~\Figref{fig:heatmap-NORMAL-MBERT} shows that we have similar trends of the improvements, however using meta-learning boost performance on all languages in the XNLI dataset up to $3.6\%$, while the largest improvement without meta-learning is $2.5\%$.
\begin{figure}[t]
\centering
    \includegraphics[width=1\textwidth]{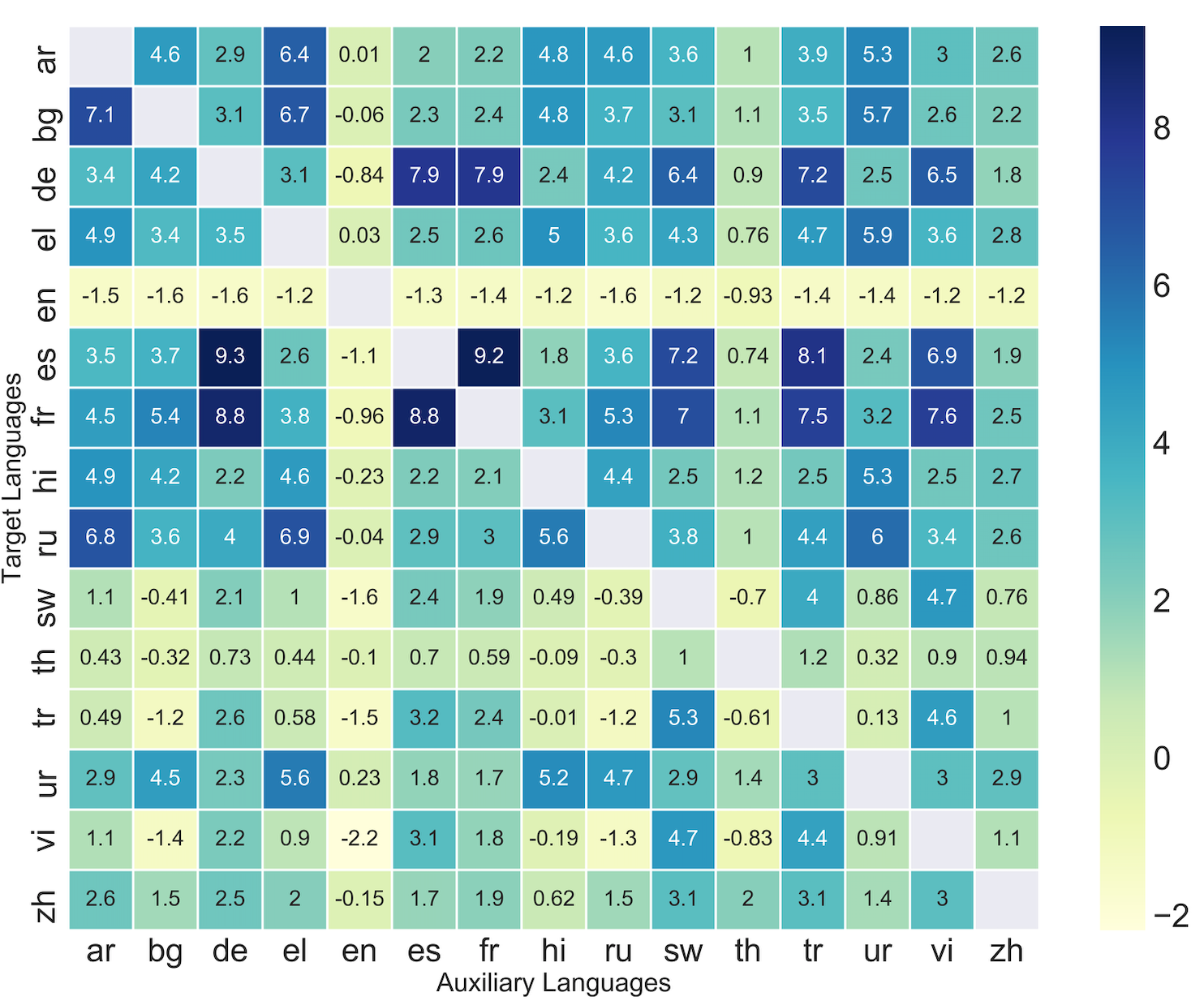}
    \caption[Differences in performance in terms of accuracy scores on the test set for zero-shot X-MAML on XNLI using the En-BERT (English) model.]{Differences in performance in terms of accuracy scores on the test set for zero-shot X-MAML on XNLI using the En-BERT (English) model. Rows correspond to target and columns to auxiliary languages used in X-MAML. Numbers on the off-diagonal indicate performance differences  between X-MAML and the baseline model in the same row. The coloring scheme indicates the differences in performance (\egc blue for large improvement).}
    \label{fig:heatmap-MAML-EBERT}
\end{figure}

\paragraph{Zero-shot X-MAML with En-BERT}
Similar to the previous section, as the first training step (\iec pre-training on a high-resource language, see Step~\ref{step1} in~\secref{sec:mode_xmaml} for more information) in X-MAML for XNLI, we fine-tune En-BERT on the MultiNLI dataset (English) to obtain the initial model parameters $\theta$ for each experiment.
Then, we apply the second and third steps of X-MAML in the zero-shot scenario. Figure  \ref{fig:heatmap-MAML-EBERT} and Table \ref{tab:xnli-eBERT-zeroshot} depict the results of this experiment.

We observe an improvement in accuracy by performing X-MAML on cross-lingual NLI using En-BERT (see Figure \ref{fig:heatmap-MAML-EBERT}). We further note that English as an auxiliary shows negative impact (\iec decreasing performance) in most of the cases. In the reverse setting, using any other language as an auxiliary does not lead to improvement on the English test dataset. The experiments show that the target languages such as Spanish (es), French (fr) and German (de) obtain the largest gains (i.e., improvements up to $+9.3\%$ points in terms of average accuracy), while languages such as Thai (th), Swahili (sw) and Vietnamese (vi) get the lowest gains in X-MAML on the cross-lingual NLI using En-BERT. This can possibly be attributed to the fact that the performance of En-BERT depends directly on word piece overlap, as denoted by \cite{Pires_2019}. For the exact accuracy scores, we refer to Table \ref{tab:xnli-eBERT-zeroshot}.

\begin{table}[t]
\centering
\resizebox{1\linewidth}{!}{
\begin{tabular}{lccccccccccccccc|r}
\toprule
 &\multicolumn{15}{c|}{\textbf{Auxiliary lang.}} & \textbf{baseline}\\
{} &     \textbf{ar} &     \textbf{bg} &     \textbf{de} &     \textbf{el} &     \textbf{en} &     \textbf{es} &     \textbf{fr} &     \textbf{hi} &     \textbf{ru} &     \textbf{sw} &     \textbf{th} &     \textbf{tr} &     \textbf{ur} &     \textbf{vi} &     \textbf{zh}  & \\
\midrule
\textbf{ar} &   - &  39.09 &  37.32 &  40.90 &  34.48 &  36.49 &  36.65 &  39.24 &  39.10 &  38.09 &  35.48 &  38.36 &  39.79 &  37.46 &  37.03 & 34.47\\
\textbf{bg} &  42.33 &   - &  38.29 &  41.92 &  35.17 &  37.55 &  37.58 &  40.04 &  38.93 &  38.32 &  36.37 &  38.72 &  40.90 &  37.81 &  37.41 & 35.23\\
\textbf{de} &  41.88 &  42.77 &   - &  41.59 &  37.68 &  46.41 &  46.43 &  40.90 &  42.70 &  44.89 &  39.42 &  45.70 &  41.05 &  45.03 &  40.30 & 38.52\\
\textbf{el} &  40.08 &  38.50 &  38.70 &   - &  35.18 &  37.65 &  37.80 &  40.15 &  38.72 &  39.42 &  35.91 &  39.82 &  41.06 &  38.73 &  37.99 & 35.15\\
\textbf{en} &  81.95 &  81.87 &  81.89 &  82.22 &   - &  82.12 &  82.05 &  82.23 &  81.88 &  82.23 &  82.52 &  82.01 &  82.03 &  82.29 &  82.27 & 83.45\\
\textbf{es} &  47.41 &  47.64 &  53.24 &  46.59 &  42.86 &   - &  53.18 &  45.79 &  47.56 &  51.10 &  44.69 &  52.04 &  46.30 &  50.83 &  45.87 & 43.95\\
\textbf{fr} &  45.55 &  46.40 &  49.81 &  44.81 &  40.08 &  49.89 &   - &  44.14 &  46.30 &  48.05 &  42.13 &  48.54 &  44.24 &  48.67 &  43.58 & 41.04\\
\textbf{hi} &  39.61 &  38.91 &  36.91 &  39.32 &  34.46 &  36.87 &  36.78 &   - &  39.08 &  37.14 &  35.88 &  37.15 &  39.98 &  37.20 &  37.40 & 34.69\\
\textbf{ru} &  41.87 &  38.73 &  39.10 &  41.98 &  35.05 &  38.02 &  38.13 &  40.73 &   - &  38.89 &  36.11 &  39.51 &  41.12 &  38.51 &  37.69 & 35.09\\
\textbf{sw} &  39.05 &  37.55 &  40.07 &  39.00 &  36.41 &  40.33 &  39.85 &  38.45 &  37.57 &   - &  37.26 &  42.01 &  38.82 &  42.70 &  38.72 & 37.96\\
\textbf{th} &  36.16 &  35.41 &  36.46 &  36.17 &  35.63 &  36.43 &  36.32 &  35.64 &  35.43 &  36.74 &   - &  36.91 &  36.05 &  36.63 &  36.67 & 35.73\\
\textbf{tr} &  39.33 &  37.62 &  41.44 &  39.42 &  37.34 &  42.07 &  41.26 &  38.83 &  37.63 &  44.12 &  38.23 &   - &  38.97 &  43.42 &  39.86 & 38.84\\
\textbf{ur} &  36.85 &  38.46 &  36.27 &  39.55 &  34.16 &  35.72 &  35.63 &  39.09 &  38.64 &  36.80 &  35.33 &  36.94 &   - &  36.91 &  36.85 & 33.93 \\
\textbf{vi} &  41.85 &  39.35 &  42.97 &  41.62 &  38.53 &  43.85 &  42.52 &  40.53 &  39.38 &  45.46 &  39.89 &  45.11 &  41.63 &   - &  41.84 & 40.72\\
\textbf{zh} &  37.21 &  36.09 &  37.18 &  36.68 &  34.48 &  36.33 &  36.55 &  35.25 &  36.16 &  37.73 &  36.64 &  37.70 &  35.99 &  37.66 &   - & 34.63 \\
\bottomrule
\end{tabular}
}
\caption[The performance in terms of average test accuracy for the zero-shot setting over 10 runs of X-MAML on the XNLI dataset using En-BERT.]{The performance in terms of average test accuracy for the zero-shot setting over 10 runs of X-MAML on the XNLI dataset using En-BERT (monolingual), as base model. Each column corresponds to the performance of the En-BERT system after meta-learning with a single auxiliary language, and evaluation on the target language of the XNLI test set. The auxiliary language is not included during the evaluation phase. Results of the En-BERT model without X-MAML (baseline) are also reported.}
\label{tab:xnli-eBERT-zeroshot}
\end{table}

\subsubsection{Few-Shot Learning}
\begin{figure}[t]
\includegraphics[width=1\textwidth]{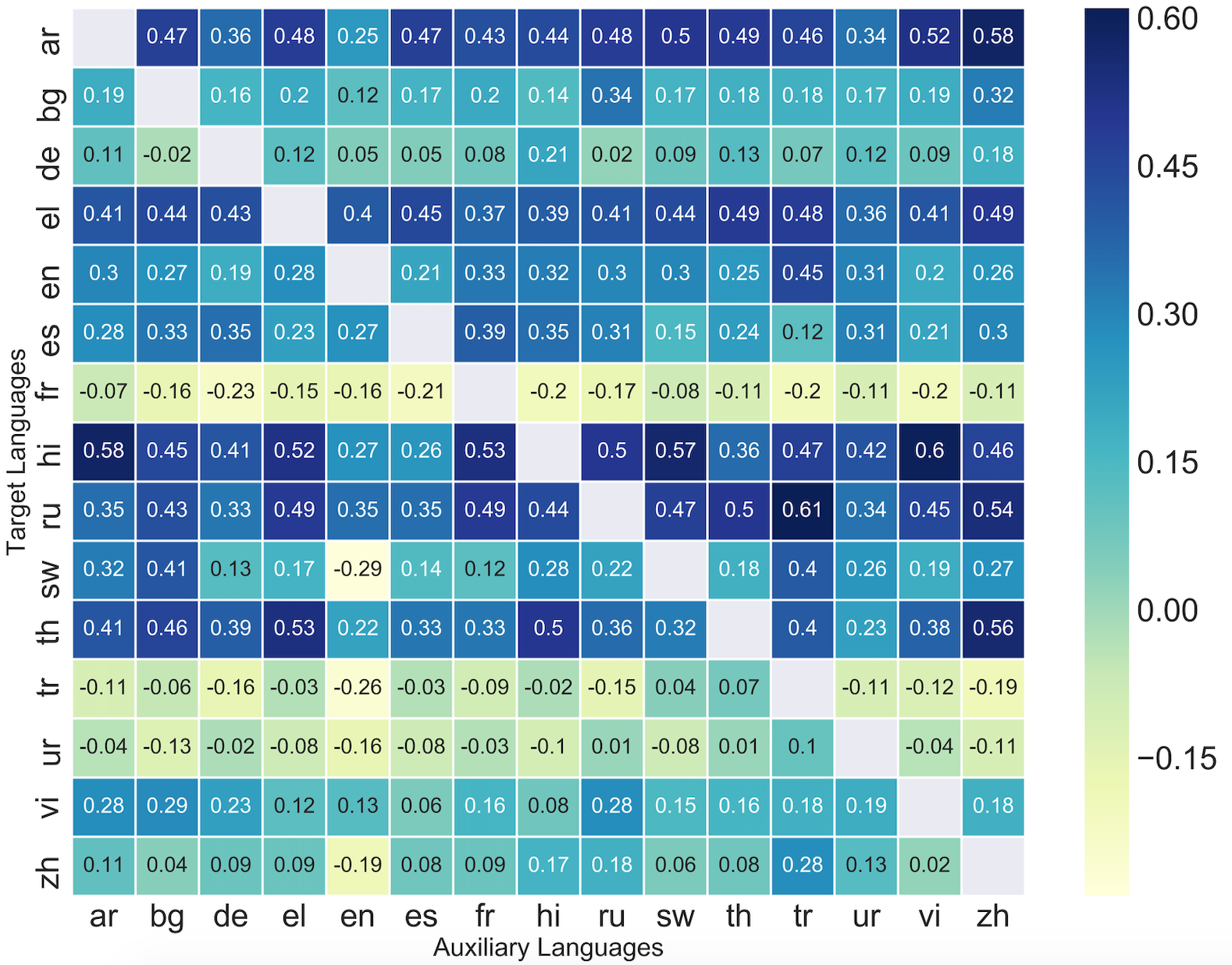}
    \caption[Differences in performance in terms of accuracy scores on the test set for few-shot X-MAML on XNLI using the Multi-BERT model]{Differences in performance in terms of accuracy scores on the test set for few-shot X-MAML on XNLI using the Multi-BERT model. Rows correspond to target and columns to auxiliary languages used in X-MAML. Numbers on the off-diagonal indicate performance differences  between X-MAML and the baseline model in the same row. The coloring scheme indicates the differences in performance (\egc blue for large improvement).}
    \label{fig:ffew-shot-heatmap-MAML-MBERT}
\end{figure}

For few-shot learning, following the steps in X-MAML, we perform fine-tuning on the development set (2.5k instances) of the target languages, and then evaluate on the test set (Step~\ref{step3} in~\secref{sec:mode_xmaml}).
We employ Multi-BERT as the underlying model $\texttt{M}$ in this scenario.
Detailed ablation results are presented in  \tabref{tab:xnli-mBERT-fewshot} and \Figref{fig:ffew-shot-heatmap-MAML-MBERT}.

Overall, these results demonstrate that we have a positive impact on most of the low-resource target languages. However, the improvements in the few-shot X-MAML are lower compared to the zero-shot setting (i.e., improvements up to $+0.61\%$ points in terms of average accuracy for few-shot X-MAML on XNLI using the Multi-BERT model). Target languages such as Hindi (hi), Russian (ru), Thai (th), Arabic (ar) and Greek (el) benefit from X-MAML with Multi-BERT. At the same time, the few-shot X-MAML with Multi-BERT provides negative impacts for French (fr), Turkish (tr) and Urdu (ur) as target languages.
\begin{table}[t]
\resizebox{1\linewidth}{!}{
\begin{tabular}{lccccccccccccccc|r}
\toprule
 &\multicolumn{15}{c|}{\textbf{Auxiliary language}} & \textbf{baseline}\\
{} &     \textbf{ar} &     \textbf{bg} &     \textbf{de} &     \textbf{el} &     \textbf{en} &     \textbf{es} &     \textbf{fr} &     \textbf{hi} &     \textbf{ru} &     \textbf{sw} &     \textbf{th} &     \textbf{tr} &     \textbf{ur} &     \textbf{vi} &     \textbf{zh}  & \\
\midrule
\textbf{ar} &   - &  67.84 &  67.73 &  67.85 &  67.62 &  67.84 &  67.80 &  67.81 &  67.85 &  67.87 &  67.86 &  67.83 &  67.71 &  67.89 &  67.95 & 67.37 \\
\textbf{bg} &  71.79 &   - &  71.76 &  71.80 &  71.72 &  71.77 &  71.80 &  71.74 &  71.94 &  71.77 &  71.78 &  71.78 &  71.77 &  71.79 &  71.92 & 71.60\\
\textbf{de} &  73.36 &  73.23 &   - &  73.37 &  73.30 &  73.30 &  73.33 &  73.46 &  73.27 &  73.34 &  73.38 &  73.32 &  73.37 &  73.34 &  73.43 & 73.25 \\
\textbf{el} &  69.95 &  69.98 &  69.97 &   - &  69.94 &  69.99 &  69.91 &  69.93 &  69.95 &  69.98 &  70.03 &  70.02 &  69.90 &  69.95 &  70.03 & 69.54 \\
\textbf{en} &  82.24 &  82.21 &  82.13 &  82.22 &   - &  82.15 &  82.27 &  82.26 &  82.24 &  82.24 &  82.19 &  82.39 &  82.25 &  82.14 &  82.20& 81.94 \\
\textbf{es} &  76.07 &  76.12 &  76.14 &  76.02 &  76.06 &   - &  76.18 &  76.14 &  76.10 &  75.94 &  76.03 &  75.91 &  76.10 &  76.00 &  76.09 &75.79\\
\textbf{fr} &  75.32 &  75.23 &  75.16 &  75.24 &  75.23 &  75.18 &   - &  75.19 &  75.22 &  75.31 &  75.28 &  75.19 &  75.28 &  75.19 &  75.28 & 75.39\\
\textbf{hi} &  64.95 &  64.82 &  64.78 &  64.89 &  64.64 &  64.63 &  64.90 &   - &  64.87 &  64.94 &  64.73 &  64.84 &  64.79 &  64.97 &  64.83 & 64.37\\
\textbf{ru} &  71.19 &  71.27 &  71.17 &  71.33 &  71.19 &  71.19 &  71.33 &  71.28 &   - &  71.31 &  71.34 &  71.45 &  71.18 &  71.29 &  71.38& 70.84\\
\textbf{sw} &  58.14 &  58.23 &  57.95 &  57.99 &  57.53 &  57.97 &  57.94 &  58.10 &  58.04 &   - &  58.00 &  58.22 &  58.08 &  58.01 &  58.09 &57.82\\
\textbf{th} &  61.59 &  61.64 &  61.57 &  61.71 &  61.40 &  61.51 &  61.51 &  61.68 &  61.54 &  61.50 &   - &  61.58 &  61.41 &  61.56 &  61.74 & 61.18\\
\textbf{tr} &  64.74 &  64.79 &  64.69 &  64.82 &  64.59 &  64.82 &  64.76 &  64.83 &  64.70 &  64.89 &  64.92 &   - &  64.74 &  64.73 &  64.66 &64.85 \\
\textbf{ur} &  63.67 &  63.58 &  63.69 &  63.63 &  63.55 &  63.63 &  63.68 &  63.61 &  63.72 &  63.63 &  63.72 &  63.81 &   - &  63.67 &  63.60 &63.71\\
\textbf{vi} &  73.51 &  73.52 &  73.46 &  73.35 &  73.36 &  73.29 &  73.39 &  73.31 &  73.51 &  73.38 &  73.39 &  73.41 &  73.42 &   - &  73.41 &73.23 \\
\textbf{zh} &  74.04 &  73.97 &  74.02 &  74.02 &  73.74 &  74.01 &  74.02 &  74.10 &  74.11 &  73.99 &  74.01 &  74.21 &  74.06 &  73.95 &   - &73.93 \\
\bottomrule
\end{tabular}
}
\captionof{table}[The performance in terms of average test accuracy for the few-shot setting over 10 runs of X-MAML on the XNLI dataset using Multi-BERT.]{The performance in terms of average test accuracy for the few-shot setting over 10 runs of X-MAML on the XNLI dataset using Multi-BERT (multilingual BERT), as base model. Each column corresponds to the performance of the Multi-BERT system after meta-learning with a single auxiliary language, and evaluation on the target language of the XNLI test set. The auxiliary language is not included during the evaluation phase. Results of the Multi-BERT model without X-MAML (baseline) are also reported.}
\label{tab:xnli-mBERT-fewshot}
\end{table}

In~\tabref{tab:originalBERT}, we compare X-MAML results with one and two auxiliary languages to the external and internal baselines. We detect that using two auxiliary languages in the meta-learning step (Step \ref{step2} in Section ~\secref{sec:mode_xmaml}) leads to similar conclusions as before (i.e., using two auxiliary languages leads the largest benefits in the few-shot X-MAML with Multi-BERT).

In contrast to the zero-shot X-MAML with Multi-BERT, we observe that Swahili (sw) acts as the overall most effective auxiliary language for meta-learning with Multi-BERT in the few-shot learning setting (see results in the few-shot learning section in Table \ref{tab:originalBERT}).

Moving on, in Table \ref{tab:originalBERT} we report results from  \cite{DBLP:journals/corr/abs-1810-04805} that use machine translation at test time (TRANSLATE-TEST) and results from  \cite{DBLP:journals/corr/abs-1904-09077} that use machine translation at training time (TRANSLATE-TRAIN), where they have been shown to be strong baselines in previous work. In TRANSLATE-TRAIN, the English training data is translated to the target language, and the model is fine-tuned using the translated data. While in TRANSLATE-TEST, the target language test data is translated into English ,and then the model is fine-tuned on the translated version.

Note that, using X-MAML, we can alleviate the machine translation step (TRANSLATE-TEST) from the target language into English. The results in Table \ref{tab:originalBERT} also indicate that X-MAML boosts Multi-BERT performance on XNLI. It is worthwhile mentioning that Multi-BERT in the TRANSLATE-TRAIN setup outperforms our few-shot X-MAML. However, we only use 2k development examples from the target languages, whereas work mentioned above, 433k translated sentences are used for fine-tuning.

\subsection{Zero-Shot Cross-Lingual QA} \label{X-MAML:QA}
Here, we attempt to answer our research questions RQ \ref{rq.5.2} and \ref{rq.5.3} in the cross-lingual QA. To understand whether our framework is model- and task- agnostic and can apply to other tasks and models besides NLI and BERT, we conduct additional experiments for the question answering task. We investigate the impact of X-MAML on other pre-trained language models, namely XLM and XLM-RoBERTa (XLM-R) (see Section \ref{xmaml:nlu-models}).
We use these models as the base model \texttt{M} in X-MAML for our QA experiments.
We employ the XLM-15 version of XLM, XLM-R$_{base}$ and XLM-R$_{large}$ versions of XLM-R (see Section \ref{xmaml:nlu-models}).
The SQuAD v1.1 training data (see Section \ref{xmaml-qa-data}) is used in the pre-training step of X-MAML (see Step~\ref{step1} in ~\secref{sec:mode_xmaml}). We use the cross-lingual development and test splits provided in the MLQA dataset (Table \ref{tbl:mlqa-dataset}) for meta-learning and evaluation steps, respectively.
\label{ch5:qa}
\begin{table}[t]
\centering
\resizebox{1\textwidth}{!}{
\begin{tabular}{c|c|l|ccccccc|c}
\toprule
\multicolumn{3}{c|}{Model} & en    & ar                   & de      & es                   & hi                   & vi                   & zh                   & avg                  \\
\toprule
\parbox[t]{3mm}{\multirow{6}{*}{\rotatebox{90}{XLM}}} & &
Our baseline & \textbf{69.80 }  &	48.95   & 52.64    &	\textbf{58.15 } &	46.67   & 48.46   & 42.64   & 52.47  \\

\cmidrule(r){2-11}
& \parbox[t]{3mm}{\multirow{4}{*}{\rotatebox{90}{\textbf{X-MAML}}}} &{{\it (One aux. lang.)}}  &  69.39      &	 48.45      &  53.04      &	57.68     &	 46.90      &  49.79     &  44.36     	& \multirow{2}{*}{52.80}  \\

& & $l\rightarrow X$ &  \textit{ar}  &	    \textit{hi}  & \textit{es}  & \textit{en} &	  \textit{zh}  &     \textit{zh} &     \textit{hi} 	&   \\
\cmidrule(r){3-11}
& &{{\it (Two aux. lang.)}}  & 68.88   &\textbf{49.76}   & \textbf{53.18}  &	58.00  &	\textbf{48.43}   & \textbf{50.86} & \textbf{45.44} 	&  \multirow{2}{*}{\textbf{53.51}} \\

& & $(l_1,l_2)\rightarrow X$ &  \textit{(es,ar)}  &  \textit{(vi,zh)}  &   \textit{(vi,zh)} &	  \textit{(en,zh)} &	  \textit{(vi,zh)}  &   \textit{(hi,zh)} & \ \textit{(es,hi)}	& \\

\midrule
\midrule
\parbox[t]{3mm}{\multirow{6}{*}{\rotatebox{90}{XLM-R$_{base}$}}} &&
\cite{Liang2020XGLUEAN} & {80.1  } & {56.4 } & {62.1 } & {67.9  } & {60.5 } & {67.1 } & {61.4 } & { 65.1 } \\
& & Our baseline & \textbf{80.38}  &	57.23  & 63.08 &	{67.91 } &	61.46   & 67.14  & 62.73  	& 65.70  \\
\cmidrule(r){2-11}
& \parbox[t]{3mm}{\multirow{4}{*}{\rotatebox{90}{\textbf{X-MAML}}}}& {\it (One aux. lang.)}  & {80.19 }  & {57.97 } & { 63.57 } & {67.46 } & {61.70 } & {67.97 }&  {64.01} & \multirow{2}{*}{66.12} \\

 && $l\rightarrow X$ & \textit{vi} &  \textit{hi}& \textit{ar}&  \textit{vi}&  \textit{vi}&  \textit{hi}&   \textit{hi}& \\
\cmidrule(r){3-11}
& &{\it (Two aux. lang.)} & 80.31  & \textbf{58.14}  & \textbf{64.07 }  &  \textbf{68.08 } & \textbf{62.67 }  & \textbf{68.82 }  & \textbf{64.06 } & \multirow{2}{*}{\textbf{66.59} } \\

&& $(l_1,l_2)\rightarrow X$ &  \textit{(ar,vi)} &   \textit{(hi,vi)}&   \textit{(ar,hi)}&    \textit{(ar,hi)}&   \textit{(es,ar)}&   \textit{(ar,hi) }&   \textit{(ar,hi)}&   \\
\midrule
\midrule
\parbox[t]{3mm}{\multirow{6}{*}{\rotatebox{90}{XLM-R$_{large}$}}}&&
\cite{hu2020xtreme}& {83.5 } & {66.6 } & {70.1 } & {74.1 } & {70.6 } & {74}   & {62.1} & {71.6 } \\

&& Our baseline & {83.95 }  &	66.09  & 70.62 &	{74.59 }  &	70.64  & 74.13  & 69.80	& 72.83 \\
\cmidrule(r){2-11}
&\parbox[t]{3mm}{\multirow{4}{*}{\rotatebox{90}{\textbf{X-MAML}}}}& {\it (One aux. lang.)} & 84.31   & {66.61}    & {70.84 }   & 74.32   & \textbf{70.94 }  & \textbf{74.84}   & \textbf{70.74} & \multirow{2}{*}{73.23} \\
 && $l\rightarrow X$ &  \textit{ar} &    \textit{hi} &   \textit{ar} &   \textit{hi} &  \textit{vi}&   \textit{ar} &  \textit{hi}&    \\
\cmidrule(r){3-11}
&& {\it (Two aux. lang.)} & \textbf{84.60}  & \textbf{66.95}  & \textbf{71.00} & \textbf{74.62}  & 70.93  & 74.73   & 70.29    & \multirow{2}{*}{\textbf{74.30}} \\

& &$(l_1,l_2)\rightarrow X$  &   \textit{(hi,vi)}&  \textit{(hi,vi)}&   \textit{(ar,vi)}&   \textit{(en,vi)}&   \textit{(ar,vi)}&   \textit{(es,hi)}&  \textit{(en,vi)} &  \\
\bottomrule
\end{tabular}}
\caption[F1 scores (average over 10 runs) on the MLQA test set using zero-shot X-MAML.]{F1 scores (average over 10 runs) on the MLQA test set using zero-shot X-MAML. Columns indicate the target languages. The avg column indicates row-wise  average F1 score. We also report the most beneficial auxiliary language/s for X-MAML in improving the test F1 of each target language.}
\label{tab:mlqa_results}
\end{table}
We use a similar approach for cross-lingual QA on the MLQA dataset.

Table \ref{tab:mlqa_results} shows the results of zero-shot X-MAML for the MLQA dataset. We compare our results on the MLQA dataset to those reported in two benchmark papers, \cite{hu2020xtreme} and \cite{Liang2020XGLUEAN}. We also report our own baseline for the task. The baselines are provided by training each base model on the SQuAD v1.1 train set (see Step~\ref{step1} in ~\secref{sec:mode_xmaml}) and evaluating on the test set of MLQA.
In Table \ref{tab:mlqa_results}, we consider only the best auxiliary languages for each target language, and then compute the average F$_1$ score.

We observe that all of the target languages benefit from at least one of the auxiliary languages by adapting the models using X-MAML, highlighting the benefits of this method. We find that:
\begin{enumerate*}[label=(\roman*)]
\item our zero-shot results with X-MAML improve on those without meta-learning (\iec baselines);
\item performing X-MAML with two auxiliary languages provides the largest gains for the models in cross-lingual QA.
\end{enumerate*}
Overall, zero-shot learning models with X-MAML outperform both internal and external baselines.
The improvement is $+1.04\%$, $+0.89\%$ and $+1.47\%$ in average F$_1$ score compared to XLM-15, XLM-R$_{base}$ and XLM-R$_{large}$, respectively.

\section{Discussion and Analysis} \label{ch5:error}
Somewhat surprisingly, we find that cross-lingual transfer with meta-learning yields improved results even when languages strongly differ (i.e., in terms of language family) from one another.
For instance, for zero-shot meta-learning on XNLI, we observe gains for almost all auxiliary languages, except for Swahili (sw).
This indicates that the meta-parameters learned with X-MAML are sufficiently language agnostic, as we otherwise would not expect to see any benefits in transferring from, \egc Russian (ru) to Hindi (hi) (one of the strongest results in~\Figref{fig:heat-map-mBERT-zero}).
This is dependent on having access to a pre-trained multilingual model such as BERT; however, using monolingual BERT (En-BERT) yields overwhelmingly positive gains in some target/auxiliary settings  (see results in~\Figref{fig:heatmap-MAML-EBERT}).
For few-shot learning when we only have access to a handful of training instances, our findings are similar, as almost all combinations of auxiliary and target languages lead to improvements when using Multi-BERT (\Figref{fig:ffew-shot-heatmap-MAML-MBERT}). Therefore, we try to shed light on the behavior of our proposed model and answer our last research question (i.e., RQ \ref{rq.5.4}) in the following section.

\subsection{Typological Correlations}
In order to better explain our results for cross-lingual zero-shot and few-shot learning, we investigate typological features, and their overlap between target and auxiliary languages.
We evaluate on the World Atlas of Language Structure (WALS, \cite{wals}), which is the largest openly available typological database.
It comprises approximately 200 linguistic features with annotations for more than 2500 languages, which have been made by expert typologists through the study of grammars and field work.
We draw inspiration from previous works \parencite{bjerva-augenstein-2018-phonology,bjerva_augenstein:18:iwclul} that attempt to predict typological features based on language representations learned under various NLP tasks.
Similarly, we experiment with the two following conditions:
\begin{enumerate}[label=(\roman*)]
\item We attempt to predict typological features based on the mutual gain/loss in performance using X-MAML.
\item We investigate whether sharing between two typologically similar languages is beneficial for performance using X-MAML.
\end{enumerate}
We train one simple logistic regression classifier per condition above, for each WALS feature.
In the first condition (i), the task is to predict the exact WALS feature value of a language, given the change in accuracy in combination with other languages.
In the second condition (ii), the task is to predict whether a main and auxiliary language have the same WALS feature value, given the change in accuracy when the two languages are used in X-MAML.
We compare with two simple baselines, one based on always predicting the most frequent feature value in the training set, and one based on predicting feature values with respect to the distribution of feature values in the training set.
\begin{figure}[t]
    \centering
    \includegraphics[width=1\textwidth]{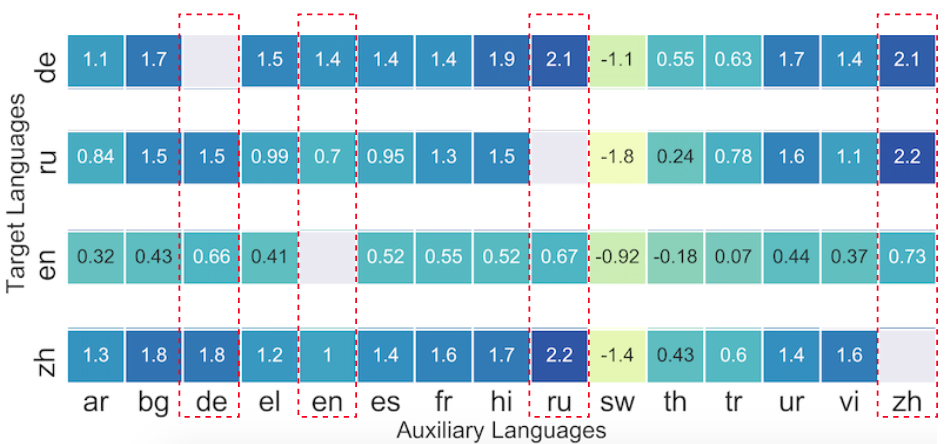}
    \caption[The mutual gains among English (en), German (de), Russian (ru), and Chinese (zh) languages in zero-shot X-MAML with Multi-BERT.]{The mutual gains among English (en), German (de), Russian (ru), and Chinese (zh) languages in zero-shot X-MAML with Multi-BERT. Rows correspond to target and columns to auxiliary languages used in X-MAML. Numbers indicate performance differences  between X-MAML and the baseline model in the same row. The coloring scheme indicates the differences in performance (\egc blue for large improvement). }
    \label{fig:mutual-typology}
 \end{figure}
We then investigate whether any features could be consistently predicted above baseline levels, given different test-training splits.
We apply a simple paired t-test to compare our model predictions to the baselines.
As we are running a large number of tests (one per WALS feature), we apply Bonferroni correction, changing our cut-off $p$-value from $p=0.05$ to $p=0.00025$.

We first investigate a few-shot X-MAML, when using Multi-BERT, as reported in ~\tabref{tab:xnli-mBERT-fewshot}.
We find that languages sharing the feature value for WALS feature \textit{67A The Future Tense} are beneficial to each other.
This feature encodes whether or not a language has an inflectional marking of the future tense, and can be considered to be a morphosyntactic feature.
We next look at zero-shot X-MAML with Multi-BERT, as reported in ~\tabref{tab:xnli-mBERT-zeroshot}.
For this case, we find that languages sharing a feature value for the WALS feature \textit{25A Locus of Marking: Whole-language Typology} typically help each other. This feature describes whether the morphosyntactic marking in a language is on the syntactic heads or dependents of a phrase. For example English (en),
German (de), Russian (ru), and Chinese (zh) are `dependent-marking' in this feature.  Moreover, if we look at the results in~\Figref{fig:heat-map-mBERT-zero}, they have the largest mutual gains from each other during the zero-shot X-MAML, as shown in ~\Figref{fig:mutual-typology}.
In both cases, we thus find that languages with similar morphosyntactic properties can be beneficial to one another when using X-MAML.

\section{Summary}
In this chapter, we show that meta-learning allows us to leverage training data from auxiliary languages and genres, to perform the zero-shot and few-shot cross-lingual and cross-genre transfer. We achieve competitive performance compared to a machine translation baseline (for XNLI), and propose a method that requires no training instances for the target task in the target language. Experiments with different models show that our method is model agnostic, and can be used to extend any pre-existing model.
We evaluated this on two challenging NLU tasks (NLI and QA), and on a total of 15 languages.
We can improve the performance of strong baseline models for (i) zero-shot XNLI, and (ii) zero-shot QA on the MLQA dataset.
Furthermore, we show in a typological analysis that languages which share certain morphosyntactic features tend to benefit from this type of transfer.

To summarize, the contribution of this chapter (detailed in ~\secref{sec:mode_xmaml}) is four-fold.
Concretely, we:
\begin{enumerate}[label=(\roman*)]
    \item exploit the use of meta-learning methods for two different natural language understanding tasks (\iec NLI, QA);
    \item evaluate the performance of the proposed architecture on various scenarios: cross-genre, cross-lingual, standard supervised, and zero-shot, across a total of 15 languages (\iec 15 languages in XNLI and 7 languages in MLQA);
    \item observe consistent improvements of our cross-lingual meta-learning architecture (X-MAML) over the previous models on various cross-lingual benchmarks (\iec improving the Multilingual BERT model by $+3.65\%$ and $+1.04\%$ points in terms of average accuracy on zero-shot and few-shot XNLI, respectively, and boosting the XLM-R$_{large}$ by $+1.47\%$ in terms of average F$_1$ score on zero-shot QA);
    \item perform an error analysis, which reveals that typological commonalities between languages can partially explain the cross-lingual trends.
\end{enumerate}

    \chapter{Conclusion and Future work}
\label{sec:sixth}
This thesis investigates methods for dealing with low-resource scenarios in information extraction and natural language understanding tasks.
To this end, we study distant supervision and sequential transfer learning in various low-resource settings.
We develop and analyze models to explore three essential questions concerning NLP tasks with minimal or no training data which cut across several of the chapters in this thesis (see Table \ref{tbl:RQs} for an overview that maps the general research questions to individual chapters and sub-questions):
\begin{Question}
    What is the impact of different input representations in neural low-resource NLP?
\end{Question}
\begin{Question}
    How can we incorporate domain knowledge in low-resource NLP?
\end{Question}   
\begin{Question}
    How can we address challenges of low-resource scenarios using transfer learning techniques?
\end{Question} 
\noindent During the course of this thesis we have made contributions in low-resource NLP in four different areas: domain-specific embeddings (Chapter \ref{sec:second}), named entity recognition (Chapter \ref{sec:third}), relation extraction and classification (Chapter \ref{sec:fourth}), and cross-genre and cross-lingual natural language understanding (Chapter \ref{sec:fifth}).
In the following, we describe our proposed methods and findings (Section \ref{conclud:methods}). Our main contributions are summarized in Section \ref{conclud:contributions}, and we provide an outlook into future directions in Section \ref{conclud:future}.  

\begin{table}[t]
	\centering
	\begin{tabular}{l|l|c}
		\toprule
	\head{Question} & \head{Chapter}& \head{Sub-Question} \\
	\midrule
	\multirow{2}{*}{\parbox{5cm}{RQ \ref{RQ.1}: \emph{\small What is the impact of different input representations in neural low-resource NLP?}}}
	& Chapter \ref{sec:second}&RQ \ref{rq.2.1} \\
	\cmidrule(ll){2-3}
	&\multirow{3}{*}{Chapter \ref{sec:fourth}}&RQ \ref{rq.4.1}\\
	&&RQ \ref{rq.4.2}\\
	&&RQ \ref{rq.4.3}\\
	\midrule
\multirow{2}{*}{\parbox{5cm}{RQ \ref{RQ.2}: \emph{\small How can we incorporate domain knowledge in low-resource NLP?}}}
 &Chapter \ref{sec:second} &RQ \ref{rq.2.2} \\
\cmidrule(ll){2-3}
&\multirow{2}{*}{Chapter \ref{sec:third}}&RQ \ref{rq.3.1}\\
	&&RQ \ref{rq.3.2}\\
	\midrule
\multirow{2}{*}{\parbox{5cm}{RQ \ref{RQ.3}: \emph{\small How can we address challenges of low-resource scenarios using transfer learning techniques?}}}
&Chapter \ref{sec:second} &RQ \ref{rq.2.1}\\
	\cmidrule(ll){2-3}
	&\multirow{2}{*}{Chapter \ref{sec:fourth}}&RQ \ref{rq.4.1}\\
	&&RQ \ref{rq.4.2}\\
	\cmidrule(ll){2-3}
&\multirow{2}{*}{Chapter \ref{sec:fifth}}&RQ \ref{rq.5.1}\\
&&RQ \ref{rq.5.2}\\
\bottomrule
	\end{tabular}
	\caption{Overview of the research questions and related chapters.}
	\label{tbl:RQs}
\end{table}
\section{Proposed Methods and Findings}\label{conclud:methods}
Previous research shows that automatically learning transferable representations in terms of word embeddings boosts NLP models' performance in various down-stream tasks.
The following research question addresses a central challenge that needs to be answered for embeddings in a technical domain:
\begin{itemize}
    \item \textbf{RQ \ref{rq.2.1}.} \textit{Can word embedding models capture domain-specific semantic relations even when trained with a considerably smaller corpus size?}
\end{itemize}
To answer this question, we here focus on a new and relatively unexplored technical domain: the oil and gas domain, train domain-specific embeddings on this technical low-resource domain. We further construct a domain-specific evaluation dataset, including a corpus and a query inventory for the oil and gas domain (Section \ref{slb}). We evaluate, in Sections \ref{exp1} and \ref{ch2:extrinsic}, the effectiveness of domain-specific models using intrinsic and extrinsic evaluations.
In Section \ref{exp1}, empirical intrinsic evaluations reveal that domain-specific trained embeddings perform better than general domain embeddings trained on much larger input data. Furthermore, in Section \ref{Serror}, the in-depth manual analysis shows the ability of the domain model to discover semantic relations such as (co)hyponymy, hypernymy, and relatedness, giving insight into these models beyond the intrinsic evaluation dataset.

In our target domain, in addition to text, there exists a domain-specific knowledge resource (i.e., Schlumberger oilfield glossary) created by domain experts to facilitate information processing. We here pose the following research question: 
\begin{itemize}
    \item \textbf{RQ \ref{rq.2.2}.} \textit{How can we take advantage of existing domain-specific knowledge resources to enhance the resulting models?}
\end{itemize}
We enhance, in Section \ref{retrofit}, the domain embeddings by incorporating domain knowledge from the oilfield glossary and constructing embedding representations for infrequent technical terms. We find that the domain embeddings and their enhanced versions can be useful resources to support a downstream domain-specific NLP task (Section \ref{ch2:extrinsic-exps}).
The results on a multi-label domain-specific sentence classification task show that the enhanced domain embeddings provide higher performance and aid the model in label assignment.

NER is a central task in NLP and one that often requires domain-specific, annotated data. In Chapter \ref{sec:third}, we focus on the named entity recognition task in several low-resource domains. We here pose the following research questions:
\begin{itemize} 
    \item \textbf{RQ \ref{rq.3.1}.} \textit{How can we address the problem of low-resource NER using distantly supervised data?} 
    \item \textbf{RQ \ref{rq.3.2}.} \textit{How can we exploit a reinforcement learning approach to improve NER in low-resource scenarios?}
\end{itemize}
We introduce a framework to address the common challenges of distantly supervised datasets for low-resource NER. The main concerns in distantly supervised NER are false positive and false negative instances. Our framework combines a neural NER model with a partial-CRF layer and a policy-based reinforcement learning component (Section \ref{NER+PA+RL}). The partial-CRF component (PA in Section \ref{NER+PA}) is designed to deal with the false negatives, while the reinforcement-based module (RL in Section \ref{NER+RL}) handles the false positives instances.
We quantify, in Section \ref{ch3:comparision}, the impact of each component in our proposed framework. We further in Section \ref{ch3:size-data}, investigate the performance of our model under settings using different sizes of human-annotated data. 
The ablation studies determine the efficiency of the partial-CRF and policy reinforcement modules in fixing the problems in the distantly annotated NER datasets.
Overall, our final system, a combination of NER, PA, and RL, achieves an improvement of +2.75
and +11.85 F1 on the BC5CDR and LaptopReview respectively over the baseline system.
Furthermore, we observe that our model can deliver relatively good performance with a small set of gold data. Our final method achieves a performance of 83.18 and 63.50 with only 2\% of the annotated dataset in the BC5CDR and LaptopReview domains, respectively. In contrast, the base NER model requires almost 45\% of the ground truth sentences to reach the same performance.

In Section \ref{ch3:comparision}, we aim to answer:
\begin{itemize}
\item \textbf{RQ \ref{rq.3.3}.} \textit{Is the proposed solution beneficial for different low-resource scenarios?}
\end{itemize}
We evaluate our model across four diverse datasets from different domains (i.e., biomedical, e-commerce, technical reviews, and news) and languages (English and Chinese). Experimental results show that our approach can boost the performance of the neural NER system in resource-poor settings and achieve higher F1 scores on the different datasets compared to previous work \footnote{At the time of publishing the results were state-of-the-art.}.

Another central IE task is relation extraction. In Chapter \ref{sec:fourth}, we introduce an adapted neural framework incorporating domain-specific embeddings and syntactic structure to address low-resource relation extraction tasks in the SemEval 2018 task 7. 
Our framework is based on a CNN architecture over the shortest dependency paths between entity pairs for relation extraction and classification in scientific text (Section \ref{ch4:model}). 
The framework leverages knowledge from both domain-specific embeddings and syntactic representations to help the low-resource relation extraction task. It ranks third in all three sub-tasks of the SemEval 2018 task.
With this, we attempt to answer the following questions: 
\begin{itemize}
    \item \textbf{RQ \ref{rq.4.1}.} \textit{Are domain-specific input representations beneficial for relation extraction task?} 
    \item \textbf{RQ \ref{rq.4.2}.} \textit{What is the impact of syntactic dependency representations in low-resource neural relation extraction?}
\end{itemize}
We first, in Section \ref{modelvariant}, investigate the utility of domain-specific word embeddings to our neural relation extraction model. The sensitivity analysis study confirms, what we already found in Chapter \ref{sec:second}, the positive impact of domain-specific embeddings by providing higher performance gains when used in our model.
By inspecting the performance of the model with and without the dependency paths in Section \ref{ch4:syntactic-effect}, we affirm the influence of syntactic structure compared to a syntax-agnostic approach in this setting. We find that the effect of syntactic structure varies between different relation types. However, the syntactic representation has a clear positive impact on all the relation types, ranging from improvements of 20 to 45 percentage points depending on the specific relation.
We further ask the following question:
\begin{itemize}
\item \textbf{RQ \ref{rq.4.3}} \textit{Which kind of syntactic dependency representation is most beneficial for neural relation extraction and classification?}
\end{itemize}
Thus, in Section \ref{ch4:different-reps}, we examine the influence of incorporating various dependency representations in our neural model.  We contrast the use of three input representations for our relation extraction model employing the widely used CoNLL, Stanford Basic (SB), and Universal Dependencies (UD) schemes. We compare the effectiveness of specific inputs to our neural relation extraction model by inspecting the effect of various syntactic representations. Furthermore, we observe that the widely used Universal Dependencies scheme consistently provides somewhat lower results in both relation classification and extraction tasks.
We, therefore, opted for manual inspection of a set of incorrect predictions provided by the Universal Dependencies-based model, which are correctly predicted by the two other systems (CoNLL and Stanford Basic -based models). Overall, our results and analysis show that the particular choice of syntactic representation has clear consequences in downstream processing. We observe that the UD paths are generally shorter, and the entities often reside
within a prepositional phrase. Whereas the SB and CoNLL paths explicitly represent the preposition in the path, the UD representation does not. We note that the system benefits from the explicit inclusion of prepositions in the path, and that the UD treatment of prepositions as dependent case markers, as well as the copula construction, is problematic in our system design.

There are several other dimensions of variation for natural language texts, as discussed in Section \ref{ch1:domain} in Chapter \ref{sec:first}. In Chapter \ref{sec:fifth} we go on to study low-resource settings in cross-genre and cross-lingual natural language understanding tasks.
We explore the use of meta-learning by leveraging training data from an auxiliary genre or language, to perform the zero-shot and few-shot cross-lingual and cross-genre transfer in two different natural language understanding (NLU) tasks: natural language inference (NLI) and question answering (QA). 
We here attempt to answer the following questions: 
\begin{itemize}
\item \textbf{RQ \ref{rq.5.1}.} \textit{Can meta-learning assist us in coping with low-resource settings in natural language understanding (NLU) tasks?}
\item \textbf{RQ \ref{rq.5.2}.} \textit{What is the impact of meta-learning on the performance of pre-trained language models such as BERT, XLM, and XLM-RoBERTa in cross-lingual NLU tasks?}
\item \textbf{RQ \ref{rq.5.3}.} \textit{Can meta-learning provide a model- and task-agnostic framework in low-resource NLU tasks?}
\end{itemize}
We propose, in Section \ref{sec:mode_xmaml}, a cross-lingual meta-learning framework for low-resource NLU tasks. We evaluate our framework on various scenarios, including cross-genre and cross-lingual NLI in zero- and few-shot settings across 15 languages (Section \ref{ch5:xnli}). We further, in Section \ref{ch5:qa}, investigate the model- and task-agnostic properties of our proposed framework by conducting experiments for the cross-lingual QA task. 
The experiments show that our cross-lingual meta-learning architecture (X-MAML) consistently improves the strong baseline models. It improves the multilingual BERT  by +3.65 and +1.04 percentage points in terms of average accuracy on zero-shot and few-shot XNLI, respectively.
Furthermore, it boosts the XLM-RoBERTa by +1.47 percentage points in terms of the average F1 score on zero-shot QA. In Section \ref{ch5:error}, we aim to answer:
\begin{itemize}
\item \textbf{RQ \ref{rq.5.4}.} \textit{Are typological commonalities among languages beneficial for the performance of cross-lingual meta-learning?}
\end{itemize}
Thus, we conduct an error analysis to explore the impact of typological sharing between languages in our framework. We evaluate on the World Atlas of Language Structure (WALS) as the largest openly available typological database. We attempt to predict typological features based on the mutual gain/loss in performance using our meta-learning framework. We further investigate whether the target and auxiliary languages have the same WALS feature value, given the change in accuracy when the two languages are used in cross-lingual meta-learning.
This indicates that languages with similar morphosyntactic properties can be beneficial to one another in our meta-learning framework. For instance, we observe that languages sharing a feature value for the WALS feature \emph{25A Locus of Marking:  Whole-language Typology} typically help each other in zero-shot cross-lingual meta-learning with Multi-BERT.

\section{Contributions}\label{conclud:contributions}
We here summarize the main contributions of the thesis:
\begin{enumerate}[label=(\roman*)]
\item Make use of  sequential transfer learning in terms of non-contextualized word embeddings to address the problem of low-resource domains in downstream tasks in NLP (see Chapters \ref{sec:second},\ref{sec:third} and \ref{sec:fourth}).
\item Enhance domain-specific embeddings using a domain-specific knowledge resource and present a benchmark dataset for intrinsic and extrinsic evaluation of domain embeddings (Chapter \ref{sec:second}). 
\item Propose a hybrid model that combines a reinforcement learning algorithm with partial annotation learning to clean the noisy, distantly supervised data for low-resource NER in different domains and languages (see Chapter \ref{sec:third}).
\item Design a neural architecture with syntactic input representation to alleviate domain impact in low-resource relation extraction (see Chapter \ref{sec:fourth}).
\item Introduce a cross-lingual meta-learning framework that provides further improvements in low-resource cross-lingual NLU tasks in various settings and languages (see Chapter \ref{sec:fifth}).
\end{enumerate}

\section{Future directions}\label{conclud:future}
Even though the proposed methods achieve competitive performance compared to previous work in the respective low-resource NLP tasks, there are several potential avenues for future research.
In the following, we will look into some of the future research directions that can alleviate some of the limitations of the proposed methods and low-resource NLP in general. 

Sequential transfer learning through pre-trained word embeddings has brought significant improvements for many low-resource NLP tasks.
The pre-trained word embeddings that we employed in chapters \ref{sec:second},\ref{sec:third} and \ref{sec:fourth} provide a single static representation for each word and have limitations that are already discussed in Section \ref{sec:emb} of Chapter \ref{sec:first}. The immediate idea for improving the proposed models in this thesis is to exploit the use of contextualized embeddings such as BERT, ELMo, and GPT. 
Some domain-specific versions of BERT are available, which are trained or fine-tuned on in-domain text, including SciBERT \parencite{DBLP:journals/corr/abs-1903-10676}, BioBERT \parencite{10.1093/bioinformatics/btz682} and ClinicalBERT \parencite{alsentzer-etal-2019-publicly} and  can be used in low-resource NER and relation extraction tasks on some of the target domains. However, there is still a need to train the contextualized embedding models in other domains. This remains a challenge since it requires large amounts of training data.

Even though the contextualized embeddings handle rare words implicitly using techniques such as byte-pair encoding and WordPiece embeddings, they still struggle with small corpora and with providing good representations for unseen words \parencite{DBLP:journals/corr/abs-1910-07181}. In chapter \ref{sec:second}, we incorporate a knowledge resource to augment the trained non-contextual embeddings by providing vector representations for infrequent and unseen technical terms. However, the proposed solution is limited in two respects: 
\begin{enumerate*}[label=(\roman*)]
    \item the target word must appear in the knowledge resource, and
    \item its neighbors must be part of the vocabulary of the embeddings model.
\end{enumerate*}
One way to overcome this limitation and improve embeddings of uncommon words is to jointly incorporate surface-form and context information directly from the textual content as described in \cite{DBLP:conf/aaai/SchickS19} and \cite{DBLP:journals/corr/abs-1904-01617}. The former combines an embedding based on n-grams with an embedding obtained from averaging over all context words. Whereas, the latter introduces an attentive mimicking model that computes an embedding by giving access not only to a word’s surface form, but also to all available contexts. The attentive mimicking model learns to attend to the most informative and reliable contexts.

The pre-trained language models can
be further enhanced by leveraging knowledge accumulated by humans in terms of knowledge resources such as WordNet \parencite{Miller:1995:WLD:219717.219748}, ConceptNet \parencite{speer2016conceptnet}, FrameNet \parencite{Baker:1998:BFP:980451.980860}, DBpedia \parencite{dbpedia-swj}. Work on incorporating knowledge resources into pre-trained language models has shown some promise on several NLP tasks \parencite{zhang2020SemBERT, peters-etal-2019-knowledge,wang2019kepler, zhang-etal-2019-ernie}. It would also be interesting to investigate the impact of jointly applying both of these research directions, i.e., surface-context information and knowledge-representations, on pre-trained language models.

A limitation of work in this thesis is the use of conventional neural architectures such as CNNs and BiLSTM in low-resource named entity recognition and relation extraction tasks, respectively. Transformer-based models such as BERT, GPT, XLM, and XLM-RoBERTa, which are proposed as one system for all tasks, might be more appropriate on low-resource NLP settings. Adapting the transformer-based model (see Section \ref{ch1:adaptation} in Chapter \ref{sec:first}) by the task specific fine-tuning, mitigates the need for having task-specific models and it transfers a pre-trained language model directly to a target task through minimal modifications, usually by modifying the last layer. \parencite{EMB-NLP01}. 

For low-resource NER, we envision numerous directions for future research. For instance, we deal with false positive instances at the sentence level via a reinforcement model. However, our method still has some challenges, and the false positive problem is still a bottleneck for the performance. We want to modify our approach to treat false positives at the entity type level, rather than treating these at the sentence level.
Moreover, we can expand our work to other types of reinforcement learning techniques such as imitation learning. It has been shown that the algorithmic expert in imitation learning allows direct policy learning. At the same time, the learned policies transfer successfully between domains and languages, improving the performance of low-resource NLP tasks  \parencite{Du_2019,liu-etal-2018-learning}.

Another limitation is the use of supervised learning algorithms throughout. The current neural models in chapters \ref{sec:third} and \ref{sec:fourth} require a set of training examples to provide good generalization. Future work can be to extend the study to improve the performance of the models in an unsupervised fashion.

We believe that there is room for further improvement in low-resource relation extraction, as presented in Chapter \ref{sec:fourth}.
A limitation of this work is that we cannot say that syntactic representations are more helpful in a resource-poor setting than in a resource-rich. This is an interesting future direction.
Another possible area of improvement would be to extend the study to neural dependency parsers. Graph-based neural dependency parser has been shown to provide more accurate parses \parencite{dozat-etal-2017-stanfords, kiperwasser-goldberg-2016-simple,song-etal-2019-leveraging}.
Moreover, we can study the problem of relation extraction in resource-poor settings by open information extraction (Open IE) techniques. Although the idea of Open IE has been investigated in many recent works \parencite{cui-etal-2018-neural,stanovsky-dagan-2016-creating,Gao2020NeuralSF,wu-etal-2019-open,Han_2019,Hu2020SelfORESR}, there are still a lot of open research questions. Most Open IE approaches focus on the English language and general domains, leaving aside other settings. The applicability and transferability of previously proposed Open IE approaches to other languages and domains will be an interesting direction for future work.

In this thesis, we study the impact of typology sharing among languages in our cross-lingual meta-learning framework. It would be interesting to investigate how NLP and linguistic typology can interact and benefit from each other in low-resource scenarios and extend our work to other cross-lingual NLP tasks and more languages.

Overall, the real world applications of NLP models are still challenging, and our contribution has been a step on the way, but there is more to do. We hope that our research in this thesis serves as a stepping stone for future research and inspires others to study open research questions in the area of low-resource NLP.

    \backmatter         

    \printbibliography

@inproceedings{CravenKumlien:99,
  author    = {Mark Craven and
               Johan Kumlien},
  editor    = {Thomas Lengauer and
               Reinhard Schneider and
               Peer Bork and
               Douglas L. Brutlag and
               Janice I. Glasgow and
               Hans{-}Werner Mewes and
               Ralf Zimmer},
  title     = {Constructing Biological Knowledge Bases by Extracting Information
               from Text Sources},
  booktitle = {Proceedings of the Seventh International Conference on Intelligent
               Systems for Molecular Biology},
  pages     = {77--86},
  publisher = {Association for the Advancement of Artificial Intelligence (AAAI) Press},
  year      = {1999},
  url       = {http://www.aaai.org/Library/ISMB/1999/ismb99-010.php},
  bibsource = {dblp computer science bibliography, https://dblp.org},
}

@inproceedings{baroni-dinu2014,
      author    = {Baroni, Marco  and  Dinu, Georgiana  and  Kruszewski, Germ\'{a}n},
      title     = {Don't count, predict! A systematic comparison of context-counting vs. context-predicting semantic vectors},
      booktitle = {Proceedings of the 52nd Annual Meeting of the Association for Computational Linguistics},
      year      = {2014},
      publisher = {Association for Computational Linguistics},
      pages     = {238--247},
      url       = {http://www.aclweb.org/anthology/P14-1023},
}

@inproceedings{BollegalaMK15a,
    Author = {Danushka Bollegala and Takanori Maehara and Ken-ichi Kawarabayashi},
    Title = {Learning Word Representations from Relational Graphs},
    Booktitle = {Proceedings of 53rd Annual Meeting of the Association for Computational Linguistics, and the 7th International Joint Conference on Natural Language Processing of the Asian Federation of Natural Language Processing},
    year = {2015},
    publisher = {Association for Computational Linguistics},
    pages = {730-740},
}

@inproceedings{ChiuACLBio,
    title = "How to Train good Word Embeddings for Biomedical {NLP}",
    author = "Chiu, Billy  and
      Crichton, Gamal  and
      Korhonen, Anna  and
      Pyysalo, Sampo",
    booktitle = "Proceedings of the 15th Workshop on Biomedical Natural Language Processing",
    year = "2016",
  %%address = "Berlin, Germany",
    publisher = "Association for Computational Linguistics",
    url = "https://www.aclweb.org/anthology/W16-2922",
    doi = "10.18653/v1/W16-2922",
    pages = "166--174",
}

@inproceedings{ChiuBillyRepEval2016,
   title = "Intrinsic Evaluation of Word Vectors Fails to Predict Extrinsic Performance",
    author = "Chiu, Billy  and
      Korhonen, Anna  and
      Pyysalo, Sampo",
    booktitle = "Proceedings of the 1st Workshop on Evaluating Vector-Space Representations for {NLP}",
    year = "2016",
    %%address = "Berlin, Germany",
    publisher = "Association for Computational Linguistics",
    url = "https://www.aclweb.org/anthology/W16-2501",
    doi = "10.18653/v1/W16-2501",
    pages = "1--6"
}

@article{DBLP:journals/corr/abs-1103-0398,
  author    = {Ronan Collobert and
               Jason Weston and
               L{\'{e}}on Bottou and
               Michael Karlen and
               Koray Kavukcuoglu and
               Pavel P. Kuksa},
  title     = {Natural Language Processing (almost) from Scratch},
  journal   = {Journal of Machine Learning Research},
  volume    = {12},
  pages     = {2493--2537},
  year      = {2011},
}

@inproceedings{faruqui:2014:NIPS-DLRLW,
  author    = {Manaal Faruqui and
               Jesse Dodge and
               Sujay Kumar Jauhar and
               Chris Dyer and
               Eduard H. Hovy and
               Noah A. Smith},
  title     = {Retrofitting Word Vectors to Semantic Lexicons},
  booktitle = {The 2015 Conference of the North American Chapter
               of the Association for Computational Linguistics: Human Language Technologies},
  publisher = {Association for Computational Linguistics},
  year      = {2015},
  pages     = {1606--1615},

}

@inproceedings{Gladkova2016NAACL,
    title = "Analogy-based detection of morphological and semantic relations with word embeddings: what works and what doesn{'}t.",
    author = "Gladkova, Anna  and
      Drozd, Aleksandr  and
      Matsuoka, Satoshi",
    booktitle = "Proceedings of the {NAACL} Student Research Workshop",
    year = "2016",
    %address = "San Diego, California",
    publisher = "Association for Computational Linguistics",
    url = "https://www.aclweb.org/anthology/N16-2002",
    doi = "10.18653/v1/N16-2002",
    pages = "8--15",
}

@inproceedings{HamiltonCLJ16,
   author    = {William L. Hamilton and
               Kevin Clark and
               Jure Leskovec and
               Dan Jurafsky},
  title     = {Inducing Domain-Specific Sentiment Lexicons from Unlabeled Corpora},
  booktitle = {Proceedings of the 2016 Conference on Empirical Methods in Natural Language Processing},
  publisher = {Association for Computational Linguistics},
  pages     = {595--605},
  year      = {2016},
}

@article{HillRK14,
   title = "{S}im{L}ex-999: Evaluating Semantic Models With (Genuine) Similarity Estimation",
    author = "Hill, Felix  and
      Reichart, Roi  and
      Korhonen, Anna",
    journal = "Computational Linguistics",
    volume = "41",
    number = "4",
    year = "2015",
    url = "https://www.aclweb.org/anthology/J15-4004",
    doi = "10.1162/COLI_a_00237",
    pages = "665--695",
}

@article{Leeuwenberg2016,
  author    = {Artuur Leeuwenberg and
               Mihaela Vela and
               Jon Dehdari and
               Josef van Genabith},
  title     = {A Minimally Supervised Approach for Synonym Extraction with Word Embeddings},
  journal   = {The Prague Bulletin of Mathematical Linguistics},
  volume    = {105},
  pages     = {111--142},
  number    = {1},
  year      = {2016},
  url       = {http://ufal.mff.cuni.cz/pbml/105/art-leeuwenberg-et-al.pdf},
  bibsource = {dblp computer science bibliography, https://dblp.org},
  publisher={Sciendo},

}

@article{LevyGD15,
  author    = {Omer Levy and
               Yoav Goldberg and
               Ido Dagan},
  title     = {Improving Distributional Similarity with Lessons Learned from Word
               Embeddings},
  journal   = {Transactions of the Association for Computational Linguistics},
  volume    = {3},
  pages     = {211--225},
  year      = {2015},
  url       = {https://tacl2013.cs.columbia.edu/ojs/index.php/tacl/article/view/570},
  bibsource = {dblp computer science bibliography, https://dblp.org},
}

@inproceedings{manning-EtAl:2014:P14-5,
 author    = {Christopher D. Manning and
               Mihai Surdeanu and
               John Bauer and
               Jenny Rose Finkel and
               Steven Bethard and
               David McClosky},
  title     = {The {Stanford} {CoreNLP} Natural Language Processing Toolkit},
  booktitle = {Proceedings of the 52nd Annual Meeting of the Association for Computational Linguistics},
  publisher = {Association for Computer Linguistics},
  pages     = {55--60},
  year      = {2014},

}

@inproceedings{MikolovYZ13,
  author    = {Tomas Mikolov and
               Wen{-}tau Yih and
               Geoffrey Zweig},
  title     = {Linguistic Regularities in Continuous Space Word Representations},
  booktitle = {Human Language Technologies: Conference of the North American Chapter of the Association of Computational Linguistics},
  publisher = {Association for Computational Linguistics},
  pages     = {746--751},
  year      = {2013},

}

@inproceedings{pilehvar-collier:2016:BioNLP16,
  author    = {Pilehvar, Mohammad Taher  and  Collier, Nigel},
  title     = {Improved Semantic Representation for Domain-Specific Entities},
  booktitle = {Proceedings of the 15th Workshop on Biomedical Natural Language Processing},
  publisher = {Association for Computational Linguistics},
  year      = {2016},
  pages     = {12--16},
}

@inproceedings{gensim,
    author = {Radim {\v R}eh{\r u}{\v r}ek and Petr Sojka},
    title = {Software Framework for Topic Modelling with Large Corpora},
   booktitle = {Proceedings of LREC 2010 workshop New Challenges for NLP Frameworks},
   pages = {46--50},
  publisher = {European Language Resources Association (ELRA)},
   year = {2010},
}

@inproceedings{schnabel2015eval,
   author    = {Tobias Schnabel and
               Igor Labutov and
               David M. Mimno and
               Thorsten Joachims},
  title     = {Evaluation methods for unsupervised word embeddings},
  booktitle = {Proceedings of the 2015 Conference on Empirical Methods in Natural Language Processing},
  publisher = {Association for Computational Linguistics},
  pages     = {298--307},
  year      = {2015},
}

@inproceedings{vanderPlas:2006,
  author    = {Lonneke van der Plas and
               J{\"{o}}rg Tiedemann},
  title     = {Finding Synonyms Using Automatic Word Alignment and Measures of Distributional Similarity},
  booktitle = {21st International Conference on Computational Linguistics and 44th Annual Meeting of the Association for Computational Linguistics},
  year = {2006},
 pages = {866--873},
 publisher = {Association for Computational Linguistics},

}

@inproceedings{yu2014improving,
  title = "Improving Lexical Embeddings with Semantic Knowledge",
    author = "Yu, Mo  and
      Dredze, Mark",
    booktitle = "Proceedings of the 52nd Annual Meeting of the Association for Computational Linguistics (Volume 2: Short Papers)",
    year = "2014",
    %address = "Baltimore, Maryland",
    publisher = "Association for Computational Linguistics",
    url = "https://www.aclweb.org/anthology/P14-2089",
    doi = "10.3115/v1/P14-2089",
    pages = "545--550",
}

@inproceedings{YaghoobzadehS16a,
    title = "Intrinsic Subspace Evaluation of Word Embedding Representations",
    author = {Yaghoobzadeh, Yadollah  and
          Sch{\"u}tze, Hinrich},
    booktitle = "Proceedings of the 54th Annual Meeting of the Association for Computational Linguistics (Volume 1: Long Papers)",
    year = "2016",
    %address = "Berlin, Germany",
    publisher = "Association for Computational Linguistics",
    url = "https://www.aclweb.org/anthology/P16-1023",
    doi = "10.18653/v1/P16-1023",
    pages = "236--246",
}

@inproceedings{Agirre:2009,
 author = {Agirre, Eneko and Alfonseca, Enrique and Hall, Keith and Kravalova, Jana and Pa\c{s}ca, Marius and Soroa, Aitor},
 title = {A Study on Similarity and Relatedness Using Distributional and WordNet-based Approaches},
 booktitle = {Proceedings of Human Language Technologies: The 2009 Annual Conference of the North American Chapter of the Association for Computational Linguistics},
 series = {NAACL '09},
 year = {2009},
 isbn = {978-1-932432-41-1},
 %location= {Boulder, Colorado},
 pages = {19--27},
 url = {http://dl.acm.org/citation.cfm?id=1620754.1620758},
 acmid = {1620758},
 publisher = {Association for Computational Linguistics},
 %address = {Stroudsburg, PA, USA},
 }

@inproceedings{Bruni:2012,
 author = {Bruni, Elia and Boleda, Gemma and Baroni, Marco and Tran, Nam-Khanh},
 title = {Distributional Semantics in Technicolor},
 booktitle = {Proceedings of the 50th Annual Meeting of the Association for Computational Linguistics: Long Papers - Volume 1},
 series = {ACL '12},
 year = {2012},
 %location= {Jeju Island, Korea},
 pages = {136--145},
 url = {http://dl.acm.org/citation.cfm?id=2390524.2390544},
 acmid = {2390544},
 publisher = {Association for Computational Linguistics},
 %address = {Stroudsburg, PA, USA},
}

@inproceedings{DBLP:journals/corr/FriedD14,
  author    = {Daniel Fried and
               Kevin Duh},
  title     = {Incorporating Both Distributional and Relational Semantics in Word
               Representations},
  booktitle = {3rd International Conference on Learning Representations, {ICLR}, Workshop Track Proceedings},
  year      = {2015},
  url       = {http://arxiv.org/abs/1412.5836},
  bibsource = {dblp computer science bibliography, https://dblp.org}
}

@inproceedings{DBLP:journals/corr/PilehvarC16,
  title = "De-Conflated Semantic Representations",
    author = "Pilehvar, Mohammad Taher  and
      Collier, Nigel",
    booktitle = "Proceedings of the 2016 Conference on Empirical Methods in Natural Language Processing",
    year = "2016",
    %address = "Austin, Texas",
    publisher = "Association for Computational Linguistics",
    url = "https://www.aclweb.org/anthology/D16-1174",
    doi = "10.18653/v1/D16-1174",
    pages = "1680--1690",
}

@article{Miller:1995:WLD:219717.219748,
 author = {Miller, George A.},
 title = {WordNet: A Lexical Database for English},
 journal = {COMMUNICATIONS OF THE ACM},
 issue_date = {Nov. 1995},
 volume = {38},
 number = {11},
 year = {1995},
 issn = {0001-0782},
 pages = {39--41},
 url = {http://doi.acm.org/10.1145/219717.219748},
 doi = {10.1145/219717.219748},
 acmid = {219748},
 publisher = {Association for Computing Machinery},
 %address = {New York, NY, USA},
}

@inproceedings{ganitkevitch2013paraphrase,
    title = "{PPDB}: The Paraphrase Database",
    author = "Ganitkevitch, Juri  and
      Van Durme, Benjamin  and
      Callison-Burch, Chris",
    booktitle = "Proceedings of the 2013 Conference of the North {A}merican Chapter of the Association for Computational Linguistics: Human Language Technologies",
    year = "2013",
    %address = "Atlanta, Georgia",
    publisher = "Association for Computational Linguistics",
    url = "https://www.aclweb.org/anthology/N13-1092",
    pages = "758--764",
}

@inproceedings{Baker:1998:BFP:980451.980860,
 author = {Baker, Collin F. and Fillmore, Charles J. and Lowe, John B.},
 title = {The Berkeley FrameNet Project},
 booktitle = {Proceedings of the 17th International Conference on Computational Linguistics - Volume 1},
 series = {COLING '98},
 year = {1998},
 %location= {Montreal, Quebec, Canada},
 pages = {86--90},
 url = {https://doi.org/10.3115/980451.980860},
 doi = {10.3115/980451.980860},
 acmid = {980860},
 publisher = {Association for Computational Linguistics},
 %address = {Stroudsburg, PA, USA},
}

@book{zipf1972human,
  title={Human Behavior and the Principle of Least Effort: An Introduction to Human Ecology},
  author={Zipf, G.K.},
  url={https://books.google.no/books?id=gLwknQEACAAJ},
  year={1972},
  publisher={Hafner},
}

@inproceedings{Botha:2014:CMW:3044805.3045104,
    author = {Botha, Jan A. and Blunsom, Phil},
    title = {Compositional Morphology for Word Representations and Language Modelling},
    year = {2014},
    publisher = {JMLR.org},
    booktitle = {Proceedings of the 31st International Conference on International Conference on Machine Learning - Volume 32},
    pages = {1899--1907},
    %location= {Beijing, China},
    series = {ICML},
}

@inproceedings{DBLP:conf/naacl/SoricutO15,
    title = "Unsupervised Morphology Induction Using Word Embeddings",
    author = "Soricut, Radu  and Och, Franz",
    booktitle = "Proceedings of the 2015 Conference of the North {A}merican Chapter of the Association for Computational Linguistics: Human Language Technologies",
    year = "2015",
    %address = "Denver, Colorado",
    publisher = "Association for Computational Linguistics",
    url = "https://www.aclweb.org/anthology/N15-1186",
    doi = "10.3115/v1/N15-1186",
    pages = "1627--1637",
}

@article{DBLP:journals/tacl/BojanowskiGJM17,
  author    = {Piotr Bojanowski and
               Edouard Grave and
               Armand Joulin and
               Tomas Mikolov},
  title     = {Enriching Word Vectors with Subword Information},
  journal =  {Transactions of the Association for Computational Linguistics},
  volume    = {5},
  pages     = {135--146},
  year      = {2017},
  url       = {https://transacl.org/ojs/index.php/tacl/article/view/999},
  bibsource = {dblp computer science bibliography, https://dblp.org},
}

@inproceedings{E17-2062,
  author = 	"Pilehvar, Mohammad Taher
		and Collier, Nigel",
  title = 	"Inducing Embeddings for Rare and Unseen Words by Leveraging Lexical Resources",
  booktitle = 	"Proceedings of the 15th Conference of the European Chapter of the Association for Computational Linguistics: Volume 2, Short Papers",
  year = 	"2017",
  publisher = 	"Association for Computational Linguistics",
  pages = 	"388--393",
  %location= 	"Valencia, Spain",
  url = 	"http://aclweb.org/anthology/E17-2062",
}

@article{Haveliwala:2003:TPC:1435677.858989,
 author = {Haveliwala, Taher H.},
  journal={IEEE Transactions on Knowledge and Data Engineering},
  title={Topic-sensitive PageRank: a context-sensitive ranking algorithm for Web search},
  year={2003},
  volume={15},
  number={4},
  pages={784-796},
}

@inproceedings{Agirre:2009:PPW:1609067.1609070,
 author = {Agirre, Eneko and Soroa, Aitor},
 title = {Personalizing PageRank for Word Sense Disambiguation},
 booktitle = {Proceedings of the 12th Conference of the European Chapter of the Association for Computational Linguistics},
 series = {EACL '09},
 year = {2009},
 %location= {Athens, Greece},
 pages = {33--41},
 url = {http://dl.acm.org/citation.cfm?id=1609067.1609070},
 acmid = {1609070},
 publisher = {Association for Computational Linguistics},
 %address = {Stroudsburg, PA, USA},
}

@inproceedings{10.1007/978-3-319-41754-7_7,
    author="Nooralahzadeh, Farhad
            and Lopez, C{\'e}dric
            and Cabrio, Elena
            and Gandon, Fabien
            and Segond, Fr{\'e}d{\'e}rique",
    title="Adapting Semantic Spreading Activation to Entity Linking in Text",
    booktitle="Natural Language Processing and Information Systems",
    year="2016",
    publisher="Springer International Publishing",
    %address="Cham",
    pages="74--90",
    isbn="978-3-319-41754-7",
}

@article{Pilehvar:2015:ST:2827893.2828161,
     author = {Pilehvar, Mohammad Taher and Navigli, Roberto},
     title = {From Senses to Texts: An all-in-one graph-based approach for measuring semantic similarity},
     journal = {Artificial Intelligence},
     issue_date = {November 2015},
     volume = {228},
     number = {C},
     year = {2015},
     issn = {0004-3702},
     pages = {95--128},
     url = {https://doi.org/10.1016/j.artint.2015.07.005},
     doi = {10.1016/j.artint.2015.07.005},
     acmid = {2828161},
     publisher = {Elsevier Science Publishers Ltd.},
     %address = {Essex, UK},
}

@article{Brin:1998:ALH:297810.297827,
    title = "The anatomy of a large-scale hypertextual Web search engine",
    journal = "Computer Networks and ISDN Systems",
    volume = "30",
    number = "1",
    pages = "107--117",
    year = "1998",
    note = "Proceedings of the Seventh International World Wide Web Conference",
    issn = "0169-7552",
    doi = "https://doi.org/10.1016/S0169-7552(98)00110-X",
    url = "http://www.sciencedirect.com/science/article/pii/S016975529800110X",
    author = "Sergey Brin and Lawrence Page",
}

@article{Rubenstein:1965,
 author = {Rubenstein, Herbert and Goodenough, John B.},
 title = {Contextual Correlates of Synonymy},
 journal = {COMMUNICATIONS OF THE ACM},
 issue_date = {Oct. 1965},
 volume = {8},
 number = {10},
 year = {1965},
 issn = {0001-0782},
 pages = {627--633},
 numpages = {7},
 url = {http://doi.acm.org/10.1145/365628.365657},
 doi = {10.1145/365628.365657},
 acmid = {365657},
 publisher = {Association for Computing Machinery},
 %address = {New York, NY, USA},
}

@article{miller1991contextual,
  author = {Miller, George A and Charles, Walter G},
  journal = {Language and Cognitive Processes},
  number = {1},
  pages = {1--28},
  publisher = {Psychology Press},
  title = {Contextual correlates of semantic similarity},
  url = {http://eric.ed.gov/ERICWebPortal/recordDetail?accno=EJ431389},
  volume = {6},
  year = {1991},
}

@inproceedings{Finkelstein:2001,
 author = {Finkelstein, Lev and Gabrilovich, Evgeniy and Matias, Yossi and Rivlin, Ehud and Solan, Zach and Wolfman, Gadi and Ruppin, Eytan},
 title = {Placing Search in Context: The Concept Revisited},
 booktitle = {Proceedings of the 10th International Conference on World Wide Web},
 series = {WWW '01},
 year = {2001},
 isbn = {1-58113-348-0},
 %location= {Hong Kong, Hong Kong},
 pages = {406--414},
 numpages = {9},
 url = {http://doi.acm.org/10.1145/371920.372094},
 doi = {10.1145/371920.372094},
 acmid = {372094},
 publisher = {Association for Computing Machinery},
 %address = {New York, NY, USA},
 keywords = {context, invisible web, search, semantic processing, statistical natural language processing},
}

@inproceedings{Yang06verbsimilarity,
    author = {Dongqiang Yang and David M. W. Powers},
    title = {Verb Similarity on the Taxonomy of Wordnet},
    booktitle = {In the 3rd International WordNet Conference (GWC-06)},
    year = {2006},
    publisher={Masaryk University},
}

@inproceedings{Radinsky:2011,
 author = {Radinsky, Kira and Agichtein, Eugene and Gabrilovich, Evgeniy and Markovitch, Shaul},
 title = {A Word at a Time: Computing Word Relatedness Using Temporal Semantic Analysis},
 booktitle = {Proceedings of the 20th International Conference on World Wide Web},
 series = {WWW '11},
 year = {2011},
 isbn = {978-1-4503-0632-4},
 %location= {Hyderabad, India},
 pages = {337--346},
 numpages = {10},
 url = {http://doi.acm.org/10.1145/1963405.1963455},
 doi = {10.1145/1963405.1963455},
 acmid = {1963455},
 publisher = {Association for Computing Machinery},
 %address = {New York, NY, USA},
 keywords = {semantic analysis, semantic similarity, temporal dynamics, temporal semantics, word relatedness},
}

@inproceedings{Halawi2012,
    author = {Halawi, Guy and Dror, Gideon and Gabrilovich, Evgeniy and Koren, Yehuda},
    title = {Large-Scale Learning of Word Relatedness with Constraints},
    year = {2012},
    isbn = {9781450314626},
    publisher = {Association for Computing Machinery},
    %address = {New York, NY, USA},
    url = {https://doi.org/10.1145/2339530.2339751},
    doi = {10.1145/2339530.2339751},
    booktitle = {Proceedings of the 18th ACM SIGKDD International Conference on Knowledge Discovery and Data Mining},
    pages = {1406–1414},
    numpages = {9},
    keywords = {semantic similarity, word relatedness},
    %location= {Beijing, China},
    series = {KDD ’12}
}

@inproceedings{Luong-etal2013,
 title = "Better Word Representations with Recursive Neural Networks for Morphology",
    author = "Luong, Thang  and
      Socher, Richard  and
      Manning, Christopher",
    booktitle = "Proceedings of the Seventeenth Conference on Computational Natural Language Learning",
    year = "2013",
    %address = "Sofia, Bulgaria",
    publisher = "Association for Computational Linguistics",
    url = "https://www.aclweb.org/anthology/W13-3512",
    pages = "104--113",
}

@inproceedings{BakerRK14,
    title = "An Unsupervised Model for Instance Level Subcategorization Acquisition",
    author = "Baker, Simon  and
      Reichart, Roi  and
      Korhonen, Anna",
    booktitle = "Proceedings of the 2014 Conference on Empirical Methods in Natural Language Processing ({EMNLP})",
    year = "2014",
    %address = "Doha, Qatar",
    publisher = "Association for Computational Linguistics",
    url = "https://www.aclweb.org/anthology/D14-1034",
    doi = "10.3115/v1/D14-1034",
    pages = "278--289"
}

@article{Landauer97asolution,
    author = {Thomas K Landauer and Susan T. Dutnais},
    title = {A solution to Plato’s problem: The latent semantic analysis theory of acquisition, induction, and representation of knowledge},
    journal = {PSYCHOLOGICAL REVIEW},
    year = {1997},
    volume = {104},
    number = {2},
    pages = {211--240}
}

@phdthesis{Almuhareb06,
  author    = {Abdulrahman Almuhareb},
  title     = {Attributes in lexical acquisition},
  school    = {University of Essex, Colchester, {UK}},
  year      = {2006},
  url       = {http://ethos.bl.uk/OrderDetails.do?uin=uk.bl.ethos.428974},
  bibsource = {dblp computer science bibliography, http://dblp.org}
}

@inproceedings{Baroni2008,
  title={Lexical Semantics Bridging the gap between semantic theory and computational simulations},
  author={Marco Baroni and Stefan Evert and Alessandro Lenci},
  year = {2008},
  booktitle={Proceedings of 20th European Summer School in Logic, Language and Information (ESSLLI 2008)}
}

@article{Baroni:2010,
    title = "Distributional Memory: A General Framework for Corpus-Based Semantics",
    author = "Baroni, Marco  and
      Lenci, Alessandro",
    journal = "Computational Linguistics",
    publisher = "MIT Press",
    volume = "36",
    number = "4",
    year = "2010",
    url = "https://www.aclweb.org/anthology/J10-4006",
    doi = "10.1162/coli_a_00016",
    pages = "673--721",
}

@phdthesis{Pado07,
    author = {Ulrike Pad\'o},
    title = {The integration of syntax and semantic plausibility in a wide-coverage model of human sentence processing},
    school = {Universit\"{a}t des Saarlandes},
    year = {2007},
    %address = {Postfach 151141, 66041 Saarbr\"{u}cken},
}

@article{McRae1998283,
    title = "Modeling the Influence of Thematic Fit (and Other Constraints) in On-line Sentence Comprehension ",
    journal = "Journal of Memory and Language ",
    volume = "38",
    number = "3",
    pages = "283 - 312",
    year = "1998",
    issn = "0749-596X",
    doi = "http://dx.doi.org/10.1006/jmla.1997.2543",
    url = "www.sciencedirect.com/science/article/pii/S0749596X97925432",
    author = "Ken McRae and Michael J. Spivey-Knowlton and Michael K. Tanenhaus"
}

@inproceedings{camacho2016find,
    title = "Find the word that does not belong: A Framework for an Intrinsic Evaluation of Word Vector Representations",
    author = "Camacho-Collados, Jos{\'e}  and
      Navigli, Roberto",
    booktitle = "Proceedings of the 1st Workshop on Evaluating Vector-Space Representations for {NLP}",
    year = "2016",
    %address = "Berlin, Germany",
    publisher = "Association for Computational Linguistics",
    url = "https://www.aclweb.org/anthology/W16-2508",
    doi = "10.18653/v1/W16-2508",
    pages = "43--50",
}

@inproceedings{qvec:enmlp:15,
    title = "Evaluation of Word Vector Representations by Subspace Alignment",
    author = "Tsvetkov, Yulia  and
      Faruqui, Manaal  and
      Ling, Wang  and
      Lample, Guillaume  and
      Dyer, Chris",
    booktitle = "Proceedings of the 2015 Conference on Empirical Methods in Natural Language Processing",
    year = "2015",
    %address = "Lisbon, Portugal",
    publisher = "Association for Computational Linguistics",
    url = "https://www.aclweb.org/anthology/D15-1243",
    doi = "10.18653/v1/D15-1243",
    pages = "2049--2054",
}

@inproceedings{TsvetkovFD16,
  title = "Correlation-based Intrinsic Evaluation of Word Vector Representations",
    author = "Tsvetkov, Yulia  and
      Faruqui, Manaal  and
      Dyer, Chris",
    booktitle = "Proceedings of the 1st Workshop on Evaluating Vector-Space Representations for {NLP}",
    year = "2016",
    %address = "Berlin, Germany",
    publisher = "Association for Computational Linguistics",
    url = "https://www.aclweb.org/anthology/W16-2520",
    doi = "10.18653/v1/W16-2520",
    pages = "111--115",
}

@article{Hardoon:2004,
 author = {Hardoon, David R. and Szedmak, Sandor R. and Shawe-taylor, John R.},
 title = {Canonical Correlation Analysis: An Overview with Application to Learning Methods},
 journal= {Neural Computation},
 volume = {16},
 number = {12},
 year = {2004},
 issn = {0899-7667},
 pages = {2639--2664},
 numpages = {26},
 acmid = {1119703},
 publisher = {MIT Press},
 %address = {Cambridge, MA, USA},
}

@article{Marcus:1993,
    title = "Building a Large Annotated Corpus of {E}nglish: The {P}enn {T}reebank",
    author = "Marcus, Mitchell P.  and Santorini, Beatrice  and Marcinkiewicz, Mary Ann",
    journal = "Computational Linguistics",
    volume = "19",
    number = "2",
    year = "1993",
    url = "https://www.aclweb.org/anthology/J93-2004",
    pages = "313--330",
}

@inproceedings{GHANNAY16.392,
  author = {Sahar Ghannay and Benoit Favre and Yannick Estève and Nathalie Camelin},
  title = {Word Embedding Evaluation and Combination},
  booktitle = {Proceedings of the Tenth International Conference on Language Resources and Evaluation (LREC 2016)},
  year = {2016},
  %location= {Portorož, Slovenia},
  publisher = {European Language Resources Association (ELRA)},
  %address = {Paris, France},
  isbn = {978-2-9517408-9-1},
 }

@inproceedings{NayakRepEval2016,
   title = "Evaluating Word Embeddings Using a Representative Suite of Practical Tasks",
    author = "Nayak, Neha  and
      Angeli, Gabor  and
      Manning, Christopher D.",
    booktitle = "Proceedings of the 1st Workshop on Evaluating Vector-Space Representations for {NLP}",
    year = "2016",
    %address = "Berlin, Germany",
    publisher = "Association for Computational Linguistics",
    url = "https://www.aclweb.org/anthology/W16-2504",
    doi = "10.18653/v1/W16-2504",
    pages = "19--23",
}

@inproceedings{TjongKimSang:2000,
 author = {Tjong Kim Sang, Erik F. and Buchholz, Sabine},
 title = {Introduction to the CoNLL-2000 Shared Task: Chunking},
 booktitle = {Proceedings of the 2nd Workshop on Learning Language in Logic and the 4th Conference on Computational Natural Language Learning - Volume 7},
 series = {ConLL '00},
 year = {2000},
 %location= {Lisbon, Portugal},
 pages = {127--132},
 url = {http://dx.doi.org/10.3115/1117601.1117631},
 doi = {10.3115/1117601.1117631},
 acmid = {1117631},
 publisher = {Association for Computational Linguistics},
 %address = {Stroudsburg, PA, USA},
}

@inproceedings{TjongKimSang:2003,
  title = "Introduction to the {C}o{NLL}-2003 Shared Task: Language-Independent Named Entity Recognition",
    author = "Tjong Kim Sang, Erik F.  and
      De Meulder, Fien",
    booktitle = "Proceedings of the Seventh Conference on Natural Language Learning at {HLT}-{NAACL} 2003",
 publisher = {Association for Computational Linguistics},
    year = "2003",
    url = "https://www.aclweb.org/anthology/W03-0419",
    pages = "142--147",
}

@inproceedings{socher-EtAl:2013,
  author    = {Socher, Richard  and  Perelygin, Alex  and  Wu, Jean  and  Chuang, Jason  and  Manning, Christopher D.  and  Ng, Andrew  and  Potts, Christopher},
  title     = {Recursive Deep Models for Semantic Compositionality Over a Sentiment Treebank},
  booktitle = {Proceedings of the 2013 Conference on Empirical Methods in Natural Language Processing},
  year      = {2013},
  %address   = {Seattle, Washington, USA},
  publisher = {Association for Computational Linguistics},
  pages     = {1631--1642},
  url       = {http://www.aclweb.org/anthology/D13-1170},
}

@inproceedings{Li:2002,
 author = {Li, Xin and Roth, Dan},
 title = {Learning Question Classifiers},
 booktitle = {Proceedings of the 19th International Conference on Computational Linguistics - Volume 1},
 series = {COLING '02},
 year = {2002},
 %location= {Taipei, Taiwan},
 pages = {1--7},
 url = {http://dx.doi.org/10.3115/1072228.1072378},
 doi = {10.3115/1072228.1072378},
 acmid = {1072378},
 publisher = {Association for Computational Linguistics},
 %address = {Stroudsburg, PA, USA},
}

@inproceedings{ganitkevitch2013ppdb,
  title = {{PPDB}: The Paraphrase Database},
  author = {Ganitkevitch, Juri and {Van Durme}, Benjamin and
    Callison-Burch, Chris},
  booktitle = {Proceedings of NAACL-HLT},
  pages = {758--764},
  year={2013},
  %address = {Atlanta, Georgia},
  publisher = {Association for Computational Linguistics},
  url = {http://cs.jhu.edu/~ccb/publications/ppdb.pdf},
}

@inproceedings{maas2011,
  author    = {Maas, Andrew L.  and  Daly, Raymond E.  and  Pham, Peter T.  and  Huang, Dan  and  Ng, Andrew Y.  and  Potts, Christopher},
  title     = {Learning Word Vectors for Sentiment Analysis},
  booktitle = {Proceedings of the 49th Annual Meeting of the Association for Computational Linguistics: Human Language Technologies},
  year      = {2011},
  %address   = {Portland, Oregon, USA},
  publisher = {Association for Computational Linguistics},
  pages     = {142--150},
  url       = {http://www.aclweb.org/anthology/P11-1015},
}

@inproceedings{Turian:2010:WRS:1858681.1858721,
 author = {Turian, Joseph and Ratinov, Lev and Bengio, Yoshua},
 title = {Word Representations: A Simple and General Method for Semi-supervised Learning},
 booktitle = {Proceedings of the 48th Annual Meeting of the Association for Computational Linguistics},
 series = {ACL '10},
 year = {2010},
 %location= {Uppsala, Sweden},
 pages = {384--394},
 url = {http://dl.acm.org/citation.cfm?id=1858681.1858721},
 acmid = {1858721},
 publisher = {Association for Computational Linguistics},
 %address = {Stroudsburg, PA, USA},
}

@inproceedings{DBLP:journals/corr/Kim14f,
  author    = {Yoon Kim},
  title     = {Convolutional Neural Networks for Sentence Classification},
  booktitle = {Proceedings of the 2014 Conference on Empirical Methods in Natural Language Processing},
  publisher = {Association for Computational Linguistics},
  pages     = {1746--1751},
  year      = {2014},
}

@inproceedings{P16-1200,
  author = 	"Lin, Yankai
		and Shen, Shiqi
		and Liu, Zhiyuan
		and Luan, Huanbo
		and Sun, Maosong",
  title = 	"Neural Relation Extraction with Selective Attention over Instances",
  booktitle = 	"Proceedings of the 54th Annual Meeting of the Association for      Computational Linguistics (Volume 1: Long Papers)    ",
  year = 	{2016},
  publisher = 	{Association for Computational Linguistics},
  pages = 	{2124--2133},
}

@inproceedings{DBLP:conf/acl/ZhouSTQLHX16,
  author    = {Peng Zhou and
               Wei Shi and
               Jun Tian and
               Zhenyu Qi and
               Bingchen Li and
               Hongwei Hao and
               Bo Xu},
  title     = {Attention-Based Bidirectional Long Short-Term Memory Networks for
               Relation Classification},
  booktitle = {Proceedings of the 54th Annual Meeting of the Association for Computational Linguistics},
  year      = {2016},
  publisher = 	{Association for Computational Linguistics},
}

@inproceedings{W15-1506,
  author = 	"Nguyen, Thien Huu
		and Grishman, Ralph",
  title = 	"Relation Extraction: Perspective from Convolutional Neural Networks",
  booktitle = 	"Proceedings of the 1st Workshop on Vector Space Modeling for {NLP} ",
  year = 	"2015",
  publisher = 	"Association for Computational Linguistics",
  pages = 	"39--48",
  %location= 	"Denver, Colorado",
  doi = 	"10.3115/v1/W15-1506",
  url = 	"http://www.aclweb.org/anthology/W15-1506"
}

@inproceedings{DBLP:journals/corr/LiuWLJZW15,
    title = "A Dependency-Based Neural Network for Relation Classification",
    author = "Liu, Yang  and
      Wei, Furu  and
      Li, Sujian  and
      Ji, Heng  and
      Zhou, Ming  and
      Wang, Houfeng",
    booktitle = "Proceedings of the 53rd Annual Meeting of the Association for Computational Linguistics and the 7th International Joint Conference on Natural Language Processing (Volume 2: Short Papers)",
    year = "2015",
    %address = "Beijing, China",
    publisher = "Association for Computational Linguistics",
    url = "https://www.aclweb.org/anthology/P15-2047",
    doi = "10.3115/v1/P15-2047",
    pages = "285--290",

}

@inproceedings{DBLP:journals/corr/XuMLCPJ15,
  title = "Classifying Relations via Long Short Term Memory Networks along Shortest Dependency Paths",
    author = "Xu, Yan  and
      Mou, Lili  and
      Li, Ge  and
      Chen, Yunchuan  and
      Peng, Hao  and
      Jin, Zhi",
    booktitle = "Proceedings of the 2015 Conference on Empirical Methods in Natural Language Processing",
    year = "2015",
    %address = "Lisbon, Portugal",
    publisher = "Association for Computational Linguistics",
    url = "https://www.aclweb.org/anthology/D15-1206",
    doi = "10.18653/v1/D15-1206",
    pages = "1785--1794",
}

@inproceedings{Bunescu:2005:SPD:1220575.1220666,
 author = {Bunescu, Razvan C. and Mooney, Raymond J.},
 title = {A Shortest Path Dependency Kernel for Relation Extraction},
 booktitle = {Proceedings of the Conference on Human Language Technology and Empirical Methods in Natural Language Processing},
 series = {HLT '05},
 year = {2005},
 %location= {Vancouver, British Columbia, Canada},
 pages = {724--731},
 numpages = {8},
 url = {https://doi.org/10.3115/1220575.1220666},
 doi = {10.3115/1220575.1220666},
 acmid = {1220666},
 publisher = {Association for Computational Linguistics},
 %address = {Stroudsburg, PA, USA},
}

@inproceedings{Boh:Niv:12,
    title = "A Transition-Based System for Joint Part-of-Speech Tagging and Labeled Non-Projective Dependency Parsing",
    author = "Bohnet, Bernd  and
      Nivre, Joakim",
    booktitle = "Proceedings of the 2012 Joint Conference on Empirical Methods in Natural Language Processing and Computational Natural Language Learning",
    year = "2012",
    %address = "Jeju Island, Korea",
    publisher = "Association for Computational Linguistics",
    url = "https://www.aclweb.org/anthology/D12-1133",
    pages = "1455--1465"
}

@inproceedings{Joh:Nug:07,
  title = "Extended Constituent-to-Dependency Conversion for {E}nglish",
    author = "Johansson, Richard  and
      Nugues, Pierre",
    booktitle = "Proceedings of the 16th Nordic Conference of Computational Linguistics ({NODALIDA} 2007)",
    year = "2007",
    %address = "Tartu, Estonia",
    publisher = "University of Tartu, Estonia",
    url = "https://www.aclweb.org/anthology/W07-2416",
    pages = "105--112",
}

@inproceedings{de-marneffe-etal-2014-universal,
    title = "Universal {S}tanford dependencies: A cross-linguistic typology",
    author = "{de Marneffe}, Marie-Catherine  and
      Dozat, Timothy  and
      Silveira, Natalia  and
      Haverinen, Katri  and
      Ginter, Filip  and
      Nivre, Joakim  and
      Manning, Christopher D.",
    booktitle = "Proceedings of the Ninth International Conference on Language Resources and Evaluation ({LREC}'14)",
    year = "2014",
    %address = "Reykjavik, Iceland",
    publisher = "European Language Resources Association (ELRA)",
    url = "http://www.lrec-conf.org/proceedings/lrec2014/pdf/1062_Paper.pdf",
    pages = "4585--4592",
}

@inproceedings{Mikolov2013a,
  author    = {Tomas Mikolov and
               Kai Chen and
               Greg Corrado and
               Jeffrey Dean},
  title     = {Efficient Estimation of Word Representations in Vector Space},
  booktitle = {1st International Conference on Learning Representations, {ICLR} 2013, Workshop Track Proceedings},
  year      = {2013},
  url       = {http://arxiv.org/abs/1301.3781},
  bibsource = {dblp computer science bibliography, https://dblp.org},
}

@Inbook{Prechelt:1998:ES:645754.668392,
    author="Prechelt, Lutz",
    title="Early Stopping --- But When?",
    bookTitle="Neural Networks: Tricks of the Trade: Second Edition",
    year="2012",
    publisher="Springer Berlin Heidelberg",
    %address="Berlin, Heidelberg",
    pages="53--67",
    isbn="978-3-642-35289-8"
}

@inproceedings{Rink:2010:UCS:1859664.1859721,
 title = "{UTD}: Classifying Semantic Relations by Combining Lexical and Semantic Resources",
    author = "Rink, Bryan  and
      Harabagiu, Sanda",
    booktitle = "Proceedings of the 5th International Workshop on Semantic Evaluation",
    year = "2010",
    %address = "Uppsala, Sweden",
    publisher = "Association for Computational Linguistics",
    url = "https://www.aclweb.org/anthology/S10-1057",
    pages = "256--259",
}

@inproceedings{Mar:Fro:Tit:17,
  author = 	"Diego Marcheggiani and Anton Frolov and Ivan Titov",
  title = 	"A Simple and Accurate Syntax-Agnostic Neural Model for Dependency-based Semantic Role Labeling",
  booktitle = 	"Proceedings of the 21st Conference on Computational Natural Language Learning",
  year = 	"2017",
  publisher = 	"Association for Computational Linguistics",
  pages = 	"411--420",
}

@inproceedings{Soc:Per:Wu:13,
  title = "Recursive Deep Models for Semantic Compositionality Over a Sentiment Treebank",
    author = "Socher, Richard  and
      Perelygin, Alex  and
      Wu, Jean  and
      Chuang, Jason  and
      Manning, Christopher D.  and
      Ng, Andrew  and
      Potts, Christopher",
    booktitle = "Proceedings of the 2013 Conference on Empirical Methods in Natural Language Processing",
    year = "2013",
    %address = "Seattle, Washington, USA",
    publisher = "Association for Computational Linguistics",
    url = "https://www.aclweb.org/anthology/D13-1170",
    pages = "1631--1642",
}

@inproceedings{Wu:Zha:Yan:17,
    title = "Sequence-to-Dependency Neural Machine Translation",
    author = "Wu, Shuangzhi  and
      Zhang, Dongdong  and
      Yang, Nan  and
      Li, Mu  and
      Zhou, Ming",
    booktitle = "Proceedings of the 55th Annual Meeting of the Association for Computational Linguistics (Volume 1: Long Papers)",
    year = "2017",
    %address = "Vancouver, Canada",
    publisher = "Association for Computational Linguistics",
    url = "https://www.aclweb.org/anthology/P17-1065",
    doi = "10.18653/v1/P17-1065",
    pages = "698--707",
}

@inproceedings{Hendrickx:2010:STM:1859664.1859670,
 author = {Hendrickx, Iris and Kim, Su Nam and Kozareva, Zornitsa and Nakov, Preslav and S\'{e}aghdha, Diarmuid \'{O}. and Pad\'{o}, Sebastian and Pennacchiotti, Marco and Romano, Lorenza and Szpakowicz, Stan},
 title = {SemEval-2010 Task 8: Multi-way Classification of Semantic Relations Between Pairs of Nominals},
 booktitle = {Proceedings of the 5th International Workshop on Semantic Evaluation},
 year = {2010},
 pages = {33--38},
 publisher = {Association for Computational Linguistics}
}

@inproceedings{Socher:2012:SCT:2390948.2391084,
 author = {Socher, Richard and Huval, Brody and Manning, Christopher D. and Ng, Andrew Y.},
 title = {Semantic Compositionality Through Recursive Matrix-vector Spaces},
 booktitle = {Proceedings of the 2012 Joint Conference on Empirical Methods in Natural Language Processing and Computational Natural Language Learning},
 year = {2012},
 pages = {1201--1211},
 numpages = {11},
 publisher = {Association for Computational Linguistics}
}

@article{DBLP:journals/corr/ZhangW15a,
  author    = {Dongxu Zhang and
               Dong Wang},
  title     = {Relation Classification via Recurrent Neural Network},
  journal   = {CoRR},
  volume    = {abs/1508.01006},
  year      = {2015},
  url       = {http://arxiv.org/abs/1508.01006},
  archivePrefix = {arXiv},
  eprint    = {1508.01006},
  bibsource = {dblp computer science bibliography, https://dblp.org},
}

@inproceedings{DBLP:journals/corr/LeeDS17,
  title = "{MIT} at {S}em{E}val-2017 Task 10: Relation Extraction with Convolutional Neural Networks",
    author = "Lee, Ji Young  and
      Dernoncourt, Franck  and
      Szolovits, Peter",
    booktitle = "Proceedings of the 11th International Workshop on Semantic Evaluation ({S}em{E}val-2017)",
    year = "2017",
    %address = "Vancouver, Canada",
    publisher = "Association for Computational Linguistics",
    url = "https://www.aclweb.org/anthology/S17-2171",
    doi = "10.18653/v1/S17-2171",
    pages = "978--984",
}

@inproceedings{DBLP:journals/corr/XuFHZ15,
  title = "Semantic Relation Classification via Convolutional Neural Networks with Simple Negative Sampling",
    author = "Xu, Kun  and
      Feng, Yansong  and
      Huang, Songfang  and
      Zhao, Dongyan",
    booktitle = "Proceedings of the 2015 Conference on Empirical Methods in Natural Language Processing",
    year = "2015",
    %address = "Lisbon, Portugal",
    publisher = "Association for Computational Linguistics",
    url = "https://www.aclweb.org/anthology/D15-1062",
    doi = "10.18653/v1/D15-1062",
    pages = "536--540",
}

@inproceedings{McD:Niv:Qui:13,
title = "Universal Dependency Annotation for Multilingual Parsing",
author = {McDonald, Ryan  and
  Nivre, Joakim  and
  Quirmbach-Brundage, Yvonne  and
  Goldberg, Yoav  and
  Das, Dipanjan  and
  Ganchev, Kuzman  and
  Hall, Keith  and
  Petrov, Slav  and
  Zhang, Hao  and
  T{\"a}ckstr{\"o}m, Oscar  and
  Bedini, Claudia  and
  Bertomeu Castell{\'o}, N{\'u}ria  and
  Lee, Jungmee},
booktitle = "Proceedings of the 51st Annual Meeting of the Association for Computational Linguistics (Volume 2: Short Papers)",
month = aug,
year = "2013",
address = "Sofia, Bulgaria",
publisher = "Association for Computational Linguistics",
url = "https://www.aclweb.org/anthology/P13-2017",
pages = "92--97",
}

@inproceedings{Niv:Mar:Gin:16,
    title = "Universal Dependencies v1: A Multilingual Treebank Collection",
    author = "Nivre, Joakim  and
      de Marneffe, Marie-Catherine  and
      Ginter, Filip  and
      Goldberg, Yoav  and
      Haji{\v{c}}, Jan  and
      Manning, Christopher D.  and
      McDonald, Ryan  and
      Petrov, Slav  and
      Pyysalo, Sampo  and
      Silveira, Natalia  and
      Tsarfaty, Reut  and
      Zeman, Daniel",
    booktitle = "Proceedings of the Tenth International Conference on Language Resources and Evaluation ({LREC}'16)",
    year = "2016",
    %address = "Portoro{\v{z}}, Slovenia",
    publisher = "European Language Resources Association (ELRA)",
    url = "https://www.aclweb.org/anthology/L16-1262",
    pages = "1659--1666",
  }

@inproceedings{Noo:Ovr:Lon:18,
  title = "{SIRIUS}-{LTG}-{U}i{O} at {S}em{E}val-2018 Task 7: Convolutional Neural Networks with Shortest Dependency Paths for Semantic Relation Extraction and Classification in Scientific Papers",
    author = "Nooralahzadeh, Farhad  and
      {\O}vrelid, Lilja  and
      L{\o}nning, Jan Tore",
    booktitle = "Proceedings of The 12th International Workshop on Semantic Evaluation",
    year = "2018",
    %address = "New Orleans, Louisiana",
    publisher = "Association for Computational Linguistics",
    url = "https://www.aclweb.org/anthology/S18-1128",
    doi = "10.18653/v1/S18-1128",
    pages = "805--810",
}

@inproceedings{Sch:Abe:Rap:12,
   title = "Learnability-Based Syntactic Annotation Design",
    author = "Schwartz, Roy  and
      Abend, Omri  and
      Rappoport, Ari",
    booktitle = "Proceedings of the 24th International Conference on Computational Linguistics (Coling 2012)",
    year = "2012",
    %address = "Mumbai, India",
    publisher = "The COLING 2012 Organizing Committee",
    url = "https://www.aclweb.org/anthology/C12-1147",
    pages = "2405--2422",

}

@inproceedings{de-marneffe-etal-2006-generating,
    title = "Generating Typed Dependency Parses from Phrase Structure Parses",
    author = "{de Marneffe}, Marie-Catherine  and
      MacCartney, Bill  and
      Manning, Christopher D.",
    booktitle = "Proceedings of the Fifth International Conference on Language Resources and Evaluation ({LREC}{'}06)",
    year = "2006",
    %address = "Genoa, Italy",
    publisher = "European Language Resources Association (ELRA)",
    url = "http://www.lrec-conf.org/proceedings/lrec2006/pdf/440_pdf.pdf",
}

@inproceedings{Miy:Sae:Sag:08,
   title = "Task-oriented Evaluation of Syntactic Parsers and Their Representations",
    author = "Miyao, Yusuke  and
      S{\ae}tre, Rune  and
      Sagae, Kenji  and
      Matsuzaki, Takuya  and
      Tsujii, Jun{'}ichi",
    booktitle = "Proceedings of the 46th Annual Meeting of the Association for Computational Linguistics: Human Language Technologies",
    year = "2008",
    %address = "Columbus, Ohio",
    publisher = "Association for Computational Linguistics",
    url = "https://www.aclweb.org/anthology/P08-1006",
    pages = "46--54",
}

@inproceedings{Oep:Ovr:Bjo:17,
  author = {Oepen, Stephan and {\O}vrelid, Lilja and
            Bj{\"o}rne, Jari and Johansson, Richard and Lapponi, Emanuele and
            Ginter, Filip and Velldal, Erik},
  title = {The 2017 {S}hared {T}ask on {E}xtrinsic {P}arser {E}valuation. {T}owards a Reusable Community Infrastructure},
  booktitle = {Proceedings of the 2017 Shared Task on Extrinsic Parser Evaluation at the Fourth International Conference on Dependency Linguistics and the 15th International Conference on Parsing Technologies},
  %address = {Pisa, Italy},
 publisher={Association for Computational Linguistics (ACL)},
  pages = {1--12},
  year = 2017
}

@inproceedings{Elm:Joh:Kle:13,
    title = "Down-stream effects of tree-to-dependency conversions",
    author = "Elming, Jakob  and
      Johannsen, Anders  and
      Klerke, Sigrid  and
      Lapponi, Emanuele  and
      Martinez Alonso, Hector  and
      S{\o}gaard, Anders",
    booktitle = "Proceedings of the 2013 Conference of the North {A}merican Chapter of the Association for Computational Linguistics: Human Language Technologies",
    year = "2013",
    %address = "Atlanta, Georgia",
    publisher = "Association for Computational Linguistics",
    url = "https://www.aclweb.org/anthology/N13-1070",
    pages = "617--626",
}

@inproceedings{Sechidis:2011:SMD:2034161.2034172,
 author = {Sechidis, Konstantinos and Tsoumakas, Grigorios and Vlahavas, Ioannis},
 title = {On the Stratification of Multi-label Data},
 booktitle = {Proceedings of the 2011 European Conference on Machine Learning and Knowledge Discovery in Databases - Volume Part III},
 series = {ECML PKDD'11},
 year = {2011},
 isbn = {978-3-642-23807-9},
 %location= {Athens, Greece},
 pages = {145--158},
 url = {http://dl.acm.org/citation.cfm?id=2034161.2034172},
 acmid = {2034172},
 publisher = {Springer-Verlag},
 %address = {Berlin, Heidelberg},
}

@inproceedings{SemEval2018Task7,
 title = "{S}em{E}val-2018 Task 7: Semantic Relation Extraction and Classification in Scientific Papers",
    author = {G{\'a}bor, Kata  and
      Buscaldi, Davide  and
      Schumann, Anne-Kathrin  and
      QasemiZadeh, Behrang  and
      Zargayouna, Ha{\"\i}fa  and
      Charnois, Thierry},
    booktitle = "Proceedings of The 12th International Workshop on Semantic Evaluation",
    year = "2018",
    %address = "New Orleans, Louisiana",
    publisher = "Association for Computational Linguistics",
    url = "https://www.aclweb.org/anthology/S18-1111",
    doi = "10.18653/v1/S18-1",
    pages = "679--688",
}

@inproceedings{Ebrahimi2015ChainBR,
    title = "Chain Based {RNN} for Relation Classification",
    author = "Ebrahimi, Javid  and
      Dou, Dejing",
    booktitle = "Proceedings of the 2015 Conference of the North {A}merican Chapter of the Association for Computational Linguistics: Human Language Technologies",

    year = "2015",
    %address = "Denver, Colorado",
    publisher = "Association for Computational Linguistics",
    url = "https://www.aclweb.org/anthology/N15-1133",
    doi = "10.3115/v1/N15-1133",
    pages = "1244--1249",
}

@inproceedings{Kok:Pot:17,
  author    = {Filippos Kokkinos and
               Alexandros Potamianos},
  title     = {Structural Attention Neural Networks for improved sentiment analysis},
  booktitle = {Proceedings of the 15th Conference of the European Chapter of the
               Association for Computational Linguistics, {EACL} 2017, Volume 2: Short Papers},
  pages     = {586--591},
  publisher = {Association for Computational Linguistics},
  year      = {2017},
  url       = {https://doi.org/10.18653/v1/e17-2093},
  doi       = {10.18653/v1/e17-2093},
  bibsource = {dblp computer science bibliography, https://dblp.org}
}

@article{DBLP:journals/corr/abs-1012-2599,
  author    = {Eric Brochu and
               Vlad M. Cora and
               Nando de Freitas},
  title     = {A Tutorial on Bayesian Optimization of Expensive Cost Functions, with
               Application to Active User Modeling and Hierarchical Reinforcement
               Learning},
  journal   = {CoRR},
  volume    = {abs/1012.2599},
  year      = {2010},
  url       = {http://arxiv.org/abs/1012.2599},
  archivePrefix = {arXiv},
  eprint    = {1012.2599},
  bibsource = {dblp computer science bibliography, https://dblp.org}
}

@inproceedings{DBLP:journals/corr/abs-1805-09927,
  title = "Robust Distant Supervision Relation Extraction via Deep Reinforcement Learning",
    author = "Qin, Pengda  and
      Xu, Weiran  and
      Wang, William Yang",
    booktitle = "Proceedings of the 56th Annual Meeting of the Association for Computational Linguistics (Volume 1: Long Papers)",
    year = "2018",
    %address = "Melbourne, Australia",
    publisher = "Association for Computational Linguistics",
    url = "https://www.aclweb.org/anthology/P18-1199",
    doi = "10.18653/v1/P18-1199",
    pages = "2137--2147",
}

@inproceedings{tsuboi-etal-2008-training,
    title = "Training Conditional Random Fields Using Incomplete Annotations",
    author = "Tsuboi, Yuta  and
      Kashima, Hisashi  and
      Mori, Shinsuke  and
      Oda, Hiroki  and
      Matsumoto, Yuji",
    booktitle = "Proceedings of the 22nd International Conference on Computational Linguistics (Coling 2008)",
    year = "2008",
    %address = "Manchester, UK",
    publisher = "Coling 2008 Organizing Committee",
    url = "https://www.aclweb.org/anthology/C08-1113",
    pages = "897--904",
}

@inproceedings{yang-etal-2018-distantly,
    title = "Distantly Supervised {NER} with Partial Annotation Learning and Reinforcement Learning",
    author = "Yang, Yaosheng  and
      Chen, Wenliang  and
      Li, Zhenghua  and
      He, Zhengqiu  and
      Zhang, Min",
    booktitle = "Proceedings of the 27th International Conference on Computational Linguistics",
    year = "2018",
    %address = "Santa Fe, New Mexico, USA",
    publisher = "Association for Computational Linguistics",
    url = "https://www.aclweb.org/anthology/C18-1183",
    pages = "2159--2169",
    }

@inproceedings{Sutton:1999:PGM:3009657.3009806,
author = {Sutton, Richard and McAllester, David and Singh, Satinder and Mansour, Yishay},
title = {Policy Gradient Methods for Reinforcement Learning with Function Approximation},
year = {1999},
publisher = {MIT Press},
%address = {Cambridge, MA, USA},
booktitle = {Proceedings of the 12th International Conference on Neural Information Processing Systems},
pages = {1057–1063},
numpages = {7},
%location= {Denver, CO},
series = {NIPS’99}
}

@inproceedings{DBLP:journals/corr/LampleBSKD16,
    title = "Neural Architectures for Named Entity Recognition",
    author = "Lample, Guillaume  and
      Ballesteros, Miguel  and
      Subramanian, Sandeep  and
      Kawakami, Kazuya  and
      Dyer, Chris",
    booktitle = "Proceedings of the 2016 Conference of the North {A}merican Chapter of the Association for Computational Linguistics: Human Language Technologies",
    year = "2016",
    %address = "San Diego, California",
    publisher = "Association for Computational Linguistics",
    url = "https://www.aclweb.org/anthology/N16-1030",
    doi = "10.18653/v1/N16-1030",
    pages = "260--270",
}

@article{Habibi2017DeepLW,
    author = {Habibi, Maryam and Weber, Leon and Neves, Mariana and Wiegandt, David Luis and Leser, Ulf},
    title = "{Deep learning with word embeddings improves biomedical named entity recognition}",
    journal = {Bioinformatics},
    volume = {33},
    number = {14},
    pages = {i37-i48},
    year = {2017},

    issn = {1367-4803},
    doi = {10.1093/bioinformatics/btx228},
    url = {https://doi.org/10.1093/bioinformatics/btx228},
}

@inproceedings{DBLP:journals/corr/abs-1709-04109,
  title={Empower sequence labeling with task-aware neural language model},
  author={Liu, Liyuan and Shang, Jingbo and Ren, Xiang and Xu, Frank Fangzheng and Gui, Huan and Peng, Jian and Han, Jiawei},
  booktitle={Thirty-Second AAAI Conference on Artificial Intelligence},
  Publisher = {Association for the Advancement of Artificial Intelligence (AAAI) Press},
  year={2018}
}

@article{Williams1992,
    author="Williams, Ronald J.",
    title="Simple statistical gradient-following algorithms for connectionist reinforcement learning",
    journal="Machine Learning",
    year="1992",
    day="01",
    volume="8",
    number="3",
    pages="229--256",
    issn="1573-0565",
    doi="10.1007/BF00992696",
    url="https://doi.org/10.1007/BF00992696"
}

@inproceedings{shang2018learning,
  title = "Learning Named Entity Tagger using Domain-Specific Dictionary",
    author = "Shang, Jingbo  and
      Liu, Liyuan  and
      Gu, Xiaotao  and
      Ren, Xiang  and
      Ren, Teng  and
      Han, Jiawei",
    booktitle = "Proceedings of the 2018 Conference on Empirical Methods in Natural Language Processing",
    year = "2018",
    %address = "Brussels, Belgium",
    publisher = "Association for Computational Linguistics",
    url = "https://www.aclweb.org/anthology/D18-1230",
    doi = "10.18653/v1/D18-1230",
    pages = "2054--2064",
}

@inproceedings{Pyysalo:2013b,
     author = {Pyysalo, S. and Ginter, F. and Moen, H. and Salakoski, T. and Ananiadou, S.},
      title = {Distributional Semantics Resources for Biomedical Text Processing},
  booktitle = {Proceedings of LBM 2013},
       year = {2013},
      pages = {39-44},
        url = {http://lbm2013.biopathway.org/lbm2013proceedings.pdf}
}

@article{DBLP:journals/corr/abs-1801-09851,
    title={Cross-type biomedical named entity recognition with deep multi-task learning},
    author={Wang, Xuan and Zhang, Yu and Ren, Xiang and Zhang, Yuhao and Zitnik, Marinka and Shang, Jingbo and Langlotz, Curtis and Han, Jiawei},
    journal={Bioinformatics},
    volume={35},
    number={10},
    pages={1745--1752},
    year={2019},
    publisher={Oxford University Press}
}

@inproceedings{DBLP:journals/corr/abs-1903-10676,
   title = "{S}ci{BERT}: A Pretrained Language Model for Scientific Text",
    author = "Beltagy, Iz  and
      Lo, Kyle  and
      Cohan, Arman",
    booktitle = "Proceedings of the 2019 Conference on Empirical Methods in Natural Language Processing and the 9th International Joint Conference on Natural Language Processing (EMNLP-IJCNLP)",
    year = "2019",
    %address = "Hong Kong, China",
    publisher = "Association for Computational Linguistics",
    url = "https://www.aclweb.org/anthology/D19-1371",
    doi = "10.18653/v1/D19-1371",
    pages = "3615--3620",
}

@inproceedings{ma-hovy-2016-end,
    title = "End-to-end Sequence Labeling via Bi-directional {LSTM}-{CNN}s-{CRF}",
    author = "Ma, Xuezhe  and
      Hovy, Eduard",
    booktitle = "Proceedings of the 54th Annual Meeting of the Association for Computational Linguistics (Volume 1: Long Papers)",
    year = "2016",
    %address = "Berlin, Germany",
    publisher = "Association for Computational Linguistics",
    url = "https://www.aclweb.org/anthology/P16-1101",
    doi = "10.18653/v1/P16-1101",
    pages = "1064--1074",
}

@article{shang2018automated,
  author    = {Jingbo Shang and
               Jialu Liu and
               Meng Jiang and
               Xiang Ren and
               Clare R. Voss and
               Jiawei Han},
  title     = {Automated Phrase Mining from Massive Text Corpora},
  journal   = {{IEEE} Transactions on Knowledge and Data Engineering},
  volume    = {30},
  number    = {10},
  pages     = {1825--1837},
  year      = {2018},
}

@article{DBLP:journals/corr/Fries0RR17,
  title     = {SwellShark: {A} Generative Model for Biomedical Named Entity Recognition
               without Labeled Data},
  author={Fries, Jason and Wu, Sen and Ratner, Alex and R{\'e}, Christopher},
  journal   = {CoRR},
  volume    = {abs/1704.06360},
  year      = {2017},
  url       = {http://arxiv.org/abs/1704.06360},
  archivePrefix = {arXiv},
  eprint    = {1704.06360},
  bibsource = {dblp computer science bibliography, https://dblp.org}
}

@inproceedings{DBLP:journals/corr/PassosKM14,
  title = "Lexicon Infused Phrase Embeddings for Named Entity Resolution",
    author = "Passos, Alexandre  and
      Kumar, Vineet  and
      McCallum, Andrew",
    booktitle = "Proceedings of the Eighteenth Conference on Computational Natural Language Learning",
    year = "2014",
    %address = "Ann Arbor, Michigan",
    publisher = "Association for Computational Linguistics",
    url = "https://www.aclweb.org/anthology/W14-1609",
    doi = "10.3115/v1/W14-1609",
    pages = "78--86",
}

@inproceedings{Ratinov:2009:DCM:1596374.1596399,
 author = {Ratinov, Lev and Roth, Dan},
 title = {Design Challenges and Misconceptions in Named Entity Recognition},
 booktitle = {Proceedings of the Thirteenth Conference on Computational Natural Language Learning},
 series = {CoNLL '09},
 year = {2009},
 isbn = {978-1-932432-29-9},
 %location= {Boulder, Colorado},
 pages = {147--155},
 numpages = {9},
 url = {http://dl.acm.org/citation.cfm?id=1596374.1596399},
 acmid = {1596399},
 publisher = {Association for Computational Linguistics},
 %address = {Stroudsburg, PA, USA},
}

@inproceedings{Mintz:2009:DSR:1690219.1690287,
 author = {Mintz, Mike and Bills, Steven and Snow, Rion and Jurafsky, Dan},
 title = {Distant Supervision for Relation Extraction Without Labeled Data},
 booktitle = {Proceedings of the Joint Conference of the 47th Annual Meeting of the ACL and the 4th International Joint Conference on Natural Language Processing of the AFNLP: Volume 2 - Volume 2},
 series = {ACL '09},
 year = {2009},
 isbn = {978-1-932432-46-6},
 %location= {Suntec, Singapore},
 pages = {1003--1011},
 numpages = {9},
 url = {http://dl.acm.org/citation.cfm?id=1690219.1690287},
 acmid = {1690287},
 publisher = {Association for Computational Linguistics},
 %address = {Stroudsburg, PA, USA},
}

@inproceedings{Riedel:2010:MRM:1889788.1889799,
 author = {Riedel, Sebastian and Yao, Limin and McCallum, Andrew},
 title = {Modeling Relations and Their Mentions Without Labeled Text},
 booktitle = {Proceedings of the 2010 European Conference on Machine Learning and Knowledge Discovery in Databases: Part III},
 series = {ECML PKDD'10},
 year = {2010},
 %location= {Barcelona, Spain},
 pages = {148--163},
 numpages = {16},
 url = {http://dl.acm.org/citation.cfm?id=1889788.1889799},
 acmid = {1889799},
 publisher = {Springer-Verlag},
 %address = {Berlin, Heidelberg},
}

@inproceedings{conf/kdd/RenEWTVH15,
  author = {Ren, Xiang and El-Kishky, Ahmed and Wang, Chi and Tao, Fangbo and Voss, Clare R. and Han, Jiawei},
  booktitle = {International Conference on Knowledge Discovery \& Data Mining},
  pages = {995-1004},
  publisher = {Association for Computing Machinery},
  title = {ClusType: Effective Entity Recognition and Typing by Relation Phrase-Based Clustering.},
  year = 2015
}

@inproceedings{DBLP:journals/corr/abs-1802-05365,
  title = "Deep Contextualized Word Representations",
    author = "Peters, Matthew  and
      Neumann, Mark  and
      Iyyer, Mohit  and
      Gardner, Matt  and
      Clark, Christopher  and
      Lee, Kenton  and
      Zettlemoyer, Luke",
    booktitle = "Proceedings of the 2018 Conference of the North {A}merican Chapter of the Association for Computational Linguistics: Human Language Technologies, Volume 1 (Long Papers)",
    year = "2018",
    %address = "New Orleans, Louisiana",
    publisher = "Association for Computational Linguistics",
    url = "https://www.aclweb.org/anthology/N18-1202",
    doi = "10.18653/v1/N18-1202",
    pages = "2227--2237",
}

@inproceedings{akbik-etal-2018-contextual,
    title = "Contextual String Embeddings for Sequence Labeling",
    author = "Akbik, Alan  and
      Blythe, Duncan  and
      Vollgraf, Roland",
    booktitle = "Proceedings of the 27th International Conference on Computational Linguistics",
    year = "2018",
    %address = "Santa Fe, New Mexico, USA",
    publisher = "Association for Computational Linguistics",
    url = "https://www.aclweb.org/anthology/C18-1139",
    pages = "1638--1649",
    abstract = "Recent advances in language modeling using recurrent neural networks have made it viable to model language as distributions over characters. By learning to predict the next character on the basis of previous characters, such models have been shown to automatically internalize linguistic concepts such as words, sentences, subclauses and even sentiment. In this paper, we propose to leverage the internal states of a trained character language model to produce a novel type of word embedding which we refer to as contextual string embeddings. Our proposed embeddings have the distinct properties that they (a) are trained without any explicit notion of words and thus fundamentally model words as sequences of characters, and (b) are contextualized by their surrounding text, meaning that the same word will have different embeddings depending on its contextual use. We conduct a comparative evaluation against previous embeddings and find that our embeddings are highly useful for downstream tasks: across four classic sequence labeling tasks we consistently outperform the previous state-of-the-art. In particular, we significantly outperform previous work on English and German named entity recognition (NER), allowing us to report new state-of-the-art F1-scores on the CoNLL03 shared task. We release all code and pre-trained language models in a simple-to-use framework to the research community, to enable reproduction of these experiments and application of our proposed embeddings to other tasks: https://github.com/zalandoresearch/flair",
}

@inproceedings{DBLP:conf/ekaw/AugensteinMC14,
    author="Augenstein, Isabelle
        and Maynard, Diana
        and Ciravegna, Fabio",
    title="Relation Extraction from the Web Using Distant Supervision",
    booktitle="Knowledge Engineering and Knowledge Management",
    year="2014",
    publisher="Springer International Publishing",
    %address="Cham",
    pages="26--41"
}

@inproceedings{DBLP:journals/corr/abs-1808-08013,
  title={Reinforcement learning for relation classification from noisy data},
  author={Feng, Jun and Huang, Minlie and Zhao, Li and Yang, Yang and Zhu, Xiaoyan},
  booktitle={Proceedings of the Thirty-Second AAAI Conference on Artificial Intelligence},
  publisher = {Association for the Advancement of Artificial Intelligence ({AAAI}) Press},
  year={2018}
}

@book{Goodfellow-et-al-2016,
    title={Deep Learning},
    author={Ian Goodfellow and Yoshua Bengio and Aaron Courville},
    publisher={MIT Press, Cambridge, MA, United States},
    year={2016},
}

@inproceedings{pontiki-etal-2014-semeval,
    title = "{S}em{E}val-2014 Task 4: Aspect Based Sentiment Analysis",
    author = "Pontiki, Maria  and
      Galanis, Dimitris  and
      Pavlopoulos, John  and
      Papageorgiou, Harris  and
      Androutsopoulos, Ion  and
      Manandhar, Suresh",
    booktitle = "Proceedings of the 8th International Workshop on Semantic Evaluation ({S}em{E}val 2014)",
    year = "2014",
    %address = "Dublin, Ireland",
    publisher = "Association for Computational Linguistics",
    url = "https://www.aclweb.org/anthology/S14-2004",
    doi = "10.3115/v1/S14-2004",
    pages = "27--35",
}

@inproceedings{Wang:2011:LAR:2020408.2020505,
 author = {Hongning Wang and Yue Lu and ChengXiang Zhai},
 title = {Latent Aspect Rating Analysis Without Aspect Keyword Supervision},
 booktitle = {Proceedings of the 17th ACM SIGKDD International Conference on Knowledge Discovery and Data Mining},
 series = {KDD '11},
 year = {2011},
 isbn = {978-1-4503-0813-7},
 %location= {San Diego, California, USA},
 pages = {618--626},
 numpages = {9},
 url = {http://doi.acm.org/10.1145/2020408.2020505},
 doi = {10.1145/2020408.2020505},
 acmid = {2020505},
 publisher = {Association for Computing Machinery},
 %address = {New York, NY, USA},
 keywords = {aspect identification, latent rating analysis, review mining},
}

@inproceedings{giannakopoulos-etal-2017-unsupervised,
    title = "Unsupervised Aspect Term Extraction with B-{LSTM} \& {CRF} using Automatically Labelled Datasets",
    author = "Giannakopoulos, Athanasios  and
      Musat, Claudiu  and
      Hossmann, Andreea  and
      Baeriswyl, Michael",
    booktitle = "Proceedings of the 8th Workshop on Computational Approaches to Subjectivity, Sentiment and Social Media Analysis",
    year = "2017",
    %address = "Copenhagen, Denmark",
    publisher = "Association for Computational Linguistics",
    url = "https://www.aclweb.org/anthology/W17-5224",
    doi = "10.18653/v1/W17-5224",
    pages = "180--188",
}

@inproceedings{chinchor98,
  added-at = {2007-12-12T23:57:17.000+0100},
  %address = {Fairfax, VA},
  author = {Chinchor, Nancy A.},
  biburl = {https://www.bibsonomy.org/bibtex/2774fb97f2a81eb6557254ec6a0fba7f3/deynard},
  booktitle = {Proceedings of the Seventh Message Understanding Conference (MUC-7)},
  interhash = {a610982840fd68df3da74d6c1937f783},
  intrahash = {774fb97f2a81eb6557254ec6a0fba7f3},
  keywords = {annotation semantic},
  note = {version 3.5, {\footnotesize http://www.itl.nist.gov/iaui/894.02/related\_projects/muc/}},
  pages = {21 pages},
  timestamp = {2007-12-12T23:57:17.000+0100},
  title = {{{Proceedings of the Seventh Message Understanding Conference (MUC-7)}} Named Entity Task Definition},
  url = {http://acl.ldc.upenn.edu/muc7/ne_task.html},
  year = {1998},
}

@article{10.1093/database/baw068,
    author = {Li, Jiao and Sun, Yueping and Johnson, Robin J. and Sciaky, Daniela and Wei, Chih-Hsuan and Leaman, Robert and Davis, Allan Peter and Mattingly, Carolyn J. and Wiegers, Thomas C. and Lu, Zhiyong},
    title = "{BioCreative V CDR task corpus: a resource for chemical disease relation extraction}",
    journal = {Database},
    volume = {2016},
    year = {2016},

    issn = {1758-0463},
    doi = {10.1093/database/baw068},
    url = {https://doi.org/10.1093/database/baw068},
    note = {baw068},
    eprint = {http://oup.prod.sis.lan/database/article-pdf/doi/10.1093/database/baw068/8224483/baw068.pdf},
}

@article{DBLP:journals/corr/abs-1812-09449,
   title={A Survey on Deep Learning for Named Entity Recognition},
   ISSN={2326-3865},
   url={http://dx.doi.org/10.1109/tkde.2020.2981314},
   DOI={10.1109/tkde.2020.2981314},
   journal={IEEE Transactions on Knowledge and Data Engineering},
   publisher={Institute of Electrical and Electronics Engineers (IEEE)},
   author={Li, Jing and Sun, Aixin and Han, Jianglei and Li, Chenliang},
   volume={},
   number={},
   year={2020},
   pages={1–1}
}

@inproceedings{DBLP:journals/corr/abs-1810-04805,
    title = "{BERT}: Pre-training of Deep Bidirectional Transformers for Language Understanding",
    author = "Devlin, Jacob  and
      Chang, Ming-Wei  and
      Lee, Kenton  and
      Toutanova, Kristina",
    booktitle = "Proceedings of the 2019 Conference of the North {A}merican Chapter of the Association for Computational Linguistics: Human Language Technologies, Volume 1 (Long and Short Papers)",
    year = "2019",
    %address = "Minneapolis, Minnesota",
    publisher = "Association for Computational Linguistics",
    url = "https://www.aclweb.org/anthology/N19-1423",
    doi = "10.18653/v1/N19-1423",
    pages = "4171--4186",

}

@inproceedings{Lafferty:2001:CRF:645530.655813,
     author = {Lafferty, John D. and McCallum, Andrew and Pereira, Fernando C. N.},
     title = {Conditional Random Fields: Probabilistic Models for Segmenting and Labeling Sequence Data},
     booktitle = {Proceedings of the Eighteenth International Conference on Machine Learning},
     series = {ICML '01},
     year = {2001},
     isbn = {1-55860-778-1},
     pages = {282--289},
     numpages = {8},
     url = {http://dl.acm.org/citation.cfm?id=645530.655813},
     acmid = {655813},
     publisher = {Morgan Kaufmann Publishers Inc.},
     %address = {San Francisco, CA, USA},
}

@article{Hochreiter:1997:LSM:1246443.1246450,
     author = {Hochreiter, Sepp and Schmidhuber, J\"{u}rgen},
     title = {Long Short-Term Memory},
     journal = {Neural Computation},
     issue_date = {November 15, 1997},
     volume = {9},
     number = {8},
     year = {1997},
     issn = {0899-7667},
     pages = {1735--1780},
     numpages = {46},
     url = {http://dx.doi.org/10.1162/neco.1997.9.8.1735},
     doi = {10.1162/neco.1997.9.8.1735},
     acmid = {1246450},
     publisher = {MIT Press},
     %address = {Cambridge, MA, USA},
}

@article{schuster1997bidirectional,
  acmid = {2205129},
  added-at = {2018-04-26T20:41:11.000+0200},
  %address = {Piscataway, NJ, USA},
  author = {Schuster, M. and Paliwal, K.K.},
  biburl = {https://www.bibsonomy.org/bibtex/2dce8863d36658307fb4d3c2b84c1683a/nosebrain},
  description = {Bidirectional recurrent neural networks},
  doi = {10.1109/78.650093},
  issn = {1053-587X},
  issue_date = {November 1997},
  journal = {IEEE Transactions on Signal Processing},
  number = {11},
  pages = {2673--2681},
  publisher = {Institute of Electrical and Electronics Engineers (IEEE)},
  timestamp = {2018-04-26T20:41:11.000+0200},
  title = {Bidirectional Recurrent Neural Networks},
  url = {http://dx.doi.org/10.1109/78.650093},
  volume = {45},
  year = {1997},
}

@inproceedings{ling-etal-2015-finding,
    title = "Finding Function in Form: Compositional Character Models for Open Vocabulary Word Representation",
    author = "Ling, Wang  and
      Dyer, Chris  and
      Black, Alan W  and
      Trancoso, Isabel  and
      Fermandez, Ram{\'o}n  and
      Amir, Silvio  and
      Marujo, Lu{\'\i}s  and
      Lu{\'\i}s, Tiago",
    booktitle = "Proceedings of the 2015 Conference on Empirical Methods in Natural Language Processing",
    year = "2015",
    %address = "Lisbon, Portugal",
    publisher = "Association for Computational Linguistics",
    doi = "10.18653/v1/D15-1176",
    pages = "1520--1530",
}

@inproceedings{conneau2019cross,
  title = {Cross-lingual Language Model Pretraining},
    author = {Conneau, Alexis and Lample, Guillaume},
    booktitle = {Advances in Neural Information Processing Systems 32},
    pages = {7059--7069},
    year = {2019},
    publisher = {Curran Associates, Inc.},
    url = {http://papers.nips.cc/paper/8928-cross-lingual-language-model-pretraining.pdf}
}

@book{wals,
  editor    = {Matthew S. Dryer and Martin Haspelmath},
  publisher = {Max Planck Institute for Evolutionary Anthropology, Leipzig},
  title     = {{WALS Online}},
  url       = {http://wals.info/},
  year      = {2013},
}

@inproceedings{bjerva-augenstein-2018-phonology,
    title = "From Phonology to Syntax: Unsupervised Linguistic Typology at Different Levels with Language Embeddings",
    author = "Bjerva, Johannes  and
      Augenstein, Isabelle",
    booktitle = "Proceedings of the 2018 Conference of the North {A}merican Chapter of the Association for Computational Linguistics: Human Language Technologies, Volume 1 (Long Papers)",
    year = "2018",
    %address = "New Orleans, Louisiana",
    publisher = "Association for Computational Linguistics",
    url = "https://www.aclweb.org/anthology/N18-1083",
    doi = "10.18653/v1/N18-1083",
    pages = "907--916",
}

@inproceedings{banko2007open,
  title = "{T}ext{R}unner: Open Information Extraction on the Web",
    author = "Yates, Alexander  and
      Banko, Michele  and
      Broadhead, Matthew  and
      Cafarella, Michael  and
      Etzioni, Oren  and
      Soderland, Stephen",
    booktitle = "Proceedings of Human Language Technologies: The Annual Conference of the North {A}merican Chapter of the Association for Computational Linguistics ({NAACL}-{HLT})",
    year = "2007",
    %address = "Rochester, New York, USA",
    publisher = "Association for Computational Linguistics",
    url = "https://www.aclweb.org/anthology/N07-4013",
    pages = "25--26",
}

@inproceedings{faruqui2015multilingual,
  author = 	"Faruqui, Manaal
		and Kumar, Shankar",
  title = 	"{Multilingual Open Relation Extraction Using Cross-lingual Projection}",
  booktitle = "Proceedings of the 2015 Conference of the North American Chapter of the  Association for Computational Linguistics: Human Language Technologies",
  year = 	"2015",
  publisher = 	"Association for Computational Linguistics",
  pages = 	"1351--1356",
  %location= 	"Denver, Colorado",
  doi = 	"10.3115/v1/N15-1151",
  url = 	"http://aclweb.org/anthology/N15-1151",
}

@inproceedings{verga2015multilingual,
  author = 	"Verga, Patrick
		and Belanger, David
		and Strubell, Emma
		and Roth, Benjamin
		and McCallum, Andrew",
  title = 	{{Multilingual Relation Extraction using Compositional Universal Schema}},
  booktitle = 	"Proceedings of the 2016 Conference of the North American Chapter of the      Association for Computational Linguistics: Human Language Technologies    ",
  year = 	"2016",
  publisher = 	"Association for Computational Linguistics",
  pages = 	"886--896",
  %location= 	"San Diego, California",
  doi = 	"10.18653/v1/N16-1103",
  url = 	"http://aclweb.org/anthology/N16-1103"
}

@inproceedings{conneau2018xnli,
  author = 	"Conneau, Alexis
		and Rinott, Ruty
		and Lample, Guillaume
		and Williams, Adina
		and Bowman, Samuel
		and Schwenk, Holger
		and Stoyanov, Veselin",
  title = 	{{XNLI: Evaluating Cross-lingual Sentence Representations}},
  booktitle = 	"Proceedings of the 2018 Conference on Empirical Methods in Natural Language Processing",
  year = 	"2018",
  publisher = 	"Association for Computational Linguistics",
  pages = 	"2475--2485",
  %location= 	"Brussels, Belgium",
  url = 	"http://aclweb.org/anthology/D18-1269"
}

@inproceedings{agic2017baselines,
    title = "Baselines and Test Data for Cross-Lingual Inference",
    author = "Agi{\'c}, {\v{Z}}eljko  and
      Schluter, Natalie",
    booktitle = "Proceedings of the Eleventh International Conference on Language Resources and Evaluation ({LREC} 2018)",
    year = "2018",
    %address = "Miyazaki, Japan",
    publisher = "European Language Resources Association (ELRA)",
    url = "https://www.aclweb.org/anthology/L18-1614",
}

@inproceedings{plank2016multilingual,
  author    = {Barbara Plank and
               Anders S{\o}gaard and
               Yoav Goldberg},
  title     = {Multilingual Part-of-Speech Tagging with Bidirectional Long Short-Term
               Memory Models and Auxiliary Loss},
  booktitle = {Proceedings of the 54th Annual Meeting of the Association for Computational
               Linguistics, {ACL} 2016, Volume 2: Short Papers},
  publisher = {The Association for Computer Linguistics},
  year      = {2016},
  url       = {https://doi.org/10.18653/v1/p16-2067},
  doi       = {10.18653/v1/p16-2067},
  bibsource = {dblp computer science bibliography, https://dblp.org}
}

@inproceedings{sutton1998reinforcement,
  title={Reinforcement learning: An introduction},
  author={Sutton, Richard and Barto, Andrew G and Bach, Francis and others},
  year={1998},
  publisher={MIT press},
  address={Cambridge, MA, United States},
  }

@inproceedings{pennington2014glove,
    title = "{G}love: Global Vectors for Word Representation",
    author={Pennington, Jeffrey and Socher, Richard and Manning, Christopher},
    booktitle = "Proceedings of the 2014 Conference on Empirical Methods in Natural Language Processing ({EMNLP})",
    year = "2014",
    %address = "Doha, Qatar",
    publisher = "Association for Computational Linguistics",
    url = "https://www.aclweb.org/anthology/D14-1162",
    doi = "10.3115/v1/D14-1162",
    pages = "1532--1543",
}

@inproceedings{dou-etal-2019-investigating,
    title = "Investigating Meta-Learning Algorithms for Low-Resource Natural Language Understanding Tasks",
    author = "Dou, Zi-Yi  and
      Yu, Keyi  and
      Anastasopoulos, Antonios",
    booktitle = "Proceedings of the 2019 Conference on Empirical Methods in Natural Language Processing and the 9th International Joint Conference on Natural Language Processing (EMNLP-IJCNLP)",
    year = "2019",
    %address = "Hong Kong, China",
    publisher = "Association for Computational Linguistics",
    url = "https://www.aclweb.org/anthology/D19-1112",
    doi = "10.18653/v1/D19-1112",
    pages = "1192--1197",
    abstract = "Learning general representations of text is a fundamental problem for many natural language understanding (NLU) tasks. Previously, researchers have proposed to use language model pre-training and multi-task learning to learn robust representations. However, these methods can achieve sub-optimal performance in low-resource scenarios. Inspired by the recent success of optimization-based meta-learning algorithms, in this paper, we explore the model-agnostic meta-learning algorithm (MAML) and its variants for low-resource NLU tasks. We validate our methods on the GLUE benchmark and show that our proposed models can outperform several strong baselines. We further empirically demonstrate that the learned representations can be adapted to new tasks efficiently and effectively.",
}

@inproceedings{abdou-etal-2019-x,
    title = {{X-{W}iki{RE}: A Large, Multilingual Resource for Relation Extraction as Machine Comprehension}},
    author = "Abdou, Mostafa  and
      Sas, Cezar  and
      Aralikatte, Rahul  and
      Augenstein, Isabelle  and
      S{\o}gaard, Anders",
    booktitle = "Proceedings of the 2nd Workshop on Deep Learning Approaches for Low-Resource NLP (DeepLo 2019)",
    year = "2019",
    %address = "Hong Kong, China",
    publisher = "Association for Computational Linguistics",
    url = "https://www.aclweb.org/anthology/D19-6130",
    doi = "10.18653/v1/D19-6130",
    pages = "265--274",
    abstract = "Although the vast majority of knowledge bases (KBs) are heavily biased towards English, Wikipedias do cover very different topics in different languages. Exploiting this, we introduce a new multilingual dataset (X-WikiRE), framing relation extraction as a multilingual machine reading problem. We show that by leveraging this resource it is possible to robustly transfer models cross-lingually and that multilingual support significantly improves (zero-shot) relation extraction, enabling the population of low-resourced KBs from their well-populated counterparts.",
}

@article{DBLP:journals/corr/abs-1804-07461,
  author    = {Alex Wang and
               Amanpreet Singh and
               Julian Michael and
               Felix Hill and
               Omer Levy and
               Samuel R. Bowman},
  title     = {{GLUE:} {A} Multi-Task Benchmark and Analysis Platform for Natural
               Language Understanding},
  journal   = {CoRR},
  volume    = {abs/1804.07461},
  year      = {2018},
  url       = {http://arxiv.org/abs/1804.07461},
  archivePrefix = {arXiv},
  eprint    = {1804.07461},
  bibsource = {dblp computer science bibliography, https://dblp.org}
}

@inproceedings{gu-etal-2018-meta,
    title = "Meta-Learning for Low-Resource Neural Machine Translation",
    author = "Gu, Jiatao  and
      Wang, Yong  and
      Chen, Yun  and
      Li, Victor O. K.  and
      Cho, Kyunghyun",
    booktitle = "Proceedings of the 2018 Conference on Empirical Methods in Natural Language Processing",
    year = "2018",
    %address = "Brussels, Belgium",
    publisher = "Association for Computational Linguistics",
    url = "https://www.aclweb.org/anthology/D18-1398",
    doi = "10.18653/v1/D18-1398",
    pages = "3622--3631",
    abstract = "In this paper, we propose to extend the recently introduced model-agnostic meta-learning algorithm (MAML, Finn, et al., 2017) for low-resource neural machine translation (NMT). We frame low-resource translation as a meta-learning problem where we learn to adapt to low-resource languages based on multilingual high-resource language tasks. We use the universal lexical representation (Gu et al., 2018b) to overcome the input-output mismatch across different languages. We evaluate the proposed meta-learning strategy using eighteen European languages (Bg, Cs, Da, De, El, Es, Et, Fr, Hu, It, Lt, Nl, Pl, Pt, Sk, Sl, Sv and Ru) as source tasks and five diverse languages (Ro,Lv, Fi, Tr and Ko) as target tasks. We show that the proposed approach significantly outperforms the multilingual, transfer learning based approach (Zoph et al., 2016) and enables us to train a competitive NMT system with only a fraction of training examples. For instance, the proposed approach can achieve as high as 22.04 BLEU on Romanian-English WMT{'}16 by seeing only 16,000 translated words ({\textasciitilde}600 parallel sentences)",
}

@inproceedings{DBLP:journals/corr/abs-1904-09077,
    title = "Beto, Bentz, Becas: The Surprising Cross-Lingual Effectiveness of {BERT}",
    author = "Wu, Shijie  and
      Dredze, Mark",
    booktitle = "Proceedings of the 2019 Conference on Empirical Methods in Natural Language Processing and the 9th International Joint Conference on Natural Language Processing (EMNLP-IJCNLP)",
    year = "2019",
    %address = "Hong Kong, China",
    publisher = "Association for Computational Linguistics",
    url = "https://www.aclweb.org/anthology/D19-1077",
    doi = "10.18653/v1/D19-1077",
    pages = "833--844",
}

@inproceedings{williams-etal-2018-broad,
    title = "A Broad-Coverage Challenge Corpus for Sentence Understanding through Inference",
    author = "Williams, Adina  and
      Nangia, Nikita  and
      Bowman, Samuel",
    booktitle = "Proceedings of the 2018 Conference of the North {A}merican Chapter of the Association for Computational Linguistics: Human Language Technologies, Volume 1 (Long Papers)",
    year = "2018",
    %address = "New Orleans, Louisiana",
    publisher = "Association for Computational Linguistics",
    url = "https://www.aclweb.org/anthology/N18-1101",
    doi = "10.18653/v1/N18-1101",
    pages = "1112--1122",
    abstract = "This paper introduces the Multi-Genre Natural Language Inference (MultiNLI) corpus, a dataset designed for use in the development and evaluation of machine learning models for sentence understanding. At 433k examples, this resource is one of the largest corpora available for natural language inference (a.k.a. recognizing textual entailment), improving upon available resources in both its coverage and difficulty. MultiNLI accomplishes this by offering data from ten distinct genres of written and spoken English, making it possible to evaluate systems on nearly the full complexity of the language, while supplying an explicit setting for evaluating cross-genre domain adaptation. In addition, an evaluation using existing machine learning models designed for the Stanford NLI corpus shows that it represents a substantially more difficult task than does that corpus, despite the two showing similar levels of inter-annotator agreement.",
}

@misc{Adam,
  author = {Kingma, Diederik P. and Ba, Jimmy},
  description = {Adam: A Method for Stochastic Optimization},
  title = {Adam: A Method for Stochastic Optimization},
  url = {http://arxiv.org/abs/1412.6980},
  year = 2014
}

@inproceedings{vinyals:16,
  author    = {Oriol Vinyals and
               Charles Blundell and
               Tim Lillicrap and
               Koray Kavukcuoglu and
               Daan Wierstra},

  title     = {Matching Networks for One Shot Learning},
  booktitle = {Advances in Neural Information Processing Systems 29: Annual Conference
               on Neural Information Processing Systems 2016},
  pages     = {3630--3638},
  year      = {2016},
  url       = {http://papers.nips.cc/paper/6385-matching-networks-for-one-shot-learning},
  bibsource = {dblp computer science bibliography, https://dblp.org}
}

@inproceedings{ravi:17,
  author    = {Sachin Ravi and
               Hugo Larochelle},
  title     = {Optimization as a Model for Few-Shot Learning},
  booktitle = {5th International Conference on Learning Representations, {ICLR} 2017, Conference Track Proceedings},
  publisher = {OpenReview.net},
  year      = {2017},
  url       = {https://openreview.net/forum?id=rJY0-Kcll},
  bibsource = {dblp computer science bibliography, https://dblp.org}
}

@inproceedings{finn:17,
  author    = {Chelsea Finn and
               Pieter Abbeel and
               Sergey Levine},
  editor    = {Doina Precup and
               Yee Whye Teh},
  title     = {Model-Agnostic Meta-Learning for Fast Adaptation of Deep Networks},
  booktitle = {Proceedings of the 34th International Conference on Machine Learning,
               {ICML} 2017},
  series    = {Proceedings of Machine Learning Research},
  volume    = {70},
  pages     = {1126--1135},
  publisher = {{PMLR}},
  year      = {2017},
  url       = {http://proceedings.mlr.press/v70/finn17a.html},
  bibsource = {dblp computer science bibliography, https://dblp.org}
}

@inproceedings{han:18,
  author    = {Xu Han and
               Hao Zhu and
               Pengfei Yu and
               Ziyun Wang and
               Yuan Yao and
               Zhiyuan Liu and
               Maosong Sun},
  title     = {FewRel: {A} Large-Scale Supervised Few-shot Relation Classification
               Dataset with State-of-the-Art Evaluation},
  booktitle = {Proceedings of the 2018 Conference on Empirical Methods in Natural
               Language Processing, Brussels, Belgium, October 31 - November 4, 2018},
  pages     = {4803--4809},
  publisher = {Association for Computational Linguistics},
  year      = {2018},
  url       = {https://doi.org/10.18653/v1/d18-1514},
  doi       = {10.18653/v1/d18-1514},
  bibsource = {dblp computer science bibliography, https://dblp.org}
}

@inproceedings{yu:18,
  title = "Diverse Few-Shot Text Classification with Multiple Metrics",
    author = "Yu, Mo  and
      Guo, Xiaoxiao  and
      Yi, Jinfeng  and
      Chang, Shiyu  and
      Potdar, Saloni  and
      Cheng, Yu  and
      Tesauro, Gerald  and
      Wang, Haoyu  and
      Zhou, Bowen",
    booktitle = "Proceedings of the 2018 Conference of the North {A}merican Chapter of the Association for Computational Linguistics: Human Language Technologies, Volume 1 (Long Papers)",
    year = "2018",
    %address = "New Orleans, Louisiana",
    publisher = "Association for Computational Linguistics",
    url = "https://www.aclweb.org/anthology/N18-1109",
    doi = "10.18653/v1/N18-1109",
    pages = "1206--1215",
}

@inproceedings{andrychowicz:16,
  author = {Andrychowicz, Marcin and Denil, Misha and Colmenarejo, Sergio G\'{o}mez and Hoffman, Matthew W. and Pfau, David and Schaul, Tom and Shillingford, Brendan and de Freitas, Nando},
    title = {Learning to Learn by Gradient Descent by Gradient Descent},
    year = {2016},
    isbn = {9781510838819},
    publisher = {Curran Associates Inc.},
    %address = {Red Hook, NY, USA},
    booktitle = {Proceedings of the 30th International Conference on Neural Information Processing Systems},
    pages = {3988–3996},
    numpages = {9},
    %location= {Barcelona, Spain},
    series = {NIPS’16}
}

@inproceedings{santoro:16,
  author    = {Adam Santoro and
               Sergey Bartunov and
               Matthew Botvinick and
               Daan Wierstra and
               Timothy P. Lillicrap},

  title     = {Meta-Learning with Memory-Augmented Neural Networks},
  booktitle = {Proceedings of the 33nd International Conference on Machine Learning,
               {ICML} 2016},
  series    = {{JMLR} Workshop and Conference Proceedings},
  volume    = {48},
  pages     = {1842--1850},
  publisher = {JMLR.org},
  year      = {2016},
  url       = {http://proceedings.mlr.press/v48/santoro16.html},
  bibsource = {dblp computer science bibliography, https://dblp.org}
}

@inproceedings{koch:15,
  title={Siamese neural networks for one-shot image recognition},
  author={Koch, Gregory and Zemel, Richard and Salakhutdinov, Ruslan},
  booktitle={Deep Learning Workshop at the International Conference on Machine Learning (ICML)},
  volume={2},
  year={2015}
}

@article{nichol:18,
  author    = {Alex Nichol and
               Joshua Achiam and
               John Schulman},
  title     = {On First-Order Meta-Learning Algorithms},
  journal   = {CoRR},
  volume    = {abs/1803.02999},
  year      = {2018},
  url       = {http://arxiv.org/abs/1803.02999},
  archivePrefix = {arXiv},
  eprint    = {1803.02999},
  bibsource = {dblp computer science bibliography, https://dblp.org}
}

@misc{ethnologue:19,
      author={Eberhard, David M. and Gary F. Simons, and Charles D. Fennig},
      title = {Ethnologue: Languages of the World},
      year={2019},
      howpublished = {\url{https://www.ethnologue.com/statistics/size}},
      note = {Accessed: 2019-05-25}
}

@inproceedings{qian-yu-2019-domain,
    title = "Domain Adaptive Dialog Generation via Meta Learning",
    author = "Qian, Kun  and
      Yu, Zhou",
    booktitle = "Proceedings of the 57th Annual Meeting of the Association for Computational Linguistics",
    year = "2019",
    %address = "Florence, Italy",
    publisher = "Association for Computational Linguistics",
    url = "https://www.aclweb.org/anthology/P19-1253",
    doi = "10.18653/v1/P19-1253",
    pages = "2639--2649",
}

@inproceedings{levy-etal-2017-zero,
    title = "Zero-Shot Relation Extraction via Reading Comprehension",
    author = "Levy, Omer  and
      Seo, Minjoon  and
      Choi, Eunsol  and
      Zettlemoyer, Luke",
    booktitle = "Proceedings of the 21st Conference on Computational Natural Language Learning ({C}o{NLL} 2017)",
    year = "2017",
    %address = "Vancouver, Canada",
    publisher = "Association for Computational Linguistics",
    url = "https://www.aclweb.org/anthology/K17-1034",
    doi = "10.18653/v1/K17-1034",
    pages = "333--342",
    abstract = "We show that relation extraction can be reduced to answering simple reading comprehension questions, by associating one or more natural-language questions with each relation slot. This reduction has several advantages: we can (1) learn relation-extraction models by extending recent neural reading-comprehension techniques, (2) build very large training sets for those models by combining relation-specific crowd-sourced questions with distant supervision, and even (3) do zero-shot learning by extracting new relation types that are only specified at test-time, for which we have no labeled training examples. Experiments on a Wikipedia slot-filling task demonstrate that the approach can generalize to new questions for known relation types with high accuracy, and that zero-shot generalization to unseen relation types is possible, at lower accuracy levels, setting the bar for future work on this task.",
}

@misc{bender2019benderrule,
  title={{The\# BenderRule: On Naming the Languages We Study and Why It Matters}},
  author={Bender, Emily M},
  year={2019},
  howpublished={\url{https://thegradient.pub/the-benderrule-on-naming-the-languages-we-study-and-why-it-matters/}},
}

@inproceedings{DBLP:journals/corr/ChenZLWJ16,
    title = "Enhanced {LSTM} for Natural Language Inference",
    author = "Chen, Qian  and
      Zhu, Xiaodan  and
      Ling, Zhen-Hua  and
      Wei, Si  and
      Jiang, Hui  and
      Inkpen, Diana",
    booktitle = "Proceedings of the 55th Annual Meeting of the Association for Computational Linguistics (Volume 1: Long Papers)",
    year = "2017",
    %address = "Vancouver, Canada",
    publisher = "Association for Computational Linguistics",
    url = "https://www.aclweb.org/anthology/P17-1152",
    doi = "10.18653/v1/P17-1152",
    pages = "1657--1668",
}

@inproceedings{liu-etal-2018-efficient-contextualized,
    title = "Efficient Contextualized Representation: Language Model Pruning for Sequence Labeling",
    author = "Liu, Liyuan  and
      Ren, Xiang  and
      Shang, Jingbo  and
      Gu, Xiaotao  and
      Peng, Jian  and
      Han, Jiawei",
    booktitle = "Proceedings of the 2018 Conference on Empirical Methods in Natural Language Processing",
    year = "2018",
    %address = "Brussels, Belgium",
    publisher = "Association for Computational Linguistics",
    url = "https://www.aclweb.org/anthology/D18-1153",
    doi = "10.18653/v1/D18-1153",
    pages = "1215--1225",
    abstract = "Many efforts have been made to facilitate natural language processing tasks with pre-trained language models (LMs), and brought significant improvements to various applications. To fully leverage the nearly unlimited corpora and capture linguistic information of multifarious levels, large-size LMs are required; but for a specific task, only parts of these information are useful. Such large-sized LMs, even in the inference stage, may cause heavy computation workloads, making them too time-consuming for large-scale applications. Here we propose to compress bulky LMs while preserving useful information with regard to a specific task. As different layers of the model keep different information, we develop a layer selection method for model pruning using sparsity-inducing regularization. By introducing the dense connectivity, we can detach any layer without affecting others, and stretch shallow and wide LMs to be deep and narrow. In model training, LMs are learned with layer-wise dropouts for better robustness. Experiments on two benchmark datasets demonstrate the effectiveness of our method.",
}

@article{PanY09TKDE,
  author = {Pan, S.J. and Yang, Q.},
  journal = {IEEE Transactions on Knowledge and Data Engineering},
  number = {10},
  pages = {1345--1359},
  title = {{A Survey on Transfer Learning}},
  volume = {22},
  year = {2010},
}

@inproceedings{howard-ruder-2018-universal,
    title = "Universal Language Model Fine-tuning for Text Classification",
    author = "Howard, Jeremy  and
      Ruder, Sebastian",
    booktitle = "Proceedings of the 56th Annual Meeting of the Association for Computational Linguistics (Volume 1: Long Papers)",
    year = "2018",
    %address = "Melbourne, Australia",
    publisher = "Association for Computational Linguistics",
    url = "https://www.aclweb.org/anthology/P18-1031",
    doi = "10.18653/v1/P18-1031",
    pages = "328--339",
    abstract = "Inductive transfer learning has greatly impacted computer vision, but existing approaches in NLP still require task-specific modifications and training from scratch. We propose Universal Language Model Fine-tuning (ULMFiT), an effective transfer learning method that can be applied to any task in NLP, and introduce techniques that are key for fine-tuning a language model. Our method significantly outperforms the state-of-the-art on six text classification tasks, reducing the error by 18-24{\%} on the majority of datasets. Furthermore, with only 100 labeled examples, it matches the performance of training from scratch on 100 times more data. We open-source our pretrained models and code.",
}

@inproceedings{ruder-etal-2019-transfer,
    title = "Transfer Learning in Natural Language Processing",
    author = "Ruder, Sebastian  and
      Peters, Matthew E.  and
      Swayamdipta, Swabha  and
      Wolf, Thomas",
    booktitle = "Proceedings of the 2019 Conference of the North {A}merican Chapter of the Association for Computational Linguistics: Tutorials",
    year = "2019",
    %address = "Minneapolis, Minnesota",
    publisher = "Association for Computational Linguistics",
    url = "https://www.aclweb.org/anthology/N19-5004",
    doi = "10.18653/v1/N19-5004",
    pages = "15--18",
    abstract = "The classic supervised machine learning paradigm is based on learning in isolation, a single predictive model for a task using a single dataset. This approach requires a large number of training examples and performs best for well-defined and narrow tasks. Transfer learning refers to a set of methods that extend this approach by leveraging data from additional domains or tasks to train a model with better generalization properties. Over the last two years, the field of Natural Language Processing (NLP) has witnessed the emergence of several transfer learning methods and architectures which significantly improved upon the state-of-the-art on a wide range of NLP tasks. These improvements together with the wide availability and ease of integration of these methods are reminiscent of the factors that led to the success of pretrained word embeddings and ImageNet pretraining in computer vision, and indicate that these methods will likely become a common tool in the NLP landscape as well as an important research direction. We will present an overview of modern transfer learning methods in NLP, how models are pre-trained, what information the representations they learn capture, and review examples and case studies on how these models can be integrated and adapted in downstream NLP tasks.",
}

@inproceedings{DBLP:journals/corr/abs-1902-05309,
   title={Transfer Learning for Sequence Labeling Using Source Model and Target Data},
   booktitle={Proceedings of the AAAI Conference on Artificial Intelligence},
   volume={33},
   pages={6260--6267},
   year={2019},
   publisher={Association for the Advancement of Artificial Intelligence (AAAI) Press},
   author={Chen, Lingzhen and Moschitti, Alessandro},
}

@inproceedings{DBLP:journals/corr/abs-1906-08237,
 author    = {Zhilin Yang and
               Zihang Dai and
               Yiming Yang and
               Jaime G. Carbonell and
               Ruslan Salakhutdinov and
               Quoc V. Le},
  title     = {XLNet: Generalized Autoregressive Pretraining for Language Understanding},
  booktitle = {Advances in Neural Information Processing Systems 32: Annual Conference
               on Neural Information Processing Systems 2019, NeurIPS 2019},
  publisher = {Curran Associates, Inc.},
  pages     = {5754--5764},
  year      = {2019},
  url       = {http://papers.nips.cc/paper/8812-xlnet-generalized-autoregressive-pretraining-for-language-understanding},
  bibsource = {dblp computer science bibliography, https://dblp.org}
}

@article{OpenAI-GP2,
author= {Alec Radford
        and  Karthik Narasimhan
        and  Tim Salimans
        and Ilya Sutskever},
year      = {2018},
title     = {Improving language understanding with unsupervised learning},
 journal={Technical report, OpenAI},
}

@inproceedings{liu-etal-2019-linguistic,
    title = "Linguistic Knowledge and Transferability of Contextual Representations",
    author = "Liu, Nelson F.  and
      Gardner, Matt  and
      Belinkov, Yonatan  and
      Peters, Matthew E.  and
      Smith, Noah A.",
    booktitle = "Proceedings of the 2019 Conference of the North {A}merican Chapter of the Association for Computational Linguistics: Human Language Technologies, Volume 1 (Long and Short Papers)",
    year = "2019",
    %address = "Minneapolis, Minnesota",
    publisher = "Association for Computational Linguistics",
    url = "https://www.aclweb.org/anthology/N19-1112",
    doi = "10.18653/v1/N19-1112",
    pages = "1073--1094",
}

@article{DBLP:journals/corr/VaswaniSPUJGKP17,
  author    = {Ashish Vaswani and
               Noam Shazeer and
               Niki Parmar and
               Jakob Uszkoreit and
               Llion Jones and
               Aidan N. Gomez and
               Lukasz Kaiser and
               Illia Polosukhin},
  title     = {Attention Is All You Need},
  journal   = {CoRR},
  volume    = {abs/1706.03762},
  year      = {2017},
  url       = {http://arxiv.org/abs/1706.03762},
  archivePrefix = {arXiv},
  eprint    = {1706.03762},
  bibsource = {dblp computer science bibliography, https://dblp.org}
}

@book{Yoav_book,
  author = {Goldberg, Yoav},
  title = {Neural Network Methods in Natural Language Processing},
  year = {2017},
  isbn = {1627052984},
  publisher = {Morgan \& Claypool Publishers, San Rafael, CA},
}

@book{EMB-NLP01,
author = {Mohammad Taher Pilehvar and Jos{\'e} Camacho-Collados},
title = {Embeddings in Natural Language Processing},
year = {2020},
publisher = {Morgan \& Claypool Publishers, San Rafael, CA},
}

@inproceedings{DBLP:journals/corr/ZhangW15b,
   title = "A Sensitivity Analysis of (and Practitioners{'} Guide to) Convolutional Neural Networks for Sentence Classification",
    author = "Zhang, Ye  and
      Wallace, Byron",
    booktitle = "Proceedings of the Eighth International Joint Conference on Natural Language Processing (Volume 1: Long Papers)",
    year = "2017",
    %address = "Taipei, Taiwan",
    publisher = "Asian Federation of Natural Language Processing",
    url = "https://www.aclweb.org/anthology/I17-1026",
    pages = "253--263",
}

@incollection{NIPS2012_4824,
title = {ImageNet Classification with Deep Convolutional Neural Networks},
author = {Alex Krizhevsky and Sutskever, Ilya and Hinton, Geoffrey E},
booktitle = {Advances in Neural Information Processing Systems 25},
pages = {1097--1105},
year = {2012},
publisher = {Curran Associates, Inc.},
url = {http://papers.nips.cc/paper/4824-imagenet-classification-with-deep-convolutional-neural-networks.pdf}
}

@inbook{10.5555/303568.303704,
author = {LeCun, Yann and Bengio, Yoshua},
title = {Convolutional Networks for Images, Speech, and Time Series},
year = {1998},
isbn = {0262511029},
publisher = {MIT Press},
%address = {Cambridge, MA, USA},
booktitle = {The Handbook of Brain Theory and Neural Networks},
pages = {255–258},
numpages = {4}
}

@article{DBLP:journals/corr/abs-1708-02709,
  title={Recent trends in deep learning based natural language processing},
  author={Young, Tom and Hazarika, Devamanyu and Poria, Soujanya and Cambria, Erik},
  journal={IEEE Computational intelligenCe magazine},
  volume={13},
  number={3},
  pages={55--75},
  year={2018},
  publisher={IEEE}
}

@incollection{reason:RumHinWil86a,
  added-at = {2016-11-26T13:19:29.000+0100},
  %address = {Cambridge, MA},
  author = {Rumelhart, D. E. and Hinton, G. E. and Williams, R. J.},
  biburl = {https://www.bibsonomy.org/bibtex/2419301764b73818c3e2e251162d870d9/machinelearning},
  booktitle = {Parallel Distributed Processing},
  chapter = 8,
  pages = {318--362},
  publisher = {MIT Press},
  title = {Learning Internal Representations by Error Propagation},
  year = 1986
}

@article{Elman90findingstructure,
    author = {Elman, Jeffrey L.},
    journal = {Cognitive Science},
    number = 2,
    pages = {179-211},
    title = {Finding Structure in Time.},
    url = {http://dblp.uni-trier.de/db/journals/cogsci/cogsci14.html#Elman90},
    volume = 14,
    year = 1990
}

@inproceedings{DBLP:journals/corr/ChoMBB14,
    title = "On the Properties of Neural Machine Translation: Encoder{--}Decoder Approaches",
    author = {Cho, Kyunghyun  and
      van Merri{\"e}nboer, Bart  and
      Bahdanau, Dzmitry  and
      Bengio, Yoshua},
    booktitle = "Proceedings of {SSST}-8, Eighth Workshop on Syntax, Semantics and Structure in Statistical Translation",
    year = "2014",
    %address = "Doha, Qatar",
    publisher = "Association for Computational Linguistics",
    url = "https://www.aclweb.org/anthology/W14-4012",
    doi = "10.3115/v1/W14-4012",
    pages = "103--111",
}

@inproceedings{DBLP:journals/corr/LuongPM15,
    title = "Effective Approaches to Attention-based Neural Machine Translation",
    author = "Luong, Thang  and
      Pham, Hieu  and
      Manning, Christopher D.",
    booktitle = "Proceedings of the 2015 Conference on Empirical Methods in Natural Language Processing",
    year = "2015",
    %address = "Lisbon, Portugal",
    publisher = "Association for Computational Linguistics",
    url = "https://www.aclweb.org/anthology/D15-1166",
    doi = "10.18653/v1/D15-1166",
    pages = "1412--1421",
}

@inproceedings{DBLP:journals/corr/BahdanauCB14,
  author    = {Dzmitry Bahdanau and
               Kyunghyun Cho and
               Yoshua Bengio},
  title     = {Neural Machine Translation by Jointly Learning to Align and Translate},
  booktitle = {3rd International Conference on Learning Representations, {ICLR} 2015, Conference Track Proceedings},
  year      = {2015},
  url       = {http://arxiv.org/abs/1409.0473},
  bibsource = {dblp computer science bibliography, https://dblp.org}
}

@inproceedings{DBLP:journals/corr/HeZRS15,
  author    = {Kaiming He and
               Xiangyu Zhang and
               Shaoqing Ren and
               Jian Sun},
  booktitle={2016 IEEE Conference on Computer Vision and Pattern Recognition (CVPR)},
  title   = {Deep Residual Learning for Image Recognition},
  publisher = {{IEEE} Computer Society},
  year={2016},
  volume={},
  number={},
  pages={770-778}
}

@article{DBLP:journals/corr/BaKH16,
  author    = {Lei Jimmy Ba and
               Jamie Ryan Kiros and
               Geoffrey E. Hinton},
  title     = {Layer Normalization},
  journal   = {CoRR},
  volume    = {abs/1607.06450},
  year      = {2016},
  url       = {http://arxiv.org/abs/1607.06450},
  archivePrefix = {arXiv},
  eprint    = {1607.06450},
  bibsource = {dblp computer science bibliography, https://dblp.org}
}

@phdthesis{Nakashole2012,
author = {Nakashole, Ndapandula},
title = {Automatic Extraction of Facts, Relations, and Entities for Web-Scale Knowledge Base Population},
school = {Universit{\"a}t des Saarlandes},
year = {2012}
}

@inproceedings{Gerber2011,
  author = {Gerber, Daniel and {Ngonga Ngomo}, Axel-Cyrille},
  booktitle = {1st Workshop on Web Scale Knowledge Extraction @ ISWC 2011},
  title = {Bootstrapping the Linked Data Web},
  year = {2011}
}

@inproceedings{Nakashole:2011,
    author = {Nakashole, Ndapandula and Theobald, Martin and Weikum, Gerhard},
    title = {Scalable Knowledge Harvesting with High Precision and High Recall},
    year = {2011},
    publisher = {Association for Computing Machinery},
    %address = {New York, NY, USA},
    booktitle = {Proceedings of the Fourth ACM International Conference on Web Search and Data Mining},
    pages = {227–236},
    numpages = {10},
}

@inproceedings{Xu:2010,
    title = "Boosting Relation Extraction with Limited Closed-World Knowledge",
    author = "Xu, Feiyu  and
      Uszkoreit, Hans  and
      Krause, Sebastian  and
      Li, Hong",
    booktitle = "Proceedings of the 23rd International Conference on Computational Linguistics: Posters",
    year = "2010",
    %address = "Beijing, China",
    publisher = "Coling 2010 Organizing Committee",
    url = "https://www.aclweb.org/anthology/C10-2155",
    pages = "1354--1362",
}

@inproceedings{Fader:2011,
 author = {Fader, Anthony and Soderland, Stephen and Etzioni, Oren},
 title = {Identifying Relations for Open Information Extraction},
 booktitle = {Proceedings of the Conference on Empirical Methods in Natural Language Processing},
 series = {EMNLP '11},
 year = {2011},
 isbn = {978-1-937284-11-4},
 %location= {Edinburgh, United Kingdom},
 pages = {1535--1545},
 numpages = {11},
 url = {http://dl.acm.org/citation.cfm?id=2145432.2145596},
 acmid = {2145596},
 publisher = {Association for Computational Linguistics},
 %address = {Stroudsburg, PA, USA},
}

@inproceedings{2015angeli-openie,
  author = {Mausam and Schmitz, Michael and Bart, Robert and Soderland, Stephen and Etzioni, Oren},
 title = {Open Language Learning for Information Extraction},
 booktitle = {Proceedings of the 2012 Joint Conference on Empirical Methods in Natural Language Processing and Computational Natural Language Learning},
 series = {EMNLP-CoNLL '12},
 year = {2012},
 %location= {Jeju Island, Korea},
 pages = {523--534},
 numpages = {12},
 url = {http://dl.acm.org/citation.cfm?id=2390948.2391009},
 acmid = {2391009},
 publisher = {Association for Computational Linguistics},
 %address = {Stroudsburg, PA, USA},
}

@inproceedings{Lin15learningentity,
    author = {Lin, Yankai and Liu, Zhiyuan and Sun, Maosong and Liu, Yang and Zhu, Xuan},
 title = {Learning Entity and Relation Embeddings for Knowledge Graph Completion},
 booktitle = {Proceedings of the Twenty-Ninth AAAI Conference on Artificial Intelligence},
 series = {AAAI'15},
 year = {2015},
 isbn = {0-262-51129-0},
 %location= {Austin, Texas},
 pages = {2181--2187},
 numpages = {7},
 url = {http://dl.acm.org/citation.cfm?id=2886521.2886624},
 acmid = {2886624},
 publisher = {Association for the Advancement of Artificial Intelligence (AAAI) Press}
}

@incollection{NIPS2013_5028,
title = {Reasoning With Neural Tensor Networks for Knowledge Base Completion},
author = {Socher, Richard and Chen, Danqi and Manning, Christopher D and Ng, Andrew},
booktitle = {Advances in Neural Information Processing Systems 26},
pages = {926--934},
year = {2013},
publisher = {Curran Associates, Inc.},
url = {http://papers.nips.cc/paper/5028-reasoning-with-neural-tensor-networks-for-knowledge-base-completion.pdf}
}

@inproceedings{NakasholeTW13,
  title = "Fine-grained Semantic Typing of Emerging Entities",
    author = "Nakashole, Ndapandula  and
      Tylenda, Tomasz  and
      Weikum, Gerhard",
    booktitle = "Proceedings of the 51st Annual Meeting of the Association for Computational Linguistics (Volume 1: Long Papers)",
    year = "2013",
    %address = "Sofia, Bulgaria",
    publisher = "Association for Computational Linguistics",
    url = "https://www.aclweb.org/anthology/P13-1146",
    pages = "1488--1497",
}

@inproceedings{CorroAGW15,
   title = "{FINET}: Context-Aware Fine-Grained Named Entity Typing",
    author = "Del Corro, Luciano  and
      Abujabal, Abdalghani  and
      Gemulla, Rainer  and
      Weikum, Gerhard",
    booktitle = "Proceedings of the 2015 Conference on Empirical Methods in Natural Language Processing",
    year = "2015",
    %address = "Lisbon, Portugal",
    publisher = "Association for Computational Linguistics",
    url = "https://www.aclweb.org/anthology/D15-1103",
    doi = "10.18653/v1/D15-1103",
    pages = "868--878",
}

@article{10.1162/089976600300015015,
    author = {Gers, Felix A. and Schmidhuber, J\"{u}rgen A. and Cummins, Fred A.},
    title = {Learning to Forget: Continual Prediction with LSTM},
    year = {2000},
    issue_date = {October 2000},
    publisher = {MIT Press},
    %address = {Cambridge, MA, USA},
    volume = {12},
    number = {10},
    issn = {0899-7667},
    url = {https://doi.org/10.1162/089976600300015015},
    doi = {10.1162/089976600300015015},
    journal = {Neural Computation},
    pages = {2451–2471},
}

@article{DBLP:journals/corr/Rong14,
  author    = {Xin Rong},
  title     = {word2vec Parameter Learning Explained},
  journal   = {CoRR},
  volume    = {abs/1411.2738},
  year      = {2014},
  url       = {http://arxiv.org/abs/1411.2738},
  archivePrefix = {arXiv},
  eprint    = {1411.2738},
  bibsource = {dblp computer science bibliography, https://dblp.org},
}

@inproceedings{Oep:Far:Ovr:18,
  title = "The 2018 Shared Task on Extrinsic Parser Evaluation: On the Downstream Utility of {E}nglish Universal Dependency Parsers",
    author = {Fares, Murhaf  and
      Oepen, Stephan  and
      {\O}vrelid, Lilja  and
      Bj{\"o}rne, Jari  and
      Johansson, Richard},
    booktitle = "Proceedings of the {C}o{NLL} 2018 Shared Task: Multilingual Parsing from Raw Text to Universal Dependencies",
    year = "2018",
    %address = "Brussels, Belgium",
    publisher = "Association for Computational Linguistics",
    url = "https://www.aclweb.org/anthology/K18-2002",
    doi = "10.18653/v1/K18-2002",
    pages = "22--33",
}

@inproceedings{GBOR16.870,
  author = {Kata G\'{a}bor and Haifa Zargayouna and Davide Buscaldi and Isabelle Tellier and Thierry Charnois},
  title = {Semantic Annotation of the ACL Anthology Corpus for the Automatic Analysis of Scientific Literature},
  booktitle = {Proceedings of the Tenth International Conference on Language Resources and Evaluation (LREC 2016)},
  year = {2016},
  %location= {Portorož, Slovenia},
  publisher = {European Language Resources Association (ELRA)},
  %address = {Paris, France},
  isbn = {978-2-9517408-9-1},
 }

@phdthesis{duong2017natural,
  title={Natural language processing for resource-poor languages},
  author={Duong, Long},
  year={2017},
  school = {University of Melbourne},
}

@article{10.1145/3241741,
    author = {Smirnova, Alisa and Cudr\'{e}-Mauroux, Philippe},
    title = {Relation Extraction Using Distant Supervision: A Survey},
    year = {2018},
    issue_date = {January 2019},
    publisher = {Association for Computing Machinery},
    %address = {New York, NY, USA},
    volume={51},
    number={5},
    pages={1--35},
    issn = {0360-0300},
    url = {https://doi.org/10.1145/3241741},
    doi = {10.1145/3241741},
    journal = {ACM Computing Survey},
}

@phdthesis{tsvetkov2016linguistic,
    title={Linguistic Knowledge in Data-Driven Natural Language Processing},
    author={Tsvetkov, Yulia},
    year={2016},
    school={Carnegie Mellon University},
}

@phdthesis{king2015practical,
  title={Practical Natural Language Processing for Low-Resource Languages},
  author={King, Benjamin Philip},
  year={2015},
  school={University of Melbourne}
}

@inproceedings{Kann_2019,
    title = "Towards Realistic Practices In Low-Resource Natural Language Processing: The Development Set",
    author = "Kann, Katharina  and
      Cho, Kyunghyun  and
      Bowman, Samuel R.",
    booktitle = "Proceedings of the 2019 Conference on Empirical Methods in Natural Language Processing and the 9th International Joint Conference on Natural Language Processing (EMNLP-IJCNLP)",
    year = "2019",
    %address = "Hong Kong, China",
    publisher = "Association for Computational Linguistics",
    url = "https://www.aclweb.org/anthology/D19-1329",
    doi = "10.18653/v1/D19-1329",
    pages = "3342--3349",
}

@book{ws-2018-deep,
    title = "Proceedings of the Workshop on Deep Learning Approaches for Low-Resource {NLP}",
    author = "Haffari, Reza  and Cherry, Colin  and Foster, George  and Khadivi, Shahram  and Salehi, Bahar",
    year = "2018",
    %address = "Melbourne",
    publisher = "Association for Computational Linguistics, Melbourne, Australia",
    url = "https://www.aclweb.org/anthology/W18-3400",
}

@book{emnlp-2019-deep,
    title = "Proceedings of the 2nd Workshop on Deep Learning Approaches for Low-Resource NLP (DeepLo 2019)",
    author = "Cherry, Colin  and  Durrett, Greg  and Foster, George  and Haffari, Reza  and Khadivi, Shahram  and Peng, Nanyun  and Ren, Xiang  and Swayamdipta, Swabha",
    year = "2019",
    %address = "Hong Kong, China",
    publisher = "Association for Computational Linguistics, Hong Kong, China",
    url = "https://www.aclweb.org/anthology/D19-6100",
}

@inproceedings{cui-etal-2018-neural,
    title = "Neural Open Information Extraction",
    author = "Cui, Lei  and
      Wei, Furu  and
      Zhou, Ming",
    booktitle = "Proceedings of the 56th Annual Meeting of the Association for Computational Linguistics (Volume 2: Short Papers)",
    year = "2018",
    %address = "Melbourne, Australia",
    publisher = "Association for Computational Linguistics",
    url = "https://www.aclweb.org/anthology/P18-2065",
    doi = "10.18653/v1/P18-2065",
    pages = "407--413",
}

@inproceedings{zeng-etal-2018-extracting,
    title = "Extracting Relational Facts by an End-to-End Neural Model with Copy Mechanism",
    author = "Zeng, Xiangrong  and
      Zeng, Daojian  and
      He, Shizhu  and
      Liu, Kang  and
      Zhao, Jun",
    booktitle = "Proceedings of the 56th Annual Meeting of the Association for Computational Linguistics (Volume 1: Long Papers)",
    year = "2018",
    %address = "Melbourne, Australia",
    publisher = "Association for Computational Linguistics",
    url = "https://www.aclweb.org/anthology/P18-1047",
    doi = "10.18653/v1/P18-1047",
    pages = "506--514",
    abstract = "The relational facts in sentences are often complicated. Different relational triplets may have overlaps in a sentence. We divided the sentences into three types according to triplet overlap degree, including Normal, EntityPairOverlap and SingleEntiyOverlap. Existing methods mainly focus on Normal class and fail to extract relational triplets precisely. In this paper, we propose an end-to-end model based on sequence-to-sequence learning with copy mechanism, which can jointly extract relational facts from sentences of any of these classes. We adopt two different strategies in decoding process: employing only one united decoder or applying multiple separated decoders. We test our models in two public datasets and our model outperform the baseline method significantly.",
}

@inproceedings{gupta-etal-2016-table,
    title = "Table Filling Multi-Task Recurrent Neural Network for Joint Entity and Relation Extraction",
    author = {Gupta, Pankaj  and
      Sch{\"u}tze, Hinrich  and
      Andrassy, Bernt},
    booktitle = "Proceedings of {COLING} 2016, the 26th International Conference on Computational Linguistics: Technical Papers",
    year = "2016",
    %address = "Osaka, Japan",
    publisher = "The COLING 2016 Organizing Committee",
    url = "https://www.aclweb.org/anthology/C16-1239",
    pages = "2537--2547",
    abstract = "This paper proposes a novel context-aware joint entity and word-level relation extraction approach through semantic composition of words, introducing a Table Filling Multi-Task Recurrent Neural Network (TF-MTRNN) model that reduces the entity recognition and relation classification tasks to a table-filling problem and models their interdependencies. The proposed neural network architecture is capable of modeling multiple relation instances without knowing the corresponding relation arguments in a sentence. The experimental results show that a simple approach of piggybacking candidate entities to model the label dependencies from relations to entities improves performance. We present state-of-the-art results with improvements of 2.0{\%} and 2.7{\%} for entity recognition and relation classification, respectively on CoNLL04 dataset.",
}

@inproceedings{zheng-etal-2017-joint,
    title = "Joint Extraction of Entities and Relations Based on a Novel Tagging Scheme",
    author = "Zheng, Suncong  and
      Wang, Feng  and
      Bao, Hongyun  and
      Hao, Yuexing  and
      Zhou, Peng  and
      Xu, Bo",
    booktitle = "Proceedings of the 55th Annual Meeting of the Association for Computational Linguistics (Volume 1: Long Papers)",
    year = "2017",
    %address = "Vancouver, Canada",
    publisher = "Association for Computational Linguistics",
    url = "https://www.aclweb.org/anthology/P17-1113",
    doi = "10.18653/v1/P17-1113",
    pages = "1227--1236",
    abstract = "Joint extraction of entities and relations is an important task in information extraction. To tackle this problem, we firstly propose a novel tagging scheme that can convert the joint extraction task to a tagging problem.. Then, based on our tagging scheme, we study different end-to-end models to extract entities and their relations directly, without identifying entities and relations separately. We conduct experiments on a public dataset produced by distant supervision method and the experimental results show that the tagging based methods are better than most of the existing pipelined and joint learning methods. What{'}s more, the end-to-end model proposed in this paper, achieves the best results on the public dataset.",
}

@article{chiu-nichols-2016-named,
    title = "Named Entity Recognition with Bidirectional {LSTM}-{CNN}s",
    author = "Chiu, Jason P.C.  and
      Nichols, Eric",
    journal = "Transactions of the Association for Computational Linguistics",
    volume = "4",
    year = "2016",
    url = "https://www.aclweb.org/anthology/Q16-1026",
    doi = "10.1162/tacl_a_00104",
    pages = "357--370",
}

@inproceedings{jiang-etal-2019-improving,
    title = "Improving Open Information Extraction via Iterative Rank-Aware Learning",
    author = "Jiang, Zhengbao  and
      Yin, Pengcheng  and
      Neubig, Graham",
    booktitle = "Proceedings of the 57th Annual Meeting of the Association for Computational Linguistics",
    year = "2019",
    %address = "Florence, Italy",
    publisher = "Association for Computational Linguistics",
    url = "https://www.aclweb.org/anthology/P19-1523",
    doi = "10.18653/v1/P19-1523",
    pages = "5295--5300",
    abstract = "Open information extraction (IE) is the task of extracting open-domain assertions from natural language sentences. A key step in open IE is confidence modeling, ranking the extractions based on their estimated quality to adjust precision and recall of extracted assertions. We found that the extraction likelihood, a confidence measure used by current supervised open IE systems, is not well calibrated when comparing the quality of assertions extracted from different sentences. We propose an additional binary classification loss to calibrate the likelihood to make it more globally comparable, and an iterative learning process, where extractions generated by the open IE model are incrementally included as training samples to help the model learn from trial and error. Experiments on OIE2016 demonstrate the effectiveness of our method. Code and data are available at https://github.com/jzbjyb/oie{\_}rank.",
}

@inproceedings{stanovsky-etal-2018-supervised,
    title = "Supervised Open Information Extraction",
    author = "Stanovsky, Gabriel  and
      Michael, Julian  and
      Zettlemoyer, Luke  and
      Dagan, Ido",
    booktitle = "Proceedings of the 2018 Conference of the North {A}merican Chapter of the Association for Computational Linguistics: Human Language Technologies, Volume 1 (Long Papers)",
    year = "2018",
    %address = "New Orleans, Louisiana",
    publisher = "Association for Computational Linguistics",
    url = "https://www.aclweb.org/anthology/N18-1081",
    doi = "10.18653/v1/N18-1081",
    pages = "885--895",
    abstract = "We present data and methods that enable a supervised learning approach to Open Information Extraction (Open IE). Central to the approach is a novel formulation of Open IE as a sequence tagging problem, addressing challenges such as encoding multiple extractions for a predicate. We also develop a bi-LSTM transducer, extending recent deep Semantic Role Labeling models to extract Open IE tuples and provide confidence scores for tuning their precision-recall tradeoff. Furthermore, we show that the recently released Question-Answer Meaning Representation dataset can be automatically converted into an Open IE corpus which significantly increases the amount of available training data. Our supervised model outperforms the existing state-of-the-art Open IE systems on benchmark datasets.",
}

@inproceedings{zhang-etal-2017-mt,
    title = "{MT}/{IE}: Cross-lingual Open Information Extraction with Neural Sequence-to-Sequence Models",
    author = "Zhang, Sheng  and
      Duh, Kevin  and
      Van Durme, Benjamin",
    booktitle = "Proceedings of the 15th Conference of the {E}uropean Chapter of the Association for Computational Linguistics: Volume 2, Short Papers",
    year = "2017",
    %address = "Valencia, Spain",
    publisher = "Association for Computational Linguistics",
    url = "https://www.aclweb.org/anthology/E17-2011",
    pages = "64--70",
    abstract = "Cross-lingual information extraction is the task of distilling facts from foreign language (e.g. Chinese text) into representations in another language that is preferred by the user (e.g. English tuples). Conventional pipeline solutions decompose the task as machine translation followed by information extraction (or vice versa). We propose a joint solution with a neural sequence model, and show that it outperforms the pipeline in a cross-lingual open information extraction setting by 1-4 BLEU and 0.5-0.8 F1.",
}

@inproceedings{Yukunma-etal-2016-label,
    title = "Label Embedding for Zero-shot Fine-grained Named Entity Typing",
    author = "Ma, Yukun  and
      Cambria, Erik  and
      Gao, Sa",
    booktitle = "Proceedings of the 26th International Conference on Computational Linguistics: Technical Papers",
    year = "2016",
    %address = "Osaka, Japan",
    publisher = "Association for Computational Linguistics",
    url = "https://www.aclweb.org/anthology/C16-1017",
    pages = "171--180",
}

@article{DBLP:journals/corr/abs-1904-10503,
  author    = {Cihan Dogan and
               Aimore Dutra and
               Adam Gara and
               Alfredo Gemma and
               Lei Shi and
               Michael Sigamani and
               Ella Walters},
  title     = {Fine-Grained Named Entity Recognition using ELMo and Wikidata},
  journal   = {CoRR},
  volume    = {abs/1904.10503},
  year      = {2019},
  url       = {http://arxiv.org/abs/1904.10503},
  archivePrefix = {arXiv},
  eprint    = {1904.10503},
  bibsource = {dblp computer science bibliography, https://dblp.org}
}

@inproceedings{mai-etal-2018-empirical,
    title = "An Empirical Study on Fine-Grained Named Entity Recognition",
    author = "Mai, Khai  and
      Pham, Thai-Hoang  and
      Nguyen, Minh Trung  and
      Nguyen, Tuan Duc  and
      Bollegala, Danushka  and
      Sasano, Ryohei  and
      Sekine, Satoshi",
    booktitle = "Proceedings of the 27th International Conference on Computational Linguistics",
    year = "2018",
    %address = "Santa Fe, New Mexico, USA",
    publisher = "Association for Computational Linguistics",
    url = "https://www.aclweb.org/anthology/C18-1060",
    pages = "711--722",
}

@inproceedings{DBLP:journals/corr/Camacho-Collados17aa,
    title = "On the Role of Text Preprocessing in Neural Network Architectures: An Evaluation Study on Text Categorization and Sentiment Analysis",
    author = "Camacho-Collados, Jos{\'e}  and
      Pilehvar, Mohammad Taher",
    booktitle = "Proceedings of the 2018 {EMNLP} Workshop {B}lackbox{NLP}: Analyzing and Interpreting Neural Networks for {NLP}",
    year = "2018",
    %address = "Brussels, Belgium",
    publisher = "Association for Computational Linguistics",
    url = "https://www.aclweb.org/anthology/W18-5406",
    doi = "10.18653/v1/W18-5406",
    pages = "40--46",
}

@inproceedings{Velldal,
    title = "Word vectors, reuse, and replicability: Towards a community repository of large-text resources",
    author = "Fares, Murhaf  and
      Kutuzov, Andrey  and
      Oepen, Stephan  and
      Velldal, Erik",
    booktitle = "Proceedings of the 21st Nordic Conference on Computational Linguistics",
    year = "2017",
    %address = "Gothenburg, Sweden",
    publisher = "Association for Computational Linguistics",
    url = "https://www.aclweb.org/anthology/W17-0237",
    pages = "271--276",
}

@inproceedings{rotsztejn-etal-2018-eth,
   title = "{ETH}-{DS}3{L}ab at {S}em{E}val-2018 Task 7: Effectively Combining Recurrent and Convolutional Neural Networks for Relation Classification and Extraction",
    author = "Rotsztejn, Jonathan  and
      Hollenstein, Nora  and
      Zhang, Ce",
    booktitle = "Proceedings of The 12th International Workshop on Semantic Evaluation",
    year = "2018",
    %address = "New Orleans, Louisiana",
    publisher = "Association for Computational Linguistics",
    url = "https://www.aclweb.org/anthology/S18-1112",
    doi = "10.18653/v1/S18-1112",
    pages = "689--696"
}

@inproceedings{luan-etal-2018-uwnlp,
    title = "The {UWNLP} system at {S}em{E}val-2018 Task 7: Neural Relation Extraction Model with Selectively Incorporated Concept Embeddings",
    author = "Luan, Yi  and
      Ostendorf, Mari  and
      Hajishirzi, Hannaneh",
    booktitle = "Proceedings of The 12th International Workshop on Semantic Evaluation",
    year = "2018",
    %address = "New Orleans, Louisiana",
    publisher = "Association for Computational Linguistics",
    url = "https://www.aclweb.org/anthology/S18-1125",
    doi = "10.18653/v1/S18-1125",
    pages = "788--792",
    abstract = "This paper describes our submission for SemEval 2018 Task 7 shared task on semantic relation extraction and classification in scientific papers. Our model is based on the end-to-end relation extraction model of (Miwa and Bansal, 2016) with several enhancements such as character-level encoding attention mechanism on selecting pretrained concept candidate embeddings. Our official submission ranked the second in relation classification task (Subtask 1.1 and Subtask 2 Senerio 2), and the first in the relation extraction task (Subtask 2 Scenario 1).",
}

@inproceedings{pratap-etal-2018-talla,
    title = "Talla at {S}em{E}val-2018 Task 7: Hybrid Loss Optimization for Relation Classification using Convolutional Neural Networks",
    author = "Pratap, Bhanu  and
      Shank, Daniel  and
      Ositelu, Oladipo  and
      Galbraith, Byron",
    booktitle = "Proceedings of The 12th International Workshop on Semantic Evaluation",
    year = "2018",
    %address = "New Orleans, Louisiana",
    publisher = "Association for Computational Linguistics",
    url = "https://www.aclweb.org/anthology/S18-1139",
    doi = "10.18653/v1/S18-1139",
    pages = "863--867",
    abstract = "This paper describes our approach to SemEval-2018 Task 7 {--} given an entity-tagged text from the ACL Anthology corpus, identify and classify pairs of entities that have one of six possible semantic relationships. Our model consists of a convolutional neural network leveraging pre-trained word embeddings, unlabeled ACL-abstracts, and multiple window sizes to automatically learn useful features from entity-tagged sentences. We also experiment with a hybrid loss function, a combination of cross-entropy loss and ranking loss, to boost the separation in classification scores. Lastly, we include WordNet-based features to further improve the performance of our model. Our best model achieves an F1(macro) score of 74.2 and 84.8 on subtasks 1.1 and 1.2, respectively.",
}

@inproceedings{lewis2019mlqa,
    title = "MLQA: Evaluating Cross-lingual Extractive Question Answering",
    author = "Lewis, Patrick and O\u{g}uz, Barlas and Rinott, Ruty and Riedel, Sebastian and Schwenk, Holger",
    booktitle = "Proceedings of the 58th Annual Meeting of the Association for Computational Linguistics",
    year = "2020",
    publisher = "Association for Computational Linguistics",
}

@article{hu2020xtreme,
    title={XTREME: A massively multilingual multi-task benchmark for evaluating cross-lingual generalization},
    author={Hu, Junjie and Ruder, Sebastian and Siddhant, Aditya and Neubig, Graham and Firat, Orhan and Johnson, Melvin},
    year={2020},
    journal = {CoRR},
    volume = {abs/2003.11080},
    url={https://arxiv.org/abs/2003.11080},

}

@article{Liang2020XGLUEAN,
  title={XGLUE: A New Benchmark Dataset for Cross-lingual Pre-training, Understanding and Generation},
  author={Liang, Yaobo and Duan, Nan and Gong, Yeyun and Wu, Ning and Guo, Fenfei and Qi, Weizhen and Gong, Ming and Shou, Linjun and Jiang, Daxin and Cao, Guihong and others},
  year={2020},
  journal = {CoRR},
  volume = {abs/2004.01401},
  url={https://arxiv.org/abs/2004.01401},
}

@inproceedings{conneau2019unsupervised,
    title = "Unsupervised cross-lingual representation learning at scale",
    author = "Conneau, Alexis and Khandelwal, Kartikay and Goyal, Naman and Chaudhary, Vishrav and Wenzek, Guillaume and Guzm\'{a}n, Francisco and Grave, Edouard and Ott, Myle and Zettlemoyer, Luke and Stoyanov, Veselin",
    booktitle = "Proceedings of the 58th Annual Meeting of the Association for Computational Linguistics",
    year = "2020",
    publisher = "Association for Computational Linguistics",
}

@phdthesis{bjerva:2017,
author = {Bjerva, Johannes},
school = {University of Groningen},
title = {{One Model to Rule them all -- Multitask and Multilingual Modelling for Lexical Analysis}},
year = {2017}
}

@inproceedings{delhoneux:2018,
    title = "Parameter sharing between dependency parsers for related languages",
    author = "{de Lhoneux}, Miryam  and
      Bjerva, Johannes  and
      Augenstein, Isabelle  and
      S{\o}gaard, Anders",
    booktitle = "Proceedings of the 2018 Conference on Empirical Methods in Natural Language Processing",
    year = "2018",
    %address = "Brussels, Belgium",
    publisher = "Association for Computational Linguistics",
    url = "https://www.aclweb.org/anthology/D18-1543",
    doi = "10.18653/v1/D18-1543",
    pages = "4992--4997",
}

@inproceedings{bjerva_augenstein:18:iwclul,
    title = "Tracking Typological Traits of Uralic Languages in Distributed Language Representations",
    author = "Bjerva, Johannes  and
      Augenstein, Isabelle",
    booktitle = "Proceedings of the Fourth International Workshop on Computational Linguistics of Uralic Languages",
    year = "2018",
    %address = "Helsinki, Finland",
    publisher = "Association for Computational Linguistics",
    url = "https://www.aclweb.org/anthology/W18-0207",
    doi = "10.18653/v1/W18-0207",
    pages = "76--86",
}

@inproceedings{levow-2006-third,
    title = "The Third International {C}hinese Language Processing Bakeoff: Word Segmentation and Named Entity Recognition",
    author = "Levow, Gina-Anne",
    booktitle = "Proceedings of the Fifth {SIGHAN} Workshop on {C}hinese Language Processing",
    year = "2006",
    %address = "Sydney, Australia",
    publisher = "Association for Computational Linguistics",
    url = "https://www.aclweb.org/anthology/W06-0115",
    pages = "108--117",
}

@inproceedings{wang-etal-2019-extracting,
    title = "Extracting Multiple-Relations in One-Pass with Pre-Trained Transformers",
    author = "Wang, Haoyu  and
      Tan, Ming  and
      Yu, Mo  and
      Chang, Shiyu  and
      Wang, Dakuo  and
      Xu, Kun  and
      Guo, Xiaoxiao  and
      Potdar, Saloni",
    booktitle = "Proceedings of the 57th Annual Meeting of the Association for Computational Linguistics",
    year = "2019",
    %address = "Florence, Italy",
    publisher = "Association for Computational Linguistics",
    url = "https://www.aclweb.org/anthology/P19-1132",
    doi = "10.18653/v1/P19-1132",
    pages = "1371--1377",
    abstract = "Many approaches to extract multiple relations from a paragraph require multiple passes over the paragraph. In practice, multiple passes are computationally expensive and this makes difficult to scale to longer paragraphs and larger text corpora. In this work, we focus on the task of multiple relation extractions by encoding the paragraph only once. We build our solution upon the pre-trained self-attentive models (Transformer), where we first add a structured prediction layer to handle extraction between multiple entity pairs, then enhance the paragraph embedding to capture multiple relational information associated with each entity with entity-aware attention. We show that our approach is not only scalable but can also perform state-of-the-art on the standard benchmark ACE 2005.",
}

@article{jiang2020improving,
  title={Improving Scholarly Knowledge Representation: Evaluating BERT-based Models for Scientific Relation Classification},
  author={Jiang, Ming and D'Souza, Jennifer and Auer, S{\"o}ren and Downie, J Stephen},
  journal = {CoRR},
  volume = {abs/2004.06153},
  url={https://arxiv.org/abs/2004.06153},
  year={2020}
}

@article{navigli2012babelnet,
    author = {Navigli, Roberto and Ponzetto, Simone Paolo},
    title = {BabelNet: The Automatic Construction, Evaluation and Application of a Wide-Coverage Multilingual Semantic Network},
    year = {2012},
    issue_date = {December, 2012},
    publisher = {Elsevier Science Publishers Ltd.},
    %address = {GBR},
    volume = {193},
    issn = {0004-3702},
    url = {https://doi.org/10.1016/j.artint.2012.07.001},
    doi = {10.1016/j.artint.2012.07.001},
    journal = {Artificial Intelligence},
    pages = {217–250},
}

@inproceedings{bordea2013domain,
  title={Domain-independent term extraction through domain modelling},
  author={Bordea, Georgeta and Buitelaar, Paul and Polajnar, Tamara},
  booktitle={The 10th international conference on terminology and artificial intelligence},
  year={2013},
  organization={TIA 2013},
}

@inproceedings{liu2020hamner,
	Author = {Liu, Shifeng and Sun, Yifang and Li, Bing and Wang, Wei and Zhao, Xiang},
	Booktitle = {Proceedings of the 34th AAAI Conference on Artificial Intelligence},
	Publisher = {Association for the Advancement of Artificial Intelligence (AAAI) Presss},
	Title = {HAMNER: Headword Amplified Multi-span Distantly Supervised Method for Domain Specific Named Entity Recognition},
	Year = {2020},
}

@inproceedings{Finkel:2009,
 author = {Finkel, Jenny Rose and Manning, Christopher D.},
 title = {Hierarchical Bayesian Domain Adaptation},
 booktitle = {Proceedings of Human Language Technologies: The 2009 Annual Conference of the North American Chapter of the Association for Computational Linguistics},
 series = {NAACL '09},
 year = {2009},
 isbn = {978-1-932432-41-1},
 %location= {Boulder, Colorado},
 pages = {602--610},
 url = {http://dl.acm.org/citation.cfm?id=1620754.1620842},
 acmid = {1620842},
 publisher = {Association for Computational Linguistics},
 %address = {Stroudsburg, PA, USA},
}

@phdthesis{bplank2011phd,
  author =  {Barbara Plank},
  title = {{Domain Adaptation for Parsing}},
  school= {University of Groningen},
  year =   {2011},
  type = {Ph.D. thesis},
}

@inproceedings{aharoni2020unsupervised,
    title = "Unsupervised Domain Clusters in Pretrained Language Models",
    author = "Roee Aharoni and Yoav Goldberg",
    booktitle = "Proceedings of the 58th Annual Meeting of the Association for Computational Linguistics",
    year = "2020",
    publisher = "Association for Computational Linguistics",
}

@inproceedings{Lee2001GENRESRT,
 title={Genres, registers, text types, domains and styles: clarifying the concepts and navigating a path through the BNC jungle},
  author={Lee, David},
  booktitle={Teaching and Learning by Doing Corpus Analysis},
  pages={245--292},
  year={2002},
  publisher={Brill Rodopi}
}

@inproceedings{van-der-wees-etal-2015-whats,
    title = "What{'}s in a Domain? Analyzing Genre and Topic Differences in Statistical Machine Translation",
    author = "{van der Wees}, Marlies  and
      Bisazza, Arianna  and
      Weerkamp, Wouter  and
      Monz, Christof",
    booktitle = "Proceedings of the 53rd Annual Meeting of the Association for Computational Linguistics and the 7th International Joint Conference on Natural Language Processing (Volume 2: Short Papers)",
    year = "2015",
    %address = "Beijing, China",
    publisher = "Association for Computational Linguistics",
    url = "https://www.aclweb.org/anthology/P15-2092",
    doi = "10.3115/v1/P15-2092",
    pages = "560--566",
}

@article{DBLP:journals/corr/Plank16,
  author    = {Barbara Plank},
  title     = {What to do about non-standard (or non-canonical) language in {NLP}},
  journal   = {CoRR},
  volume    = {abs/1608.07836},
  year      = {2016},
  url       = {http://arxiv.org/abs/1608.07836},
  archivePrefix = {arXiv},
  eprint    = {1608.07836},
  bibsource = {dblp computer science bibliography, https://dblp.org}
}

@inproceedings{Pires_2019,
    title = "How Multilingual is Multilingual {BERT}?",
    author = "Pires, Telmo  and
      Schlinger, Eva  and
      Garrette, Dan",
    booktitle = "Proceedings of the 57th Annual Meeting of the Association for Computational Linguistics",
    year = "2019",
    %address = "Florence, Italy",
    publisher = "Association for Computational Linguistics",
    url = "https://www.aclweb.org/anthology/P19-1493",
    doi = "10.18653/v1/P19-1493",
    pages = "4996--5001",
}

@inproceedings{DBLP:conf/aaai/SchickS19,
  author    = {Timo Schick and
               Hinrich Sch{\"{u}}tze},
  title     = {Learning Semantic Representations for Novel Words: Leveraging Both
               Form and Context},
  booktitle = {The Thirty-Third {AAAI} Conference on Artificial Intelligence, {AAAI}
               2019, The Thirty-First Innovative Applications of Artificial Intelligence
               Conference, {IAAI} 2019, The Ninth {AAAI} Symposium on Educational
               Advances in Artificial Intelligence, {EAAI} 2019},
  pages     = {6965--6973},
  publisher = {Association for the Advancement of Artificial Intelligence (AAAI) Press},
  year      = {2019},
  url       = {https://doi.org/10.1609/aaai.v33i01.33016965},
  doi       = {10.1609/aaai.v33i01.33016965},
  bibsource = {dblp computer science bibliography, https://dblp.org}
}

@inproceedings{DBLP:journals/corr/abs-1904-01617,
  title = "Attentive Mimicking: Better Word Embeddings by Attending to Informative Contexts",
    author = {Schick, Timo  and
      Sch{\"u}tze, Hinrich},
    booktitle = "Proceedings of the 2019 Conference of the North {A}merican Chapter of the Association for Computational Linguistics: Human Language Technologies, Volume 1 (Long and Short Papers)",
    year = "2019",
    %address = "Minneapolis, Minnesota",
    publisher = "Association for Computational Linguistics",
    url = "https://www.aclweb.org/anthology/N19-1048",
    doi = "10.18653/v1/N19-1048",
    pages = "489--494",
}

@inproceedings{DBLP:journals/corr/abs-1910-07181,
    author    = "Timo Schick and
               Hinrich Sch{\"{u}}tze",
    title     = "{BERTRAM:} Improved Word Embeddings Have Big Impact on Contextualized
               Model Performance",
    booktitle = "Proceedings of the 58th Annual Meeting of the Association for Computational Linguistics",
    year = "2020",
    publisher = "Association for Computational Linguistics",
}

@article{10.1093/bioinformatics/btz682,
    title={BioBERT: a pre-trained biomedical language representation model for biomedical text mining},
  author={Lee, Jinhyuk and Yoon, Wonjin and Kim, Sungdong and Kim, Donghyeon and Kim, Sunkyu and So, Chan Ho and Kang, Jaewoo},
  journal={Bioinformatics},
  volume={36},
  number={4},
  pages={1234--1240},
  year={2020},
  publisher={Oxford University Press}
}

@inproceedings{alsentzer-etal-2019-publicly,
    title = "Publicly Available Clinical {BERT} Embeddings",
    author = "Alsentzer, Emily  and
      Murphy, John  and
      Boag, William  and
      Weng, Wei-Hung  and
      Jindi, Di  and
      Naumann, Tristan  and
      McDermott, Matthew",
    booktitle = "Proceedings of the 2nd Clinical Natural Language Processing Workshop",
    year = "2019",
    %address = "Minneapolis, Minnesota, USA",
    publisher = "Association for Computational Linguistics",
    url = "https://www.aclweb.org/anthology/W19-1909",
    doi = "10.18653/v1/W19-1909",
    pages = "72--78",
}

@inproceedings{speer2016conceptnet,
  author = {Speer, Robyn and Chin, Joshua and Havasi, Catherine},
    title = {ConceptNet 5.5: An Open Multilingual Graph of General Knowledge},
    year = {2017},
    publisher = {Association for the Advancement of Artificial Intelligence (AAAI) Press},
    booktitle = {Proceedings of the Thirty-First AAAI Conference on Artificial Intelligence},
    pages = {4444–4451},
    numpages = {8},
    %location= {San Francisco, California, USA},
    series = {AAAI’17}
}

@article{dbpedia-swj,
  author = {Lehmann, Jens and Isele, Robert and Jakob, Max and Jentzsch, Anja and Kontokostas, Dimitris and Mendes, Pablo N. and Hellmann, Sebastian and Morsey, Mohamed and van Kleef, Patrick and Auer, S\"{o}ren and Bizer, Christian},
  journal = {Semantic Web Journal},
  number = 2,
  pages = {167--195},
  title = {{DBpedia} - A Large-scale, Multilingual Knowledge Base Extracted from Wikipedia},
  url = {http://jens-lehmann.org/files/2015/swj_dbpedia.pdf},
  volume = 6,
  year = 2015
}

@inproceedings{zhang2020SemBERT,
  author    = {Zhuosheng Zhang and
               Yuwei Wu and
               Hai Zhao and
               Zuchao Li and
               Shuailiang Zhang and
               Xi Zhou and
               Xiang Zhou},
  title     = {Semantics-Aware {BERT} for Language Understanding},
  booktitle = {The Thirty-Fourth {AAAI} Conference on Artificial Intelligence, {AAAI}
               2020, The Thirty-Second Innovative Applications of Artificial Intelligence
               Conference, {IAAI} 2020, The Tenth {AAAI} Symposium on Educational
               Advances in Artificial Intelligence, {EAAI} 2020},
  pages     = {9628--9635},
  publisher = {Association for the Advancement of Artificial Intelligence (AAAI) Press},
  year      = {2020},
  url       = {https://aaai.org/ojs/index.php/AAAI/article/view/6510},
  bibsource = {dblp computer science bibliography, https://dblp.org},
}

@inproceedings{peters-etal-2019-knowledge,
    title = "Knowledge Enhanced Contextual Word Representations",
    author = "Peters, Matthew E.  and
      Neumann, Mark  and
      Logan, Robert  and
      Schwartz, Roy  and
      Joshi, Vidur  and
      Singh, Sameer  and
      Smith, Noah A.",
    booktitle = "Proceedings of the 2019 Conference on Empirical Methods in Natural Language Processing and the 9th International Joint Conference on Natural Language Processing (EMNLP-IJCNLP)",
    year = "2019",
    %address = "Hong Kong, China",
    publisher = "Association for Computational Linguistics",
    url = "https://www.aclweb.org/anthology/D19-1005",
    doi = "10.18653/v1/D19-1005",
    pages = "43--54",
}

@article{wang2019kepler,
  title={KEPLER: A Unified Model for Knowledge Embedding and Pre-trained Language Representation},
  author={Wang, Xiaozhi and Gao, Tianyu and Zhu, Zhaocheng and Liu, Zhiyuan and Li, Juanzi and Tang, Jian},
  year={2019},
  journal   = {CoRR},
  volume    = {abs/1911.06136},
  url       = {https://arxiv.org/abs/1911.06136},
}

@inproceedings{zhang-etal-2019-ernie,
    title = "{ERNIE}: Enhanced Language Representation with Informative Entities",
    author = "Zhang, Zhengyan  and
      Han, Xu  and
      Liu, Zhiyuan  and
      Jiang, Xin  and
      Sun, Maosong  and
      Liu, Qun",
    booktitle = "Proceedings of the 57th Annual Meeting of the Association for Computational Linguistics",
    year = "2019",
    %address = "Florence, Italy",
    publisher = "Association for Computational Linguistics",
    url = "https://www.aclweb.org/anthology/P19-1139",
    doi = "10.18653/v1/P19-1139",
    pages = "1441--1451",
}

@inproceedings{liu-etal-2018-learning,
    title = "Learning How to Actively Learn: A Deep Imitation Learning Approach",
    author = "Liu, Ming  and
      Buntine, Wray  and
      Haffari, Gholamreza",
    booktitle = "Proceedings of the 56th Annual Meeting of the Association for Computational Linguistics (Volume 1: Long Papers)",
    year = "2018",
    %address = "Melbourne, Australia",
    publisher = "Association for Computational Linguistics",
    url = "https://www.aclweb.org/anthology/P18-1174",
    doi = "10.18653/v1/P18-1174",
    pages = "1874--1883",
}

@inproceedings{Du_2019,
    title = "An Empirical Comparison on Imitation Learning and Reinforcement Learning for Paraphrase Generation",
    author = "Du, Wanyu  and
      Ji, Yangfeng",
    booktitle = "Proceedings of the 2019 Conference on Empirical Methods in Natural Language Processing and the 9th International Joint Conference on Natural Language Processing (EMNLP-IJCNLP)",
    year = "2019",
    %address = "Hong Kong, China",
    publisher = "Association for Computational Linguistics",
    url = "https://www.aclweb.org/anthology/D19-1619",
    doi = "10.18653/v1/D19-1619",
    pages = "6012--6018",
}

@inproceedings{dozat-etal-2017-stanfords,
    title = "{S}tanford{'}s Graph-based Neural Dependency Parser at the {C}o{NLL} 2017 Shared Task",
    author = "Dozat, Timothy  and
      Qi, Peng  and
      Manning, Christopher D.",
    booktitle = "Proceedings of the {C}o{NLL} 2017 Shared Task: Multilingual Parsing from Raw Text to Universal Dependencies",
    year = "2017",
    %address = "Vancouver, Canada",
    publisher = "Association for Computational Linguistics",
    url = "https://www.aclweb.org/anthology/K17-3002",
    doi = "10.18653/v1/K17-3002",
    pages = "20--30",
}

@article{kiperwasser-goldberg-2016-simple,
    author    = {Eliyahu Kiperwasser and
               Yoav Goldberg},
    title     = {Simple and Accurate Dependency Parsing Using Bidirectional {LSTM}
               Feature Representations},
    journal = "Transactions of the Association for Computational Linguistics",
    volume    = {4},
    pages     = {313--327},
    year      = {2016},
    url       = {https://transacl.org/ojs/index.php/tacl/article/view/885},
    bibsource = {dblp computer science bibliography, https://dblp.org}

}

@inproceedings{song-etal-2019-leveraging,
    title = "Leveraging Dependency Forest for Neural Medical Relation Extraction",
    author = "Song, Linfeng  and
      Zhang, Yue  and
      Gildea, Daniel  and
      Yu, Mo  and
      Wang, Zhiguo  and
      Su, Jinsong",
    booktitle = "Proceedings of the 2019 Conference on Empirical Methods in Natural Language Processing and the 9th International Joint Conference on Natural Language Processing (EMNLP-IJCNLP)",
    year = "2019",
    %address = "Hong Kong, China",
    publisher = "Association for Computational Linguistics",
    url = "https://www.aclweb.org/anthology/D19-1020",
    doi = "10.18653/v1/D19-1020",
    pages = "208--218",
}

@inproceedings{Gao2020NeuralSF,
    title = "Neural Snowball for Few-Shot Relation Learning",
    author = "Gao, Tianyu and Han, Xu and Xie, Ruobing and Liu, Zhiyuan and Lin, Fen and Lin, Leyu and Sun, Maosong",
    booktitle = "Proceedings the Thirty-fourth AAAI Conference on Artificial Intelligence",
    publisher = "Association for the Advancement of Artificial Intelligence (AAAI) Press",
    year = "2020"
}

@inproceedings{wu-etal-2019-open,
    title = "Open Relation Extraction: Relational Knowledge Transfer from Supervised Data to Unsupervised Data",
    author = "Wu, Ruidong  and
      Yao, Yuan  and
      Han, Xu  and
      Xie, Ruobing  and
      Liu, Zhiyuan  and
      Lin, Fen  and
      Lin, Leyu  and
      Sun, Maosong",
    booktitle = "Proceedings of the 2019 Conference on Empirical Methods in Natural Language Processing and the 9th International Joint Conference on Natural Language Processing (EMNLP-IJCNLP)",
    year = "2019",
    %address = "Hong Kong, China",
    publisher = "Association for Computational Linguistics",
    url = "https://www.aclweb.org/anthology/D19-1021",
    doi = "10.18653/v1/D19-1021",
    pages = "219--228",
}

@inproceedings{Han_2019,
    title = "{O}pen{NRE}: An Open and Extensible Toolkit for Neural Relation Extraction",
    author = "Han, Xu  and
      Gao, Tianyu  and
      Yao, Yuan  and
      Ye, Deming  and
      Liu, Zhiyuan  and
      Sun, Maosong",
    booktitle = "Proceedings of the 2019 Conference on Empirical Methods in Natural Language Processing and the 9th International Joint Conference on Natural Language Processing (EMNLP-IJCNLP): System Demonstrations",
    year = "2019",
    %address = "Hong Kong, China",
    publisher = "Association for Computational Linguistics",
    url = "https://www.aclweb.org/anthology/D19-3029",
    doi = "10.18653/v1/D19-3029",
    pages = "169--174",
}

@article{Hu2020SelfORESR,
  title={SelfORE: Self-supervised Relational Feature Learning for Open Relation Extraction},
  author={Hu, Xuming and Wen, Lijie and Xu, Yusong and Zhang, Chenwei and Yu, Philip S},
  year={2020},
  journal   = {CoRR},
  volume    = {abs/2004.02438},
  url       = {https://arxiv.org/abs/2004.02438},
}

@inproceedings{stanovsky-dagan-2016-creating,
    title = "Creating a Large Benchmark for Open Information Extraction",
    author = "Stanovsky, Gabriel  and
      Dagan, Ido",
    booktitle = "Proceedings of the 2016 Conference on Empirical Methods in Natural Language Processing",
    year = "2016",
    %address = "Austin, Texas",
    publisher = "Association for Computational Linguistics",
    url = "https://www.aclweb.org/anthology/D16-1252",
    doi = "10.18653/v1/D16-1252",
    pages = "2300--2305",
}

@book{eisenstein2019introduction,
  title={Introduction to Natural Language Processing},
  author={Eisenstein, J.},
  isbn={9780262042840},
  lccn={2018059552},
  series={Adaptive Computation and Machine Learning series},
  url={https://books.google.no/books?id=72yuDwAAQBAJ},
  year={2019},
  publisher={MIT Press, Cambridge, MA, United States},
}

@inproceedings{rajpurkar-etal-2016-squad,
    title = "{SQ}u{AD}: 100,000+ Questions for Machine Comprehension of Text",
    author = "Rajpurkar, Pranav  and
      Zhang, Jian  and
      Lopyrev, Konstantin  and
      Liang, Percy",
    booktitle = "Proceedings of the 2016 Conference on Empirical Methods in Natural Language Processing",
    year = "2016",
    %address = "Austin, Texas",
    publisher = "Association for Computational Linguistics",
    url = "https://www.aclweb.org/anthology/D16-1264",
    doi = "10.18653/v1/D16-1264",
    pages = "2383--2392",
}

@inproceedings{sennrich-etal-2016-neural,
    title = "Neural Machine Translation of Rare Words with Subword Units",
    author = "Sennrich, Rico  and
      Haddow, Barry  and
      Birch, Alexandra",
    booktitle = "Proceedings of the 54th Annual Meeting of the Association for Computational Linguistics (Volume 1: Long Papers)",
    year = "2016",
    %address = "Berlin, Germany",
    publisher = "Association for Computational Linguistics",
    url = "https://www.aclweb.org/anthology/P16-1162",
    doi = "10.18653/v1/P16-1162",
    pages = "1715--1725",
}

@article{DBLP:journals/corr/abs-1907-11692,
  author    = {Yinhan Liu and
               Myle Ott and
               Naman Goyal and
               Jingfei Du and
               Mandar Joshi and
               Danqi Chen and
               Omer Levy and
               Mike Lewis and
               Luke Zettlemoyer and
               Veselin Stoyanov},
  title     = {RoBERTa: {A} Robustly Optimized {BERT} Pretraining Approach},
  journal   = {CoRR},
  volume    = {abs/1907.11692},
  year      = {2019},
  url       = {http://arxiv.org/abs/1907.11692},
  archivePrefix = {arXiv},
  eprint    = {1907.11692},
  bibsource = {dblp computer science bibliography, https://dblp.org}
}

@inproceedings{kudo-richardson-2018-sentencepiece,
    title = "{S}entence{P}iece: A simple and language independent subword tokenizer and detokenizer for Neural Text Processing",
    author = "Kudo, Taku  and
      Richardson, John",
    booktitle = "Proceedings of the 2018 Conference on Empirical Methods in Natural Language Processing: System Demonstrations",
    year = "2018",
    %address = "Brussels, Belgium",
    publisher = "Association for Computational Linguistics",
    url = "https://www.aclweb.org/anthology/D18-2012",
    doi = "10.18653/v1/D18-2012",
    pages = "66--71",
}

@inproceedings{8578229,
  author={F. {Sung} and Y. {Yang} and L. {Zhang} and T. {Xiang} and P. H. S. {Torr} and T. M. {Hospedales}},
  booktitle={2018 IEEE/CVF Conference on Computer Vision and Pattern Recognition},
  title={Learning to Compare: Relation Network for Few-Shot Learning},
  year={2018},
  pages={1199-1208},
  publisher={IEEE}

}

@inproceedings{zoph-etal-2016-transfer,
    title = "Transfer Learning for Low-Resource Neural Machine Translation",
    author = "Zoph, Barret  and
      Yuret, Deniz  and
      May, Jonathan  and
      Knight, Kevin",
    booktitle = "Proceedings of the 2016 Conference on Empirical Methods in Natural Language Processing",
    year = "2016",
    %address = "Austin, Texas",
    publisher = "Association for Computational Linguistics",
    url = "https://www.aclweb.org/anthology/D16-1163",
    doi = "10.18653/v1/D16-1163",
    pages = "1568--1575",
}

@inproceedings{obamuyide-vlachos-2019-model,
    title = "Model-Agnostic Meta-Learning for Relation Classification with Limited Supervision",
    author = "Obamuyide, Abiola  and
      Vlachos, Andreas",
    booktitle = "Proceedings of the 57th Annual Meeting of the Association for Computational Linguistics",
    year = "2019",
    %address = "Florence, Italy",
    publisher = "Association for Computational Linguistics",
    url = "https://www.aclweb.org/anthology/P19-1589",
    doi = "10.18653/v1/P19-1589",
    pages = "5873--5879",
}

@inproceedings{Mi2019MetaLearningFL,
  title     = {Meta-Learning for Low-resource Natural Language Generation in Task-oriented Dialogue Systems},
  author    = {Mi, Fei and Huang, Minlie and Zhang, Jiyong and Faltings, Boi},
  booktitle = {Proceedings of the Twenty-Eighth International Joint Conference on
               Artificial Intelligence, {IJCAI-19}},
  publisher = {International Joint Conferences on Artificial Intelligence Organization},
  pages     = {3151--3157},
  year      = {2019},
  doi       = {10.24963/ijcai.2019/437},
  url       = {https://doi.org/10.24963/ijcai.2019/437},
}
\end{document}